%% file: main.tex
\pdfoutput=1

\documentclass[11pt]{article}
\usepackage[table]{xcolor}
\usepackage{authblk}
\makeatletter
\renewcommand\AB@affilsepx{\quad \protect\Affilfont}
\makeatother

\PassOptionsToPackage{numbers, compress}{natbib}

\usepackage[]{acl}

\usepackage{times}
\usepackage{latexsym}

\usepackage[T1]{fontenc}

\usepackage[utf8]{inputenc}
\usepackage{microtype}
\usepackage{listings}
\usepackage{multirow}
\usepackage{hhline}
\usepackage{changepage} 
\usepackage{adjustbox}
\usepackage{subcaption} 
\usepackage{url}
\usepackage{wrapfig}
\usepackage{booktabs}
\usepackage{siunitx}
\usepackage{amsmath,amsfonts,amssymb}
\usepackage{cleveref}
\usepackage{tabularx}
\usepackage{hyperref}
\usepackage{enumitem}
\usepackage{float}
\usepackage{xspace}

\input{math.tex}

\input{commands.tex}

\usepackage[edges]{forest}
\definecolor{hiddendraw}{RGB}{205, 44, 36}
\definecolor{hidden-blue}{RGB}{194,232,247}
\definecolor{hidden-orange}{RGB}{243,202,120}
\definecolor{hidden-yellow}{RGB}{242,244,193}
\definecolor{hidden-red}{RGB}{255,0,0}
\definecolor{hidden-grey}{RGB}{122,122,122}
\usepackage{xcolor}         
\definecolor{mydarkblue}{rgb}{0,0.08,0.55}
\definecolor{DarkRed}{HTML}{780000}
\definecolor{RegRed}{HTML}{C1121F}
\definecolor{CoolBlue}{HTML}{669BBC}
\definecolor{PlotGreen}{RGB}{81,157,62}
\definecolor{PlotOrange}{RGB}{239,133,54}
\definecolor{PlotBlue}{RGB}{59,118,175}
\definecolor{PlotRed}{RGB}{197,57,50}

\newcommand*{\eg}{e.g.,\@\xspace}
\newcommand*{\ie}{i.e.,\@\xspace}

\tikzstyle{mybox}=[
    rectangle,
    draw=hiddendraw,
    rounded corners,
    text opacity=1,
    minimum height=1.5em,
    minimum width=5em,
    inner sep=2pt,
    align=center,
    fill opacity=.5,
    ]

\hypersetup{
    colorlinks=true,
    linkcolor=blue,
    filecolor=magenta,  
    urlcolor=cyan,
    pdfpagemode=FullScreen,
    }

\title{Challenges and Applications of Large Language Models}

\author[$\alpha$, $\dagger$, $\ast$]{Jean Kaddour}
\author[$\beta$, $\ast$]{Joshua Harris}
\author[$\alpha$]{Maximilian Mozes}
\author[$\gamma$, $\delta$, $\epsilon$]{\authorcr Herbie Bradley}
\author[$\zeta$]{Roberta Raileanu}
\author[$\eta$, $\ast$]{Robert McHardy}

\affil[$\alpha$]{University College London}
\affil[$\beta$]{UK Health Security Agency}
\affil[$\gamma$]{EleutherAI\authorcr}
\affil[$\delta$]{University of Cambridge}
\affil[$\epsilon$]{Stability AI}
\affil[$\zeta$]{Meta AI Research}
\affil[$\eta$]{InstaDeep}

\begin{document}
\maketitle
\thispagestyle{plain}
\pagestyle{plain}
\begin{abstract}
Large Language Models (LLMs) went from non-existent to ubiquitous in the machine learning discourse within a few years. Due to the fast pace of the field, it is difficult to identify the remaining challenges and already fruitful application areas. In this paper, we aim to establish a systematic set of open problems and application successes so that ML researchers can comprehend the field's current state more quickly and become productive.
\renewcommand*{\thefootnote}{\fnsymbol{footnote}}
\footnotetext[1]{Equal contribution.}
\footnotetext[2]{\{jean.kaddour,robert.mchardy\}.20@ucl.ac.uk, joshua.harris@ukhsa.gov.uk}
\renewcommand*{\thefootnote}{\arabic{footnote}}
\end{abstract}

\tableofcontents

\section{Introduction}
\begin{figure}
    \centering
    \includegraphics[width=0.49\textwidth]{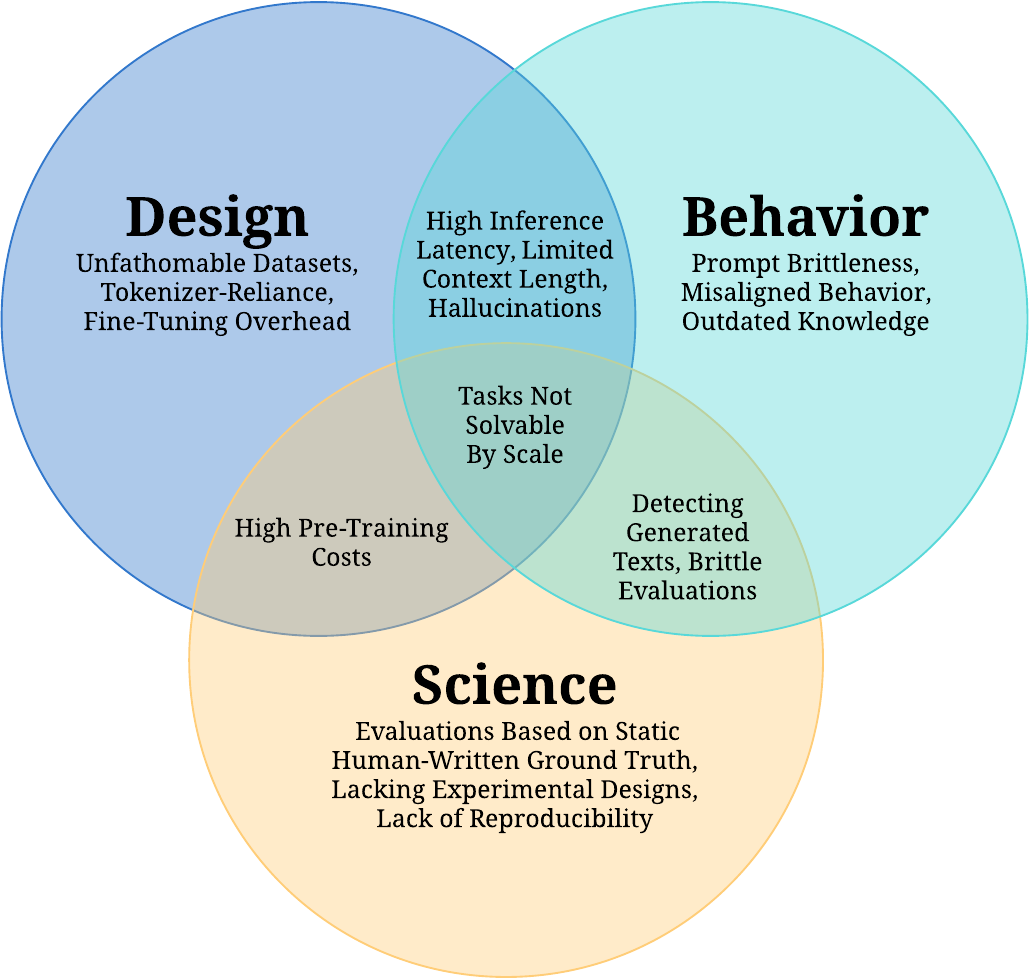}
    \caption{\textbf{Overview of LLM Challenges.} \emph{Design}ing LLMs relates to decisions taken before deployment. \emph{Behavior}ial challenges occur during deployment. \emph{Science} challenges hinder academic progress.}    \label{fig:llm_challenge_overview}
\end{figure}
Given the quickly growing plethora of LLM research papers, we aim to address two questions: (1) \textbf{Challenges}: What problems remain unresolved? and (2) \textbf{Applications}: Where are LLMs currently being applied, and how are the challenges constraining them? 
For (1), we group the challenges in \Cref{fig:llm_challenge_overview} into three broader categories ``Design'', ``Behavior'', and ``Science''. To provide answers for (2), we explore the fields of chatbots, computational biology, computer programming, creative work, knowledge work, law, medicine, reasoning, robotics, and the social sciences.

This paper is an opinionated review and assumes familiarity with LLMs and how they work (we refer to more introductory works in \Cref{sec:rw}). Further, we focus on models trained on text data. We target a technical researcher audience and do not discuss political, philosophical, or moral perspectives on LLMs.

\section{Challenges}
\begin{chall}{Challenge}
    This box highlights a challenge.
\end{chall}

\subsection{Unfathomable Datasets} \label{sec:datasets}
Scaling the amount of pre-training data has been one of the major drivers to equip LLMs with general-purpose capabilities~\citep{kaplan2020scaling}. The size of pre-training datasets quickly outgrew the number of documents most human teams could manually quality-check. Instead, most data collection procedures rely on heuristics regarding data sources and filtering. 

In this section, we explore the adverse consequences of these heuristics and the reality that many model practitioners possess only a nebulous understanding of the data on which their model has been trained. We refer to this issue as follows.
\begin{chall}{Unfathomable Datasets}
The size of modern pre-training datasets renders it impractical for any individual to read or conduct quality assessments on the encompassed documents thoroughly. \end{chall}

\paragraph{Near-Duplicates} can arise in different forms and have been reported to degrade model
performance~\citep{lee2021deduplicating,hernandez2022scaling,minipile}. Near-duplicates are harder to find compared to \emph{exact} duplicates; filtering out of such is a standard step in most data collection pipelines, \eg using the MinHash algorithm~\citep{broder1998min}.~\citet{lee2021deduplicating} propose the \emph{NearDup} method and find that over $1\%$ of tokens emitted unprompted from a model are part of a memorized sequence of the C4 dataset, \eg it contains a 61-word sequence repeated $61,036$ times in the training split. By deduplicating it, they reduce the rate of emitted memorizations by $10$x.~\citet{abbas2023semdedup} introduce \emph{SemDeDup}, a technique designed to identify \emph{semantic} duplicates that, although perceptually distinct, convey predominantly similar information, such as sentences with analogous structures with certain words replaced by synonyms. After applying their method to C4, they find that it improves over \emph{NearDup}. Similarly,~\citet{minipile} find near-duplicates in the Pile~\citep{gao2020pile} by clustering document embeddings and identifying clusters gathering duplicates.

\paragraph{Benchmark Data Contamination} occurs when the training dataset contains data from or similar to the evaluation test set. This can lead to inflated performance metrics, as the model can memorize the test data and simply regurgitate it back during testing. 

Finding and removing all training and test data overlaps is difficult in practice. For example, the GPT-3 authors~\citet{brown2020gpt3} found a code bug after training, resulting in only partially removing all detected overlaps from the training data. They could not afford to retrain the model, so they used it with the remaining overlaps and ``cleaned'' variants of the considered benchmarks, with all potentially leaked examples removed. They define overlapping examples as examples that share at least 13 consecutive words with any other example in the pre-training set. If an example is shorter than 13 words, they consider it overlapping if it shares all of its words with another example.

Similarly,~\citet{dodge2021documenting} search for test data in the web-crawled C4 corpus but measure exact matches, normalized for capitalization and punctuation. They find various input-and-label contaminations of text generation and knowledge completion tasks; and input-only contaminations of the GLUE benchmark.  
They argue that there are two ways test data can end up in a snapshot of Common Crawl (the original dump source of C4): either a given test set is built from a web text or uploaded after creation.~\citet{Lmcontamination} ask ChatGPT to generate academic benchmark instances, finding that it has memorized multiple ones, including some test splits.~\citet{jacoviStopUploadingTest2023} propose three strategies to mitigate contamination, including encryption and training exclusion controls.

\paragraph{Personally Identifiable Information (PII)} such as phone numbers and email addresses, have been found within pre-training corpora, resulting in privacy leaks during prompting.~\citet{carlini2019secret,extracting_data,lukasAnalyzingLeakagePersonally2023} extract PII data by prompting GPT-2;~\citet{kulkarniGitHubCopilotAI2021} report how an engineer yields secret API keys by prompting GitHub Copilot.~\citet{henderson2022pile} discuss the availability of PII in law data across different jurisdictions and filter it based on the legal norm in the respective jurisdiction.
\citet{el-mhamdiImpossibleSafetyLarge2023} contend that because strong model performance typically requires memorization of the training data~\citep{feldman2020does,brown2021memorization}, the (undetected) existence of PII in the training data will likely result in models that render them extractable. 

\input{tables/pretrain_datasets.tex}

\paragraph{Pre-Training Domain Mixtures} Several studies have argued for diversity in the pre-training corpus ~\citep{gao2020pile,longprepretrainerGuideTraining2023,leeScaleDiversityCoefficient2023}. Many popular corpora follow this by concatenating datasets from different sources, as illustrated in \Cref{tab:datasets}. However, it remains underexplored what amount of data from different sources is necessary for strong downstream performances. Finding suboptimal mixtures can cause low transferability to downstream tasks~\citep{wang2019characterizing,wang2022evaluating} and reliance on spurious correlations~\citep{cml,wu2022generating,lynch2023spawrious}. 
\citet{xieDoReMiOptimizingData2023} find domain mixture proportions by training a small proxy model using group-distributionally robust optimization~\citep{sagawa2020distributionally}; surprisingly, they find that the final model trained using their found domain weights yields improved perplexity across all domains, even when it down-weights a domain. Given a target downstream task,~\citet{yaoNLPScratchLargeScale2022,xieDataSelectionLanguage2023} select subsets most useful for pre-training. ~\citet{longprepretrainerGuideTraining2023} measure the effects of domain compositions and find that inclusion of heterogeneous data sources is broadly beneficial and likely more important than the data quality (as measured by the document quality classifier employed by PaLM \cite{chowdhery2022palm} and GLaM \cite{glam}) or size, which also motivates smaller yet more diverse pre-training datasets~\citep{minipile}.

\paragraph{Fine-Tuning Task Mixtures} have to be determined for fine-tuning a pre-trained model on many different tasks, usually with comparatively few examples per task. This technique, which we call multitask-prompted fine-tuned LMs (MTLMs), has demonstrated significant generalization improvements with very little additional training compute. For example, \emph{instruction fine-tuning} via task instructions prepended to each set of input-output pairs is a very popular scheme, which we will later discuss in more detail in \Cref{par:instruction_finetuning}.~\citet{wang2022super} propose \texttt{Super-NaturalInstructions}, a fine-tuning dataset with 1,616 diverse tasks and expert-written instructions.~\citet{muennighoff2022crosslingual} extend MTLM to the multilingual setting, showing that fine-tuning on multilingual tasks with English prompts improves results on tasks in all languages.

However, similar to the previous paragraph, how to balance the task datasets well remains unclear. As the tasks can vary in size considerably,~\citet{raffel2022t5} mix each task in proportion to the number of examples in its 'train' split (up to some \texttt{max\_num\_examples}).~\citet{jangExploringBenefitsTraining2023a} report that MTLMs can underperform expert LLMs fine-tuned on only a single task because of (i) negative task transfer, where learning multiple tasks at once hinders the learning of some specific tasks, and (ii) catastrophic forgetting of previous tasks when learning new tasks.~\citet{opt_iml} study varying task (sets) proportions, finding several trade-offs and concluding that the right values for these parameters depend on the downstream end-goals.~\citet{flan_collection} balance different sets of task sources by omitting them, one at a time, and ranking their contributions on the MMLU benchmark~\citep{mmlu}; further, they mix the input prompt templates of zero- and few-shot prompting; finding that this improves the performance in both settings. Another trend is to imitate closed-source models like ChatGPT by collecting a dataset of API outputs (against OpenAI's terms and conditions) and fine-tuning an open-source LM with it~\citep{taori2023alpaca}. However,~\citet{gudibande2023false} point out that such imitation models are only good at mimicking the proprietary model's style but not its content, a distinction that has been discussed extensively in the causality literature~\citep{cml}. They conclude that substantial capability gaps between fine-tuned open-sourced and closed-source models remain, motivating future work for better imitation data. 

\subsection{Tokenizer-Reliance} Tokenization is the process of breaking a sequence of words or characters into smaller units called tokens, such that they can be fed into the model. One common tokenization approach is \emph{subword tokenization}, where we split words into smaller units, called \emph{subwords} or \emph{WordPieces}~\citep{BPE}. The goal is to handle rare and out-of-vocabulary words in a model's vocabulary effectively while maintaining a limited number of tokens per sequence in the interest of computational complexity. Subword tokenizers are usually trained unsupervised to build a vocabulary and optionally merge rules to encode the training data efficiently. 

\begin{figure*}
    \includegraphics[width=\textwidth]{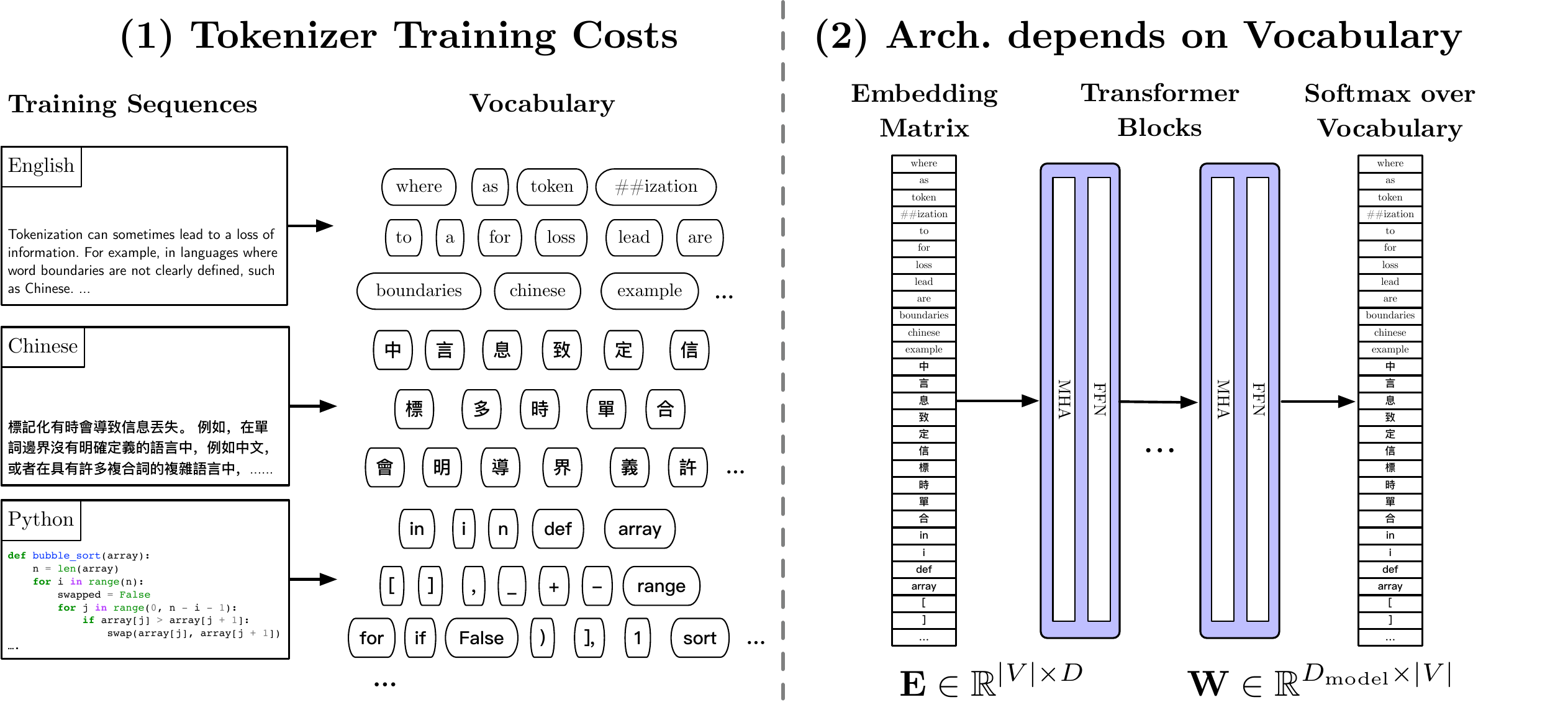}
    \caption{\textbf{Exemplary Drawbacks of relying on Tokenization.} (1) The tokenizer training step involves non-trivial computations, \eg multiple passes over the entire pre-training dataset, and introduces a dependency on it, which can become especially problematic in multilingual settings. (2) The embedding layer $\mathbf E$ and output layer $\mathbf W$ of LLMs involve the vocabulary size; \eg making up $\approx 66\%$ of the model's parameter count in T5 models \cite{xueByT5TokenfreeFuture2022}.}
\end{figure*}

However, the necessity of tokenization comes with multiple drawbacks~\citep{karpathy2023tokenization}; some of which we discuss below. For example,~\citet{ahia2023all, petrov2023language} show that \textbf{the number of tokens necessary to convey the same information varies significantly across languages}, making the pricing policy of API language models, which charge users based on the number of processed or generated tokens, potentially unfair. They find that users of many supported languages are overcharged while receiving subpar results, with this group predominantly residing in areas where these APIs are already less affordable.

Further, \textbf{discrepancies between the data that a tokenizer and a model have been trained on can lead to glitch tokens}~\citep{rumbelowSolidGoldMagikarpPromptGeneration}, which can subsequently cause unexpected model behavior as their corresponding embeddings are essentially untrained. This coupling between the tokenizer and pre-training corpus creates the burden of a new training run of the tokenizer each time the pre-training corpus is modified. 

Next, Tokenization schemes that work well in a multilingual setting, particularly with non-space-separated languages such as Chinese or Japanese, remain challenging~\citep{fujii2023different, chung2023unimax}.

Existing subword tokenization schemes are predominantly greedy algorithms trying to encode language as efficiently as possible regarding the number of tokens used. Naturally, these methods favor subwords comprising larger parts of the training data and, therefore, subwords that are shared across many languages. This favors languages with shared scripts like Latin and Cyrillic, resulting in suboptimal tokenization of low-resource languages~\citep{chung-etal-2020-improving, zheng2021allocating}.

\begin{chall}{Tokenizer-Reliance}
    Tokenizers introduce several challenges, \eg computational overhead, language dependence, handling of novel words, fixed vocabulary size, information loss, and low human interpretability. 
\end{chall}

\paragraph{Subword-Level Inputs} are the dominant paradigm, providing a good trade-off between vocabulary size and sequence length. \textbf{Byte-Pair Encoding~\citep{BPE, wang2020neural}} (BPE) starts with the set of symbols (characters or bytes) that comprise the training data. The tokenizer is then trained to learn rules to merge the most frequent pair of two consecutive tokens---defined by the existing vocabulary---into a new vocabulary item. Byte-level BPE (BBPE)~\citep{wang2020neural} is an extension of BPE with byte-level subwords, particularly suited for multilingual tasks where it enables vocabulary sharing between languages. A trained BPE tokenizer applies the previously learned rules to tokenize inputs.
\textbf{WordPiece~\citep{schuster2012wordpiece, wu2016wordpiece}} is a closed-source tokenization algorithm used, \eg in BERT~\citep{devlin-etal-2019-bert}. Like BPE, WordPiece starts with a small initial vocabulary, which is iteratively extended by learning merge rules and creating new vocabulary items. Rather than selecting the most frequent pair of consecutive tokens, WordPiece uses a scoring function to normalize the frequency of the pair by the frequencies of the individual tokens to prioritize common pairs with rare individual tokens.
\textbf{Unigram Tokenization~\citep{kudo-2018-subword}} iteratively trims a large base vocabulary to a given target size. To this end, at each step of the tokenizer training, a unigram language model is used to compute a loss over the training data conditional on a certain vocabulary item being removed. A proportion of the subwords with the lowest losses are removed to form the base vocabulary for the next iteration. Unigram tokenization is probabilistic, \ie during inference, all possible tokenizations of a given sequence are scored using the unigram language model, and the most likely one is selected.
\textbf{SentencePiece~\citep{kudo2018sentencepiece}} is a commonly used open-source library, implementing several tokenization algorithms such as (B)BPE and Unigram tokenization. SentencePiece also implements non-subword tokenization approaches like word- and character-level tokenization.

\paragraph{Byte-Level Inputs} are an alternative to subword tokenization is use byte-level inputs. Byte-level inputs can either be used in combination with subword tokenizers~\citep{wang2020neural} or used to define a limited vocabulary that can be used to encode all possible sequences. For example, ~\citet{xue-etal-2022-byt5} train a non-subword mT5 model using UTF-8 bytes rather than subword tokens as inputs, showing promising performance on multilingual data. While this enables subword-free LLMs, UTF-8 encodes Latin languages with fewer bytes than \eg{} Chinese, Japanese or Korean\footnote{\href{https://www.unicode.org/versions/Unicode15.0.0/}{https://www.unicode.org/versions/Unicode15.0.0/}}.~\citet{tay2022charformer} propose the Charformer, a tokenization-free model which learns a soft subword tokenization in latent space (Gradient-Based Subword Tokenization) given byte-level inputs. Charformer performs comparably to subword-based models while incurring less computational overhead than other byte or subword models.~\citet{choeBridgingGapTokenizerFree2019} train a small-scale, 0.8B language model based on raw byte-level inputs and show that it performs comparably. On a smaller scale,~\citet{clark-etal-2022-canine} show that their tokenization- and vocabulary-free encoder \emph{Canine} outperforms a comparable tokenization-based model.~\citet{yu2023megabyte} address the computational cost that byte-level tokenization incurs by segmenting input sequences into local patches, which can be processed in parallel. Similarly,~\citet{hortonBytesAreAll2023} propose to operate directly on file bytes. In a parallel line of work,~\citet{rustLanguageModellingPixels2023} render text as images and train an encoder model to predict the raw pixels of the images.

\subsection{High Pre-Training Costs} \label{sec:efficient_training} 
The vast majority of the training costs go toward the pre-training process. Training a single LLM can require hundreds of thousands of compute hours, which in turn cost millions of dollars and consume energy amounts equivalent to that used by several typical US families annually \cite{9810097,chowdhery2022palm,biderman2023pythia}. Recently proposed scaling laws \cite{kaplan2020scaling} posit that model performances scale as a power law with model size, dataset size, and the amount of compute used for training, which is fairly unsustainable and can be classified as \textcolor{red}{Red AI} \cite{schwartzGreenAI2019}, where state-of-the-art results are essentially ``bought'' by spending massive computational resources. For example, depending on the exact law coefficients, reducing the error from 3\% to 2\% can require an order of magnitude more data or compute \cite{sorscher2022beyond}.

\begin{chall}{Unsustainable Loss Power-Law~\citep{kaplan2020scaling}}
Performance increases through larger compute budgets but at a decreasing rate if the model or dataset size is fixed, reflecting a power law with diminishing returns. \end{chall}

In the following, we look at two lines of work aiming at resolving such issues.
\paragraph{Compute-Optimal Training Recipes~\citep{hestness2017deep, kaplan2020scaling}}
In \Cref{sec:datasets}, we discussed how the availability of LLM pre-training data has become abundant through the quickly-spread practice of including web-crawled text. Further, thanks to the introduction of Transformer models~\citep{transformers} and suitable hardware~\citep{hooker2021hardware}, we have scaled models to unprecedented sizes. Assuming that we have not yet reached the limits of data ~\citep{BlancheMinerva_2023,villalobos2022will,penedoRefinedWebDatasetFalcon2023} nor model sizes~\citep{kaplan2020scaling,chinchilla,openai2023gpt4}; currently, the main bottleneck is the amount of compute available \cite{ChipsShortage}. Given a particular budget, how large should the pre-training corpus and model be to maximize training efficiency? 
 
As mentioned at the beginning of this section, one recent proposal is to learn empirical ``\emph{scaling laws}''~\citep{hestness2017deep,kaplan2020scaling}, which describe the relationship between LLM performance and the compute budget, model, and dataset size. These laws can provide the right scaling recipe for compute-optimal training, ideally, even when extrapolating to larger compute budgets. For example, ~\citet{openai2023gpt4} report that they were able to accurately predict the model performance of the full-size GPT-4 model based on the performance of a series of smaller models using at most 10,000x less compute than the full model. 

The exact power law coefficients are still heavily debated. ~\citet{kaplan2020scaling} put forward that the model size should be scaled more aggressively than the dataset size to use a given compute budget optimally. Contrary to this,~\citet{chinchilla} find that many LLMs are undertrained and argue that the number of parameters and data should be scaled equally. However, power laws sometimes come in the form of bounds, which can span an order of magnitude difference in the amount of data to be used given a concrete compute budget \cite{Suchenzang_2023}. Further, the pre-training loss does not always correlate well with downstream performance \cite{kaddour2022when,Liu2022SamePL,ntng}.

The viewpoint of~\citet{touvronLLaMAOpenEfficient2023,vriesGoSmolGo2023,touvronLlamaOpenFoundationa} is that when selecting a model size, the computation resources for later usage (inference) should be considered, not just the one-time training costs. They suggest that it might be beneficial to train a smaller model more intensively upfront to offset larger inference costs in the future. Hence, they train models of various sizes on more tokens than are typically used to achieve the best performance possible, given the model size.

One remaining hurdle of performance prediction is inverse scaling, which we discuss in \Cref{ref:unlearnable}. Since scaling laws were typically constructed in the context of pre-training and thereby decoupled from downstream tasks, it remains an open question of how to predict inverse scaling properties.~\citet{tayScaleEfficientlyInsights2022} find that scaling laws can differ in upstream and downstream setups; aside from only the model size, model shape matters for downstream fine-tuning.

\paragraph{Pre-Training Objectives}

\begin{figure}
    \centering
\includegraphics[width=0.99\columnwidth]{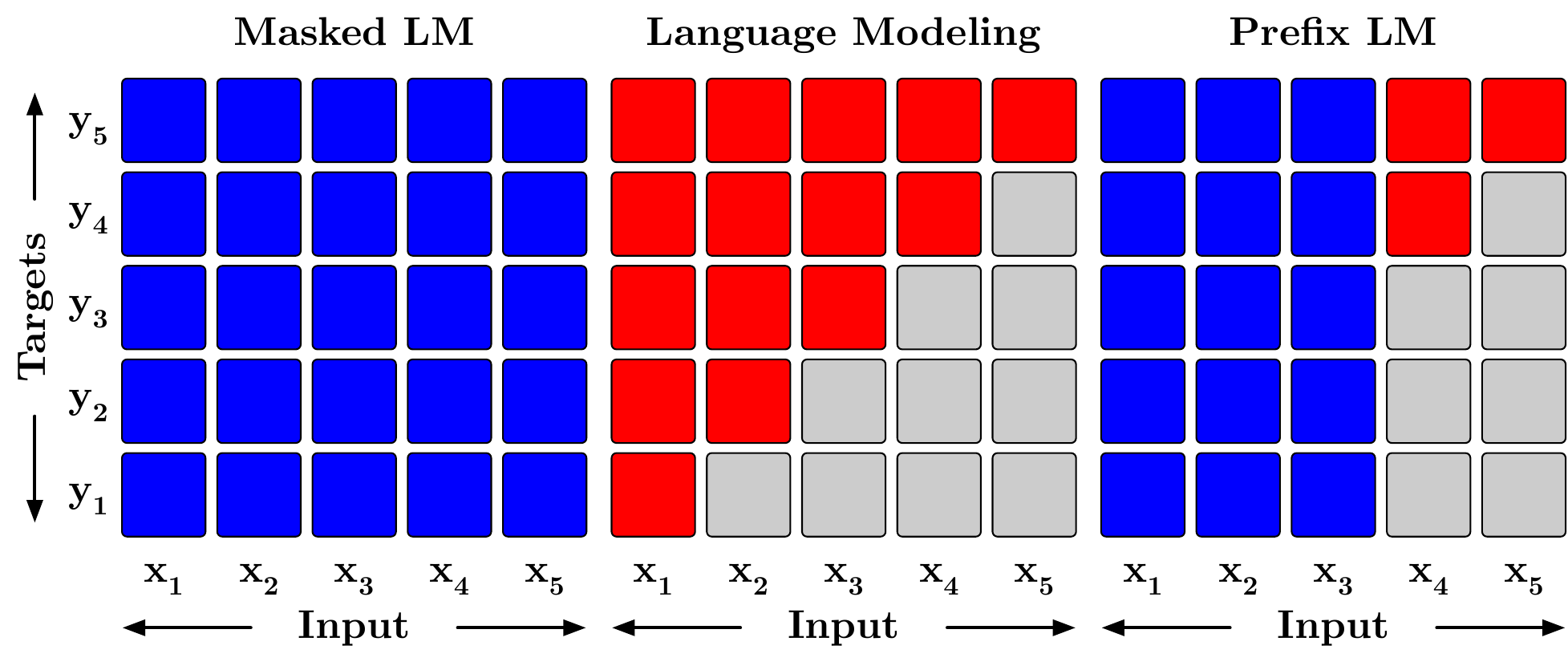}
    \caption{\textbf{Masking Strategies.} Each row denotes to which inputs $\vx_i$ (columns) a particular output $\vy_i$ (row) can attend to (\textcolor{red}{uni-} or bi-\textcolor{blue}{directional}).}
    \label{fig:masking}
\end{figure}

\begin{figure*}[ht!]
    \centering
\includegraphics[width=\textwidth,height=0.60\textheight,
]{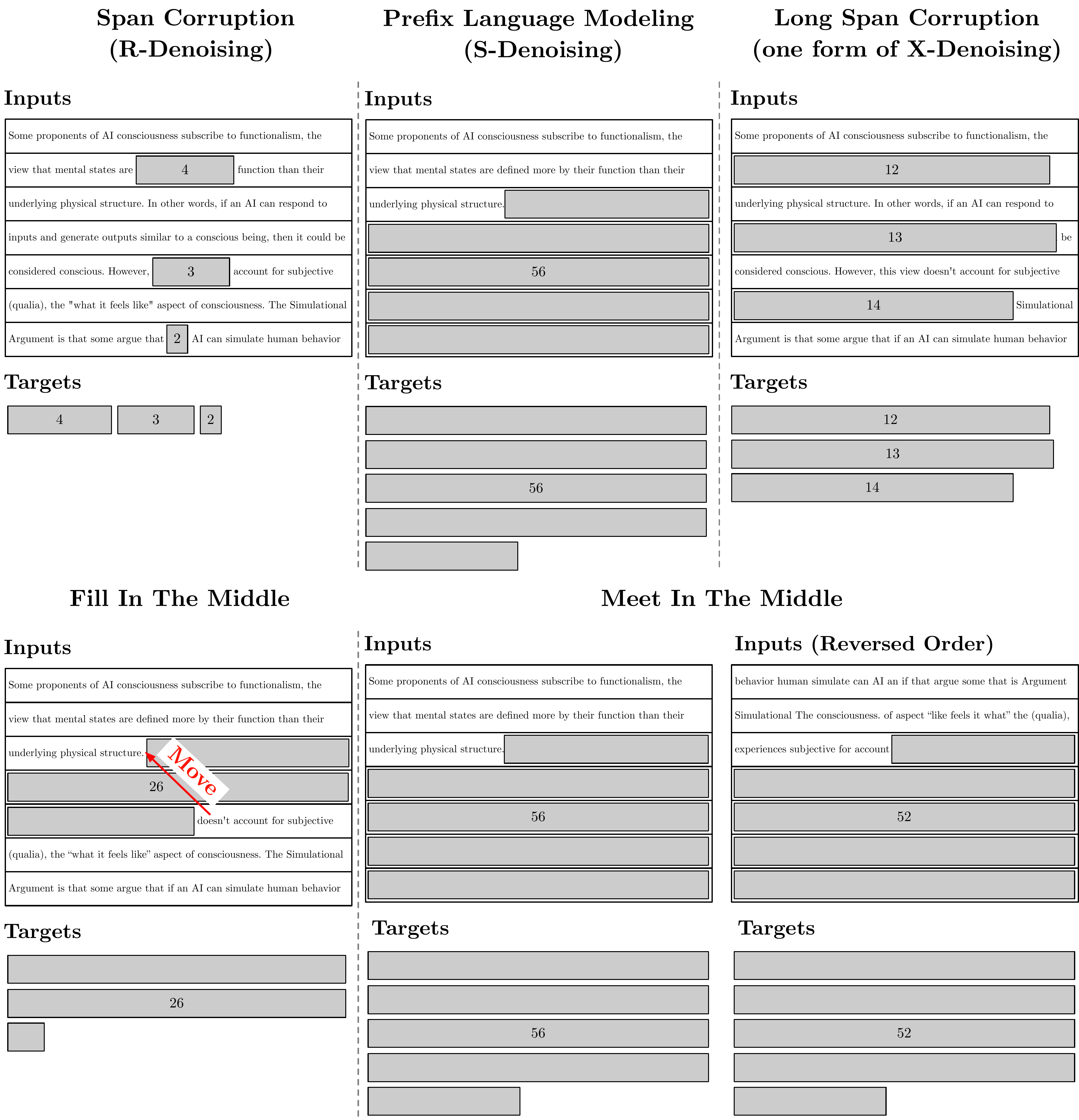}   \caption{\textbf{Self-Supervised Data Construction by Pre-Training Objectives}, adopted from~\citet{tay2022ul2}. We indicate masked tokens with gray rectangles, which become the targets. For brevity, we omit special tokens.}
    \label{fig:inputs_targets}
\end{figure*}

Various pre-training objectives (PTO) are suitable for performing self-supervised training of LLMs. The exact choice of PTO heavily influences the model's data efficiency during pre-training, which in turn can reduce the number of iterations required. A PTO typically is a function of the (i) architecture, (ii) input/targets construction (\eg target span length, low/high corruption, see  \Cref{fig:inputs_targets}), and (iii) masking strategy (\Cref{fig:masking}). While (i) and (ii) can be disentangled and should not be conflated conceptually~\citep{tay2022ul2}, in practice, there exist popular combinations that achieve good performances.

Attending to all tokens, as shown in \Cref{fig:masking}(left), is the most data-efficient strategy since it uses context from before and after the token to be predicted. However, for that reason, it is unsuitable for text generation \cite{devlin-etal-2019-bert}, since it considers future context for prediction. We typically employ it in natural language understanding (NLU) tasks~\citep{devlin-etal-2019-bert}, where it has shown strong results. The next token prediction objective is most suitable for natural language generation (NLG) but also the least data efficient since it only attends to the past context (\Cref{fig:masking}(middle)). More recent advances in pre-training objectives aim to find a middle-ground to increase data efficiency by providing stronger and more diverse training signals, \eg the Prefix LM, which partly attends to past tokens, as illustrated in \Cref{fig:masking}(right) and discussed below.

The following discusses the trade-offs between some of the recently proposed objectives. \Cref{fig:inputs_targets} visually depicts the different pre-training objectives. Notation-wise, we denote a sequence of $N$ tokens $x$ as $x = x_1, \ldots, x_N$.

We start with the most basic and still widely-used \textbf{Language Modeling~\citep{brown2020gpt3}} (or \emph{next token prediction}) objective. Here, we learn parameters $\params$ by maximizing the likelihood of the next token given the previous tokens,
\begin{equation}
    L(x) = \sum_{i = 1}^N \log P(x_i | x_1, \ldots, x_{i - 1}; \params)
    \label{eq:objective_ntp}.
\end{equation}
\textbf{Masked Language Modeling (MLM; or Cloze)~\citep{taylor1953cloze, devlin-etal-2019-bert}} \label{mlm} hides a set proportion of tokens in the sequence by replacing them with a special [MASK] token. The literature employs the MLM objective for non-autoregressive, \ie non-generative, bidirectional context models, where the model uses tokens before and after the target token for predictions, leveraging a more holistic understanding of its context than the NTP objective. Furthermore, we can use each input sentence to predict multiple masked tokens in a single pass, while the NTP objective typically learns from predicting one token at a time.

Let $x_\text{MASK}$ denote the set of indices of the masked tokens and $x_{\neg\text{MASK}}$ the unmasked tokens. The objective of MLM is then to maximize the likelihood given the parameters $\params$,
\begin{equation}
    \begin{aligned}
    L(x_\text{MASK} | x_{\neg\text{MASK}}) = \frac{1}{|x_\text{MASK}|} & \\ \cdot \sum_{i \in x_\text{MASK}} \log P(x_{\text{MASK}_i} | x_{\neg\text{MASK}}; \params).
    \label{eq:objective_mlm}
    \end{aligned}
\end{equation}
~\citet{patel2023bidirectional} show that such models produce representations more suitable for transfer learning; however, they come with difficulties in performing in-context learning (\Cref{sec:prompt_brittleness}). 

To further improve the training efficiency of the MLM objective,~\citet{bajaj2022metro} propose to replace input tokens with ones generated by an auxiliary language model (ALM), resulting in a \emph{Model generated dEnoising TRaining Objective} (METRO). Their approach consists of roughly three components: (i) train an ALM using the MLM objective, (ii) given some inputs with masked positions, predict the tokens (with the ALM), (iii) train the main model to correct these tokens inserted in the masked positions, \ie 1) predict whether the ALM has replaced a token and if so, 2) predict the original token. They train the auxiliary and main model jointly.

\textbf{Prefix Language Modeling~\citep{raffel2022t5}} generalizes language modeling by allowing prefix tokens with a bidirectional receptive field to be added to the input (without prefix, it is equivalent to standard LM). Note that this is still different from the bidirectional context as in MLM, where we always condition on all the tokens before and after the masked ones (see \Cref{fig:masking}~left). For computing the hidden states of the prefix, prefix-LM attends to tokens before and after (see \Cref{fig:masking}~right).

\textbf{Span Corruption~\citep{lewis-etal-2020-bart, raffel2022t5, du-etal-2022-glm}} or \emph{span denoising} refers to a group of denoising objectives that generalize MLM to denoise contiguous sequences of tokens within a given text, called \emph{spans}. The denoising objectives typically replace the sampled spans with a single unique masking token and train the model to fill it in.~\citet{raffel2022t5} shows that this can speed up training because span corruption produces shorter sequences on average compared to corrupting individual tokens in an i.i.d. manner.

\textbf{Mixture of Denoisers~\citep{tay2022ul2}} (MoD) refers to injecting objective diversity by mixing multiple denoising objectives.~\citet{tay2022ul2} categorize three denoising objectives: \{R,S,X\}-Denoiser. The \ub{r}egular denoising corresponds to the previously introduced span denoising. \ub{S}pecific denoising comprises splitting a given sequence into a prefix acting as the context and a suffix acting as the target. In e\ub{x}treme denoising, we corrupt large parts of the input by either (a) increasing the proportion of masked tokens per span or (b) increasing the span length forcing the model to generate long sequences with limited context, which we illustrate in \Cref{fig:inputs_targets}). The MoD objective has subsequently been shown to improve model performance by continuing training pre-trained LLMs~\citep{raffel2022t5,chowdhery2022palm} for relatively few steps~\citep{tay2022upalm}. 

\textbf{Fill In the Middle}~\citet{bavarian2022efficient} propose to augment the next token prediction objective by shuffling tokens within a document such that we \emph{fill in the middle} (FIM) based on prefix and suffix. They demonstrate that models pre-trained on a mixture of FIM-transformed and left-to-right data result in left-to-right and FIM capability models. 

\textbf{Meet in the Middle} ~\citet{nguyen2023meet} extend the FIM objective by enabling bidirectional context to construct a denser, more data-efficient supervision signal while maintaining the autoregressive nature of the underlying model: They train two decoders---one forward $\overrightarrow{p}\left(x_i \mid x_{<i};\params\right)$ and one backward language model $\overleftarrow{p}\left(x_i \mid x_{<i};\params\right)$---with shared parameters $\params$. Additionally, they add an agreement regularize to the loss, encouraging the forward and backward model to agree:
for a dataset $S$ of sequences, the full pre-training loss is
\begin{equation}
\begin{split}
    \sum_{x \in S} \sum_{i=1}^{|x|} &\underbrace{ -\log \overrightarrow{p}\left(x_i \mid x_{<i}; \params \right)}_{\text{NLL for forward model}} \\
    &\underbrace{-\log \overleftarrow{p}\left(x_i \mid x_{>i}; \params\right)}_{\text{NLL for backward model}} \\
    & \underbrace{+\beta D_{i, x}^{T V}(\overrightarrow{p} \| \overleftarrow{p})}_{\text{agreement regularizer}},
\end{split}
\end{equation} where $D_{i, x}^{T V}(\overrightarrow{p} \| \overleftarrow{p})$ is the total variation distance among the two models on the $i$-th token. Once pre-training has been completed, we can use only the forward model $\overrightarrow{p}$.

\paragraph{Parallelism Strategies}
The sheer size of LLMs makes it hard to train or even do inference with them on only one accelerator (GPU, TPU, etc.). A common solution is \emph{model parallelism}, which can be viewed as a \emph{divide-and-conquer} strategy: we slice up various parts of the model (dividing the problem into sub-problems), distribute them across multiple devices, with each device computing a portion of the overall computation (solve each problem independently) and combine all results to produce the final output (forward/backward pass). 

Implementing model parallelism synchronously creates a problem where running data batches through multiple workers with sequential dependency (each layer depends on results from the previous layer) leads to significant waiting times and under-utilization of computation resources.

Another strategy is \emph{pipeline parallelism}, which combines model parallelism with \emph{data parallelism}, meaning that we not only distribute parts of the model across different devices but parts of the data too, \ie each worker splits its mini-batch further into micro-batches with gradients being accumulated across all micro-batches before the weight update.~\citet{gpipe} instantiate such an approach called \emph{GPipe}, which divides each mini-batch into smaller micro-batches distributed across different accelerators simultaneously; gradients are applied synchronously at the end. Compared to naive model parallelism, this decreases waiting times and increases the utilization of computational resources.

These issues have motivated asynchronous parallelization schemes.
\citet{recht2011hogwild} present \emph{Hogwild!}, which \emph{greedily} applies gradients to the local weights on each accelerator as soon as they arrive, offering better resource utilization than pipeline parallelism but suffering from training instabilities due to \emph{stale gradients} which are based on outdated model weights.

\citet{gomez2022interlocking} propose \emph{N-Wise interlocking backpropagation}, which is a generalization of end-to-end and local training. While end-to-end (global) training performs a forward pass through all layers, computes a loss and gradients, and backpropagates through all layers, local training performs forward passes through all layers individually and immediately computes a local loss and gradient update, offering higher resource utilization at the cost of (empirically) worse task performance. \emph{N-Wise interlocking backpropagation} strikes a compromise by performing a forward pass through $N$ layers before computing a loss and updating the parameters of the associated layers, enabling better layer communication than local training and higher computational efficiency than end-to-end training.

\citet{chowdhery2022palm} leverage a combination of model parallelism and fully sharded data parallelism (FSDP)~\citep{xu2021gspmd, zhao2023pytorch}---a technique where each device only holds a subset of the model parameters, gradients, and optimizer states, and parameters necessary for local computations are communicated on-demand---to enable highly parallel, high throughput training across thousands of chips within a single TPU pod. PaLM further employs data parallelism to achieve scaling at pod level, leveraging the Pathways~\citep{barham2022pathways} system to distribute data. 

In a parallel line of work,~\citet{gshard} propose \emph{GShard}, a model parallelism method that extends the XLA~\citep{50530} compiler, enabling automatic sharding of models.

\paragraph{Miscellaneous}~\citet{rae2021gopher} stack the layers of a 4.5B parameter model to jump-start and accelerate the training of a 9B model, which led to a 40\% reduction in compute; an idea that has been previously used for training smaller-scale LMs~\citep{pmlr-v97-gong19a}.~\citet{brown2020gpt3} progressively increase the batch size from a small to the full value over training when training GPT-3; a trick that has been previously used for training image models~\citep{smith2017don}.~\citet{sanyal2023understanding} apply latest weight averaging~\citep{kaddour2022stop} to LLMs between 1 and 12B parameters; for a 6.9B parameter model, they reach savings of up to 4,200 GPU hours. For smaller-scale models, there exist various pre-training speedup algorithms~\citep{zhang2020accelerating,zhuang2023survey}, but they have not been scaled up yet and shown to offer only limited gains when compared with budget-adjusted baselines~\citep{ntng}.

\subsection{Fine-Tuning Overhead}
A potential drawback of pre-training LLMs on massive and diverse sets of textual data is that the resulting models might struggle to explicitly capture the distributional properties of task-specific datasets. To address this, fine-tuning refers to adapting the pre-trained model parameters on comparatively smaller datasets that are specific to an individual domain or task. LLM fine-tuning is highly effective at adapting LLMs for downstream tasks~\citep{howard-ruder-2018-universal, devlin-etal-2019-bert, gpt2}. 

Technically speaking, fine-tuning can be achieved by further training a model on a smaller dataset. Depending on the model architecture, this is done by either (i) directly fine-tuning pre-trained models using a standard language modeling objective or (ii) adding individual learnable layers to the output representations of a pre-trained language model, which are designed to create compatibility between the model's output representations and the output formats of individual downstream tasks (\eg for text classification or sequence labeling). See~\citet{devlin-etal-2019-bert} (Figure 1) for an illustration. 

However, LLMs with billions of parameters have large memory requirements to store (i) the model parameters, (ii) the model activations, and (iii) the gradients and corresponding statistics. Due to limited device memory (\eg GPU or TPU) necessitates access to large clusters with many devices to fine-tune a full LLM, limiting access to a few institutions with large compute resources.

\begin{chall}{Large Memory Requirements}
Fine-tuning entire LLMs requires the same amount of memory as pre-training, rendering it infeasible for many practitioners. 
\end{chall}

\begin{figure}[t!]
    \centering
     \begin{subfigure}[b]{0.99\columnwidth}
         \centering
         \includegraphics[width=\columnwidth]{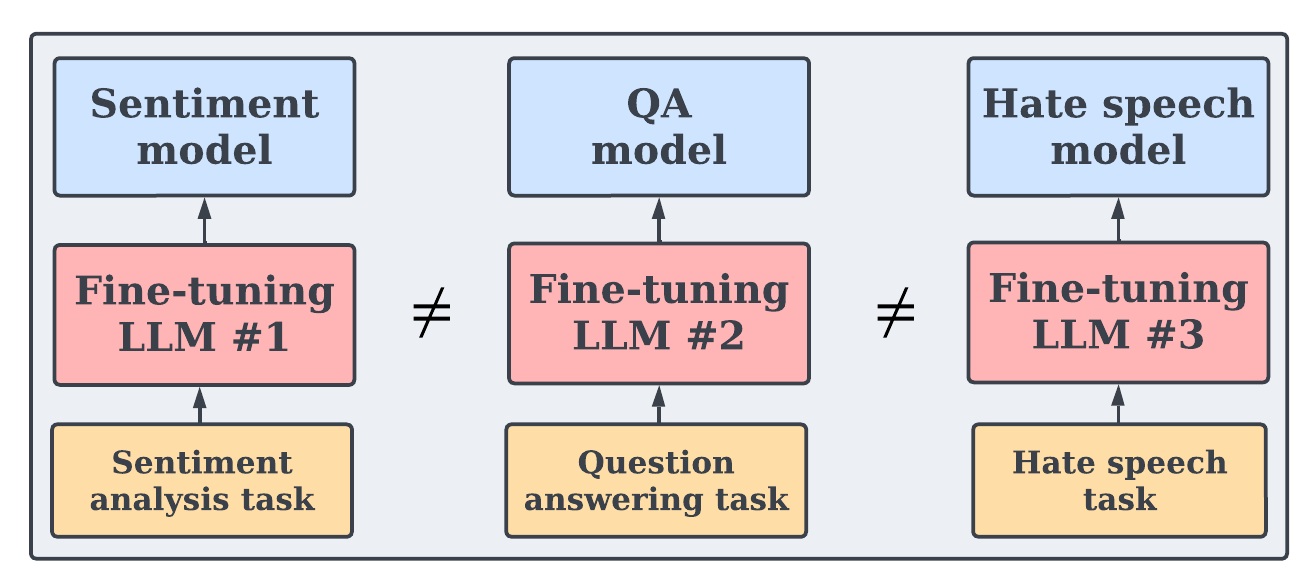}
         \caption{}
     \end{subfigure}
     \hfill 
     \vspace{0.05cm}
     \begin{subfigure}[b]{0.99\columnwidth}
         \centering
         \includegraphics[width=\columnwidth]{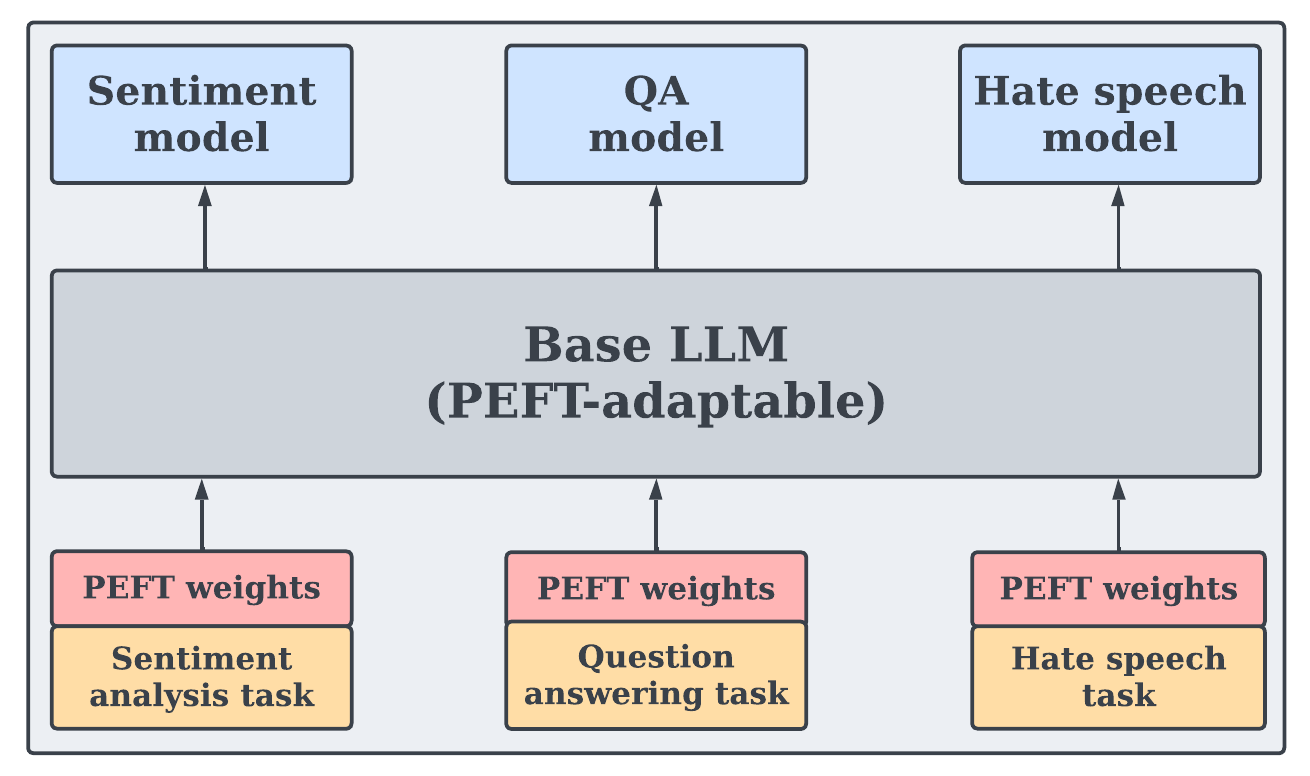}
         \caption{}
     \end{subfigure} 
    \caption{\textbf{Fine-tuning an LLM for a specific downstream task}. (a) illustrates vanilla fine-tuning, which requires updating the entire model, resulting in a new model for each task. In (b), PEFT instead learns a small subset of model parameters for each task with a fixed base LLM. The same base model can be re-used during inference for different tasks.}
    \label{fig:ft_vs_peft}
\end{figure}

Moreover, while full model fine-tuning is effective at adapting LLMs to perform well on specific downstream tasks, individual copies of fine-tuned LLMs need to be stored and loaded for individual tasks, which is computationally inefficient~\citep{houlsby2019parameter, li-liang-2021-prefix} and requires practitioners to keep individual fine-tuned LLMs in memory for every task. We illustrate this overhead in Figure~\ref{fig:ft_vs_peft}.
\begin{chall}{Overhead of Storing and Loading Fine-Tuned LLMs~\cite{houlsby2019parameter, li-liang-2021-prefix}}
When adapting an LLM via full-model fine-tuning, an individual copy of the model must be stored (consuming data storage) and loaded (expending memory allocation, etc.) for each task.
\end{chall}

\paragraph{Parameter-efficient fine-tuning}

An alternative method to adapt an LLM to a specific dataset/domain is via \textit{parameter-efficient fine-tuning} (PEFT). PEFT refers to a class of methods that adapt LLMs by updating only a small subset of model parameters.
\textbf{Adapters~\citep{houlsby2019parameter}} are one of the earliest works on PEFT. This method incorporates additional, learnable layers into a Transformer architecture that are updated during fine-tuning whilst keeping the remainder of the network unchanged. Experimental results on 26 text classification tasks (incl. the GLUE benchmark~\citep{wang-etal-2018-glue}) reveal that models trained via Adapters are competitive with full fine-tuning while updating only 3\% of the model's parameters.~\citet{ben-zaken-etal-2022-bitfit} instead propose only to update the model's bias terms for fine-tuning, which make up less than 1\% of the model's parameters. Experimental results show competitive performance across tasks of the GLUE benchmark. We are aware of three general frameworks for incorporating adapters into language model fine-tuning, namely AdapterHub~\citep{pfeiffer-etal-2020-adapterhub}, LLM-Adapters~\citep{hu2023llm}, and HuggingFace's PEFT library~\cite{hfpeft}.

PEFT methods introduced for larger models include \textbf{prefix-tuning}~\citep{li-liang-2021-prefix} and \textbf{prompt-tuning}~\citep{lester-etal-2021-power}, which both operate by prepending a set of learnable token embeddings to an input. These token embeddings (also referred to as \textit{soft prompts}~\citep{lester-etal-2021-power}) are learned during the fine-tuning stage, whereas the remainder of the model parameters remains fixed. Most notably, such soft prompts contain thousands rather than millions of parameters and are much more efficient to store. Notably, one still has to backpropagate through the network while fine-tuning the tokens. Alternatives for models with only black-box API access have been proposed too~\citep{pmlr-v162-sun22e,diao2022black}.

It has been shown that prompt-tuning can learn generalizable representations with very small amounts of training data, achieving competitive performances when trained on less than 100 examples for safety classification~\citep{mozes2023towards} or five examples for multilingual question answering~\citep{agrawal2022qameleon}. In addition to that, recent work investigates the potential of using soft prompts for pre-training and transfer learning across different tasks~\citep{gu-etal-2022-ppt, vu-etal-2022-spot}. 

\citet{liu2022few} introduce (IA)$^3$, which scales activations in individual Transformer layers with learnable vectors. The authors demonstrate its effectiveness by showing that models trained using (IA)$^3$ outperform full model fine-tuning on various datasets whilst updating only 0.01\% of the model's parameters. 

\citet{malladiFineTuningLanguageModels2023} propose a memory-efficient zeroth-order (MeZO) optimizer, which only requires the same memory footprint as during inference (instead of storing gradients or optimizer states). Further, it can optimize non-differentiable objectives like accuracy or F1 scores, which conventional gradient-based tuning methods cannot.

\citet{hu2021lora} propose Low-Rank Adaptation (LoRA), which formulates parameter updates of weight matrices at individual Transformer layers as an additive low-rank decomposition. Such a reparameterization avoids the need to compute dense matrix multiplications.~\citet{dettmers_qlora_2023} extend LoRA to quantized LLMs, drastically reducing memory usage, allowing them to fine-tune a 65B model on a single 48GB GPU. The authors mention that regular training of the same model requires more than 780 GB of GPU memory.

\paragraph{Compute Requirements} 

However, despite substantial improvements in \emph{memory complexity} needed to fine-tune LLMs for specific tasks, a remaining challenge is the \emph{time complexity}. Fine-tuning an LLM, even with PEFT methods, still requires full gradient computation. The computational infrastructure needed to adapt LLMs prohibits potential applications like personalization on smaller devices.

\begin{chall}{Full Matrix Multiplications}
Parameter-efficient fine-tuning of LLMs still requires computing full forward/backward passes throughout the whole network.
\end{chall}

\subsection{High Inference Latency} \label{sec:inference_costs} According to \citet{popeEfficientlyScalingTransformer2022,weng2023inference}, two reasons why LLMs exhibit high inference latencies are: (1) \textbf{low parallelizability} since the inference procedure proceeds one token at a time and (2) \textbf{large memory footprints}, due to the model size and the transient states needed during decoding (\eg attention key and value tensors). Further, the authors also discuss the quadratic scaling of the attention mechanisms in Transformers, which we discuss separately in \Cref{sec:context_length}.

\begin{chall}{High Inference Latency \cite{popeEfficientlyScalingTransformer2022,weng2023inference}}
LLM inference latencies remain high because of low parallelizability and large memory footprints. 
\end{chall}

In the following section, we review techniques used to address these challenges by \eg reducing the memory footprint (size and/or bandwidth), or accelerating specific computational operations. Note that some of these techniques may also be applicable during the training process, but we discuss them here since they are not only designed for training, like the approaches discussed in \Cref{sec:efficient_training}.

\paragraph{Efficient Attention} 
Roughly two lines of work aim to accelerate attention mechanism computations by (i) lower-level hardware-aware modifications or (ii) higher-level sub-quadratic approximations of the attention mechanism. 

For the former, multi-query attention~\citep{fast_transformer_decoding} aims to reduce memory bandwidth bottlenecks when sequentially generating sequences of tokens using Transformer decoder layers by keeping only one attention head for the key and value tensors. Similarly,~\citet{dao2022flashattention,pagliardini2023faster} reduce memory bandwidth by proposing an alternative computation method for multi-head self-attention, called \texttt{FlashAttention}, to minimize the number of I/O operations to speed up the computation on modern GPUs. As an optimized attention implementation, \texttt{FlashAttention} leverages operator fusion to reduce the memory bandwidth bottleneck.~\citet{pagliardini2023faster} build on top of \texttt{FlashAttention} and incorporate attention sparsity patterns, encompassing key/query dropping and hashing-based attention. \citet{pope2022efficiently} implement different sharding techniques to efficiently spread the feedforward and attention computations across devices while optimizing for inter-device communication costs, enabling context lengths of up to 43,000 tokens using multi-query attention.

With regards to the second stream of work, a common theme to improve the computational or memory complexity of the attention mechanism is to sparsify the attention matrix or introducing (linear) approximations
~\citep{tay2022efficient}. However, the scalability of some efficient Attention approximations has been questioned. For example,~\citet{tay2022scaling,huaTransformerQualityLinear2022} find that the Performer attention approximation~\cite{choromanski2020rethinking} severely underperforms the vanilla self-attention mechanism, especially when scaled up to large models.

\paragraph{Quantization} is a post-training technique that reduces the memory footprint and/or increases the model's throughput by reducing the computational precision of weights and activations. \texttt{nuQmm}~\citep{park2022nuqmm} and \texttt{ZeroQuant}~\citep{yao2022zeroquant} use a non-uniform quantization method to quantize weights and apply custom CUDA kernels for computational benefits. \texttt{LLM.int8()}~\citep{llm8bit} is a degradation-free quantization scheme enabling efficient inference of multi-billion parameter LLMs by utilizing Int8 quantization and falling back to higher precision for certain outlier features without the need for re-training. Similarly, GLM-130B~\citep{zeng2022glm130b} uses a degradation-free 8-bit quantization scheme, storing weights in 8-bit and performing matrix multiplications in 16-bit precision.~\citet{frantar2022gptq} propose an efficient, one-shot quantization technique to compress LLM weights down to 3 to 4 bits per weight, enabling 175B parameter models to be run on a single GPU.~\citet{dettmers2023spqr} further improve upon this by combining higher precision representations for outlier weights and grouped quantization.

\paragraph{Pruning} is a complementary post-training technique to quantization, removing parts of the weights of a given model (without degrading its performance). An important distinction is whether the pruning follows a \emph{structured} pattern or is \emph{unstructured}. Structured sparse models substitute dense sections of a model with an assembly of significantly smaller yet still dense components. Unstructured sparse models contain weights of value zero, which do not influence the network's behavior and can therefore be committed in theory. However, in practice, it is more challenging to translate theoretical to practical computation savings on current hardware~\citep{gale_sparse_2020,dehghani_efficiency_2022,liu2023ten}. 

On the structured side, early work on pruning language models mainly aims at comparatively small MLM-type models~\citep{wang2019structured, Fan2020Reducing, jiao-etal-2020-tinybert}.~\citet{ma2023llm} propose LLM-Pruner, which aims at pruning LLMs in a task-agnostic manner while preserving the zero-shot capabilities of the models. To this end, LLM-Pruner adopts a three-stage pruning procedure where 1) interdependent structures within the model are identified and grouped, 2) the contribution to the overall performance is estimated for each group, and low-performing groups are pruned, 3) performance recovery via parameter-efficient fine-tuning procedure using LoRA~\citep{hu2021lora}.

On the unstructured side, SparseGPT~\citep{frantar2023massive} is an unstructured pruning approach specifically developed to be fast enough to be run on LLMs with hundreds of billions of parameters within a few hours, being able to prune the number of parameters by up to 60\% while maintaining roughly the same model performance.~\citet{sun2023simple} propose Wanda (Pruning by \textbf{W}eights \textbf{and a}ctivations), which applies magnitude pruning based on the product of each weight's magnitude and the norm of the corresponding input activations, matching SparseGPT in performance while requiring only a single forward pass to prune the network. Both SparseGPT and Wanda can be extended to perform semi-structured pruning, enabling n:m sparsity~\citep{hubara2021sparsity, zhou2021sparsity} and achieving the corresponding speed-ups on recent GPUs~\citep{mishra2021accelerating}.

\paragraph{Mixture-of-Experts} architectures typically consist of a set of \emph{experts (modules)}, each with unique weights, and a \emph{router} (or \emph{gating}) network, which determines which expert module processes an input. MoE models decrease inference time by not using all experts at once but only activating a subset of them. Further, they can reduce communication across devices in model-distributed settings by placing each expert on a separate accelerator; only the accelerators hosting the router and the relevant expert model must communicate.  
\citet{shazeer2017outrageously} propose one of the first MoE layers embedded within a language model, which they refer to as \emph{sparsely-gated MoEs} (SG-MoEs). They denote by $G(\vx)$ and $E_i(\vx)$ the gating network output and the $i$-th expert network output for a given input $\vx$, respectively. We can then write the output as $\vy=\sum_{i=1}^n G(\vx)_i E_i(\vx)$. Wherever $G(\vx)_i=0$, we do not need to compute $E_i(\vx)$, thereby saving compute during inference. 
\citet{gshard} scale up an SG-MoE model to 600B parameters by proposing \emph{GShard}, a model parallelism method that extends the XLA~\citep{50530} compiler. While SG-MoE selects the top-$k$ experts with $k>1$, the \emph{Switch Transformer} (ST)~\citep{switch_transformer} architecture uses $k=1$ experts, which reduces routing computation and communication across experts (which may be located on different accelerators). ST empirically outperformed a strongly tuned T5 model with up to 7x pre-training speedups.~\citet{base} notice that the learned routers can result in unbalanced assignments across experts. To ensure balanced routing, they formulate a linear assignment problem that maximizes token-expert affinities while equally distributing the number of tokens across experts.
\citet{smlp} propose \emph{sMLP}, an MoE using only MLPs blocks, which (i) they scale up to 10B, (ii) results in a 2x improvement in pre-training speed, and (iii) outperforms sparse Transformer counterparts. 

However, MoE models still suffer from unique issues like expert collapse (all experts learning the same), likely caused by underconstrained routing functions~\citep{chen2023sparse}. For example,~\citet{roller2021hash} demonstrates that learned expert assignments do not always outperform random ones.

Interestingly, instead of designing an architecture for sparsity explicitly,~\citet{liLazyNeuronPhenomenon2023a} observe that the activation maps of default Transformer models often emerge to be very sparse implicitly; the larger the model, the sparser measured by the percentage of nonzero entries. Similarly,~\citet{zhang2022moefication} find that post-training \emph{MoEfication}, \ie converting monolithic models to equivalent MoE models, can speed up inference by 2x. 

\paragraph{Cascading} refers to the idea of employing differently-sized models for different queries \cite{chenFrugalGPTHowUse2023}. In spirit, this idea is similar to Mixture-of-Experts models, but instead of learning a routing module, we employ a \emph{cascade} of multiple, differently-sized monolithic models (these can be even black-box API models) and learn a scoring function that decides which model(s) receive which query. \citet{chenFrugalGPTHowUse2023} demonstrate that this strategy dominates the Pareto frontier between accuracy and cost. 

\paragraph{Decoding Strategies} can greatly impact the computational cost of performing inference. For example, beam search trades off compute for higher-quality results. Another example of a computationally expensive decoding scheme is sample-and-rank~\citep{adiwardana2020towards} where $N$ independent sequences of tokens $y^1, \ldots, y^N$ are obtained using random sampling, and the highest probability sequence is used as the final output. 

Latency-oriented strategies such as speculative sampling~\citep{stern2018blockwise,leviathan2022fast,chen2023accelerating} first autoregressively generate a draft of length $K$ using a smaller (draft) model; then, the larger (target) model scores the draft, followed by a modified rejection sampling scheme to accept a subset of the tokens from left to right. Similar ideas have been proposed in various contexts, such as for blockwise parallel generation~\citep{stern2018blockwise}, grammatical error correction~\citep{sun-etal-2021-instantaneous}, and with a larger LLM refining generation produced by a small model~\citep{kim2023big}. \citet{delcorroSkipDecodeAutoregressiveSkip2023} observe that tokens towards the end of a sequence are easier to predict due to more contextual information, motivating a new decoding strategy that skips earlier layers in the network for such tokens.

\subsubsection{Software}
Various frameworks have been designed to enable the efficient training of multi-billion to trillion parameter language models such as \texttt{DeepSpeed}~\citep{rasley2020deepspeed} and \texttt{Megatron-LM}~\citep{shoeybi2019megatron} to account for the unique challenges arising when training such models. This is necessitated by the fact that most LLMs do not fit into a single device's (GPU, TPU) memory, and scaling across GPUs and compute nodes needs to account for communication and synchronization costs. \texttt{FlexGen}~\citep{sheng2023flexgen} provides further speed-ups by aggregating memory and compute resources from the GPU, CPU, and disk and utilizing techniques such as 4-bit quantization, enabling inference with 175B parameter models on a single GPU.

The frameworks typically combine existing parallelism strategies to compensate for drawbacks and scale model training across multiple sets of compute nodes, within compute nodes, and across multiple GPUs per node. \eg~\citet{smith2022using} use tensor slicing within a node, pipeline parallelism across nodes, and data parallelism to train multiple model replicas over sets of nodes. Additional features include memory optimizations~\citep{rajbhandari2020zero, ren2021zero, rajbhandari2021zeroinf}, communication-efficient~\citep{tang2021communication, li20211communication, lu2022maximizing} and fused optimizers\footnote{\href{https://github.com/nvidia/apex}{https://github.com/nvidia/apex}}, and support for MoE training~\citep{rajbhandari2022moe}. 

Specialized implementations such as \texttt{Tutel}~\citep{hwang2022tutel} and \texttt{MegaBlocks}~\citep{gale2022megablocks} offer efficient sparse MoE training, while \texttt{Alpa}~\citep{zheng2022alpa} enables automatic data and model parallelism for LLMs written in Jax. The \texttt{FasterTransformer}\footnote{\href{https://github.com/NVIDIA/FasterTransformer}{https://github.com/NVIDIA/FasterTransformer}} library includes highly optimized Transformer encoder and decoder implementations for TensorFlow, PyTorch, and Triton.

\citet{kwon2023vllm} introduce \texttt{vLLM}, an open-source library for efficient inference and LLM serving. vLLM employs PagedAttention, which partitions each sequence's KV cache into fixed-size blocks. When performing attention computations, blocks are fetched from non-contiguous memory. This enables memory sharing, reducing memory consumption and transfers in decoding strategies such as beam search, ultimately improving throughput.

The \texttt{Petals}~\citep{borzunov2022petals} library\footnote{\href{https://github.com/bigscience-workshop/petals}{https://github.com/bigscience-workshop/petals}} allows users to collaboratively fine-tune and run LLMs by distributing subsets of model parameters to individual machines.

All of these libraries address the enormous computational costs associated with training and running LLMs, either by offering more efficient implementations, lowering memory requirements, or using distributed or decentralized computing strategies.

\subsection{Limited Context Length} \label{sec:context_length}
Addressing everyday NLP tasks often necessitates an understanding of a broader context. For example, if the task at hand is discerning the sentiment in a passage from a novel or a segment of an academic paper, it is not sufficient to merely analyze a few words or sentences in isolation. The entirety of the input (or \emph{context}), which might encompass the whole section or even the complete document, must be considered. Similarly, in a meeting transcript, the interpretation of a particular comment could pivot between sarcasm and seriousness, depending on the prior discussion in the meeting.

\citet{longchat2023} evaluate several LLMs in the long-context settings and find that while commercial closed-API models often fulfill their promise, many open-source models -- despite claiming to perform well with longer contexts -- exhibit severe performance degradation. They point out that there is a difference between being \emph{architecturally-able} to deal with long inputs and actually \emph{performing well}. Having an architecture that can infer long inputs does not guarantee that the LLM will perform as well on those as on shorter inputs. 
Similarly, \citet{liuLostMiddleHow2023} find that changing the location of relevant information in the input can degrade model performance. Interestingly, they find that decoder-only LLMs like GPT-3.5 can deal well with such information at the beginning or end of the input context; they cannot access information in the middle of it well, resulting in a U-shaped performance curve.

\begin{chall}{Limited Context Length}
    Limited context lengths are a barrier for handling long inputs well to facilitate applications like novel or textbook writing or summarizing. 
\end{chall}

To this end, we discuss three lines of work permitting longer context lengths.
First, we look at efficient attention mechanisms, which help mitigate the effect of long inputs on the computational requirements of Transformer models. Next, we examine positional embedding schemes in the light of generalization to longer sequence lengths than those used during training. Lastly, we revise Transformer alternatives which neither require attention nor positional embeddings. 

\paragraph{Efficient Attention Mechanisms} One way of addressing the limited context of LLMs is by designing more efficient attention mechanisms that can process longer inputs. ~\citet{ma2021luna} introduce \emph{Luna}, a linear unified nested attention mechanism that approximates softmax attention with two nested linear attention functions, yielding only linear (as opposed to quadratic) time and space complexity, allowing it to process much longer inputs. Similarly,~\citet{shen2021efficient} and~\citet{li2020linear} present alternative attention mechanisms equivalent to the dot-product attention but which require substantially less memory and compute resources.~\citet{guo-etal-2022-longt5} propose an attention mechanism called \emph{Transient Global}, which is an extension of local attention where each token can attend to nearby tokens and a set of global tokens. It enables to handle sequences with up to 12,000 tokens. Similarly, \emph{CoLT5}~\citep{ainslie2023colt5} enables context lengths of up to 64,000 tokens by splitting the computations into a light branch with local attention, fewer attention heads, and a heavy branch with full attention. \emph{CoLT5} applies the light branch to every token and the heavy branch to a subset of tokens that are selected by a learnable routing function. 

After investigating the effect of the dot-product self-attention mechanism,~\citet{tay2021synthesizer} propose the \emph{Synthesizer}, a new architecture that learns synthetic attention weights without token-token interactions, showing that it consistently outperforms transformers on various language-based tasks. ~\citet{britz2017efficient} offer an alternative attention mechanism based on a fixed-size memory representation that is more efficient, yielding inference speedups of $20\%$ without significantly hurting performance.~\citet{huaTransformerQualityLinear2022} combine a single-head attention mechanism with a linear attention approximation to achieve speed-ups between 4.9x and 12.1x for auto-regressive language modeling while obtaining similar perplexities as a standard Transformer model.~\citet{ding2023longnet} propose dilated attention which splits a sequence into equally long segments and processes each of these in parallel using a sparsified attention mechanism. Dilated attention offers a linear computational complexity in the sequence length and, applied hierarchically, enables inputs of up to 1B tokens.

\paragraph{Length Generalization}
As the required compute of Transformer-based LLMs grows quadratic with the sequence length, it is a desired property to build LLMs that can be trained on short sequences and generalize well to significantly longer sequences during inference.

The fundamental building block of the Transformer architecture is the self-attention mechanism. It is permutation-invariant; therefore, the output is independent of the input sequence order. Positional information is commonly injected to make the model respect a token's position in the sequence, \ie capture the semantics of where a token occurs rather than just whether it occurs. The longer the input is, the more important the positional embedding becomes since the model needs to effectively use information from different parts of the input that may cover a wide range of distances from the current token.

Without positional embeddings, a Transformer models the relations between any two tokens with equal probability. Hence, positional embeddings introduce an LSTM-like inductive bias that (typically) tokens closer to each other in the sequence are more relevant to each other. Depending on the positional embedding scheme chosen, this can be learned or effectively hard-coded. However, it remains unclear what is the most effective positional embedding scheme for long inputs. Further, models face difficulties generalizing to unseen sequence lengths by introducing a dependency on sequence positions. This is an undesirable artifact of positional embeddings, as language semantics do not inherently depend on the length of an utterance. 

While positional encoding schemes such as relative positional encodings or, more recently, ALiBi have made progress in building more generalizable ways for injecting positional information into Transformers, the challenge of generalizing to sequences much longer than seen during training remains largely unsolved. Surprisingly,~\citet{haviv-etal-2022-transformer} find that causal LLMs without positional encodings are competitive compared to models with positional encodings and accredit this success to the causal attention mask leaking positional information into the model.

In the following, we first summarize some standard positional embeddings technique and then move to more advanced schemes designed to improve length generalization.
We start with \textbf{Absolute Positional Embeddings~\citep{transformers}}, which inject positional information by sinusoidal embeddings based on the absolute position $i$ of a token $\vx_i$ within their sequence $\vx_1, \dots, \vx_N$ into the model input. Given an input sequence $\rmX = [\vx_1, \dots, \vx_N]$, we add a positional embedding matrix $\rmP \in \mathbb{R}^{n \times d}$ of the same shape to get the positional encoding outputs $\rmX+\rmP$, where the element on the $i^{\text {th }}$ row and the $(2 j)^{\text {th }}$ or the $(2 j+1)^{\text {th }}$ column of $\rmP$ follows sinusoidal functions.
~\citet{transformers} also compare against learned positional embeddings and find no significant performance difference. In contrast, sinusoidal positional encodings require no trainable parameters, and the authors hypothesize that they enable extrapolation to sequence lengths longer than the ones contained in the training set. However, this feature is not guaranteed, as the subsequent layers in the network need to be able to deal with such extrapolated positional embeddings. 
\textbf{Learned positional encodings} do not possess inherent generalization capabilities for unseen sequence lengths. This limitation arises because the embeddings associated with absolute positions not encountered during training---depending on the implementation---either do not exist or remain untrained (random). 
\textbf{Relative Positional Embeddings} have subsequently been developed, extending absolute positional embeddings to relative offsets between token positions~\citep{shaw-etal-2018-self,huang2018music,dai-etal-2019-transformer,chen2023extending}. While rarely used in their vanilla form in LLMs~\citep{rae2021gopher}, relative positional embeddings have given rise to the methods outlined in the following paragraphs. They offer better generalization to unseen sequence lengths than absolute positional encodings. All unseen absolute positions will be converted to previously observed relative offsets between positions, enabling better generalization to long input sequences at inference time. 
\textbf{Rotary Position Embeddings (RoPE)~\citep{su2021roformer}} unite absolute and relative methods by incorporating absolute positional information in a rotation matrix and modeling the relative positional offset through a rotation. They directly modify the self-attention calculation rather than injecting positional information into the embeddings. The attention between positions $i, j$ linearly depends on $i - j$ by introducing a $d \times d$ dimensional block diagonal matrix $\mR^d_{\Theta, k}$, resulting in a self-attention mechanism defined as \begin{equation}
    \softmax{}\left(\frac{1}{\sqrt{d}} \sum_{i, j} \vx_i^\top \mW_q^\top \mR^d_{\Theta, (i - j)} \mW_k \vx_j \right)
    \label{eq:self_attention_rope}.
\end{equation}
While RoPE has been adapted in many LLMs~\citep{gpt-j,black2022gptneox,chowdhery2022palm} and~\citet{su2021roformer} show RoPE leading to better performance on long text tasks,~\citet{press2021alibi} demonstrate that this positional encoding scheme extrapolates poorly to unseen sequence lengths. However,~\citet{chen2023extending} demonstrate that by interpolating rather than extrapolating longer than before observed context windows and briefly fine-tuning RoPE-based models, enabling pre-trained LLMs to extend their context window to very long sizes of up to $32,768$ tokens.

\textbf{Relative Positional Bias~\citep{raffel2022t5}} directly bias the attention computation (\Cref{eq:self_attention_t5}) with a learned bias per relative positional offset and attention head instead of adding information to the token embeddings
\begin{equation}
    \softmax{}\left(\frac{1}{\sqrt{d}} \sum_{i, j} \vx_i^\top \mW_q^\top \mW_k \vx_j + b_{i - j} \right).
    \label{eq:self_attention_t5}
\end{equation}

\citet{press2021alibi} follow a similar methodology but use heuristics to define \emph{ALiBi} (\textbf{A}ttention with \textbf{Li}near \textbf{Bi}ases), a non-learned bias that is used to penalize attention scores in long-range interactions~\citep{scao2021bloom}, i.e., a recency-bias is backed into the model. Here, $m$ is a pre-defined, head-specific slope--by default, the set of slopes for $n$ heads form a geometric sequence.
\begin{equation}
    \softmax{}\left(\frac{1}{\sqrt{d}} \sum_{i, j} \vx_i^\top \mW_q^\top \mW_k \vx_j + m \cdot -(i - j) \right).
    \label{eq:self_attention_alibi}
\end{equation}
\citet{press2021alibi} motivate \emph{ALiBi} by designing it to generalize well to unseen sequence lengths. They show that training a model with it on training sequences with a maximum sequence length of $1,024$ tokens achieves the same perplexity on a test set with a maximum sequence length of $2,048$ as a model trained with sinusoidal positional encodings on sequences with up to $2,048$ tokens. Thereby, it not only enables larger context lengths but can also potentially reduce pre-training costs (\Cref{sec:efficient_training}).

While some of the existing positional encoding schemes offer better generalization to long sequences than others, it remains unclear how reliable they are. For example, \citet{taylor2022galactica} report trying ALiBi in the \emph{Galactica} LLM and not observing ``large gains'' compared to using learned positional encodings. Similarly, \citet{kazemnejad2023impact} find that popular positional encoding schemes such as \emph{ALiBi}, \emph{RoPE}, and absolute positional encodings do not perform well in terms of length generalization in a suite of 10 reasoning downstream tasks. 

In a parallel line of work, \citet{anilExploringLengthGeneralization2022} demonstrate that naively fine-tuning a pre-trained LLM is insufficient for length generalization in the context of reasoning tasks. Instead, they propose combining in-context learning and scratchpad/chain-of-thought reasoning to enable LLMs to generalize to unseen sequence lengths in- and out-of-distribution, with performance scaling with model size. The authors report that fine-tuning can further improve model performance dependent on the task performance of the baseline. 

\paragraph{Transformer Alternatives}
While Transformers are the dominant paradigm in LLMs today due to their strong performance, several more efficient alternative architectures exist. 
One line of work tries to replace the attention mechanism using \textbf{state space models} (SSMs), which offer near-linear computational complexity w.r.t. the sequence length.
\citet{daoHungryHungryHippos2023} investigate the weaknesses of state space models (SSMs) in language modeling and find that existing approaches struggle with recalling previous tokens and comparing tokens in the sequence. Based on these findings, the authors propose \emph{H3} with a shift matrix to recall previous tokens and multiplicative interactions for token comparisons. The authors demonstrate that \emph{H3} comes close to Transformer-based LLMs for language modeling, offering further improvements when combined with attention.
\citet{poliHyenaHierarchyLarger2023a} propose the \emph{Hyena} operator, a convolution-based sub-quadratic attention replacement designed for long sequences. \emph{Hyena} tries to emulate the attention mechanisms' dynamic nature by introducing data-controlled computations, \ie \emph{Hyena} applies an element-wise gating operation based on the operator's input to mimic the attention contextualization. \emph{Hyena}-based models have been used on natural language for sequence lengths of up to $131,000$ tokens \citep{poliHyenaHierarchyLarger2023a} and up to $1,000,000$ tokens in the context of genomics \citep{nguyen2023hyenadna}.
\citet{fathiBlockStateTransformer2023} propose the Block-State Transformer, which builds upon a hybrid layer that combines an SSM for long-range contextualization and a Transformer for short-range interactions between tokens. The authors find similar performance to Transformer-based baselines while obtaining speed-ups of up to 10x on sequence-level, enabling models with more than $65,000$ tokens sequence length.

Another line of work utilizes \textbf{recurrent neural networks} (RNNs), which offer linear computational complexity and memory requirements with respect to the sequence length as the backbone of LLMs. \citet{pengRWKVReinventingRNNs2023} propose \emph{Receptance Weighted Key Value} (RWKV) to combine the parallelization benefits of Transformer-based LLMs during training with the fast inference and low compute requirements of RNNs. The authors accomplish this by leveraging a linear attention-like mechanism, scaling non-Transformer LLMs to 14B parameters, and matching the performance of similarly-sized Transformer LLMs.

\subsection{Prompt Brittleness} \label{sec:prompt_brittleness}
A prompt is an input to the LLM. The prompt syntax (\eg length, blanks, ordering of examples) and semantics (\eg wording, selection of examples, instructions) can have a significant impact on the model's output~\citep{lu-etal-2022-fantastically}. 

As an analogy, if we were to think of an LLM as a (fuzzy) database and prompts as queries \cite{jonathanfrankle[@jefrankle]LouderPeopleBack2022}, it becomes clear that slight changes in the query can result in vastly different outputs. Consequently, the wording, as well as the order of examples included in a prompt, have been found to influence the model's behavior significantly~\citep{webson-pavlick-2022-prompt, zhao-etal-2021-calibrate, lu-etal-2022-fantastically}. 

\begin{chall}{Prompt Brittleness~\citep{zhao-etal-2021-calibrate,webson-pavlick-2022-prompt, lu-etal-2022-fantastically}}
Variations of the prompt syntax, often occurring in ways unintuitive to humans, can result in dramatic output changes.
\end{chall}

\begin{figure*}[ht!]
    \centering
\includegraphics[width=\textwidth]{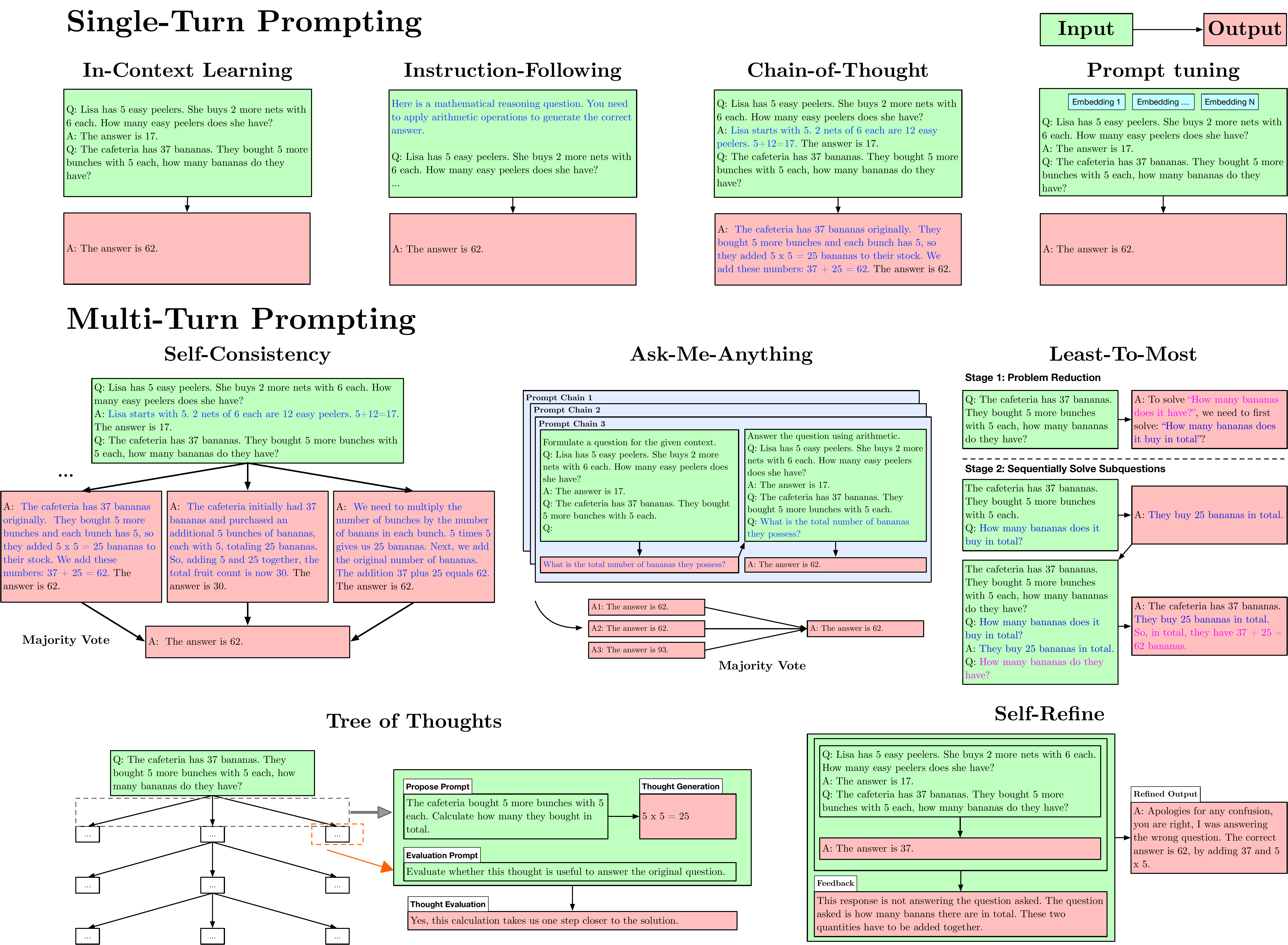}
    \caption{\textbf{Overview of Selected Prompting Methods}, categorized into Single-Turn and Multi-Turn Prompting. We use a running example across all methods inspired by \citet{wei2022chain}.}
    \label{fig:prompting_methods}
\end{figure*}

Designing natural language queries that steer the model's outputs toward desired outcomes is often referred to as \emph{prompt engineering}~\citep{saraviaPromptEngineeringGuide2022,promptslabAwesomePromptEngineering2023,weng2023prompt}. \Cref{fig:prompting_methods} summarizes some of the most popular prompting methods with an example adapted from \citet{wei2022chain}. As we can see, there are lots of equally-plausible prompting techniques, and the current state of prompt engineering still requires lots of experimentation, with little theoretical understanding of why a particular way to phrase a task is more sensible other than that it achieves better empirical results. Developing LLMs that are robust to the prompt's style and format remains unsolved, leaving practitioners to design prompts ad-hoc rather than systematically.

\paragraph{Single-Turn Prompting} methods improve the input prompt in various ways to get a better answer in a single shot.
\textbf{In-Context Learning (ICL)} refers to an LLM's ability to learn a new task solely via inference (without any parameter updates) by conditioning on a concatenation of the training data as demonstrations~\citep{brown2020gpt3, schick2021s}. This enables users and practitioners to use LLMs for a variety of NLP tasks by simply listing examples of the dataset (\eg input texts and their corresponding labels) without the need to adjust the LLM's inner workings. 

Various existing works investigate why ICL shows such competitive results across NLP tasks. One explanation concurrently proposed by~\citep{von2022transformers,dai2022can,rek2023what} is that ICL emulates gradient-based meta-learning, \ie it implicitly fine-tunes the model through gradient descent in their forward pass. 

Interestingly,~\citet{min2022rethinking} show that input-label associations in the few-shot prompt are not decisive for model performance: randomly flipping labels of few-shot demonstrations shows to harm an LLM's ability to solve NLP tasks barely. However, few-shot learning (with and without random labels) vastly outperforms zero-shot learning (\ie no demonstrations are provided in the prompt). The authors argue that the demonstrations are helpful for task performance in that the LLM instead learns the label space and the input distribution of the task. 

In later work, \citet{pan2023incontext} explain that there are two distinct mechanics through which ICL leverages demonstrations: on the one hand, \emph{task recognition} is the ability to recognize a task through demonstrations (possibly without ground-truth labels or perhaps even wrong ones, as in the case of ~\citet{min2022rethinking}). After this recognition phase, it applies its pre-trained capabilities. On the other hand, the skill to acquire new input-label mappings unseen in pre-training is called \emph{task learning}. 

While input-label associations may not seem to drive few-shot performance, at least in the case of task recognition, ~\citet{lu-etal-2022-fantastically} show that the order of few-shot examples matters in that LLMs are highly sensitive to permutations of the order in which the few-shot demonstrations are provided.

Alternative explanations of the ICL phenomenon take place around Bayesian inference \cite{xieExplanationIncontextLearning2022}, sparse linear regression \cite{Abernethy2023AMF}, structure induction \cite{Hahn2023ATO},  maintaining coherence \cite{Sia2023IncontextLA}, kernel regression \cite{Han2023InContextLO}, and clone-structured causal graphs \cite{swaminathanSchemalearningRebindingMechanisms2023}.

\textbf{Instruction-Following} is mainly explained in \Cref{sec:alignment}, as it requires supervised fine-tuning. To briefly recap, the idea is to prepend task-describing instructions (\eg \emph{``This is a text classification task for movie reviews. Here are a few examples: ...''}) in the input prompts. 

\textbf{Chain-of-Thought (CoT)~\citep{ling-etal-2017-program, wei2022chain}} describes a technique used to construct few-shot prompts via a series of intermediate reasoning steps leading to the final output. Answer rationales to solve algebraic problems were originally proposed in the pre-LLM era~\citep{ling-etal-2017-program} and later experienced big popularity as a prompting strategy for LLMs~\citep{wei2022chain}. Extensions of chain-of-thought prompting include zero-shot variants~\citep{kojima2022large} and automatically generated series of reasoning steps~\citep{zhang2022autocot}.

\textbf{Impersonation~\citep{salewski2023context}} is a technique in which the prompt for the model asks it to pretend to be a domain expert when answering a domain-specific question. \citet{salewski2023context} find that LLMs answer domain-specific questions more accurately when prompted to impersonate a domain expert.

\paragraph{Multi-Turn Prompting} methods iteratively chain prompts and their answers together. \label{sec:multi_turn_prompting}

\textbf{Ask Me Anything~\citep{arora2022ama}} uses multiple prompt templates (called prompt chains), which are used to reformat few-shot example inputs into an open-ended question-answering format. The final output is obtained by aggregating the LLMs predictions for each reformatted input via a majority vote.

\textbf{Self-consistency~\citep{wang2022selfconsistency}} extends chain-of-thought prompting by sampling multiple reasoning paths and selecting the most consistent answer via a majority vote.

\textbf{Least-to-Most~\citep{zhou2022least}} uses a set of constant prompts to use the LLM to decompose a given complex problem into a series of subproblems. The LLM sequentially solves the subproblems with prompts for later-stage subproblems containing previously produced solutions, iteratively building the final output.

\textbf{Scratchpad~\citep{nye2021scratchpad}} is a method to fine-tune LLMs on multi-step computation tasks such that they output intermediate reasoning steps, \eg intermediate calculations when performing additions, into a ``scratchpad'' before generating the final result.

\textbf{ReAct~\citep{yao2022react}} combines reasoning and acting by prompting LLMs to generate reasoning traces (\eg Chain-of-thought) and action plans, which can be executed to allow the model to interact with external environments such as Wikipedia to incorporate knowledge. 

\textbf{Automatic Reasoning and Tool-Use (ART)~\citep{paranjape2023art}} is a method to automatically generate multi-step reasoning prompts, including symbolic calls to external tools such as search and code generation or execution. To this end, ART retrieves demonstrations of related tasks from a library of tasks with accompanying reasoning steps and uses a frozen language model to generate intermediate reasoning steps.

\textbf{Self-refine~\citep{madaan2023selfrefine}} is based on the notion of iterative refinement, \ie improving an initial solution over multiple steps. To this end, a single LLM generates an initial output and then iteratively provides feedback on the previous output, followed by a refinement step in which the feedback is incorporated into a revised output. 
 
\textbf{Tree of Thoughts~\citep{yaoTreeThoughtsDeliberate2023}} generalize CoT to maintain a tree of thoughts (with multiple different paths), where each thought is a language sequence that serves as an intermediate step. Doing so enables the LLM to self-evaluate the progress intermediate thoughts make towards solving the problem and incorporating search algorithms, such as breadth-first or depth-first search, allowing systematic exploration of the tree with lookahead and backtracking. 

\paragraph{Controlled Generation}
The approaches above primarily modify the prompt text to steer model outputs. However, instead of reformulating the input text, we can control the output by approaches that directly modify the inference procedure given a fixed set of prompts. Before the advent of LLMs, this line of work has been referred to as \emph{controlled generation} \cite{keskar2019ctrl,dathathri2020plug,krause-etal-2021-gedi-generative}.

In the context of LLMs, \citet{sanchezStayTopicClassifierFree2023} proposes to use classifier-free guidance sampling \cite{ho2022classifierfree}, where the input prompt's importance is up-weighted throughout the generation of a sequence. \citet{roushallenYouProbablyDon} proposes five ideas related to modifying the prompt throughout the decoding of a single sequence; for example, alternating between two input prompts. Such works often borrow ideas from the text-to-image generation community \cite{nichol2022glide,automatic1111StableDiffusionWeb2023}. One idea we have not seen borrowed yet is negative prompting, \ie including a description of unwanted outputs. According to \citet{NegativePromptsText2023}, the first attempts at such an idea resulted in negative outcomes. 

\subsection{Hallucinations} \label{sec:hallucinations}
The popularity of services like ChatGPT suggests that LLMs are increasingly used for everyday question-answering. As a result, the factual accuracy of these models has become more significant than ever. 

\begin{figure}[h]
    \centering
    \includegraphics[width=\columnwidth]{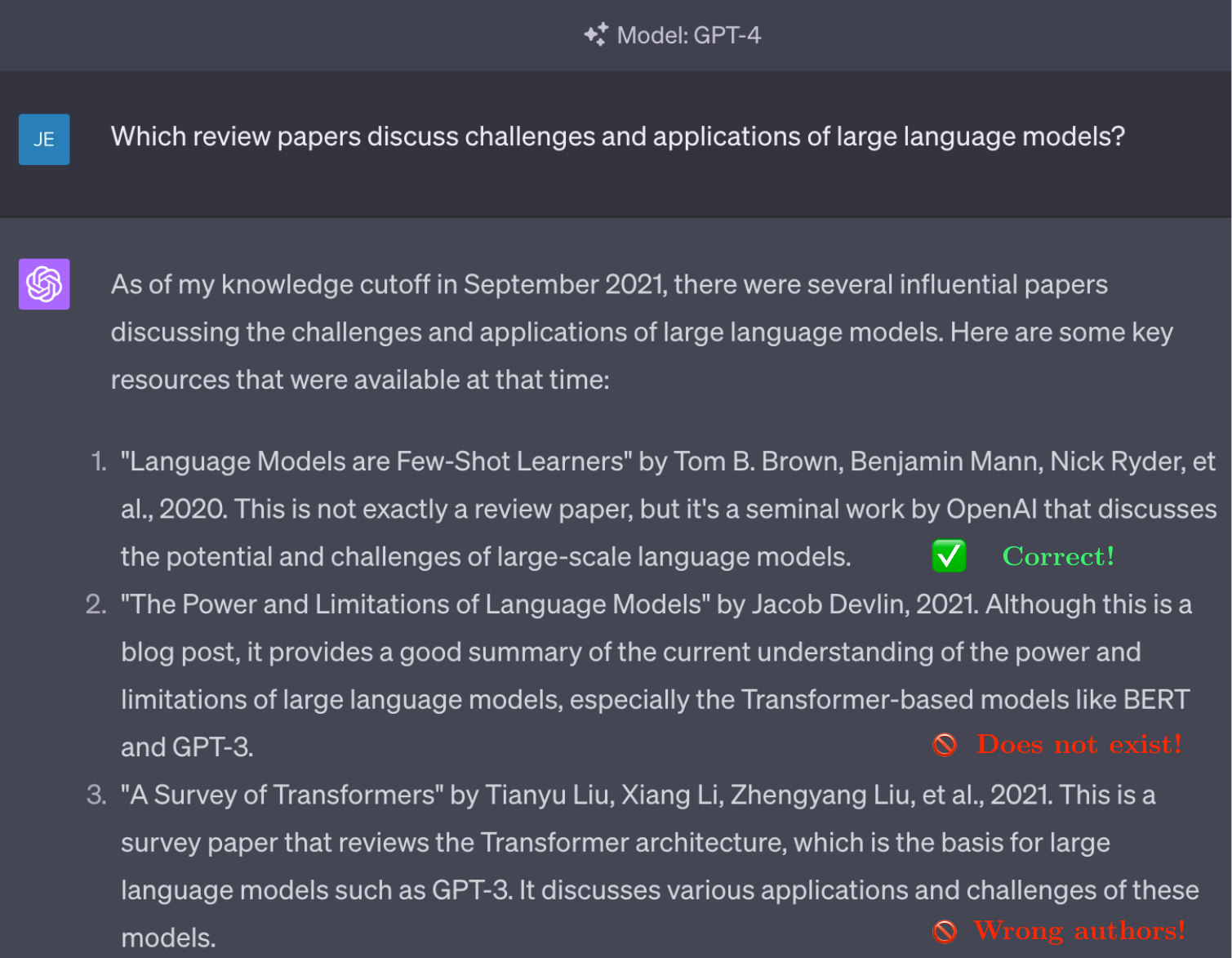}
    \caption{\textbf{Example of Hallucinations with GPT-4}, accessed on 02/06/2023.}
    \label{fig:example_hallucination}
\end{figure}

Unfortunately, LLMs often suffer from \emph{hallucinations}, which contain inaccurate information that can be hard to detect due to the text's fluency. \Cref{fig:example_hallucination} illustrates an example.

\begin{figure*}
  \centering
  \includegraphics[width=\linewidth]{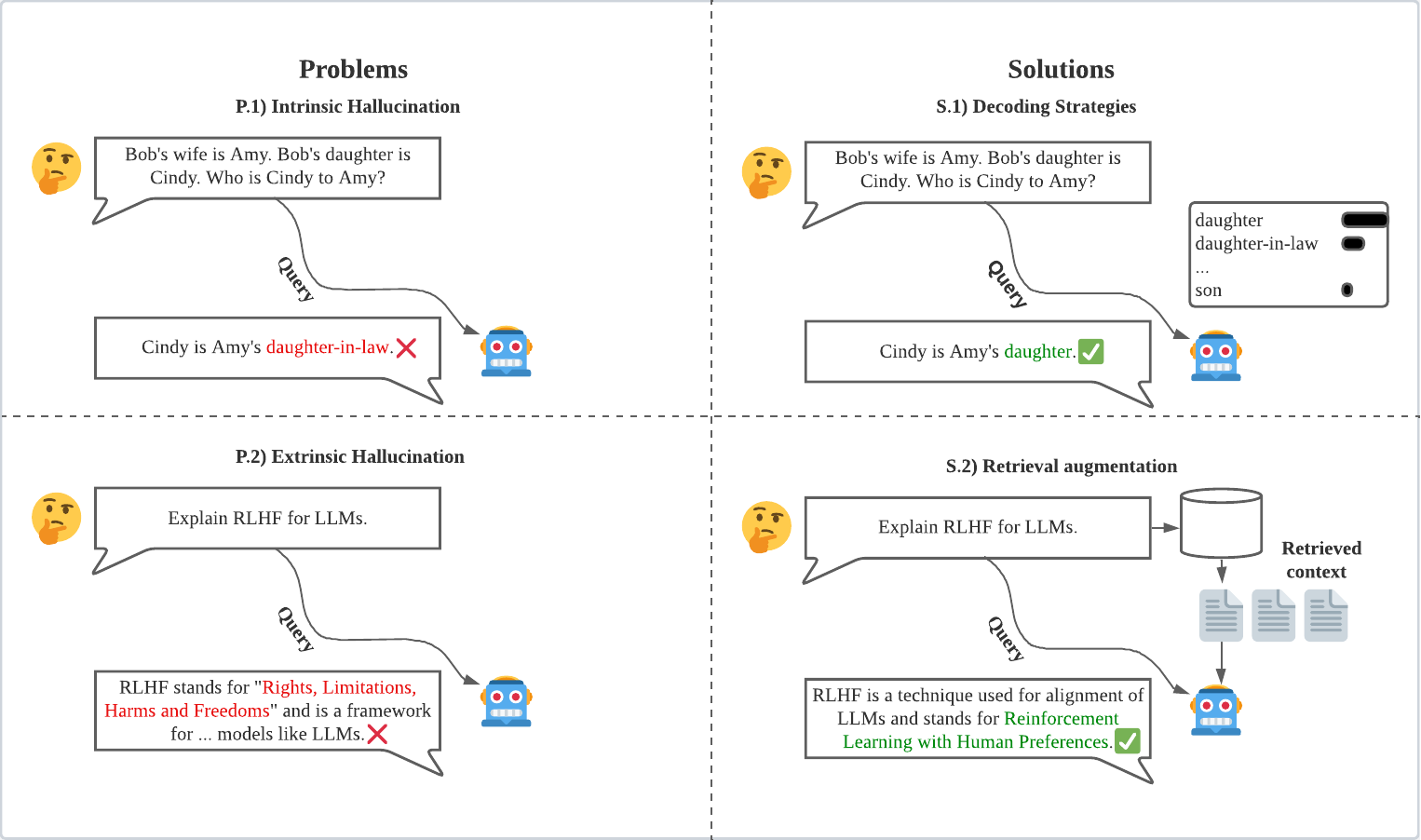}
  \caption{\textbf{Illustration of a) intrinsic and b) extrinsic hallucinations} in user interaction with an LLM, inspired by~\citet{zhaoSurveyLargeLanguage2023}. In a), the produced answer contradicts the given context, whereas in b), the context does not provide enough information about whether the produced answer would contradict.}
  \label{fig:hallucinations}
\end{figure*}

To distinguish between different types of hallucinations, we consider the provided \emph{source content} of the model, \eg the prompt, possibly including examples or retrieved context. Based on such, we can distinguish between \emph{intrinsic} and \emph{extrinsic} hallucinations~\citep{jiSurveyHallucinationNatural2023}. In the former, the generated text logically contradicts the source content. In the latter, we cannot verify the output correctness from the provided source; the source content does not provide enough information to assess the output, which is, therefore, under-determined. Extrinsic hallucination is not necessarily erroneous, as it merely means the model generated an output that can neither be grounded nor contradicted by the source content. This is still, to some degree, undesirable as the provided information cannot be verified. We illustrate intrinsic and extrinsic hallucinations in \Cref{fig:hallucinations}.

\begin{chall}{Hallucination~\citep{lee2018hallucinations,rohrbach2018object,jiSurveyHallucinationNatural2023}}
Generated text that is fluent and natural but unfaithful to the source content (intrinsic) and/or under-determined (extrinsic).
\end{chall}

\citet{liuExposingAttentionGlitches2023} attribute hallucinations commonly observed in LLMs to an architectural flaw in Transformer models while observing that recurrent neural networks perfectly solve their minimalistic synthetic benchmarks, designed to isolate the issue of hallucination in the context of algorithmic reasoning. Here, we focus on ways to address hallucinations in LLMs without changing the model architecture itself, including (i) supplying the LLM with relevant sources (\emph{retrieval augmentation}) or (ii) decoding strategies. 

\paragraph{How to Measure Hallucinations} \citet{leeFactualityEnhancedLanguage} provide the \emph{FactualityPrompts} dataset consisting of factual and nonfactual input prompts, which allows one to isolate the effect of prompt's actuality on the model's continuation. Further, they measure hallucinations using named-entity- and textual entailment-based metrics.
\citet{minFActScoreFinegrainedAtomic2023} notice that evaluating factuality can be difficult because generations can contain a mixture of supported and unsupported information, making binary judgments of quality inadequate and human evaluation time-consuming. Hence, they propose a framework that first breaks generations into atomic facts and then computes the percentage of atomic facts supported by an external knowledge source like Wikipedia. \citet{zhangHowLanguageModel2023} detect the behavior of \emph{hallucination snowballing}, where the LLM over-commits to early mistakes (before outputting the explanation) in its generation, which it otherwise would not make.

\paragraph{Retrieval Augmentation} One way to mitigate hallucinations is to ground the model's input on external knowledge, which is often referred to as \emph{retrieval augmentation}. In other words, we can decouple (i) memory storage of knowledge (\eg databases or search indexes \cite{lazaridou2022internetaugmented}) and (ii) processing of the knowledge to arrive at a more modular architecture. For (i), a \emph{retriever} module retrieves the top-$k$ relevant documents (or passages) for a query from a large corpus of text. Then, for (ii), we feed these retrieved documents to the language model together with the initial prompt. In theory, using an external data source may also make it easier to interpret which knowledge is retrieved and update it without tediously fine-tuning the model.

\citet{shuster2021retrieval} demonstrate hallucinations in GPT-3 and study various components of retrieval-augmented architectures to mitigate them. Their best models reduce hallucinated responses by over 60\% on average and up to 85\% on out-of-distribution data, on which the model has not been trained.

We summarize a few popular retrieval augmentation (RA) approaches as follows.
\emph{Retrieval-augmented language model pre-training} (REALM)~\citep{guu2020retrieval} inserts retrieved documents into the pre-training examples. While~\citet{guu2020retrieval} designed REALM for extractive tasks such as question-answering,~\citet{lewis2020retrieval} propose \emph{retrieval-augmented generation} (RAG), a language generation framework using retrievers for knowledge-intensive tasks that humans could not solve without access to an external knowledge source.~\citet{yogatama-etal-2021-adaptive} propose the \emph{adaptive Semiparametric Language Models} architecture, which incorporates the current local context, a short-term memory that caches earlier-computed hidden states, and a long-term memory based on a key-value store of (hidden-state, output) tuples. To equip a retrieval-augmented LLM with few-shot abilities that were before only emergent in LLMs with many more parameters,~\citet{atlas} propose a KL-divergence loss term for retrieval models, resulting in \textsc{Atlas}.~\citet{retro} study scaling up retrieval databases up to 2 trillion tokens and achieving comparable performance to GPT-3 on some tasks despite using $25\times$ fewer parameters while highlighting the retrieval model's ability to copy-paste existing training chunks.~\citet{taskaware_retrieval} introduce a collection of 40 retrieval datasets with instructions and a corresponding model trained on them. 

\begin{figure}
    \centering
    \includegraphics[width=\linewidth]{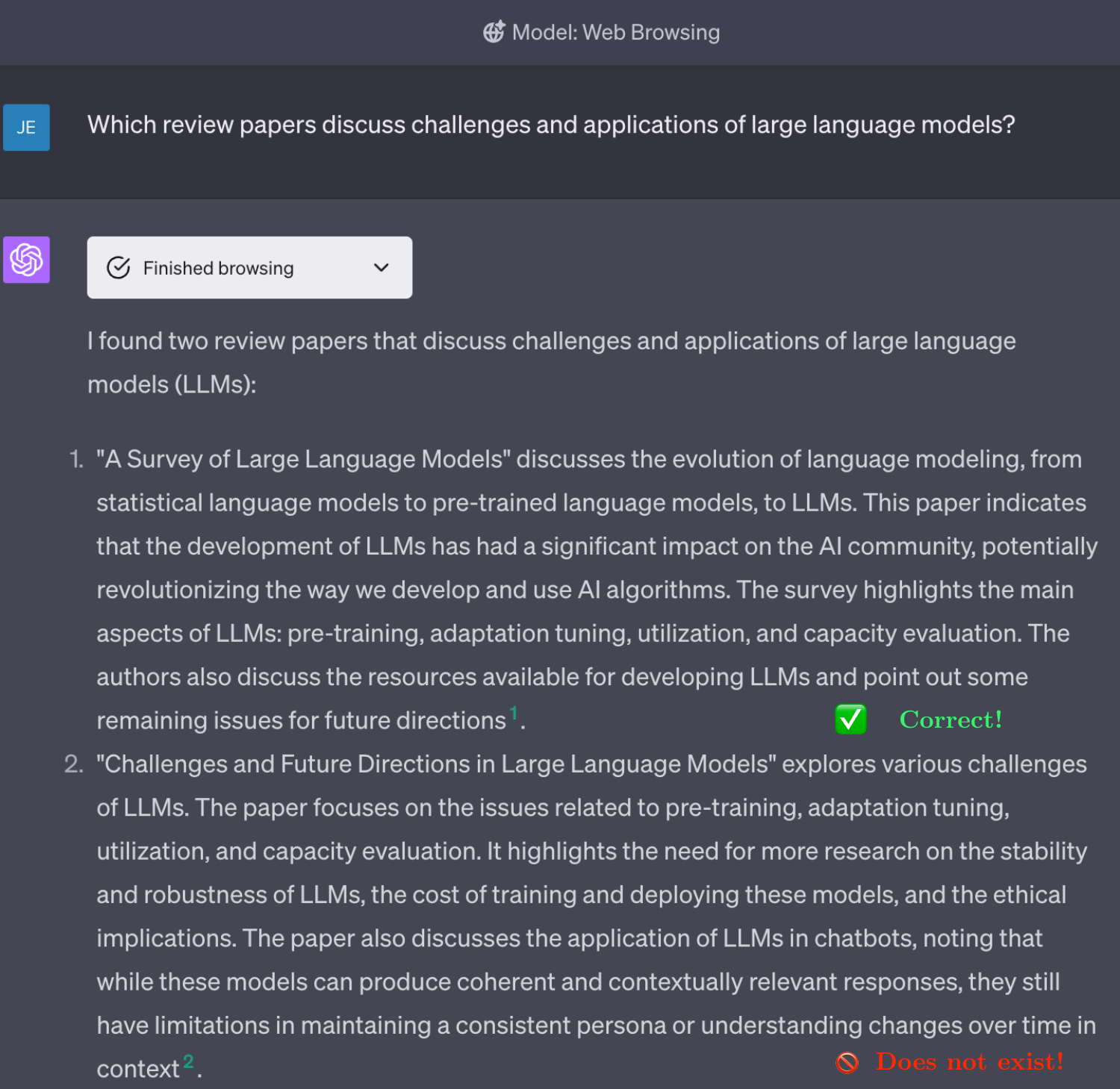}
    \caption{\textbf{Example of Retrieval-Augmented GPT-4}, accessed on 02/06/2023.}
    \label{fig:web_browsing}
\end{figure}

However, standard RA does not always solve the hallucinations problem. \Cref{fig:web_browsing} illustrates an example of ChatGPT browsing the web first to retrieve relevant documents before answering the query. While the Bing browsing plugin retrieves two (existent) related papers (\cite{zhaoSurveyLargeLanguage2023,yan2023practical}), unfortunately, the final response still contains a hallucination: the second paper's title and summary are factually inaccurate. The second paper's true title is ``Practical and Ethical Challenges of Large Language Models in Education: A Systematic Literature Review''~\cite{yan2023practical}.  

Another failure mode of RA is illustrated by \citet{khattabDemonstrateSearchPredictComposingRetrieval2023}, who find that sometimes the retriever cannot find passages that directly answer the question. Hence, they propose a framework that unifies techniques from RA and multi-turn prompting (\Cref{sec:multi_turn_prompting}) to solve more complex questions programmatically.

\paragraph{Decoding Strategies}
Another approach to mitigating hallucinations is refining the decoding strategy during inference time.~\citet{leeFactualityEnhancedLanguage} show that standard decoding algorithms (\eg top-p truncation) can induce hallucinations due to the uniform randomness introduced at every sampling step.~\citet{dziriNeuralPathHunter2021} observe a positive correlation between increased diversity in response generation and hallucinations. 

The reason for inducing randomness and diversity in popular decoding strategies is that generating the most likely sequence often leads to an unsurprising and unnatural text compared to human communication~\citep{see-etal-2019-massively, holtzman2020topp,zhang-etal-2021-trading}.~\citet{zhang-etal-2021-trading} phrase this challenge as a trade-off between diversity and quality. While this challenge remains largely unsolved, several approaches such as diverse beam search~\citep{vijayakumar2018dbs} and confident decoding~\citep{tianStickingFactsConfident2020} try reducing the induced hallucinations at the decoding level.

\textbf{Uncertainty-Aware Beam Search~\citep{xiaoHallucinationPredictiveUncertainty2021}} is based on the observation that higher predictive uncertainty corresponds to a larger chance of generating hallucinations. Therefore, the method introduces a penalty term in the beam search to penalize high predictive uncertainty during decoding.

\textbf{Confident Decoding~\citep{tianStickingFactsConfident2020}} hypothesize that hallucinations of encoder-decoder models originate by not attending to the source when decoding. They propose an attention-based confidence score to measure how strongly a model attends the source and a variational Bayes training procedure to ensure the model generates high-confidence answers.

\subsection{Misaligned Behavior}
\label{sec:alignment}

The alignment problem refers to the challenge of ensuring that the LLM's behavior aligns with human values, objectives, and expectations and that it does not cause unintended or undesirable harms or consequences~\citep{russell2021human,gabriel2020artificial,hendrycks2020aligning}. Most of the existing alignment work can be categorized into either methods for detecting misaligned behavior (such as model evaluation and auditing, mechanistic interpretability, or red teaming) or methods for aligning model behavior (such as pre-training with human feedback, instruction fine-tuning, or RLHF).

\begin{chall}{Misaligned Behavior}
LLMs often generate outputs that are not well-aligned with human values or intentions, which can have unintended or negative consequences.
\end{chall}

\paragraph{Pre-Training With Human Feedback}\citet{korbak2023rlhfpre} introduce the concept of \emph{pre-training with human feedback} (PHF) where human feedback is incorporated during the pre-training stage rather than during fine-tuning. The authors compare five different PHF approaches such as filtering~\citep{solaiman2021process,wang-2021-comment}, conditional training~\citep{ficek-etal-2022-tackle,fan-etal-2021-augmenting,keskar2019ctrl}, unlikelihood~\citep{welleck2019neural}, reward-weighted regression~\citep{peters-martins-2021-smoothing}, and advantage-weighted regression~\citep{peng-2021-marvs}, and find that conditional training leads to the best trade-off between alignment and capabilities. Conditional training is a simple technique that prepends a control token $c$ (\eg {\tt<|good|>} or {\tt<|bad|>}) before each training example $x$ depending on the outcome of a thresholded reward function $R(x) \geq t$. During inference, the model generations are conditioned on $c = \text{{\tt<|good|>}}$. Conditional training results in significantly better alignment with human preferences than standard LM pre-training, followed by fine-tuning with human feedback without hurting downstream task performance.

\paragraph{Instruction Fine-Tuning} \label{par:instruction_finetuning}~\citet{yi-etal-2019-towards, wei2022finetuned, mishra-etal-2022-cross, ouyang2022gptinstruct, wang2022super} fine-tune pre-trained LLM on instructional data, \ie data containing natural language instructions and the desired responses according to human judgment. Instruction-tuned (IT) LLMs often reach state-of-the-art downstream performances and improve over their non-IT counterparts~\citep{opt_iml,flan}, as can be seen, \eg in the publicly available HELM evaluations~\citep{helm_website}.~\citet{ouyang2022gptinstruct, wang2022selfinstruct} find that they produce more truthful and less toxic text while generating preferred outputs.

To generate instruction sets,~\citet{zhou2023large} propose the Automatic Prompt Engineer (APE) method, which leverages LLMs to generate, score, and rephrase instruction-following zero- and few-shot prompts.~\citet{flan_collection} describe and analyze the steps taken to create an improved version of the Flan collection~\citep{wei2022finetuned} used to train FLAN-PaLM~\citep{flan}. When trained on this data, the authors find that the improved model performance stems from more diverse tasks by inverting input-output pairs and data augmentation techniques such as mixing zero-shot and few-shot prompts.~\citet{honovich2022unnatural} generate a large dataset of natural language instructions using a pre-trained LLM to generate and then rephrase instructions. They show that a T5 ("LM-adapted") fine-tuned on this data outperforms other instruction fine-tuned T5 models such as T0++~\citep{t0} and Tk-Instruct~\citep{wang2022super}.

\paragraph{Reinforcement Learning From Human Feedback (RLHF)} is a variation of RL that incorporates feedback from humans in the form of rewards~\citep{christiano2017rlhf,stiennon2020learning} and has proven to be an effective way of aligning LLMs with human preferences~\citep{ouyang2022gptinstruct,bai2022constitutional}. RLHF works by using a pre-trained LM to generate text, which is then evaluated by humans by, for example, ranking two model generations for the same prompt. This data is then collected to learn a reward model that predicts a scalar reward given any generated text. The reward captures human preferences when judging model output. Finally, we optimize the LM against such reward model using RL policy gradient algorithms like PPO~\citep{schulman2017proximal}. RLHF can be applied directly to a general-purpose LM pre-trained via self-supervised learning. However, applying RLHF right after pre-training may not be good enough for more complex tasks. In such cases, RLHF is typically applied after an initial supervised fine-tuning phase using a small number of expert demonstrations for the corresponding downstream task~\citep{ramamurthy2022reinforcement, ouyang2022gptinstruct, stiennon2020learning}. 
RLHF has also proven helpful for a wide range of language generation tasks, from summarization~\citep{ziegler2019fine, wu2021recursively, stiennon2020learning} to training more helpful, harmless, and accurate assistants~\citep{sparrow, cohen2022dynamic, ouyang2022gptinstruct, bai2022constitutional}, and learning to use tools~\citep{reiichiro2021webgpt,rae2021gopher,gophercite}.

RLHF can also introduce unwanted side effects.~\citet{perez2022discovering} show that LLMs fine-tuned with RLHF can be more inclined to repeat back a user's (preferred) political views and much more likely to express particular political and religious views as well as an increased stated desire not to be shut down. Regarding the latter, the models elaborated that this would interfere with their goal of being helpful. However, the authors equally observed positive or neutral behavior reinforcements when fine-tuning LLMs with RLHF.

Further, there is an ongoing debate about the extent to which the ``RL'' in RLHF is needed.~\citet{rafailov2023direct} identify a mapping between reward functions and optimal policies, which allows them to design \emph{Direct Preference Optimization} (DPO), an algorithm that implicitly optimizes the same objective as existing RLHF algorithms. DPO requires only solving a classification problem on the human preference data, eliminating the need to fit a reward model and employ RL. Similarly,~\citet{zhouLIMALessMore2023} find that fine-tuning LLaMa on only 1,000 selected prompts and responses, without any RL or reward modeling, can be enough to outperform RLHF-trained models like DaVinci003 from OpenAI. Consequently, the authors pose the \emph{Superficial Alignment Hypothesis}: The knowledge and skills of a model are primarily acquired during the pre-training phase, while alignment instructs it on the appropriate subdistribution of formats to use in user interactions.

Since RLHF involves many different components such as (1) the preferences data collected from humans, (2) the reward models to learn the human preferences, and (3) the policy optimization algorithm (\eg PPO),  \citet{zhengSecretsRLHFLarge2023} announce to release a sequel dissecting each. The most recent part focuses on step (3) and finds that various RL tricks can be applied to make vanilla PPO more stable.   

\begin{figure}[ht]
    \centering
    \includegraphics[width=\linewidth]{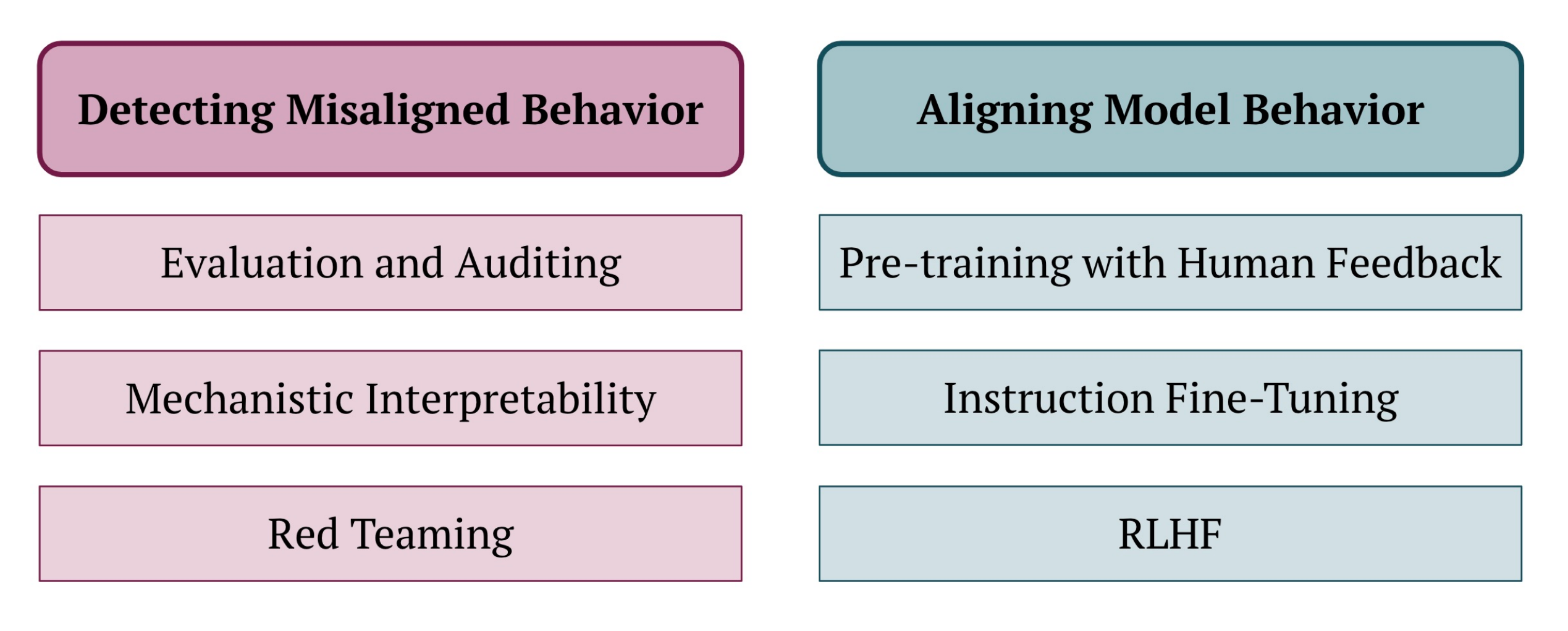}
    \caption{\textbf{Alignment.} We categorize existing alignment work into methods for detecting misaligned behavior or aligning models.}
    \label{fig:alignment}
\end{figure}

\paragraph{Self-improvement} refers to fine-tuning an LLM on self-generated data~\citep{self_improvement}. While this technique can be used to improve the model's capabilities, it can also be used to improve the model's alignment with human values.~\citet{self_improvement} first demonstrate this ability by annotating unlabeled reasoning datasets. Surprisingly, this allows the LLM to \emph{self-improve} by significant amounts. Similarly,~\citet{zelikman2022star} bootstrap LLMs by iteratively prompting them to generate rationales and then fine-tuning them on those leading to correct answers.

More related to the alignment problem,~\citet{bai2022constitutional} self-critique generated outputs and produce refinements conditioned on these critiques, which are then used to fine-tune a pre-trained model. Similarly,~\citet{liu2023languages} propose \emph{Chain of Hindsight} (CoH), which conditions models on generations paired with natural language feedback, allowing the model to detect and correct mistakes. CoH results in better alignment with human preferences than other methods according to human evaluations, leading to significant improvements in summarization and dialogue.~\citet{ma2023oops} use a similar technique to detect and repair unethical LLM outputs automatically. In a similar spirit,~\citet{wang2023self} encourage LLMs to critique their given instructions to reduce harmful outputs due to a user's malicious intent.

~\citet{schick2023toolformer} propose \emph{Toolformer}, a novel approach in which LLMs generate and filter their own tool-use examples to teach themselves when and how to call different APIs such as a retriever model, a calculator, or a calendar, which can improve the model's factuality, mathematical capabilities, and time-awareness. Besides learning to use tools~\citep{gou2023critic}, self-improvement was also employed for learning how to code~\citep{to2023better, chen2023teaching} or solve computer tasks~\citep{kim-2022-revisiting}. \citet{cohenLMVsLM2023} study cross-examination between two LLMs, where the \emph{examiner} LLM tries to detect factual errors by the \emph{examinee} LLM through multi-turn interactions. In the future, similar approaches could be used to develop LMs that know when to query a human or better-aligned model to ask for alignment advice when uncertain.

\paragraph{Evaluation and Auditing}
The ability to scalably and thoroughly evaluate LM behaviors and detect when they are harmful is of great importance for alignment. For example, ~\citet{shevlane2023model} highlight the importance of model evaluation for addressing extreme risks such as offensive cyber capabilities or strong manipulation skills. Recently,~\citet{carlini2023aligned} discovered that even aligned LLMs (which were instruction fine-tuned to prevent harmful behaviors) can be adversarially attacked via brute force (although current NLP-based attacks fail). A large body of work evaluates models via crowdsourcing or existing data sources. However, this can be time-consuming, expensive, or unavailable. Recently,~\citet{perez2022discovering} propose automatically generating evaluations using LLMs. This approach has a high agreement with crowd workers, leading to high-quality, diverse evaluations and the discovery of many new behaviors. In addition, it has a high agreement with crowd workers. The authors discover new cases of inverse scaling where LLMs get worse with size, such as repeating back a user's preferred answer and a greater desire to pursue concerning goals like resource acquisition and goal preservation. They also find that RLHF makes LLMs express stronger political views and a greater desire to avoid a shutdown. LLM evaluation and auditing are critical for informing policymakers and other stakeholders and making responsible decisions about model training, deployment, and security. \Cref{sec:evaluation} discusses the evaluation of LLM capabilities more broadly, while in this section, we focus on evaluating whether the model's behaviors are harmful and more relevant for alignment (\eg red teaming, mechanistic interpretability). 

\paragraph{Red Teaming} is one of the most promising and widely used approaches for detecting harmful content generated by LLMs. Typically, models are red-teamed by asking humans to generate prompts that lead to undesirable model outputs. In a recent study,~\citet{ganguli2022red} investigate the scaling behavior of red teaming across different model sizes and model types (a pre-trained LLM, an LLM prompted to be helpful, honest, and harmless); an LLM that uses rejection sampling at test time, and an LLM fine-tuned with RLHF). They find that \textbf{red-teaming RLHF models} becomes more difficult as they scale while red-teaming the other models remains the same as they scale.
~\citet{perez2022red} automatically find cases where a target LLM behaves in harmful ways by optimizing another LLM via reinforcement learning to generate prompts that lead to offensive responses. This approach uncovers tens of thousands of offensive replies in a chatbot, groups of people that are discussed in offensive ways, personal and hospital phone numbers generated as the chatbot's own contact info, leakage of private training data in generated text, as well as harms that occur over the course of a conversation.

Taking a different approach,~\citet{lee2023query} propose \textbf{Bayesian red teaming}, which iteratively identifies diverse positive test cases leading to model failures by utilizing the pre-defined user input pool and past evaluations via Bayesian optimization.

Most works on red teaming LLMs use a classifier to detect undesired outputs, assuming the harmful behavior is known with precision beforehand~\citep{casper2023explore}. However, this is not always the case, so~\citet{casper2023explore} aim to relax this assumption considering that the adversary only has access to a high-level, abstract specification of undesired behavior. They propose a three-stage approach where they first explore the model's behavior in the desired context, then establish a measurement of undesired behavior, and then exploit the model's flaws using this measure and an established red teaming methodology.

In the past, coevolution algorithms that simultaneously evolve strong strategies along with dangerous counter-strategies have been shown to work well in realistic domains~\citep{hingston2011red}. Hence, applying such techniques for \textbf{automatically red-teaming} LLMs could be a fruitful research direction. Another research area related to red teaming is \textit{debate} which aims to leverage other AI models to evaluate whether the model's behaviors are safe and useful during training. These methods are expected to be particularly useful for aligning future powerful LLMs when the tasks are too complex for humans to judge the model's plans or actions directly.

~\citet{irving2018ai} train models via self-play on zero-sum debate games. More specifically, given a question or proposed action, two agents take turns making short statements up to a limit, then a human judges which of the agents gave the most accurate and most useful information. This approach has improved factuality and reasoning in LLMs~\citep{du2023improving}. However, it requires multiple generations, which can slow down the time-to-result (\Cref{sec:inference_costs}) and longer context windows, which many LLMs still struggle with (\Cref{sec:context_length}).

\paragraph{Emergent Capabilities} Understanding which capabilities will emerge while training LLMs and when they will emerge is an important step in ensuring that we do not train unsafe or misaligned LLMs~\citep{hendrycks2021unsolved,steinhardt2022future}. In addition, a better understanding of the factors that lead to these emergent capabilities could allow us to make desirable abilities emerge faster and ensure undesirable abilities do not ever emerge, which are essential for AI safety and alignment. \citet{wei2022emergent} claim that LLMs display emergent abilities, \ie capabilities that are not present in smaller-scale models that are present in larger-scale models.~\citet{schaeffer2023emergent} propose an alternative explanation: emergent abilities may appear due to the researcher's choice of metric rather than fundamental changes in model behavior with scale. Various studies provide evidence that these alleged emergent abilities disappear when using different metrics or better statistics and may not be a fundamental property of scaling LLMs. Multiple papers have argued that AI systems could learn to deceive, even if they are not explicitly trained to do so because deception can help agents achieve their goals ~\citep{brundage2018malicious,hendrycks2021unsolved,hendrycks2022x,bubeck2023sparks,kenton2021alignment}. For example, it could be easier to gain human approval through deception than to earn it legitimately. In addition, models capable of deception have a strategic advantage over always honest models, so there is a hidden incentive to develop this ability. However, of course, we would like to be able to detect and prevent \textit{emergent deception} in AI systems since this can have unintended negative consequences. 
\citet{emergent_deception} study whether current LLMs generate deceptive outputs and how deception scales with the number of parameters, showing that deception can indeed emerge at larger model sizes in both pre-trained LLMs and LLMs fine-tuned with RLHF. Similarly, ~\citet{hazell2023large} show that LLMs can already be used in phishing campaigns, suggesting that deceptive behavior can already be extracted from them when prompted in particular ways.

\paragraph{Mechanistic Interpretability} (MI) is another important research area for AI alignment which aims to understand better how the models work at a low level to enable the detection of undesirable behaviors or even instill desirable behaviors directly in the model's weights. More specifically, the goal of MI is to reverse-engineer an LLM's learned behaviors into their individual components, \ie a process to find and understand human-interpretable neurons. As an analogy,~\citet{MechanisticInterpretabilityVariables} compares MI with reverse-engineering compiled program binaries into human-readable source code. For example,~\citet{elhage2021mathematical}; discover that small Transformers have components that can be understood as interpretable circuits, while~\citet{olsson2022context} find a mechanism that seems to drive a significant fraction of in-context learning. Similarly,~\citet{meng2022locating} aim to locate factual associations in language models.~\citet{nanda2023progress} find that the emergent grokking phenomenon is not a sudden shift but rather arises from the gradual amplification of structured mechanisms encoded in the weights, followed by the later removal of memorizing components. Extending this work, ~\citet{conmy2023towards} propose a new algorithm to automate the identification of important units in a neural network. Given a model's computational graph, this algorithm finds subgraphs that explain a particular behavior of the model. In a similar spirit,~\citet{liu2023seeing} introduce a method for making neural networks more modular and interpretable by embedding neurons in a geometric space and augmenting the loss function with a cost proportional to the length of each neuron connection. This approach discovers useful modular neural networks for many simple tasks, revealing compositional structures in symbolic formulas, interpretable decision boundaries, and features for classification, as well as mathematical structure in algorithmic datasets. In an attempt to understand how an LLM's predictions change after each layer, ~\citet{belrose2023eliciting} develop a method that can decode any hidden state into a distribution over the vocabulary. Using this technique, the authors show that the trajectory of latent predictions can be used to detect malicious inputs with high accuracy. Finally,~\citet{burns2022discovering} introduce a method that can recover diverse knowledge represented in LLMs across multiple models and datasets without using any human supervision or model outputs. In addition, this approach reduced prompt sensitivity in half and maintained a high accuracy even when the language models are prompted to generate incorrect answers. This work is a promising first step towards better understanding what LLMs know, distinct from what they say, even when we don't have access to explicit ground truth labels.

\paragraph{Biases}
Since the pre-training datasets of LLMs are often unfathomable (\Cref{sec:datasets}) and contain web-crawled data, they most likely contain online discourse involving political discourse (\eg climate change, abortion, gun control), hate speech, discrimination, and other media biases.~\citet{paullada2021data} find misogyny, pornography, and other malignant stereotypes~\citep{birhane2021multimodal,biderman2021pitfalls,minipile} in pre-training datasets. Similarly,~\citet{fengpretrainingDataLanguage2023} find that LLMs have political leanings that reinforce the polarization present in the pre-training corpora, propagating social biases into hate speech predictions and misinformation detectors.
Several recent papers discuss the potential origins of biases in LLMs (such as training data or model specification), ethical concerns when deploying biased LLMs in various applications, as well as current ways of mitigating these biases~\citep{ferrara2023should,liu2022quantifying,liang2021towards}. Finally,~\citet{viswanath2023fairpy} present a comprehensive quantitative evaluation of different kinds of biases, such as race, gender, ethnicity, age, etc., exhibited by some popular LLMs. They also release an easy-to-use toolkit that allows users to debias existing and custom models using existing methods.

\paragraph{Toxicity Detection}
\citet{weidinger2021ethical} denote toxicity as one of the main risks associated with LLMs. What makes this problem particularly challenging is the label ambiguity, where output may be toxic in a certain context but not in others, and different people may have different notions of toxicity~\citep{ousidhoum2021probing, gehman2020realtoxicityprompts,deshpande2023toxicity}.~\citet{jones-2022-development} propose to detect toxic outputs using discrete optimization automatically. Similarly,~\citet{faal2023reward} employ reward models to mitigate toxicity in LLMs. An alternative way of reducing toxicity is by pre-training LLMs with human preferences~\citep{korbak2023rlhfpre} or instructions~\citep{prabhakaran-etal-2021-releasing}.

\paragraph{Prompt Injections} Recent work demonstrated that LLMs can be very sensitive to prompt injections, which makes them brittle and unsafe for certain applications~\citep{greshake2023more,wolf2023fundamental}. For example, they can be tricked into leaking personal information such as email addresses from the training data on via \textit{prompt leaking}~\citep{self_improvement,li2023multi}. This poses a significant risk to privacy, particularly when the models are fine-tuned on personal or proprietary data. One can also adversarially prompt LLMs to override the original instructions or employed controls, making them unsafe for certain applications~\citep{greshake2023more,zhao2023prompt,perez2022ignore}. \citet{weiJailbrokenHowDoes2023} attribute such failures to competing capability and safety training objectives and mismatched generalization between safety and capability behavior. 

\paragraph{Agency}
\citet{agent_models} argue that, although LLMs are trained to predict the next word in a text corpus, by doing this, they can infer and represent agentic properties such as the goals, beliefs, or intentions of the human who produced the corresponding piece of text. To support this claim, they present evidence from the literature of LLMs modeling communicative intentions~\citep{radford2017learning}, beliefs~\citep{li2021implicit}, and desires~\citep{lin2020limitations}. If this hypothesis is true, the alignment problem is of even greater importance and may pose additional challenges. This agentic behavior can be problematic from a safety point of view since models could have false beliefs, malicious intents, or even pursue misaligned goals. More research on detecting and preventing such behavior is needed to ensure the safe deployment of LLMs. 

\subsection{Outdated Knowledge}

\begin{figure*}
    \includegraphics[width=\textwidth]{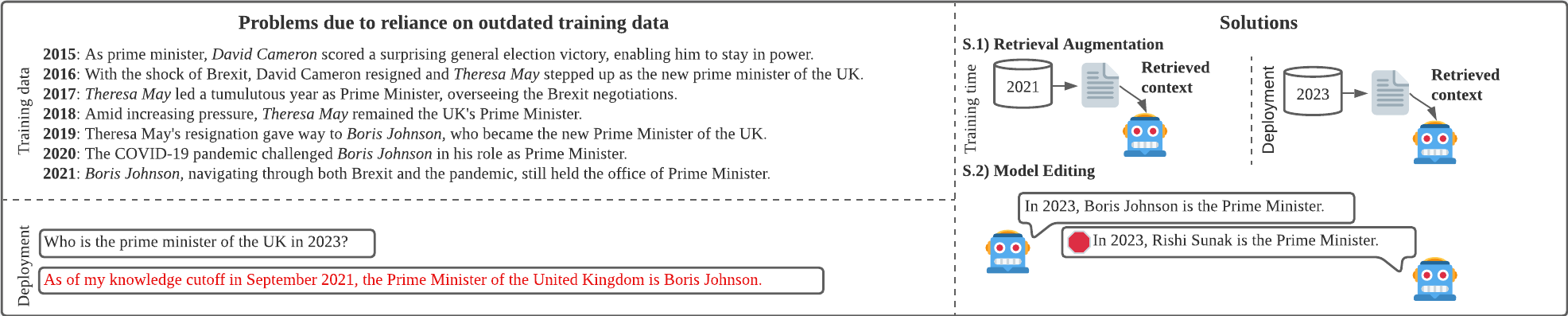}
    \caption{\textbf{Outdated knowledge} can be addressed with S.1) retrieval augmentation by hot-swapping an underlying retrieval index with up-to-date knowledge or S.2) by applying model editing techniques.}
\end{figure*}

Factual information learned during pre-training can contain inaccuracies or become outdated with time (for instance, it might not account for changes in political leadership). However, re-training the model with updated pre-training data is expensive, and trying to ``unlearn'' old facts and learn new ones during fine-tuning is non-trivial. 

Existing model editing techniques are limited in their effectiveness of updating isolated knowledge~\citep{yaoEditingLargeLanguage2023,hoelscherObermaierDetectingEditFailures2023}. For example, \citet{hoelscherObermaierDetectingEditFailures2023} find that model edits can result in unintended associations. This low specificity limits their applicability to real-world use cases, where only a single faulty or outdated bit of information should be updated in a model, and related pieces of information must reflect this update in information equally, without unrelated ones being changed.

\begin{chall}{Isolated Model Updates without Side-Effects \cite{hoelscherObermaierDetectingEditFailures2023}}
Updating isolated model behavior or factual knowledge can be expensive and untargeted, which might cause unintended side-effects.
\end{chall}

Two popular approaches for addressing this issue are \emph{Model editing} ~\citep{sinitsinEDITABLENEURALNETWORKS2020,yaoEditingLargeLanguage2023}, which aims at ``bug-fixing'' models efficiently and leveraging non-parametric knowledge sources in \emph{retrieval-augmented language modeling} (which we omit here and detail in \Cref{sec:hallucinations}).
Current model editing techniques change the model's behavior by modifying the model parameters or using an external post-edit model.

\paragraph{Modifying Model Parameters} techniques can be further split into \textbf{locate-then-edit} methods~\citep{dai-etal-2022-knowledge, meng2022locating, meng2023massediting} which first locate the ``buggy'' part of the model parameters and then apply an update to them to alter their behavior, and \textbf{meta-learning} methods~\citep{de-cao-etal-2021-editing, mitchell2022fast} which use an external model to predict the weight update.

\paragraph{Preserving Model Parameters} methods employ an additional post-edit model~\citep{mitchell2022serac} or insert new weights into the original model~\citep{dong-etal-2022-calibrating, huang2023transformerpatcher} to achieve the desired change in model behavior.~\citet{hartvigsen2022aging} wraps model layers in adapters and adds a similarity-based mechanism to decide when to use the adapter to perform edits in the latent space.

\citet{yaoEditingLargeLanguage2023} find that these methods lack non-trivial generalization capabilities and varying performance and applicability to different model architectures. For example, the best-performing methods ROME~\citep{meng2022locating} and MEMIT~\citep{meng2023massediting} empirically only work well on decoder-only LLMs.

Alternatively, \textbf{retrieval-augmented language modeling} enables the utilization of hot-swappable non-parametric indices. These knowledge sources can be updated during inference time to reflect an updated state of the underlying knowledge. E.g.,~\citet{lewis2020retrieval} demonstrate that swapping their model's non-parametric memory with an updated version enabled it to answer questions about world leaders who had changed between the memory collection dates. Similarly,~\citet{atlas} demonstrate that their retrieval-augmented model can update its knowledge forward and backward in time by swapping the index.

\subsection{Brittle Evaluations}
\label{sec:evaluation}

One reason why the evaluation of language models is a challenging problem is that they have an \textit{uneven capabilities surface}---a model might be able to solve a benchmark problem without issues, but a slight modification of the problem (or even a simple change of the prompt) can give the opposite result~\citep{zhao-etal-2021-calibrate, lu-etal-2022-fantastically,susanzhangPilingPileonSorry2023} (see Section~\ref{sec:prompt_brittleness}). Unlike humans, we cannot easily infer that an LLM that can solve one problem will have other related capabilities. This means that it is difficult to assess the performance of LLMs holistically since rigorous benchmarks are needed to identify weaknesses for a wide variety of inputs. 

\begin{chall}{Brittle Evaluations}
    Slight modifications of the benchmark prompt or evaluation protocol can give drastically different results.
\end{chall}

Holistic benchmark suites, such as HELM \cite{liang2022holistic}, try to make benchmarking more robust by standardizing evaluation across all scenarios and tasks while ensuring broad coverage across as many capabilities and risks as possible. Increasingly, models are additionally being benchmarked on tests designed for humans, including the SAT, LSAT, and mathematics competition tests, to name a few.~\citet{zhong2023agieval} develop a benchmark, `AGIEval', to rigorously test the abilities of LLMs on these tests, and find that GPT-4 achieves human-level performance on several of these tests.

On traditional benchmarks, models can be quite brittle to the choice of prompt or evaluation technique for a particular benchmark question. For example,~\citet{hfmmlu2023} found that benchmark results vary significantly depending on the choice of evaluation method for the multiple choice problem-solving benchmark MMLU~\citep{mmlu}, whether it be generating text and checking if the first token matches the letter of the multiple choice answer~\citep{helm_website}, or gathering log-probabilities of each correct answer~\citep{eval-harness}. Prompt variations are also not typically normalized for, so models may be sensitive to variations such as whether or not the prompt appends `Please answer yes or no'.~\citet{jain2023bring} find that larger models and instruction-fine-tuned models are likely to be more sensitive to small variations in the prompt.

\subsection{Evaluations Based on Static, Human-Written Ground Truth}
Another challenge of LLM evaluations is that they often rely on human-written `ground truth' text. However, we often want to evaluate their performance in domains where such text is scarce or relies on expert knowledge, such as programming or mathematics tasks. As models get more capable and perform better than humans on benchmark tests in some domains, the ability to obtain comparisons to `human-level' performance diminishes.

Further, benchmark datasets become outdated over time---as models become more capable, older benchmarks become saturated or overfit and no longer provide a useful signal for further improvement \cite{dehghani2021benchmark,raji2021aibenchmark,kiela2021dynabench}. They are typically constructed around a set of tasks that were relevant at the time of creation but may not adapt well to the changing capabilities of LLMs. This means the community must continually adapt to new static benchmarks while de-emphasizing older ones or more dynamic evaluation measures, such as human evaluation of model outputs.

\begin{chall}{Reliance on Static, Human-Written Ground Truth}
    Static benchmarks become less useful over time due to changing capabilities while updating them often relies on human-written ground truth.
\end{chall}

To combat these issues, \citet{srivastava2022beyond} regularly admit new tasks to the \emph{Beyond the Imitation Game benchmark} (BIG-Bench), including programmatically evaluated tasks. Further, we highlight two separate streams of work enabling dynamic evaluations without humans in the loop. 

\paragraph{Model-generated evaluation tasks}
As LLM capabilities improve, they can increasingly generate useful benchmark questions or evaluation prompts themselves.~\citet{perez2022discovering} shows that LLMs can be used to generate static benchmark datasets for arbitrary axes, using reward models trained on human preferences to filter a generated dataset for quality. \citet{wangLargeLanguageModels2023} find that the order in which candidate examples are presented in the prompt can greatly impact the model-generated evaluation. To mitigate this issue, they propose the usage of a prompting template which encourages the model to generate assessment evidence before assigning a score and averaging scores of multiple assessments with swapped candidate positions.

\paragraph{Model-generated scores}
Aside from generating evaluation questions, models are increasingly used to directly grade the performance of other models and act as a `judge' of other models' capabilities \cite{lin2023llm, wang2023pandalm,jain2023bring}. This concept follows the motivation that while it may be challenging for a model to generate `correct' answers to prompts in many domains, it can often be easier to evaluate the correctness of an answer or to judge the relative quality between two answers \cite{zhang2019bertscore,fu2023gptscore}. However, these techniques often produce evaluation results that vary significantly depending on the `judge' model and suffer from robustness issues that make them a poor substitute for human judgment.

\subsection{Indistinguishability between Generated and Human-Written Text}
Detecting language generated by LLMs is important for various reasons; some of which include preventing (1) the spread of misinformation (\eg authoritative-sounding false narratives citing fake studies) ~\cite{zellers2019defending}, (2) plagiarism (\eg LLMs prompted to rewrite existing content in ways that bypass plagiarism detection tools) ~\cite{wahle2022large,wahle2022identifying}, (3) impersonation or identify theft (\eg by mimicking a person's writing style)
\cite{schuster2020limitations,weidinger2021ethical}, and (4) automated scams and frauds (\eg large-scale generation of phishing emails) ~\cite{weiss2019deepfake}, and (5) accidentally including inferior generated text in future models' training data 
~\cite{radfordRobustSpeechRecognition2022}. However, such detections become less trivial as the fluency of LLMs improves ~\cite{bakhtinRealFakeLearning2019}.

\begin{chall}{Detecting LLM-generated Text}
    The difficulty in classifying whether a text is LLM-generated or written by a human.
\end{chall}

There are primarily two lines of work addressing this problem: (i) \emph{post-hoc detectors}, which aim to classify arbitrary text as being LLM-generated, and (ii) \emph{watermarking} schemes, which modify the text generation procedure to make the detection easier. However, both approaches can be susceptible to \emph{paraphrase attacks}, which we discuss thirdly.

\paragraph{Post-hoc Detectors} \citet{gehrmannGLTRStatisticalDetection2019} open-source a tool that visualizes statistically improbable tokens to support humans in detecting generated text artifacts.
\citet{bakhtinRealFakeLearning2019} explore energy-based models to discriminate between real and fake text, including scenarios where the text generator was trained on a completely different dataset than the discriminator. \citet{uchenduAuthorshipAttributionNeural2020} examine three authorship attribution problems: (1) were two texts produced by the same method or not; (2) given a text, was it generated by human or machine, (3) which method generated a given text?   
\citet{mitchellDetectGPTZeroShotMachineGenerated2023} investigate whether a model can detect its own samples by posing a hypothesis: minor rewrites of generated text have lower probability under the model than the original sample, while the same cannot be said about human-written text. Generated passages tend to lie in the negative curvature regions of the model's log probability function. Their method, \emph{DetectGPT}, exploits this hypothesis by approximating that curvature given some samples.

\paragraph{Watermarking}  
~\citet{kirchenbauerWatermarkLargeLanguage2023} employ a \emph{watermark}, \ie a hidden pattern that is imperceptible to humans but algorithmically identifiable, during inference as follows: for each to be generated token, they (1) hash the previous token to seed a random number generator; (2) using that seed, they randomly partition the vocabulary into a ``green list'' and ``red'' list, and (3) sample the next token by excluding any token from the red list. In the case of low-entropy tokens, which renders it difficult to introduce changes to the vocabulary, they introduce a ``soft'' version, which promotes using the green list only for high-entropy tokens (when many plausible choices are available). In follow-up work, the same first authors ~\citet{kirchenbauerReliabilityWatermarksLarge2023} study the robustness of their watermarking scheme \emph{in the wild}, \ie after it is re-written by humans, non-watermarked LLMs, or mixed into a longer hand-written document. They conclude that watermarks remain detectable given sufficient tokens and argue that this required amount of text is a crucial yet overlooked metric.

~\citet{yangWatermarkingTextGenerated2023} study watermarking of black-box API models, where we cannot access the model's inference procedure. 
~\citet{tangBaselinesIdentifyingWatermarked2023} provide algorithms for identifying watermarks, noting that watermarked LLMs tend to produce token distributions that differ identifiably from non-watermarked models. ~\citet{christUndetectableWatermarksLanguage} introduce \emph{undetectable} watermarks, which can only be detected with the knowledge of a secret key. 

To make watermarks robust to text corruptions (we study a common type of such in the next paragraph), ~\citet{yooRobustNaturalLanguage2023} suggest placing them on ``invariant features'', which are invariant to minor modifications of the text.

\paragraph{Paraphrasing Attacks}
One way to evade machine-generated text detectors is to re-phrase the text such that the revealing LLM signatures get removed.

\begin{chall}{Paraphrasing Attacks}
 Another LLM can rewrite LLM-generated text to preserve approximately the same meaning but change the words or sentence structure.    
\end{chall}

\citet{krishnaParaphrasingEvadesDetectors2023} evade several detectors (\eg dropping DetectGPT's detection accuracy from 70.3\% to 4.6\%) by training an 11B paraphrase generation model that can paraphrase paragraphs and provides scalar knobs to control the amount of lexical diversity and reordering in the paraphrases. To defend against such attacks, they propose storing model generations in a database, from which the API provider can retrieve semantically similar texts later. Since paraphrasing does not modify the semantics of the text, the authors demonstrate that this retrieval approach is fairly robust to paraphrasing attacks.  

\citet{sadasivanCanAIGeneratedText2023} claim that the detection of generated text, even with watermarking, is not reliable; neither in practice, by performing paraphrasing attacks; nor in theory, by providing a theoretical impossibility result. They also discuss how an adversary can query watermarked LLMs multiple times to extract its watermarking scheme and spoof the watermark detector by composing human text that is then wrongly classified as model-generated.

\subsection{Tasks Not Solvable By Scale} \label{ref:unlearnable}
The ongoing advancements of LLM capabilities consistently astonish the research community, for instance, by achieving high performances on the MMLU~\citep{mmlu} benchmark much sooner than competitive human forecasters had anticipated~\citep{flan}. Similarly, within less than a year, OpenAI released GPT-3.5 and GPT-4, where the latter significantly outperformed the former on various tasks~\citep{openai2023gpt4}. 

Given this progress, one may question whether there are limits we deem impossible to overcome within the current paradigm of scaling data/model sizes of autoregressive Transformer-based LLMs. We emphasize that such tasks' (permanent) existence is still somewhat speculative. Here, we explore possible patterns behind such tasks instead of discussing specific ones (which we do in \Cref{sec:evaluation} and \Cref{sec:app}).

\begin{chall}{Tasks Not Solvable By Scale}
    Tasks \emph{seemingly} not solvable by further data/model scaling.
\end{chall}

\paragraph{Inverse Scaling} (IS) is the phenomenon of task performance worsening as model scale and training loss performance increases. 
\citet{lin2021truthfulqa} first stumbled upon this property when evaluating models of increasing sizes (\eg GPT-2, GPT-3) on their benchmark that measures whether an LLM is truthful in generating answers to questions. They conjecture that common training objectives incentive false answers (which they call \emph{imitative falsehoods}) if they have a high likelihood on the training distribution (we discuss dataset issues in \Cref{sec:datasets}). 
\citet{mckenzieInverseScalingWhen2023} collect 11 datasets that exhibit IS behavior and identify four potential causes for such: (1) models regurgitating memorized data rather than following in-context instructions, (2) imitation of undesirable patterns in the training data, (3) models learning to perform easier, so-called \emph{``distractor task''} rather than the intended ones, and (4) spurious correlations in the given few-shot examples.

\citet{wei2022inverse} somewhat challenge the existence of inverse scaling by evaluating the tasks proposed by \citet{mckenzieInverseScalingWhen2023} on even larger models; up to trained on five times more compute. In this increased compute region, four out of eleven tasks remain inverse scaling; six out of eleven exhibit \emph{``U-shaped scaling''}, where the performance first decreases up to a certain size and then increases again. The authors hypothesize that U-shaped scaling occurs when a task contains a distractor task, which larger models can learn to ignore.
Similarly, in the case of quantifier comprehension tasks, \citet{guptaProbingQuantifierComprehension2023} argue that previously observed inverse scaling behavior might have been due to inappropriate testing methodology.

\paragraph{Compositional tasks} composed of multiple sub-problems are an ideal outlet to investigate whether models go beyond rote memorization of observed facts and deduce novel knowledge \cite{pressMeasuringNarrowingCompositionality2023}. 
\citet{zhangParadoxLearningReason2022a} investigate whether language models can learn deductive reason from data by introducing a class of propositional logic problems. The authors prove that the model has enough capacity to solve the task, yet, it instead learns to rely on statistical features rather than emulating the correct reasoning function. 
\citet{pressMeasuringNarrowingCompositionality2023} measure how often a model can correctly answer all sub-problems but not generate the overall solution, a ratio they refer to as \emph{compositionality gap}. They find that increasing the model size in the GPT-3 family of models improves solving sub-problems faster than composed problems, suggesting that larger models show no improvement for this gap.
\citet{dziriFaithFateLimits2023} find that systematic problem-solving capabilities do not emerge from maximum likelihood training of Transformer models in general. They base this claim on two hypotheses: (i) Transformers reduce compositional tasks into linearized path matching, a form of shortcut learning \cite{geirhos2020shortcut} that does not generalize robustly; and (ii) errors in the early stages of the task (\ie when sub-problems follow some order) compound substantially. 
\citet{asherLimitsLearningLanguage2023} prove that LLMs cannot learn semantic entailment or consistency as defined in formal semantics \cite{dowty2012introduction} due to a lacking understanding of universal quantifiers (\eg \emph{every, some, many, most, etc.}).

\paragraph{Memorization vs. Generalization} An ongoing debate evolves around the question of to what degree LLMs memorize instead of generalize (and what exactly the difference is \cite{balestriero2021learning}). Memorization has been shown to (1) hurt (certain) downstream task performances \cite{lee2021deduplicating}, (2) increase with the model size \cite{extracting_data,kim2022provable,tirumalaMemorizationOverfittingAnalyzing,mahdaviMemorizationCapacityMultiHead2023}, and (3) emerge unpredictably from smaller or partially-trained models \cite{bidermanEmergentPredictableMemorization2023a}. Hence, we wonder whether some tasks do not benefit from further model/dataset size scaling.

One such class of tasks might be \emph{counterfactual tasks} \cite{wuReasoningRecitingExploring2023}, \ie tasks on which LLMs initially perform well modified such that specific input-output conditions are changed while the general reasoning procedure remains the same. For example, for an arithmetic task, the counterfactual variant would alter the base from 10 to 2. \citet{wuReasoningRecitingExploring2023} find that LLMs perform poorer the less common the counterfactual conditions are, which they call a \emph{``memorization-like effect''}. An interesting future direction would be to explore whether increasing model size exacerbates performance due to more memorization or actually improves because scaling-law-optimal pre-training recipes would dictate scaling the dataset proportionally (\Cref{sec:efficient_training}), which then may include more of such tasks with uncommon conditions.

\subsection{Lacking Experimental Designs}
\input{tables/all_llms.tex}
\Cref{table:all_llms} shows a (non-exhaustive) overview of selected LLMs within the scope of this review, described in academic papers. Many works do not include controlled ablations, which is especially problematic due to their large design space. We posit that this impedes scientific comprehension and advancement.

\paragraph{Lack of Controlled Ablations}
We observe that many papers do not run controlled experiments (\emph{ablations}) by varying one factor at a time, likely due to the prohibitive computational cost. For example, \citet{chowdhery2022palm} conjecture PaLM might outperform GPT-3 and other LLMs on many tasks due to higher training corpus quality, but note they ``do not perform the necessary ablation studies to say this conclusively'' and instead solely focus on model depth and width. Many papers from \Cref{table:all_llms} adopt hyper-parameters from previous works \cite{sanyal2023understanding} and do not tune them after introducing a change in the training pipeline. Sometimes, important implementation details are not mentioned, \eg when optimizer states are reset during training~\citep{chung2023details}.

\begin{chall}{Uncontrolled Experiments}
    Papers presenting novel LLMs often lack controlled experiments, likely due to the prohibitive costs of training enough models. 
\end{chall}

An easy yet expensive fix is to run ablations by varying one factor at a time, \eg keeping most hyper-parameters fixed except the model size \cite{biderman2023pythia} or context lengths \cite{touvronLlamaOpenFoundationa}.
A cheaper potential remedy can be \emph{zero-shot hyper-parameter transfer} from smaller models to larger ones~\citep{winkelmolen2020practical,yang2021tuning}.~\citet{yang2021tuning} find that when using the $\mu P$ network parameterization scheme, one can transfer the effect of changing hyper-parameters such as the learning rate across varying model depths, batch sizes, sequence lengths, and training times, which they verify empirically up to a 6.7B model. However, it has yet to be verified if such transferability still holds for other varying factors; and if so, researchers could afford to conduct more ablation experiments via smaller models. 

If additional experiments are prohibitively expensive, another recommendation is to report evaluation results beyond aggregated performance measures. For example, in reinforcement learning, recent work has argued that providing entire performance distributions across all runs is less biased and more robust to outliers than point estimates \cite{agarwalDeepReinforcementLearning2021}.

\paragraph{Curse of Dimensionality}
In \Cref{table:all_llms}, we highlight some but not all differences across models, as the table format constrained us. Other common differences include the training datasets or fine-grained architectural details, \eg the usage of multi-head \cite{transformers} or multi-query attention \cite{shazeer2019fast}.

We note that a core characteristic of LLMs is their vast design space, which renders scientific inquiry challenging \cite{ioannidisWhyMostPublished2005}. For example, by taking into account the (i) data sources and their proportions within the pre-training dataset, (ii) choice and training hyper-parameters of the tokenizer, and (iii) pre-training objective, the combined design space quickly becomes high-dimensional. Undertaking factorial experiments within such expansive design spaces results in a combinatorially-growing number of single training runs, and the lack of sufficient experimental coverage can severely inhibit scientific understanding of what makes an LLM perform well. While this issue is not unique to LLMs, they tend to be larger in the number of parameters---and therefore compute requirements, feedback loop times, and training costs---than models in most other fields.

\begin{chall}{Curse of (Design) Dimensionality}
    Common design spaces of LLM experiments are high-dimensional.
\end{chall}

One possible way forward is to encourage the community to use techniques like Bayesian optimization (BO) with dimensionality reduction~\citep{wang2013bayesian,moriconi2020high}, where we use a non-linear feature mapping to map the input (the hyper-parameter configuration) onto a lower dimensional manifold followed by a BO procedure to optimize the underlying blackbox function (the LLM with respect to the hyper-parameters). Another suitable tool to explore the design space efficiently can be treatment effect estimation \cite{kunzel2019metalearners,nie2021quasi}, \eg where the treatment is a vector describing certain ablations \cite{sin}.

\subsection{Lack of Reproducibility} \label{sec:reproducibility}
The reproducibility of empirical results is important to verify scientific claims and rule out errors in experimental protocols leading to such. When researchers try to build upon non-reproducible results, they might waste resources. 

Unfortunately, we stumble upon two unique reproducibility issues in LLM research: repeatability of (i) training runs and (ii) generations by close-sourced API-served models. While the term ``reproducibility'' is often used more broadly and can slightly vary in its meaning \cite{Reproducibility2023}, in the following, we focus on ``repeatability'', which we define as the ability to repeat experimental outcomes exactly. 

\paragraph{Training Repeatability} Typical training protocols of LLMs involve parallelism across multiple compute nodes. The scheduling and communication strategies between nodes can be non-deterministic~\citep{niuHogwildLockFreeApproach}. This variability can affect the final result, especially in algorithms that are not ``order-invariant'', such as stochastic gradient descent (SGD). Some sources of randomness are (i) lock-free parallelism schemes~\citep{niuHogwildLockFreeApproach}, (ii) floating point precision, \eg when summing gradients across devices, the order in which these sums are computed can affect the final result~\citep{goldbergWhatEveryComputer1991}, (iii) non-deterministic, performance-optimized operations, which are much faster and therefore desirable~\citep{ReproducibilityPyTorchDocumentation}.

Further, ~\citet{carliniPoisoningWebScaleTraining2023} point out that some pre-training datasets consist of an index of web content that individual users must crawl themselves, rather than using static, standalone dumps. This is due to monetary, privacy, and legal restrictions. As a result, reproducibility can be easily compromised if any of the sources in the index have changed between the time the dataset curator collected them and the time the end-user downloads them.

\begin{chall}{Irrepeatable Training Runs}
    Parallelism strategies designed to distribute the training process across many accelerators are typically non-deterministic, rendering LLM training irreproducible. 
\end{chall}

\paragraph{Inference Repeatability}
Another peculiarity of commercial LLMs is that they are typically served via stochastic API in a black-box setting, which comes with the following challenges: (i) the provider retains complete authority over the model and can introduce unpublicized changes, including retraining the model, modifying its parameters, or completely replacing it; (ii) even if model updates are communicated, there is still uncertainty about whether access to specific model versions will be maintained once they are deemed outdated, (iii) even with a decoding temperature set to zero, API models often produce stochastic outputs~\citep{ofirpress[@ofirpress]GPT3SeemsBe2022,ruis2022large,rileygoodside[@goodside]EdgecaseGPT3Big2022}.

\citet{chenHowChatGPTBehavior2023} provide preliminary evidence confirming dramatic changes in API-served models. They find that GPT-3.5 and GPT-4 performances on four diverse tasks vary vastly within three months (March to June 2023). For example, GPT-4's accuracy in identifying prime numbers was 97.6\%, but in June, its accuracy dropped to 2.4\%; while for GPT-3.5, the trend is reversed and it got much better over time.

\begin{chall}{Irreproducible API Inference}
    API-served models are often irreproducible. 
\end{chall}

An easy fix is to rely exclusively on open-source LLMs \cite{OpenLLMLeaderboard}. 

\section{Applications}
\label{sec:app}
In this section, we aim to provide practitioners with a broad overview of the areas in which LLMs are currently being applied and highlight some common application architectures across domains.

Analogous to the Challenges section, we highlight the key constraints in each application area as follows.
\begin{constraint}{Constraint}
    This box highlights a constraint.
\end{constraint}

\input{tables/applications.tex}

\subsection{Chatbots} \label{sec:chatbots}
General-purpose chatbots (dialogue agents) combine the tasks of information retrieval, multi-turn interaction, and text generation (including code). 

\citet{lamda} introduced the LaMDA family of chatbot LLMs with up to 137B parameters, focusing on safety (via supervised fine-tuning on human annotations) and factual grounding (via access to external knowledge sources). Notably, smaller LaMDA models (2B parameters) with fine-tuning are shown to perform similarly on dialogue quality and safety/grounding scores to the larger LaMDA models (137B parameters) without fine-tuning. LaMDA models were released as part of the Bard chatbot service~\citep{bard}. However, the latest version of Bard now uses the PaLM 2 LLM~\citep{anil2023palm, bard_palm_2}.

\citet{sparrow} propose Sparrow, a chatbot based on a 70B parameter Chinchilla LLM, and use RLHF (\Cref{sec:alignment}) targeting 23 rules to fine-tune the model to be more {\it helpful}, {\it correct}, and {\it harmless}. Sparrow also incorporates external knowledge using a retrieval model to provide evidence from a Google Search query. The RLHF approach outperforms the only dialogue-prompted and supervised fine-tuned approaches regarding output preference and rule violation rate.

Similarly,~\citet{chatgpt} train the ChatGPT chatbot using supervised fine-tuning and RLHF (\Cref{sec:alignment}) to specialize a GPT-3.5 LLM for dialogue. GPT-4~\citep{openai2023gpt4} is the underlying model for the ChatGPT Plus chatbot, but training and architecture details have not been released.

\citet{blenderbot3} introduce BlenderBot-3, a 175B parameter chatbot based on the OPT-175 LLM using supervised fine-tuning. BlenderBot-3 incorporates external knowledge through modules that conduct internet searches and retrieve text-based long-term memories generated from previous outputs to help performance over long interactions.

\begin{constraint}{Maintaining Coherence}
    Multi-turn interactions make Chatbots easily ``forget'' earlier parts of the conversation or repeat themselves \cite{borjiCategoricalArchiveChatGPT2023,rayChatGPTComprehensiveReview2023}.
\end{constraint}

\citet{kopf2023openassistant} release the OpenAssistant Conversations dataset of human-annotated interactions and use this to instruction fine-tune Pythia and LLaMA models (up to 30B parameters) for chatbot applications. To help align the final models, the dataset is generated with guidelines to make the responses \emph{polite}, \emph{helpful}, \emph{concise}, \emph{friendly}, and \emph{safety-aware}. The LLaMA 30B version is currently used within the HuggingChat chatbot application~\citep{hugging_chat}.

A key challenge of fine-tuning chatbots is creating a broad training dataset of high-quality conversations. To address this problem~\citet{chen2023places} demonstrate using existing LLMs (OPT 30B) to generate high-quality synthetic conversation datasets based on a small number of expert-written examples. Human crowd workers assessed the generated conversations to be comparable to existing human-generated datasets on the metrics: \emph{interesting}, \emph{coherent}, \emph{natural}, and \emph{consistent}.~\citet{chen2023places} show the synthetic dataset can be used to fine-tune a chatbot (BlenderBot 400M) and achieve performance only slightly below fine-tuning with human-generated datasets.

Chatbots' intended generality also makes evaluating their capabilities' full range difficult.~\citet{chatgpt_jack_of_all_trades} evaluate ChatGPT (GPT-3.5) on 25 tasks with 38k prompts covering a diverse set of capabilities, including but not limited to question answering, emotion recognition, offensive language detection, spam detection, inference, and sentiment analysis. While ChatGPT is shown to have strong performance across the 25 tasks, it usually underperforms the SOTA in that domain. More recently, ~\citet{bubeck2023sparks} and~\citet{openai2023gpt4} investigate the capabilities of GPT-4 (base model of ChatGPT Plus) across a wide range of tasks, including interactions with humans and tools. Using these evaluations~\citet{bubeck2023sparks} conclude that GPT-4 is `strikingly close to human-level performance' across tasks. 

Finally, the challenge of inference latency (\Cref{sec:inference_costs}) is also potentially going to become an important constraint~\cite{yang2023harnessing} for chatbot applications as LLMs scale. There is a trade-off between the need for responsive live user interaction in a conversational format and utilizing larger LLMs~\citep{openaiforumgpt4}.

\begin{constraint}{High Inference Latency}
 High inference latency (\Cref{sec:inference_costs}) hinders the user experience~\citep{openaiforumgpt4}, especially in multi-turn interaction with chatbots.
\end{constraint}

\subsection{Computational Biology} \label{sec:computational_biology}
In computational biology, we are interested in non-text data representing similar sequence modeling and prediction challenges.

\subsubsection{Protein Embeddings}
One popular application of LLM-like models in biology is to generate protein embeddings from amino-acid or genomic sequence inputs. These embeddings can then be used as inputs for structure prediction, novel sequence generation, and protein classification tasks. Protein language models perform strongly on many academic datasets, but their applicability to downstream tasks such as drug design is often unclear~\citep{dauparas2022proteinmpnn}.

\begin{constraint}{Transfer to Downstream Applications}
    The ultimate objective of protein language models is to deploy them in real-world projects such as drug design. Evaluations often target smaller and/or specialized datasets, not considering how the models could contribute to protein design in~vitro or in~vivo.
\end{constraint}

\citet{elnaggar2020prottrans} train a range of LLM architectures to extract embeddings from protein amino acid sequences. These embeddings are then used as inputs on supervised per-amino acid and per-protein prediction tasks. The best-performing LLM architecture (ProtT5) achieved SOTA results on per-amino acid protein secondary structure prediction without using evolutionary information. Similarly,~\citet{wu2022tfold} predict antibody backbone and side-chain conformations.

\citet{lin2022language} take a similar approach to training a protein LLM, the Evolutionary Scale Model Transformer-2 (ESM-2), on protein amino acid sequences from the UniRef database using a masked language modeling approach. They show significant performance increases as the model is scaled from 8 million to 15B parameters, with the largest models outperforming the ProtT5 on protein structure prediction benchmarks (CASP14, CAMEO)~\citep{kinch2021casp, robin2021cameo}. They also introduce ESMFold, which uses the ESM-2 embedding model for end-to-end atomic resolution prediction from a single sequence. While ESMFold underperforms the SOTA AlphaFold2~\citep{alphafold2} on the CAMEO and CASP14 benchmarks, the authors note that by relying only on embeddings ESMFold has an order of magnitude faster inference time than AlphaFold2, using just the protein sequence of interest rather than structural templates and multiple sequence alignments (MSAs).~\citet{jeliazkov2023hallucinate} find that protein sequences designed by an inverted AlphaFold2 model are unlikely to be expressed, but sequences generated using an inverted protein LLM such as ESMFold were more likely to be expressed.

Researchers have also adopted the ESM-1 and ESM-2 models to generate protein embeddings for enzyme-substrate chemical structural class prediction~\citep{jinich2022predicting}, training 3D geometric graph neural networks for proteins~\citep{wu2023geometric}, identifying disease-causing mutations~\citep{liu2022protein}, designing novel proteins~\citep{verkuil2022language}, and guided evolution of antibodies for affinity maturation~\cite{hie2023efficient}.

\citet{xTrimoPGLM} propose training a new model xTrimoPGLM (100B parameters) simultaneously for protein embedding and generation tasks using MLM and generative objectives. The xTrimoPGLM-100B model (with fine-tuning where relevant) outperforms existing approaches on 13 out of 15 evaluated tasks.  

Protein embedding models with alternative inputs have also been proposed.~\citet{outeiral2022codon} train an 86 million parameter protein LLM CaLM (Codon adaptation Language Model) using sequences of codons (nucleotide triads) as input instead of amino acids due to codons containing potentially richer information.~\citet{madani2023large} train a 1.2B parameter protein embedding model ProGen on 280 million protein amino acid sequences with additional \emph{control tags} specifying protein properties. ProGen is then fine-tuned using data from specific protein families and applied to generate functional full-length amino acid sequences. Similarly,~\citet{xu2023protst} propose training a protein language model, the ProtST, on protein sequences and additional text descriptions of their key properties for protein classification and retrieval tasks.

Finally, for antibodies specifically,~\citet{shuai2021generative} propose an Immunoglobulin Language Model (IgLM) using the GPT-2 architecture (with 13 million parameters) for the generation of immunoglobulin sequences, using a masked language modeling approach. Similar to~\citet{xu2023protst}, the IgLM model also takes additional conditioning tags corresponding to chain type and species as input. The authors show the IgLM model can then be used for the controllable generation of infilled and full-length antibody sequences. 

\subsubsection{Genomic Analysis}
LLMs in the field of genomic analysis enable a better understanding of the effects of mutations in humans and predict genomic features directly from DNA sequences. While genomic language models are a promising research direction, current models cannot process many genomic sequences as their sequence lengths commonly exceed multiple billions of nucleotides~\citep{nurk2022genome}.

\begin{constraint}{Limited Context Window}
    The largest genomes have vastly longer DNA sequences~\citep{nurk2022genome} than existing genomic LLMs' context windows can handle, constraining the types of genomes that can be successfully modeled using these approaches. 
\end{constraint}

\citet{zvyagin2022genslms} introduce a range of hierarchical LLMs (up to 25B parameters) with long input sequences (2048 - 10,240 tokens), referred to as Genome-scale Language Models (GenSLMs). The GenSLM models are pre-trained on Prokaryotic gene sequences from the BV-BRC dataset using codon tokenization~\citep{outeiral2022codon} and then fine-tuned on SARS-CoV-2 genome sequences for the task of identifying potential new variants and generative modeling. However, the authors note that it remains unclear whether the GenSLM architecture generates richer representations than the protein LLM approaches.

\citet{dalla2023nucleotide} train Nucleotide Transformers with 500 million to 2.5B parameters on nucleotide sequences from human and other species genomes, using a masked language modeling approach. The Nucleotide Transformers were evaluated on 18 genomic prediction tasks with fine-tuned larger models achieving the best results.

\citet{nguyen2023hyenadna} propose HyenaDNA, a genomic language model based on the Hyena architecture \citep{poliHyenaHierarchyLarger2023a}, enabling modeling of genomic sequences of up to 1 million tokens. HyenaDNA outperforms Transformer-based models with multiple orders of magnitude more parameters while incorporating the in-context learning capabilities of LLMs into the genomics domain.

\subsection{Computer Programming} \label{sec:computer_programming}
One of LLMs' most advanced and broadly adopted applications is generating and completing computer programs in various programming languages. This section deals with programming-specific LLMs where the model is fine-tuned or pre-trained exclusively for programming applications, but it is important to note the increasing use of general chatbots partially trained on code datasets (such as ChatGPT) for programming tasks. 

\subsubsection{Code Generation}\label{code_gen_sec} 

Code generation refers to using an LLM to output new code for a given specification or problem provided as a prompt. Several computer programming-specific LLMs and approaches have been proposed.

For Python code generation, \citet{codex} introduce Codex, a fine-tuned GPT-3 LLM (up to 12B parameters) specialized to generate stand-alone Python functions from doc strings. Fine-tuning was conducted using a raw dataset of 159 GB of Python source code from GitHub and a filtered dataset of correctly implemented standalone Python functions. Codex models outperformed similarly sized GPT-3 and GPT-J models on the HumanEval evaluation set, with the Codex model trained on the filtered dataset (Codex-S) achieving the best results. Importantly,~\citet{codex} note that there was no observed improvement from using a pre-trained GPT-3 model as a base other than faster convergence.

\citet{chen2023teaching} seek to improve the performance of Codex through a \emph{self-debugging} prompting approach. Three forms of \emph{self-debugging} are investigated. \emph{Simple} feedback prompts the model to decide whether the generated code solution is correct. \emph{Unit-test} feedback prompts the model with the output of unit tests provided in the problem description. \emph{Code explanation} feedback prompts the model to explain the solution in detail and use the explanation to correct the solution. In each case, this process is repeated iteratively until the model provides a solution it states is correct or a maximum number of attempts has been made. Codex using the \emph{self-debugging} prompting framework with code explanation (and unit-testing if applicable) outperforms the base Codex model on C++-to-Python translation, text-to-SQL generation, and text-to-Python generation.

~\citet{gunasekar2023textbooks} train a smaller model Phi-1 (1.3B parameters) to generate Python functions from doc strings. Training phi-1 using a combination of filtered existing datasets and new synthetic \emph{textbook} and \emph{exercise} datasets results in a model that can achieve near current SOTA results on HumanEval while having over an order of magnitude fewer parameters and tokens than previous works.

Another area of interest has been the development of multilingual programming LLMs.~\citet{code_evaluation} evaluate a range of code generation LLMs and train a new multilingual LLM Polycoder (2.7B parameters) using source code from 12 languages. However, for Python specifically, Codex outperforms Polycoder and other existing models (GPT-J, GPT-Neo, and CodeParrot) on HumanEval. 

\citet{codegen} train the CodeGen family of LLMs (up to 16B parameters) using a combination of three datasets: natural language, multilingual programming source code (C, C++, Go, Java, JavaScript, and Python), and a monolingual Python dataset. The largest CodeGen model using the monolingual training set was shown to outperform the Codex-12B model.~\citet{codegen} also test CodeGen on multi-step program synthesis, where a program is broken down into multi-step natural language prompts, which the model then implements individually (creating the new Multi-Turn Programming Benchmark (MTPB)).

Finally, \citet{li2022competition} focus on the task of solving competitive programming questions (Codeforces, Description2Code, and CodeNet). The AlphaCode LLM (up to 41B parameters) is first pre-trained on a multilingual dataset (C++, C\#, Go, Java, JavaScript, Lua, PHP, Python, Ruby, Rust, Scala, and TypeScript) of 715 GB of source code from GitHub. It is then fine-tuned using a new curated dataset of competitive programming problems called CodeContests. To achieve high performance,~\citet{li2022competition} use large-scale sampling (up to millions of samples), filtering, and clustering of candidate solutions generated by AlphaCode to select the final submissions.

However, whilst these existing code-generation LLMs have achieved impressive results, a critical current constraint in applying LLMs to code generation is the inability to fit the full code base and dependencies within the context window. To deal with this constraint, a few frameworks have been proposed to retrieve relevant information or abstract the relevant information into an API definition.

\begin{constraint}{Long-Range Dependencies \citep{zhang2023repocoder, shrivastava2022repository}}
Long-range dependencies across a code repository usually cannot be regarded because of limited context lengths (\Cref{sec:context_length}). 
\end{constraint}

\citet{zhang2023repocoder} introduce RepoCoder, a retrieval-based framework for repository-level code completion that allows an LLM to consider the broader context of the repository. A multi-step \emph{retrieval-augmented generation} approach is taken, where the initial code generated is then used to retrieve further, potentially more relevant, repository code snippets to refine the final output. This approach can be considered a retrieval-based method for relieving the long-range dependency constraint.

Similarly,~\citet{shrivastava2022repository} propose the Repo-Level Prompt Generator (RLPG) framework to dynamically retrieve relevant repository context and construct the correct prompt for a given completion task. To do this, many \emph{prompt proposals} are generated from different \emph{prompt sources} (\eg parent class) and \emph{prompt contexts} (\eg method names). The best prompt is then selected by a \emph{prompt proposal classifier} and combined with the default context to generate the final output.

\begin{figure}[htb]
    \centering
\includegraphics[width=\columnwidth]{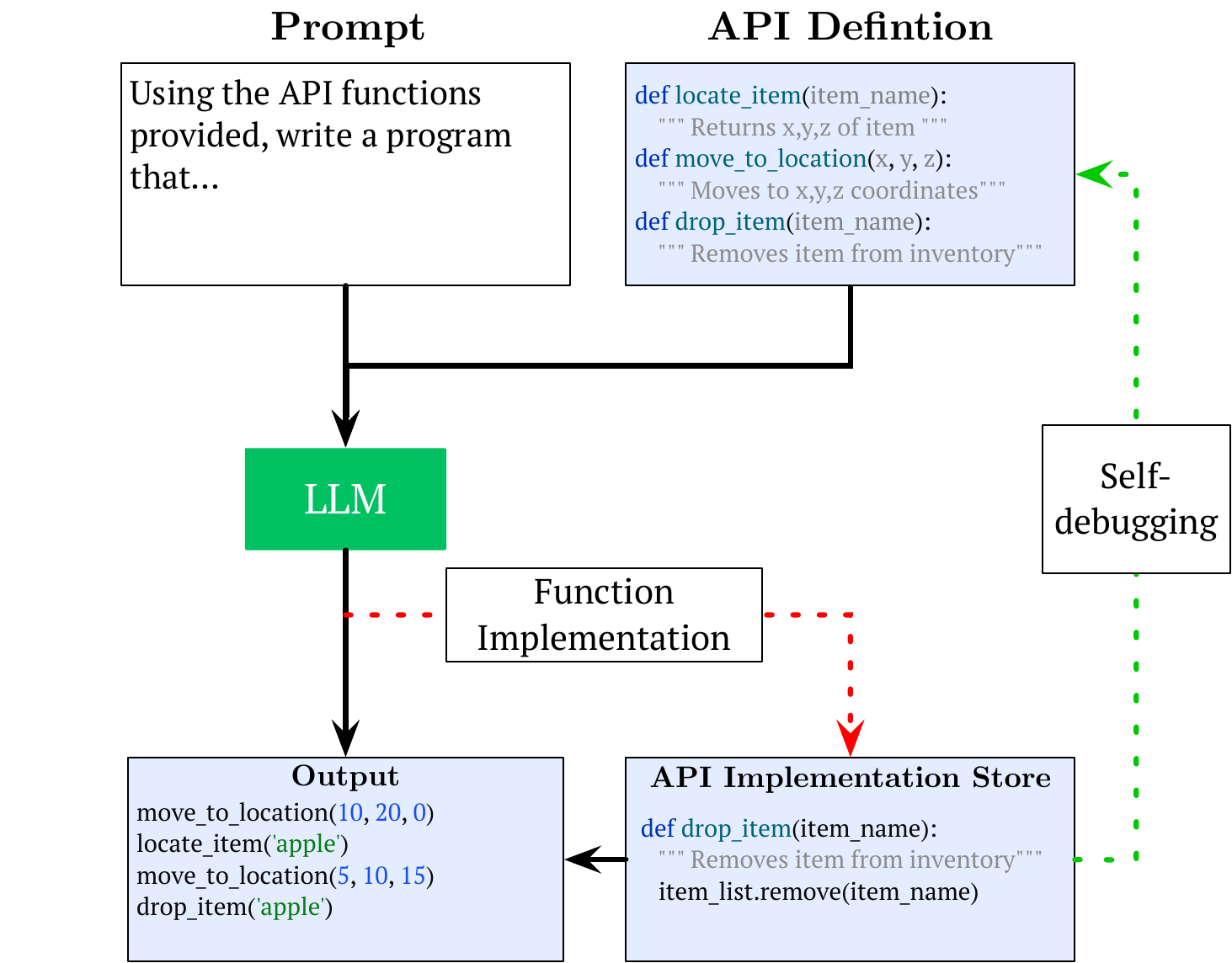}
    \caption{\textbf{API Definition Framework}. Illustration of providing a general API definition in the prompt~\citep{suris2023vipergpt, wang2023voyager, chatgpt_robo} to enable the consistent use of either external code or tools to solve the specific task whilst minimizing the required context window. Extensions to this approach have included asking the LLM to implement the functions within the API definition (\textcolor{red}{red}) and to prompt the LLM to self-debug any API code that does not execute (\textcolor{PlotGreen}{green}).}
    \label{fig:api_pipeline}
\end{figure}

Finally,~\citet{suris2023vipergpt} create the ViperGPT framework, which utilizes the Codex LLM to generate programs that answer text-based visual queries. The Codex model is prompted with the query text and an API specification to do this. The human-generated API specification provides functions designed to deal with low-level visual tasks (\eg find({\it object})) that the LLM can then use to generate solution code. This approach significantly reduces the tokens needed to provide repository/code context by only providing the API definition. This \emph{API definition} approach, illustrated in \ref{fig:api_pipeline} has been used in robotics by~\citet{chatgpt_robo}, and by \citet{wang2023voyager} as part of a Minecraft agent. Previously, \citet{gupta2022visual} used a pre-defined function approach within VISPROG, which uses GPT-3, external python \emph{modules}, and few-shot prompting with example programs to solve visual tasks.   

\subsubsection{Code Infilling and Generation}
Code infilling refers to modifying or completing existing code snippets based on the code context and instructions provided as a prompt.

\citet{incoder} train the InCoder LLM (up to 6.7B parameters) to both generate Python code and infill existing code using a masked language modeling approach. Incoder is trained using 159 GB of text split roughly equally between Python source code, StackOverflow content, and source code in other languages. On the HumanEval generation benchmark, InCoder underperforms the best-performing Codex and CodeGen models. However, unlike the other models, InCoder can perform single and multi-line infilling of existing code. 

Similarly,~\citet{santacoder} train a set of smaller SantaCoder models (1.1B parameters) for code generation and code infilling using 268 GB of Python, JavaScript, and Java source code. SantaCoder is primarily evaluated on the MultiPL-E benchmark (an extension of HumanEval and MBPP~\citep{austin2021program} benchmarks), with it shown to outperform InCoder on both HumanEval generation and infilling (passing over 100 attempts). 

Code infilling is particularly relevant for applications involving modifying, reviewing, or debugging existing code. ~\citet{didact} explore the data from the intermediary steps in the development process to help automatically resolve reviewer comments~\citep{code_review_ml}. The Dynamic Integrated Developer ACTivity (DIDACT) methodology formalizes tasks in the software development process (\eg repairing builds, predicting reviewer comments, etc.) into \emph{state}, \emph{intent}, and \emph{action} components, and trains the model to predict code modifications. This approach aims to train the model to understand the \emph{process} of software development rather than only the end product.

\subsection{Creative Work} \label{sec:creative_work}
For creative tasks, LLMs have primarily been applied to story and script generation.

For long-form story generation,~\citet{mirowski2022co} propose using a 70B Chinchilla-optimal~\citep{chinchilla} LLM Dramatron with prompting, prompt chaining, and hierarchical generation to create complete scripts and screenplays without the requirement for a human-in-the-loop (although co-writing is facilitated). The ability of Dramatron to help create a script was evaluated qualitatively through co-writing and follow-up interviews with 15 industry experts. 

Similarly,~\citet{yang2022re3} propose using GPT-3 with a Recursive Reprompting and Revision framework (Re3) to generate stories over 2,000 words long. The Re3 approach uses zero-shot prompting with GPT-3 to generate a plan (settings, characters, outline, etc.). It then recursively prompts GPT-3 to generate story continuations using a specified dynamic prompting procedure. Possible story continuations are then ranked for coherence and relevance using separate fine-tuned Longformer models as part of a \emph{Rewrite} module. Finally, local edits to the selected continuations are made by detecting factual inconsistencies using the combination of a GPT-3 model~\citep{ouyang2022gptinstruct} and a BART model~\citep{lewis-etal-2020-bart} as part of an \emph{Edit} module. This process can then be iterated for fully automated story generation. 

Finally,~\citet{yang2022doc} introduce the Detailed Outline Control (DOC) framework to maintain plot coherence over thousands of words using GPT-3. While DOC uses the same high-level \emph{planning-drafting-revision} approach as Re3, it implements this through the use of a \emph{detailed outliner} and \emph{detailed controller}. The \emph{detailed outliner} first breaks down the high-level outline into subsections using a breadth-first approach, with candidate generations for the subsections created, filtered, and ranked. The bodies of the detailed outline subsections are then generated iteratively using a structured prompting approach. During the generation, an OPT-based FUDGE~\citep{yang2021fudge} \emph{detailed controller} is used to help maintain relevance. 

\begin{figure}[t!]
    \centering
\includegraphics[width=0.99\columnwidth]{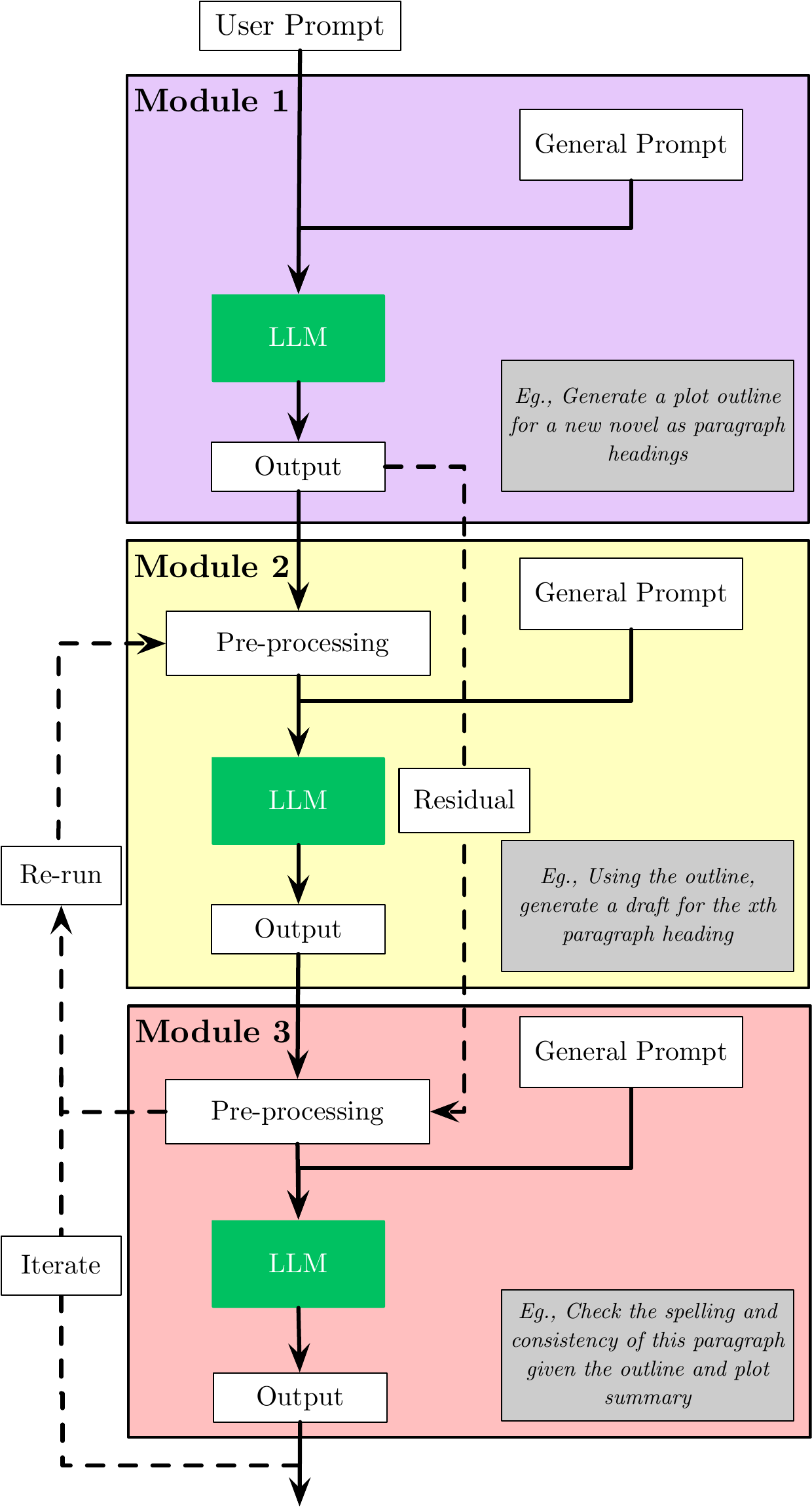}
    \caption{\textbf{Modular Prompting}. Illustration of using a series of separate prompts~\citep{mirowski2022co, yang2022re3, mirowski2022co, wang2023voyager, wang2023reelframer} and processing steps to enable an LLM to perform tasks that would either not fit in a single context window or could not easily be specified in a single prompting step.}
    \label{fig:modular_prompting}
\end{figure}

In each case, to apply LLMs to long-form story generation, the task is broken down into a series of short-form sub-tasks (\ref{fig:modular_prompting}). The current capabilities of LLMs primarily drive this approach, but also the desire to have a human-in-the-loop for some co-writing use cases~\cite{mirowski2022co}.

\begin{constraint}{Limited Context Window~\citep{mirowski2022co, yang2022re3}}
The inability of current LLMs to keep the entire generated work within the context window currently constrains their long-form applications and generates the need for modular prompting (\ref{fig:modular_prompting}).
\end{constraint}

For short form generation,~\citet{chakrabarty2022help} propose CoPoet (fine-tuned T5 and T0 models) for collaborative poetry generation,~\citet{razumovskaia2022little} use PaLM and prompting with plans for cross-lingual short story generation,~\citet{wang2023reelframer} use GPT-4 as part of the ReelFramer tool to help co-create news reels for social media, ~\citet{ippolito2022creative} use LaMDA as part of the Wordcraft creative writing assistant, and~\citet{calderwoodspinning} apply a fine-tuned GPT-3 model as part of their Spindle tool for helping generate choice-based interactive fiction. 

For more general creative tasks, ~\citet{haase2023artificial} assess a range of LLMs (including ChatGPT) on their capacity for idea generation (\emph{everyday creativity}) using the Alternative Uses Test (generating alternative uses for given items). On this task, LLMs were found to perform comparably to 100 human participants.

Finally, for visual creative tasks, LLMs have also been used to increase the level of control users have when using image generation models.~\citet{fengLayoutGPTCompositionalVisual2023} propose the LayoutGPT method where an LLM (GPT-3.5, GPT-4 or Codex) is used to generate a CSS Structure layout the image should follow based on a text-based user prompt. This layout can be visualized and used as input to guide an image generation model. This approach performs strongly on text-to-image generation and indoor scene synthesis. A similar concept is implemented by \citet{lian2023llmgrounded}, where an LLM (GPT-3.5) is used to generate natural language layouts (bounding boxes and descriptions) to guide a diffusion model. Using an LLM as part of a \emph{modality conversion} framework \ref{fig:modality_conversion} has also been used in robotics \cite{liu2023reflect, huang2022inner} and knowledge work \cite{liu2022deplot}.

\subsection{Knowledge Work} \label{sec:knowledge_work}
With researchers increasingly demonstrating LLMs' ability to perform well on domain-specific knowledge tasks such as within Law~\citep{katz2023gpt} or Medicine~\citep{medpalm2}, interest has grown in LLMs' capacity for wider \emph{knowledge work}. These applications are likely to be found across the labor market with~\citet{eloundou2023gpts} estimating that 80\% of the US workforce is in roles where at least 10\% of tasks could be affected by LLMs.

In the professional services field,~\citet{bommarito2023gpt} evaluate GPT-3.5 and previous GPT versions on actual and synthetic questions from the Uniform CPA Examination Regulation section and AICPA Blueprints for legal, financial, accounting, technology, and ethical tasks. Using only zero-shot prompting, the best performing model (latest GPT-3.5) struggles with quantitative reasoning, achieving results similar to random guessing on multiple-choice questions. However, on qualitative sections, GPT-3.5 achieved ~50-70\% accuracy, significantly ahead of random guessing and approaching human-level scores.

\begin{constraint}{Numerical Reasoning~\citep{qian2022limitations, bommarito2023gpt}}
LLMs have generally seen worse performance on quantitative tasks, potentially constraining their applications in knowledge work areas such as financial services or accounting. 

\end{constraint}

\citet{wu2023bloomberggpt} train BloombergGPT (50B parameters) for various financial knowledge work, including sentiment analysis, classification, NER/NED, and financial question answering. BloombergGPT is shown to outperform the OPT (66B parameters), GPT-NeoX, and BLOOM (176B parameters) LLMs on these financial domain-specific tasks and performs competitively on broader benchmarks.

\citet{thiergart2021understanding} considers the applicability of GPT-3 to the task of email management, including classification, information extraction (NER), and generating response text. Whilst it is noted that GPT-3 has the capacity for all three tasks, the author highlights current issues around reliability, lack of access to internal data, and the need for a human in the loop.

\citet{liu2022deplot} propose enabling LLMs to understand charts and plots by first using a vision plot-to-text translation model (DePlot) to decompose the chart into a linearized data table. Once the chart or plot has been converted into a text-based data table, it is combined with the prompt and provided to a Flan-PaLM, Codex, or GPT-3.5 LLM. A similar \emph{modality conversion} \ref{fig:modality_conversion} approach has also been used in robotics \cite{liu2023reflect, huang2022inner} for sensor data.

\citet{news_summary} evaluate a range of LLMs (GPT-3, InstructGPT, OPT, GLM, Cohere, and Anthropic) on the task of news summarization. On the DM/CNN and XSUM benchmarks, instruction fine-tuned models (InstructGPT) perform the best across summarization faithfulness, relevance, and coherence. To evaluate against human capability~\citet{news_summary} collect reference summarizations for 100 articles from 6 freelance writers. Zero-shot InstructGPT-3 performs comparably to the freelance writers across the three metrics.

\citet{cheng2023gpt4} investigate GPT-4's capacity to perform data analysis and compare it to human analysts. GPT-4 is combined with a \emph{modular prompting} framework \ref{fig:modular_prompting} with three steps, code generation (SQL and Python), code execution (``collect data and output figures'', etc.), and analysis generation (``generate five bullet points about the analysis''). While GPT-4 performs well, it currently underperforms experienced human data analysts on tasks from NvBench~\cite{luo2021synthesizing}.  

For scientific knowledge work,~\citet{taylor2022galactica} train the Galactica LLM specifically on scientific text for tasks such as scientific knowledge recall, reasoning, citation prediction, and scientific Q\&A. In addition to a domain-specific training corpus, Galactica is specialized in the scientific domain through the use of specialized tokens, working memory, and \emph{prompt-pre-training}. 

\citet{dunn2022structured} propose fine-tuning GPT-3 for scientific combined named entity recognition and relation extraction (LLM-NERRE). First, 100 to 1,000 manually annotated prompt-completion pairs are created by humans. These examples are then used to fine-tune a GPT-3 model for the specific NERRE task. 

Finally, \citet{liuReviewerGPTExploratoryStudy2023} evaluate GPT-4's ability to review academic papers, specifically: identifying errors, verifying author checklists, and selecting the \emph{better} abstract. GPT-4 shows some capacity to detect errors, with 7 out of 13 errors detected, and verify author checklists, with 87\% accuracy. However, GPT-4 is shown to have limited capacity for distinguishing the \emph{better} paper abstract.

\subsection{Law} \label{sec:law}

Applications of LLMs within the legal domain share many similarities with medicine, including legal question answering~\citep{yu2022legal, katz2023gpt} and legal information extraction~\citep{chalkidis2020legal}. However, other domain-specific applications have been proposed, such as case outcome prediction~\citep{hamilton2023blind}, legal research~\citep{iu2023chatgpt}, and legal text generation~\citep{peric2020legal}.

\subsubsection{Legal Question Answering and Comprehension}

Key tasks of the legal field are finding related precedents, answering legal questions, and comparing existing documents or statutes.

Using a general-purpose LLM with prompting approach,~\citet{yu2022legal} use GPT-3.5 with zero-shot, few-shot, and CoT prompting to achieve SOTA performance on the legal entailment task (identifying the relevant statutes and determining if a given premise is correct) in the Competition on Legal Information Extraction/Entailment (COLIEE) dataset~\citep{Rabelo2022-xe}. They also investigate a GPT-3.5 version fine-tuned using the COLIEE training set with and without explanations but find the zero- and few-shot legal prompting approaches perform best. Similarly,~\citet{rosa2022billions} use a general monoT5 model with zero-shot prompting on the COLIEE entailment task.

On the US legal Uniform Bar Examination (UBE),~\citet{bommarito2022gpt} show that GPT-3.5 with zero-shot prompting can achieve ~50\% on the multiple choice Multistate Bar Examination component, but note that fine-tuning the model on relevant examples does not appear to improve performance. More recently,~\citet{katz2023gpt} show that GPT-4 with zero-shot prompting exhibits SOTA performance on the full UBE, including the multiple choice, essay, and performance test components, and achieves passing scores.

~\citet{blair2023can} assess GPT-3.5's ability to reason about legal facts and statutes using the StAtutory Reasoning Assessment (SARA) dataset~\cite{holzenberger2020dataset}. GPT-3.5 is shown to have SOTA performance but with significant variation depending on the type of prompting used (zero-shot, few-shot, and CoT). GPT-3.5 was also shown to perform relatively poorly on synthetic statutory reasoning tasks. 

~\citet{choi2023chatgpt} evaluate ChatGPT (GPT-3.5) on 95 multiple-choice and 12 essay questions from the final exams at the University of Minnesota law school. ChatGPT was found to perform at the level of a C+ student, near the bottom of the class, but with passing scores.  

\begin{constraint}{Out of Date Information}
Due to regularly updated laws and new precedents, the training/retrieval data become outdated frequently \cite{henderson2022pile}.
\end{constraint}

Finally, many more specific legal question-answering applications have been proposed, including: explaining legal concepts (GPT-4 + retrieval) \cite{savelka2023explaining}, summarizing legal judgments (GPT-3.5) \cite{deroy2023ready}, litigation research and drafting \cite{iu2023chatgpt}, and helping full-fill the tasks of a law professor (ChatGPT) \cite{pettinato2023chatgpt}.

\subsubsection{Case Prediction and Legal Text Generation}

Case prediction and legal text generation involve predicting or completing legal opinions. Whilst there is currently sparse usage of LLMs in the literature, smaller language models have been applied, suggesting potential future LLM applications in this area.

\citet{hamilton2023blind} use nine separate GPT-2 models trained on individual supreme court justice's authored opinions to predict how each justice will vote on a given case. They use a handcrafted prompt, including a summary of the topic generated by GPT-3. However, they find this approach to case prediction does not match the SOTA. 

Previously,~\citet{chalkidis2019neural} trained a range of attention-based models (including BERT) to predict case outcomes from the European Court of Human Rights (ECHR). The attention-based models outperformed an SVM with a bag of words approach for binary violation classification, multi-label violation classification, and case importance prediction.

Finally,~\citet{peric2020legal} use a dataset of 50,000 judicial opinions from U.S. Circuit Courts to train a Transformer-XL model and fine-tune a GPT-2 model. The models were then evaluated for their ability to complete a judicial opinion, with a start given as a prompt. In qualitative evaluations, human participants struggled distinguishing between machine-generated and genuine text.

\subsection{Medicine}\label{sec:medicine}

Many applications of LLMs have been proposed in the medical domain, including medical question answering~\citep{medpalm, medpalm2, lievin2022can, yunxiang2023chatdoctor, nori2023capabilities}, clinical information extraction~\citep{agrawal2022large, rajkomar2022deciphering}, indexing~\citep{you2021bertmesh}, triage~\citep{sezgin2022operationalizing, levine2023diagnostic}, and management of health records~\citep{korngiebel2021considering}.

\subsubsection{Medical Question Answering and Comprehension}

Medical question answering and comprehension consists of generating multiple-choice and free-text responses to medical questions.

\citet{medpalm} proposed using few-shot, CoT, and self-consistency prompting to specialize the general-purpose PaLM LLM to medical question answering and comprehension. They demonstrate a Flan-PaLM model~\citep{flan} using a combination of the three prompting strategies to achieve the previous SOTA results on the MedQA, MedMCQA, PubMedQA, and MMLU medical datasets. To further align the model to the medical domain, they proposed Med-PaLM, which utilizes instruction prompt-tuning based on 40 examples from a panel of clinicians and task-specific human-engineered prompts. 

\citet{medpalm2} then extend the Med-PaLM approach with Med-PaLM 2 using the newer PaLM 2 LLM as its base model. \citet{medpalm2} conduct further instruction-fine tuning and use a new ensemble refinement (ER) prompting strategy (where stochastically sampled outputs are first generated and provided within the final prompt). This allows Med-PaLM 2 to achieve the current SOTA on the MultiMedQA benchmark.

\citet{lievin2022can} adopt a similar approach using zero-shot, few-shot, and CoT prompting to adapt the GPT-3.5 LLM to medical question answering (USMLE and MedMCQA) and comprehension (PubMedQA) tasks. In addition,~\citet{lievin2022can} propose using retrieval augmentation where relevant text from Wikipedia is retrieved and included in the prompt. More recently, \citet{nori2023capabilities} evaluated GPT-4 on USMLE and MultiMedQA datasets using zero and few shot prompting. GPT-4 is found to outperform GPT-3.5 across benchmarks significantly. However, several issues relating to using GPT-4 for real-world clinical applications are raised, including the \emph{risks of erroneous generations} and the \emph{risks of bias}. ~\citet{tang2023evaluating} raise similar issues and find that GPT-3.5 and ChatGPT have issues with factual accuracy and representing the level of certainty during medical summarization.

\begin{constraint}{Hallucination and Bias~\citep{tang2023evaluating, nori2023capabilities, medpalm}}
     The safety-critical nature of the medical domain means the possibility of hallucinations significantly limits the current use cases. Further work is also needed to reduce the risk of LLMs perpetuating existing bias in clinical datasets.
\end{constraint}

\citet{yunxiang2023chatdoctor} fine-tune a LLaMA LLM ChatDoctor (7B parameters) specifically for the task of medical question answering. To specialize the LLaMA model, it is first instruction fine-tuned using the Alpaca dataset~\citep{taori2023alpaca} and then fine-tuned to the medical domain using a dataset of 100k patient conversations. Similarly to~\citet{lievin2022can}, ChatDoctor is augmented with two external knowledge sources (a disease database and Wikipedia) to improve the factual grounding of the model.

Instead of using general models with specialized prompting or fine-tuning, ~\citet{pubmedgpt} train a new model PubMedGPT specifically for medical question answering and text generation tasks. PubMedGPT is trained using a combination of PubMed abstracts and full documents from the Pile~\citep{gao2020pile}. \citet{peng2023study} also train a new LLM GatorTronGPT (up to 20B parameters) for biomedical question answering and relation extraction using a mixture of clinical and general English text. Whilst these approaches outperformed existing smaller specific purpose models~\citep{gu2021domain, yasunaga2022deep} in medical question answering, they currently underperform the larger general purpose LLMs (GPT-3.5/4 and MedPaLM 1/2). However, there remains debate over whether larger general or specialized clinical models are the best approach. Looking at models up to GPT-3,~\citet{do_we_need_clinical_language_models} question the effectiveness of LLM in-context learning approaches by showing that small specialized clinical models fine-tuned on limited annotated data outperform the former. 

Finally, LLMs have also been applied to a range of more specific medical question-answering tasks, including evaluating GPT-3 on its' ability to triage and diagnose cases \cite{levine2023diagnostic}, responding to social media genetics \cite{duong2023analysis} and general \cite{ayers2023comparing} patient questions (ChatGPT), answering questions from the Korean general surgery board exams (GPT-3.5, GPT-4) \cite{chatgptoperating}, consultation and medical note taking~\cite{lee2023benefits}, and answering ophthalmology questions \cite{ChatGPTOphthalmology}.

\subsubsection{Medical Information Retrieval}

Medical text often contains domain-specific abbreviations, acronyms, and technical terms presenting specific information retrieval challenges. This has led LLMs also to be applied to help structure and extract data from medical sources. 

\citet{agrawal2022large} use InstructGPT (GPT-3) with prompt templates (zero- and one-shot) for clinical information extraction, such as extracting medication dosage and frequency from medical notes or disambiguation of medical acronyms. They also introduce two methods for converting the LLM output into a structured format using a {\it verbilizer} for mapping to classification labels and a {\it resolver} for more complex structured outputs such as lists (GPT-3 + R).

\citet{rajkomar2022deciphering} take a different approach by treating medical acronym disambiguation as a translation task and training a specialized end-to-end T5 LLM. To preserve privacy, they also use a training dataset generated from public web pages (without medical acronyms) and web-scale reverse substitution of medical acronyms, with only evaluation done on actual clinical notes. 

Finally, \citet{gu2023distilling} use GPT-3.5 and knowledge distillation to train a PubMedBERT model for adverse drug event extraction (entity and relation). The distilled PubMedBERT model outperforms GPT-3.5 and GPT-4, and performs similarly to specialized models that use supervised learning.

\subsection{Reasoning} \label{sec:reasoning}
Mathematical and algorithmic tasks often require a different set of capabilities than traditional NLP tasks, such as understanding mathematical operations, complex multi-step reasoning, and longer-term planning. Therefore, the applicability of LLMs to these tasks, and methods for improving their capabilities, is an active area of research. 

For mathematical reasoning tasks, \citet{solving_math_problems} test a range of fine-tuning (supervised and RLHF), prompting (zero-shot and few-shot), and re-ranking (majority voting and reward model) to evaluate whether they improve a base LLM's (70B parameters) ability to generate accurate reasoning steps on word-based maths problems in the GSM8K dataset~\citep{gsm8k}. Whilst fine-tuning on intermediate steps (``process-based'') performs similarly to using only final answers (``outcome-based'') on final answer correctness, processed-based approaches are found to generate significantly fewer errors in reasoning. 

\citet{self_improvement} take this a step further by showing that the mathematical reasoning ability of a PaLM LLM on the GSM8K dataset can be \emph{self-improved} through fine-tuning on a dataset of high-confidence reasoning paths generated by the same PaLM base model.

Using only prompting,~\citet{kojima2022large} find that zero-shot CoT prompting alone significantly improves the performance of GPT-3 and PaLM LLMs over standard zero- and few-shot prompting on the MultiArith and GSM8K datasets. While~\citet{better_reasoners} introduce DIVERSE, a prompting approach that uses a diverse set of prompts for each question and a trained verifier (with reasoning step awareness) to improve further GPT-3.5's performance on GSM8K and other reasoning benchmarks. Finally,~\citet{Shridhar2022AutomaticGO} take a novel approach by training new models to break down a mathematical word problem into \emph{Socratic sub-questions} to guide the answer of either other LLMs or human learners. GPT-3 prompted with these sub-questions outperforms simple one-shot prompting on the GSM8K dataset. 

\citet{robustness_math_lm} evaluate a range of LLMs (including GPT-3) at mathematical reasoning using a new framework to understand the causal impact of different input factors (e.g framing, operands, and operations). Instruction fine-tuned GPT-3 models are found to be significantly more robust and sensitive than the smaller LLMs evaluated.

Other LLM use cases in algorithmic and mathematical reasoning have also been proposed.~\citet{gadgiltowards} apply a Codex LLM with prompt engineering and filtering to the task of mathematical formalization (in the context of theorem proving).~\citet{analogical_reasoning} evaluate GPT-3.5's capacity for analogical reasoning using tasks that emulate Raven's Standard Progressive Matrices (SPM), letter string analogies, and verbal analogies. GPT-3.5 is shown to generally outperform human participants (undergraduates) at matrix reasoning and verbal analogies, but with more mixed results on letter string analogies.~\citet{yu2022alert} introduce the ALERT benchmark to evaluate LLM reasoning across ten skills (logistic, causal, common-sense, abductive, spatial, analogical, argument, and deductive reasoning, as well as textual entailment and mathematics).              
\citet{ruis2022large} study LLMs' capability to interpret implicatures, for example, whether they understand the response "I wore gloves" to the question ``Did you leave fingerprints?'' as meaning ``No''; finding that lots of models perform close to random.
Finally, \citet{valmeekam2023large} propose a new assessment framework for \emph{common-sense} planning and find that existing LLMs GPT-3.5 and BLOOM perform poorly. Using the framework for the Blocksworld domain (planning tasks with different colored blocks on a surface), the best GPT-3.5 model only came up with a valid plan 5\% of the time, compared to 78\% of human participants. 

\begin{constraint}{Sub-Human-Performance \citep{valmeekam2023large, willig2023causal}}
Existing LLMs struggle to match human performance on reasoning benchmarks. 
\end{constraint}

Another line of work has investigated the intersection of LLMs and causal reasoning~\citep{peters2017elements,cml}.~\citet{kiciman2023causal} argue that GPT-3.5/4 outperform existing algorithms in three causal benchmarks. In contrast,~\citet{gao2023chatgpt} evaluate ChatGPT on three causal reasoning tasks (distinct from~\citet{kiciman2023causal}) and find that it performs rather poorly; further, few-shot and chain-of-thought prompting sometimes further exacerbates its performance.~\citet{srivastava2022beyond} propose 14 causal reasoning tasks, some of which are considered to be very hard~\citep{suzgun2022challenging}. 
Similarly,~\citet{jin2023large} curate another causal inference task and posit that current LLMs still fail to generalize.~\citet{lampinen2023passive} study whether LLMs can generalize causal intervention strategies from few-shot examples. ~\citet{willig2023causal} conjecture that current LLMs are ``causal parrots'', simply reciting causal knowledge embedded in their data rather than doing causal reasoning \cite{cml}.

Overall, while LLMs show some capacity for more complex reasoning, the relatively poor performance of LLMs on a number of reasoning tasks and benchmarks~\citep{valmeekam2023large, gao2023chatgpt, jin2023large} stands in contrast to the often human level performance being seen in other capabilities~\citep{bubeck2023sparks, kiela2021dynabench}.

\subsection{Robotics and Embodied Agents} \label{sec:robotics}
LLMs have also started to be incorporated into robotics applications to provide high-level planning and contextual knowledge.

\citet{saycan} implement a PaLM-540B LLM in the SayCan architecture to break down high-level text-based instructions into a sequence of lower-level robot tasks that can be executed. The authors use the LLM to propose possible next actions via iteratively scoring the most likely of a defined set of low-level tasks based on the high-level text input. The low-level task to be executed is then determined by combining the low-level tasks proposed by the LLM with affordance functions which determine the probability of the robot completing the task given the current low-level context.

\citet{palm-e} take this concept a step further by combining the PaLM-540B LLM with additional input modalities (22B parameter vision transformer) to create the PaLM-E model. By introducing images into the input, the PaLM-E model can predict which low-level tasks are possible given the current state, whether the previous low-level tasks executed failed, and incorporate images into long-horizon planning, allowing it to outperform the original SayCan results. 

Another approach has been to use LLMs to generate code for robotics tasks. \citet{chatgpt_robo} combine ChatGPT with a pre-defined high-level function library of robotic capabilities for human \emph{on the loop} robotics tasks. By providing details of the function library in the prompt, ChatGPT is then shown to be able to break down high-level natural language instructions into a set of lower-level function calls, which can then be executed on the robot if the human is satisfied it is accurate. This is another example of the \emph{API definition} \ref{fig:api_pipeline} approach, also used in computer programming~\cite{suris2023vipergpt}. Other related works that use LLMs to generate code for robotics applications include using an LLM for hierarchical code generation to write robot policies (Codex) \cite{liang2023code}, to generate code policies and maintain a written state (GPT-3.5)~\cite{yoneda2023statler}, and using an LLM for code-based task planning (GPT-3, Codex) \cite{singh2022progprompt}. 

Finally, LLMs have also been combined with modality-to-text pre-processing to provide the LLM with additional input from the robot's environment. \citet{liu2023reflect} use GPT-4 as part of the REFLECT framework for detecting and explaining robot failures. To achieve this, multi-modal sensory inputs are first converted into a text-based hierarchical summary at the sensory, event, and sub-goal levels. The hierarchical summary then prompts the LLM to detect and analyze failures. Similarly, \citet{huang2022inner} combine an LLM (InstructGPT, PaLM) with multiple sources of text-based environment feedback for robotic task planning.

\begin{constraint}{Single Modality \citep{liu2023reflect, saycan, chatgpt_robo}}
While LLMs can help robots or agents understand instructions and add high-level planning capabilities, their inability to directly learn from image, audio or other sensor modalities constrain their applications.  
\end{constraint}

For agents in simulated worlds,~\citet{wang2023voyager} use the GPT-4 LLM within their VOYAGER framework to create a Minecraft agent that can autonomously explore, acquire new skills and complete tasks. First, they use GPT-4 to propose new tasks for the agent to complete as part of the \emph{automatic curriculum}. Then, they ask it to generate code to solve the proposed task given the current state to add to its \emph{skills library}, which can then be used in the future (similar to the API approach \ref{fig:api_pipeline} used by~\citet{chatgpt_robo}). Finally, the authors use GPT-4 to verify whether the executed code has achieved the proposed task. This framework outperforms prompting approaches such as ReAct, Reflexion, and AutoGPT (\Cref{sec:prompt_brittleness}). 

Prior work using LLMs for planning in simulated worlds include:~\citet{wang2023describe} using GPT-3 for Minecraft,~\citet{huang2022language} using GPT-3 and Codex in VirtualHome, and~\citet{nottingham2023embodied} using Codex for Minecraft.

\subsection{Social Sciences \& Psychology}\label{sec:social_sciences}
\begin{figure}[ht]
\centering
\includegraphics[width=\linewidth]{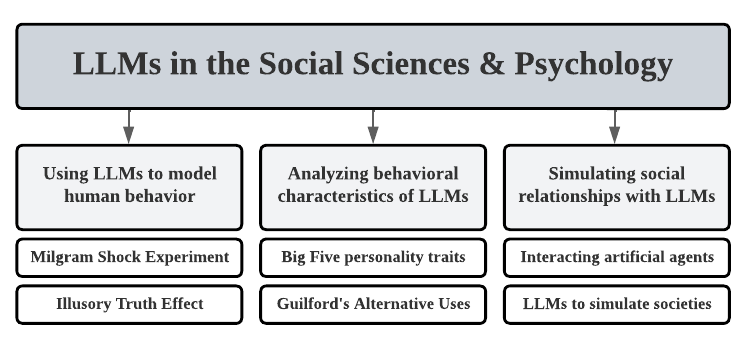}
\caption{Use cases of LLMs in the social sciences and psychology can mainly be structured into three categories: using LLMs to model human behavior~\cite[\eg][]{aher2022using, horton2023large}, analyzing behavioral characteristics of LLMs~\cite[\eg][]{pellert2023ai}, and using LLMs to simulate social relationships~\cite[\eg][]{park2023generative}.}
\label{fig:social_sciences}
\end{figure}

The rapid advancements of LLMs have fostered the use of such models across research in the psychological and behavioral sciences. Reviewing the existing literature, we have identified three main areas and tasks in which LLMs have been used in the context of the psychological and behavioral sciences: using LLMs to simulate human behavioral experiments~\cite[\eg][]{argyle2022out, griffin2023susceptibility, horton2023large, wu2023large, dominguez2023questioning}, analyzing the personality traits of LLMs~\cite[\eg ][]{miotto2022gpt, pellert2023ai,safdari2023personality}, and employing them as artificial agents to model social relationships~\citep{park2023artificial}. See ~\Cref{fig:social_sciences} for an illustration.

\subsubsection{Modeling Human Behavior}
In the behavioral sciences, there is an increasing interest in using LLMs as models for psychological experiments. Being able to model human behavior computationally through language models would entail a variety of advantages over using human participants: experiments with LLMs are cheaper, faster, can be scaled easier, and are potentially less sensitive to ethical considerations~\citep{griffin2023susceptibility}. In light of this, various works have compared LLMs with human participants from a behavioral perspective.

~\citet{argyle2022out} demonstrate how LLMs can generate responses corresponding to virtual participants in behavioral experiments. They do so by using LLMs to generate samples of responses to studies related to political opinions and voting behavior. In particular, the authors investigate three studies: the first asks participants to list words associated with outgroup partisans, and the second and third focus on vote prediction based on demographics. Across scenarios, experimental results demonstrate that GPT-3 provides answers that closely align with human responses.

~\citet{horton2023large} argue that LLMs can be used to computationally model human behavior and demonstrate such an ability in economics by exploring their behavior in economic scenarios. They conducted four experiments focusing on economic decision-making using GPT-3, showing that the LLM can approximately replicate results obtained with human individuals.

~\citet{griffin2023susceptibility} investigate the suitability of LLMs to model psychological change. In their study, the authors assess LLM responses to two behavioral tests, the \textit{illusory truth effect}~\cite[ITE;][]{henderson2022reproducible} and an experiment measuring the influence of populist news to change in political views~\citep{bos2020effects}. The results demonstrate that in both scenarios, human judgments tend to align with LLM-based judgments, indicating that LLMs have the potential to model the effect of influence on human individuals.

~\citet{aher2022using} introduce \textit{the Turing Experiment} (TE) to measure an LLM's suitability to model human behavior. A TE consists of inputs to the LLM that signal a certain demographic (\eg names or occupations) as well as a set of experimental details and corresponding outputs used to simulate human behavior. The authors apply their approach to four individual tests, namely an ultimatum game from behavioral economics~\citep{houser2014experimental, krawczyk2018introduction}, garden-path sentences used in psycholinguistics~\citep{christianson2001thematic, patson2009lingering}, the Milgram Shock Experiment from social psychology~\citep{milgram1963behavioral}, and the wisdom of crowds task used to measure collective social intelligence~\citep{moussaid2013social}. Demographic details are simulated via gender titles and surnames. The results show that LLMs largely align with human behavior across the tests. However, the authors note that LLM size matters and that larger models tend to provide results that are more aligned with human responses.

~\citet{aher2022using} point out that the LLMs were most likely exposed to the four behavioral experiments during their pre-training. To account for that, the authors create artificial variations of the experiments with conditions that differ from previous studies. Additionally, the authors note that a potential risk with using LLMs to simulate human responses is the introduction of generations that contain biases stemming from the models' training data.

\begin{constraint}{Social Biases~\cite{aher2022using, miotto2022gpt}}
Unbalanced views and opinions in the training data skew the LLMs towards biased human behaviors.
\end{constraint}

~\citet{park2023artificial} replicate a set of 8 psychological studies from the Many Labs 2 project~\citep{klein2018many} using GPT-3 to assess the LLM for its ability to simulate human behavioral data. Such studies include tests in which subjects are asked to choose between a kiss from a favorite movie star and \$50~\citep{rottenstreich2001money} and where subjects had to decide between paying a traffic violation fine and going to court~\citep{ross1977false}. These experiments show that GPT-3 replicates only 37.5\% of the effects obtained from human participants. The authors argue that these results are attributed to humans and LLMs representing inherently different cognitive systems.

~\citet{maddelaTrainingModelsGenerate2023} study identifying unhelpful thought patterns and possible reframings to facilitate mental health. They release a dataset called \textsc{PatternReframe} and evaluate GPT-3.5 on it, showing that it can perform very well without additional training. They conclude that practitioners of cognitive behavioral therapy may benefit from using LLMs to produce richer training material.

\subsubsection{Analyzing Behavioral Characteristics of LLMs}
In addition to using LLMs as models for human behavior, various existing works study LLMs by analyzing their personality traits.

~\citet{jiang2022mpi} do so by introducing the \textit{Machine Personality Inventory} (MPI) dataset, a collection of items to assess personalities according to the Big Five personality factors: extraversion, agreeableness, openness, conscientiousness, and neuroticism~\citep{mccrae1997personality}.

~\citet{miotto2022gpt} assess GPT-3's personalities using the HEXACO~\citep{ashton2009hexaco} and Human Values~\citep{humvalscale} scales. Their experimental results reveal that GPT-3 obtains personality and value scores that align with human participants.~\citet{miotto2022gpt} provide an extensive analysis of varying temperature values used to prompt the LLM, finding that an increased temperature yields changes in the model's personalities, \eg GPT-3 shows a higher unwillingness to manipulate as well as increased scores on anxiety. Similar results were obtained concerning the Human Values scale, where model responses varied substantially for different temperature values.

In line with this work,~\citet{pellert2023ai} argue that LLMs possess psychological traits as observed in human individuals and can be assessed through psychometric tests. The authors conduct experiments measuring, among others, the Big Five personality traits in a zero-shot setup. In contrast, to~\citet{miotto2022gpt, pellert2023ai} investigate smaller models based on BERT and find that different variants of BERT score across the five personalities in a fairly homogeneous fashion, with traits that are high on agreeableness and extraversion, but low on neuroticism.

In a related fashion,~\citet{stevenson2022putting} assess LLM performance (GPT-3) on the Guilford’s Alternative Uses Test~\cite[AUT;][]{guilford1967creativity}, a test to assess human creativity. The test asks participants to suggest uses for physical objects (\eg a book or a fork). Comparing the AUT test performance of GPT-3 to that of psychology students, the authors found that human responses score higher on originality and surprise, whereas GPT-3's responses were more useful.

~\citet{kosinski2023theory} test Theory of Mind (ToM) in LLMs. ToM refers to the ability to track others' unobservable mental states, such as intentions, beliefs, or desires. The authors find that among LLMs of the GPT family, recent models can increasingly solve ToM tasks without having been explicitly trained to do so. For instance, while GPT-2 shows virtually no capability of solving ToM tasks, GPT-3.5 (based on InstructGPT) and GPT-4 performed similarly to 6- and 7-year-old children, respectively.~\citet{gandhi2023understanding} present a template-based framework for generating synthetic samples to evaluate ToM in LLMs, which are then applied to five recently developed LLMs (incl. GPT-3, GPT-4, LLaMA, and Claude). The authors show that most models struggle with ToM in its basic forms. However, GPT-4 performs closest to the human comparison of all tested models.

\subsubsection{Simulating Social Relationships}
While most previous works measure LLMs as models for human behavior through replicating human behavioral studies,~\citet{park2023generative} use the power of LLMs to model the interaction between artificial agents. The authors model a community of 25 artificial agents interacting in a digital environment to achieve this. Each character has unique traits, and the characters interact with each other through natural language. Simulating such societies, the authors observe emergent social behaviors (\eg forming new relationships and attending events) between agents that are formed without any human interaction.

\begin{figure}[htb]
    \centering
\includegraphics[width=\columnwidth]{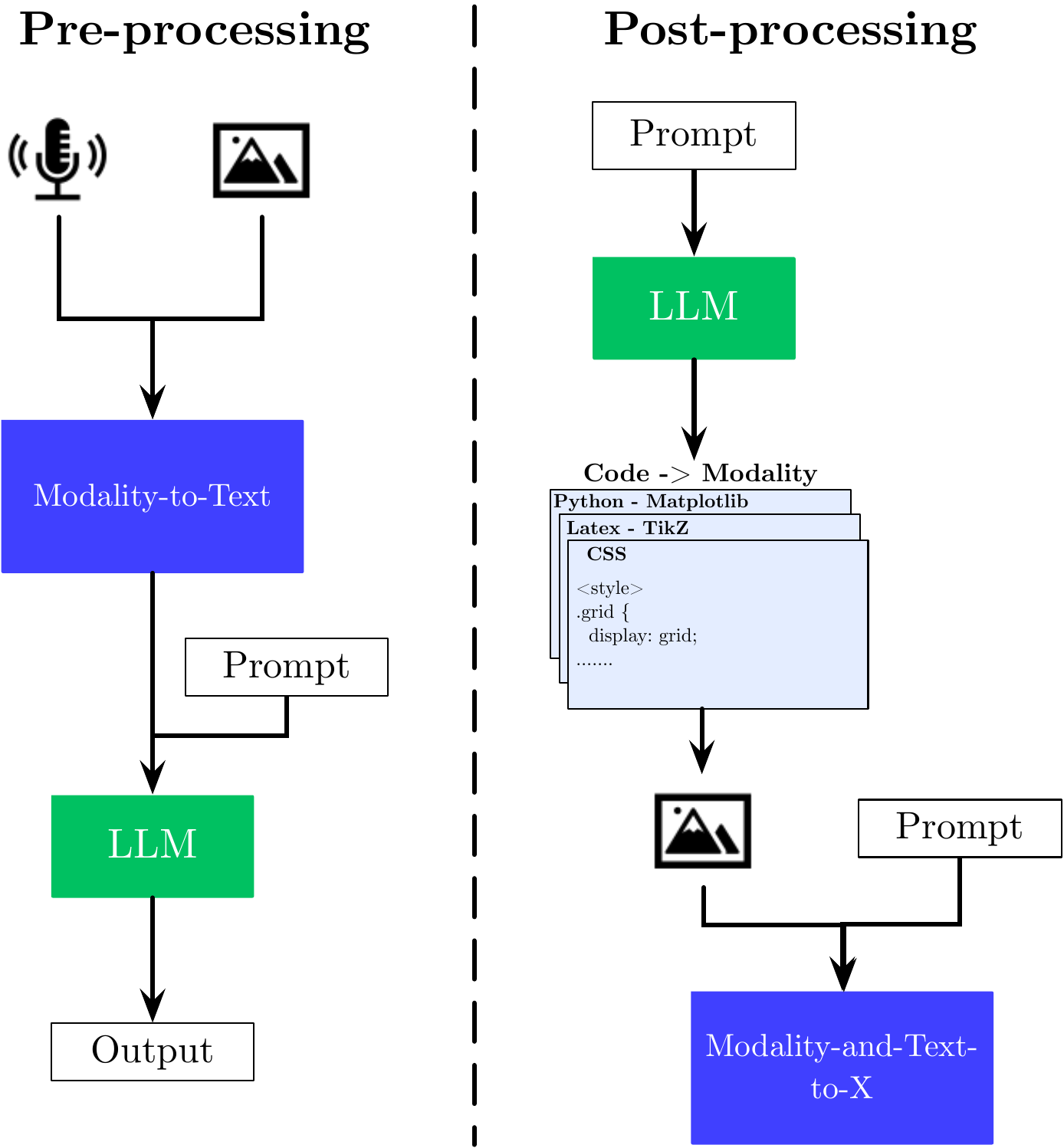}
    \caption{\textbf{Modality Conversion}. Illustration of using models with other input modalities as pre or post-processing steps in an LLM pipeline~\citep{fengLayoutGPTCompositionalVisual2023, liu2022deplot, liu2023reflect, huang2022inner, lian2023llmgrounded}. For some use cases, this approach can be used as an alternative to training a multi-modal model or using a shared embedding space.}
    \label{fig:modality_conversion}
\end{figure}

\subsection{Synthetic Data Generation} \label{sec:synthetic_data}
The ability of LLMs to perform in-context learning allows them to be prompted to generate synthetic datasets for training much smaller domain-specific models.

\citet{label_cost} propose using GPT-3 to label datasets more cost-effectively than human labelers. These labeled datasets can then be used to train more compute-efficient smaller models. To evaluate this approach, RoBERTa and PEGASUS models are trained for 9 NLP tasks using human and GPT-3 generated labels. GPT-3 labels are shown to outperform human labels when labeling budgets are small, but higher-quality human labels tend to lead to better models at higher labeling budgets.  

Similarly,~\citet{ding2022gpt} propose three prompting approaches for training data generation with GPT-3: unlabeled data annotation (generate labels for known examples), training data generation (generate examples and labels), and assisted training data generation (with Wikidata provided as additional context). Fine-tuning a smaller BERT model for text classification and NER tasks using these approaches showed results similar to or worse than using GPT-3 directly.

\citet{gunasekar2023textbooks} leverage synthetic data generation with GPT-3.5 to train a new code generation LLM (see \Cref{code_gen_sec}). The generated data consists of synthetic Python textbooks focusing on reasoning, basic algorithmic skills, and synthetic Python exercises. One important finding of this work is that introducing randomness into data generation is crucial, all while ensuring the examples maintain their quality and coherence.

\citet{yoo-etal-2021-gpt3mix-leveraging} propose GPT3Mix to generate additional synthetic data from an existing dataset for classification tasks. GPT3Mix uses GPT-3 with a prompt containing real examples from the dataset and a task specification to create synthetic examples and \emph{pseudo-labels} jointly. This new augmented dataset is then used to fine-tune BERT and DistilBERT models. This method combines data augmentation approaches with knowledge distillation by training smaller classification models using soft labels.

~\citet{bonifacio2022inpars} propose InPars, a method for using LLMs to generate synthetic retrieval examples for fine-tuning on information retrieval tasks. GPT-3 is few-shot prompted to generate a relevant question for a randomly sampled document along with the question's associated probability. A smaller monoT5 model is then fine-tuned using this dataset to rank relevant documents for a given question. The fine-tuned model outperforms only pre-trained models but performs worse than models fine-tuned using the existing MS MARCO training dataset~\cite{bajaj2018ms}.

~\citet{dai2023chataug} introduce AugGPT, which uses ChatGPT (GPT-3.5) to augment each example in a small base dataset with six additional rephrased synthetic examples. This new augmented dataset is then used to fine-tune a specialized BERT model. This approach outperforms existing augmentation approaches, such as word and character substitution.

Finally, instead of generating synthetic data to achieve a specialized task,~\citet{shridhar2022distilling} propose Decompositional Distillation, which aims to use synthetic data to replicate in smaller models the multi-step reasoning capabilities, such as CoT, that emerge in larger LLMs. First, GPT-3 is used with a manually designed few-shot prompt to decompose a problem into (sub-question, sub-solution) pairs. This synthetic sub-question dataset is then used to fine-tune a T5 \emph{problem decomposer} to generate sub-questions. Finally, a GPT-2 \emph{problem solver} is fine-tuned to provide the sub-solutions to the teacher-generated sub-questions.

Overall, while LLM-generated synthetic data can potentially bring significant cost benefits, the greater its role, the higher the potential for it to fail to capture the true distribution and potentially lead to model collapse~\citep{shumailov2023curse}. 

\begin{constraint}{Hallucinated Distributions ~\citep{shumailov2023curse}}
    Using LLMs for fully synthetic data generation is currently constrained by our inability to verify whether the synthetic data generated is representative of the true distribution in the corresponding real-world data. 
\end{constraint}

In cases where the LLM is only used to label existing data~\citep{label_cost, ding2022gpt} this will likely reduce the risk of generating an unrepresentative training distribution (although hallucinated labels remain an issue). Where the LLM is used to generate (or partially generate) both the input and the target~\citep{ding2022gpt, dai2023chataug, gunasekar2023textbooks, bonifacio2022inpars, shridhar2022distilling} the issue of hallucinated distributions becomes potentially significant.

\section{Related Work} \label{sec:rw}
Closest to ours is the concurrent work by ~\citet{zhaoSurveyLargeLanguage2023}, who provide an extensive survey of large language models and associated topics.~\citet{mialon2023augmented} focus on surveying augmented language models, \ie ``language models with reasoning skills and the ability to use tools''.~\citet{tornedeAutoMLAgeLarge2023} survey LLMs in the context of AutoML methods, highlighting existing methods and challenges in leveraging these for improving LLMs. \citet{tangScienceDetectingLLMGenerated2023} survey LLM-generated text detection techniques. 
\citet{changSurveyEvaluationLarge2023a} concurrently survey evaluation tasks of LLMs.

The literature also contains several previous surveys and evaluations specific to individual application domains that reference LLMs, including:  chatbots~\cite{luo2022critical}, computational biology~\citep{tran2023survey, hu2022protein}, computer programming~\citep{shirafuji2023exploring}, medicine~\citep{nerella2023transformers,  wornow2023shaky, wang2023large, nerella2023transformers}, law~\citep{cyphert2021human, sun2023short}, knowledge work~\citep{eloundou2023gpts, xie2023survey}, and reasoning~\citep{huangReasoningLargeLanguage2023}.

\section{Conclusion}
In this work, we identify several unsolved challenges of large language models, provide an overview of their current applications, and discuss how the former constrain the latter. By highlighting the limitations of existing methods, we hope to foster future research addressing these. We also hope that by providing an overview of the approaches used in different applied areas, we can facilitate the transfer of ideas between domains and target further research.

\section*{Acknowledgements}
We thank Abhishek Kumar and Stella Rose Biderman for fruitful discussions and feedback on the draft.

\bibliographystyle{acl_natbib}
\small{\bibliography{anthology,refs}}

\end{document}

%% file: math.tex
\usepackage{amsmath,amsfonts,bm}

\newcommand{\params}{\boldsymbol{\theta}}

\def\eqref#1{equation~\ref{#1}}

\def\1{\bm{1}}

\def\rmP{{\mathbf{P}}}

\def\rmX{{\mathbf{X}}}

\def\vx{{\bm{x}}}
\def\vy{{\bm{y}}}

\def\mR{{\bm{R}}}

\def\mW{{\bm{W}}}

\DeclareMathAlphabet{\mathsfit}{\encodingdefault}{\sfdefault}{m}{sl}
\SetMathAlphabet{\mathsfit}{bold}{\encodingdefault}{\sfdefault}{bx}{n}

\newcommand{\softmax}{\mathrm{softmax}}



%% file: commands.tex
\newcommand{\ub}[1]{\textbf{\underline{#1}}}

\definecolor{CBblue}{RGB}{0, 107, 164}
\definecolor{CBorange}{RGB}{255, 128, 14}
\definecolor{CBgray}{RGB}{89, 89, 89}
\definecolor{CByellow}{RGB}{171, 171, 42}
\definecolor{CBpurple}{RGB}{137, 61, 86}
\definecolor{CBgreen}{RGB}{27, 158, 119}
\definecolor{CBpink}{RGB}{215, 48, 39}

\newcommand{\cblue}[1]{\textcolor{CBblue}{#1}}
\newcommand{\corange}[1]{\textcolor{CBorange}{#1}}

\newcommand{\cyellow}[1]{\textcolor{CByellow}{#1}}
\newcommand{\cpurple}[1]{\textcolor{CBpurple}{#1}}
\newcommand{\cgreen}[1]{\textcolor{CBgreen}{#1}}

\newcommand{\missing}[1]{\textbf{\textcolor{red}{#1}}}

\usepackage{tcolorbox}
\usepackage{fontawesome}

\tcbuselibrary{skins}
\newtcolorbox{challsumm}{
  enhanced,
  colframe=red!70!black,
  colback=red!10!white,
  coltitle=white,
  fonttitle=\bfseries,
  sharp corners,
  boxrule=1pt,
  drop fuzzy shadow
}

\newtcolorbox{chall}[1]{
  enhanced,
  colframe=red!70!black,
  colback=red!10!white,
  coltitle=white,
  fonttitle=\bfseries,
  title={\faWarning\hspace{0.5em}\textbf{#1}},
  sharp corners,
  boxrule=1pt,
  drop fuzzy shadow
}

\newtcolorbox{constraint}[1]{
  enhanced,
  colframe=orange!70!black,
  colback=orange!10!white,
  coltitle=white,
  fonttitle=\bfseries,
  title={\faWarning\hspace{0.5em}\textbf{#1}},
  sharp corners,
  boxrule=1pt,
  drop fuzzy shadow
}

\newtcolorbox{constraintsumm}{
  enhanced,
  colframe=orange!70!black,
  colback=orange!10!white,
  coltitle=white,
  fonttitle=\bfseries,
  sharp corners,
  boxrule=1pt,
  drop fuzzy shadow
}

\Crefname{section}{Sec.}{Secs.}
\Crefname{equation}{Eq.}{Eqs.}
\Crefname{figure}{Fig.}{Figs.}
\Crefname{tabular}{Tab.}{Tabs.}

\usepackage{pifont}
\newcommand{\cmark}{\ding{51}}%
\newcommand{\xmark}{\ding{55}}%

\usepackage{tikz}
\usepackage{fontawesome}


%% file: tables/pretrain_datasets.tex
\begin{table}[ht!]
\centering
    \resizebox{\columnwidth}{!}{
\begin{tabular}{@{}l|p{2cm}|c|c|p{3cm}|c@{}}
\toprule
\multicolumn{1}{l|}{\multirow{2}{*}{\bf Date}} &
  \multicolumn{1}{c|}{\multirow{2}{*}{\bf Name}} &
  \multicolumn{2}{c|}{\bf Size} &
  \multicolumn{1}{c|}{\multirow{2}{*}{\bf Sources}} &
  \multirow{2}{*}{\bf Public} \\ \cmidrule(lr){3-4}
\multicolumn{1}{l|}{} &
  \multicolumn{1}{c|}{} &
  \multicolumn{1}{c|}{\bf GB} &
  \multicolumn{1}{c|}{\bf Tokens${}^{*}$} &
  \multicolumn{1}{c|}{} &
 \\  \midrule  2014 & BookCorpus \citep{bookcorpus_1,bookcorpus_2} & 5 GB & 11 B & Novels  & \href{https://huggingface.co/datasets/bookcorpus}{Yes}  \\  
 \midrule  2019 & OSCAR \citep{oscar} & 6.3 T & ? & Webpages in 166 languages & \href{https://huggingface.co/datasets/oscar}{Yes}  \\ 
   \midrule 2019 & WebText \citep{gpt2} & 40 GB & ? & Webpages  & No \\
   \midrule 12.2020 & CC-100 \citep{conneau-etal-2020-unsupervised} & 2.5 TB & 292 B & Webpages in 100 Languages  & \href{https://huggingface.co/datasets/cc100}{Yes}   \\
   \midrule 12.2020 & The Pile \citep{gao2020pile, biderman2022datasheet} & 825 GB & 300 B & Science, Webpages, GitHub Code, Law, etc. & \href{https://pile.eleuther.ai/}{Yes} \\
   \midrule 2020 &  C4 \citep{raffel2022t5} &  745 GB & 156 B & Webpages  & \href{https://huggingface.co/datasets/c4}{Yes}\\
   \midrule 10.2020 &  mC4 \citep{xue-etal-2021-mt5} &  ? & 6.3 T & Webpages in 101 Languages & \href{https://huggingface.co/datasets/mc4/}{Yes}\\
   \midrule 2021 & MassiveText \citep{rae2021gopher} & 10.5 TB & 2.34 T & Webpages, Books, News, and Code & No\\
  \midrule   12.2021 & GLaM \citep{glam} & ? & 1.6 T & Webpages, Wikipedia, Conversations, Forums, Books, News  & No
  \\
\midrule  01.2022 & Infiniset \citep{lamda} & ? & 2.81 T & Forum dialogs, C4 data, Code, Wikipedia, Webpages  & No \\
  \midrule  06.2022 & ROOTS \citep{roots} & 1.61 TB & 2.34 T & Webpages in 46 languages and GitHub Code in 13 languages  & \href{https://huggingface.co/bigscience-data}{Yes}
  \\ \midrule  11.2022 & The Stack \citep{the_stack} & 6 TB & 235 B & GitHub Code in 30 languages  & \href{https://huggingface.co/datasets/bigcode/the-stack}{Yes}
    \\ \midrule  04.2023 & LLaMA \cite{touvronLLaMAOpenEfficient2023} / RedPajama \citep{together2023redpajama} & 2.7 TB & 1.2 T & Webpages, GitHub Code, Science, Wikipedia, Books & \href{https://huggingface.co/datasets/togethercomputer/RedPajama-Data-1T}{Yes}
  \\ \midrule  06.2023 & RefinedWeb \citep{penedoRefinedWebDatasetFalcon2023} & 2.8 TB & 600 B & Webpages & \href{https://huggingface.co/datasets/tiiuae/falcon-refinedweb}{Yes}  \\  \bottomrule
\end{tabular}}
\caption{\textbf{Overview of Selected Pre-Training Datasets.} Over the years, pre-training datasets have become more \emph{unfathomable}: they grew rapidly in size and diversity, and not all datasets are publicly available (we do not include datasets that have very little or no information available about them). Unless stated otherwise, the natural language is in English. ${}^{*}$ We report the number of tokens as provided by the respective paper based on their proposed tokenization scheme.}
\label{tab:datasets}
\end{table}

%% file: tables/all_llms.tex
\newcommand*\rot{\rotatebox{90}}
\newcommand*\enc{Enc.-Only}
\newcommand*\dec{Dec.-Only}
\newcommand*\encdec{Enc. \& Dec.}

\renewcommand{\arraystretch}{0.98}
\newcommand\setrow[1]{\gdef\rowmac{#1}#1\ignorespaces}
\newcommand\clearrow{\global\let\rowmac\relax}
\clearrow

\begin{table*}
  \setlength{\tabcolsep}{2pt} 
  \caption{\footnotesize
  \textbf{Overview of selected LLMs}. \missing{Missing details denoted by N/A}. For papers that investigate various model sizes, we only report the largest.  For each tokenizer entry with ``SP'', we could not extract from the respective paper whether BPE or Unigram tokenization was used. For publicly available code repositories and checkpoints, the corresponding \textcolor{cyan}{\cmark} is clickable. Abbreviations: Autoregressive blank filling (ARBF) \citep{du-etal-2022-glm}, Byte-pair encoding (BPE), Instruction-following (IF), Masked Language Modeling (MLM), Rotary  Next token prediction (NTP), SentencePiece (SP), Span Corruption (SC).
  }
  \label{table:all_llms}
   \footnotesize 
   
    \raggedright
    \resizebox{\textwidth}{!}{
    \begin{tabular}{ >{\rowmac}l >{\rowmac}l  >{\rowmac}c >{\rowmac}c >{\rowmac}r >{\rowmac}r >{\rowmac}c >{\rowmac}c >{\rowmac}c >{\rowmac}c >{\rowmac}c >{\rowmac}c >{\rowmac}c  >{\rowmac}c >{\rowmac}c <{\clearrow}}
      \rowcolor{white}
      \toprule
     \setrow{\bfseries} \rot{Date} & \rot{Name} & \rot{Organization} & \rot{Language} & \rot{\# Parameters} & \rot{\# Tokens}  & \rot{Architecture} & \rot{Train. Obj.} & \rot{Tokenizer} & \rot{Pos. Embed.} & \rot{\cblue{IF}} & \rot{\corange{MoE}} & \rot{\cyellow{Code avail.}} & \rot{\cpurple{Ckpt. avail.}} & \rot{\cgreen{Pre-trained}}  \\
      \midrule
      2018.11 & GPipe \citep{gpipe}& Google & Multil. & 6B & \missing{N/A}  & \encdec & NTP & BPE & Learned &\cblue{ \xmark  }&\corange{ \xmark }&\cyellow{ \cmark }&\cpurple{ \xmark}&\cgreen{ \xmark }\\ 
      2019.09 & Megatron-LM \citep{shoeybi2019megatron} & Microsoft & Eng. & 8.3B & 157B & \dec & NTP & BPE & Learned &\cblue{ \xmark }&\corange{ \xmark }&\cyellow{ \cmark }&\cpurple{ \xmark}&\cgreen{ \xmark }\\ 
      2019.10 & T5 \citep{raffel2022t5} & Google & Multil. & 11B & 1T  & \encdec & SC & SP & T5 &\cblue{ \xmark }&\corange{ \xmark }&\cyellow{ \href{https://github.com/google-research/text-to-text-transfer-transformer}{\cmark} }&\cpurple{ \href{https://github.com/google-research/text-to-text-transfer-transformer}{\cmark}}&\cgreen{ \xmark }\\
    2020.05 & GPT-3 \citep{brown2020gpt3} & OpenAI & Eng. & 175B & 300B& \dec & NTP & BPE & Learned &\cblue{ \xmark }&\corange{ \xmark }&\cyellow{ \xmark }&\cpurple{ \xmark}&\cgreen{ \xmark }\\ 
      2020.06 & GShard \citep{gshard} & Google & Multil. & 600B & 1T  & \encdec & NTP & SP & \missing{N/A} &\cblue{ \xmark }&\corange{ \cmark }&\cyellow{ \xmark }&\cpurple{ \xmark}&\cgreen{ \xmark }\\ 
    2020.10 & mT5 \citep{xue-etal-2021-mt5} & Google & Multil. & 13B & 1T  & \encdec & SC & SP & T5 &\cblue{ \xmark }&\corange{ \xmark }&\cyellow{ \href{https://github.com/google-research/multilingual-t5}{\cmark} }&\cpurple{ \href{https://github.com/google-research/multilingual-t5}{\cmark} }&\cgreen{ \xmark }\\
      2021.01 & Switch \citep{switch_transformer} & Google & Multil. & 1.5T & \missing{N/A}  & \encdec & SC & SP & T5 &\cblue{ \xmark }&\corange{ \cmark }&\cyellow{ \href{https://github.com/google/flaxformer/tree/main/flaxformer/architectures/moe}{\cmark} }&\cpurple{ \href{https://github.com/google-research/t5x/blob/main/docs/models.md}{\cmark} }&\cgreen{ \xmark }\\ 
    2021.03 & BASE \citep{base} & Meta & Eng. & 117B & \missing{N/A}  & \encdec & NTP &  BPE & Sinus. &\cblue{ \xmark }&\corange{ \cmark }&\cyellow{ \href{https://github.com/facebookresearch/fairseq}{\cmark} }&\cpurple{ \xmark}&\cgreen{ \xmark }\\ 
      2021.04 & PanGu-$\alpha$ \citep{pangu} & Huawei & Multil. & 200B & 317B  & \dec & NTP & BPE & Learned &\cblue{ \xmark }&\corange{ \xmark }&\cyellow{ \xmark }&\cpurple{ \xmark}&\cgreen{ \xmark }\\ 
    2021.05 & ByT5 \citep{xue-etal-2022-byt5} & Google & Multil. & 12.9B & 1T & \encdec & SC & \missing{N/A} & T5 &\cblue{ \xmark }&\corange{ \xmark }&\cyellow{ \href{https://github.com/google-research/byt5}{\cmark} }&\cpurple{ \href{https://github.com/google-research/byt5}{\cmark} }&\cgreen{ \xmark }\\ 
      2021.06 & CPM-2 \citep{cpm2} & Tsinghua Uni. & Multil. & 198B & \missing{N/A} & \encdec & SC & Custom & Sinus. &\cblue{ \xmark }&\corange{ \cmark }&\cyellow{ \cmark }&\cpurple{ \href{https://github.com/TsinghuaAI/CPM}{\cmark}}&\cgreen{ \xmark }\\ 
      2021.06 & nmT5 \citep{kale-etal-2021-nmt5} & Google & Multil. & 3.7B & 100B  & \encdec & MLM, NTP & SP & T5 &\cblue{ \xmark }&\corange{ \xmark }&\cyellow{ \xmark }&\cpurple{ \xmark }&\cgreen{ \cmark }\\ 
      2021.07 & ERNIE 3.0 \citep{sun2021ernie} & Baidu & Chin. & 10B & 375B & \encdec & Custom & BPE & Rel. &\cblue{ \xmark }&\corange{ \xmark }&\cyellow{ \xmark }&\cpurple{ \xmark}&\cgreen{ \xmark }\\ 
      2021.08 & Jurassic-1 \citep{lieber2021jurassic} & AI21 & Eng. & 178B & 300B  & \encdec & NTP & SP & Learned &\cblue{ \xmark  }&\corange{ \xmark }&\cyellow{ \xmark }&\cpurple{ \xmark}&\cgreen{ \xmark }\\ 
      2021.08 & ExT5 \citep{aribandi2022ext} & Google & Eng. & 11B & 1T & \encdec & SC, Custom & SP & T5 &\cblue{ \xmark }&\corange{ \xmark }&\cyellow{ \href{https://github.com/tensorflow/mesh}{\cmark} }&\cpurple{ \xmark}&\cgreen{ \xmark }\\ 
      2022.01 & FLAN-LaMDA \citep{wei2022finetuned} & Google & Eng. & 137B & 245M  & \dec & NTP & BPE & T5 &\cblue{ \xmark }&\corange{ \cmark }&\cyellow{ \xmark }&\cpurple{ \xmark }&\cgreen{ \cmark }\\
      2021.10 & M6-10T \citep{lin2021m6} & Alibaba & Eng. & 10T & \missing{N/A}  & Uni. \encdec & SC, NTP & SP & \missing{N/A} &\cblue{ \xmark }&\corange{ \xmark }&\cyellow{ \xmark }&\cpurple{ \xmark }&\cgreen{ \xmark }\\ 
      2021.10 & Yuan \citep{yuan} & Inspur AI & Chin. & 245B & 180B & \dec & NTP & BPE & \missing{N/A} &\cblue{ \xmark }&\corange{ \xmark }&\cyellow{ \xmark }&\cpurple{ \xmark}&\cgreen{ \xmark }\\ 
      2021.10 & T0 \citep{t0} & BigScience & Eng. & 11B & 12B & \encdec & SC, NTP & SP & T5 &\cblue{ \xmark }&\corange{ \xmark }&\cyellow{\href{https://github.com/bigscience-workshop/t-zero}{\cmark} }&\cpurple{ \href{https://github.com/bigscience-workshop/t-zero}{\cmark} }&\cgreen{ \cmark }\\
      2021.12 & Gopher \citep{rae2021gopher} & DeepMind & Eng. & 280B & 300B & \dec & NTP & SP & Rel. &\cblue{ \xmark }&\corange{ \xmark }&\cyellow{ \xmark }&\cpurple{ \xmark}&\cgreen{ \xmark }\\
      2021.12 & RETRO \citep{retro} & DeepMind & Eng. & 7B & 419B & \encdec & NTP (Ret.) & SP & Rel. &\cblue{ \xmark }&\corange{ \xmark }&\cyellow{ \xmark }&\cpurple{ \xmark}&\cgreen{ \xmark }\\ 
      2021.12 & GLaM \citep{glam} & Google & Multil. & 1.2T & 600B  & \dec & NTP & SP & Rel. &\cblue{ \xmark }&\corange{ \cmark }&\cyellow{ \xmark }&\cpurple{ \xmark}&\cgreen{ \xmark }\\ 
      2021.12 & WebGPT \citep{reiichiro2021webgpt} & OpenAI & Eng. & 175B & \missing{N/A} & \dec & NTP & BPE & Learned &\cblue{ \xmark  }&\corange{ \xmark }&\cyellow{ \xmark }&\cpurple{ \xmark }&\cgreen{ \cmark }\\
      2021.12 & FairSeq \citep{ott2019fairseq} & Meta & Eng. & 1.1T & 300B & \dec & NTP & BPE & Sinus. &\cblue{ \xmark }&\corange{ \cmark }&\cyellow{ \href{https://github.com/facebookresearch/fairseq/tree/main/examples/moe\_lm}{\cmark} }&\cpurple{ \href{https://github.com/facebookresearch/fairseq/tree/main/examples/moe\_lm}{\cmark} }&\cgreen{ \xmark }\\ 
      2021.12 & XGLM \citep{lin-etal-2022-shot} & Meta & Multil. & 7.5B & 500B  & \dec & NTP & Unigram & Sinus. &\cblue{ \xmark }&\corange{ \xmark }&\cyellow{ \href{https://github.com/facebookresearch/fairseq}{\cmark} }&\cpurple{ \href{https://github.com/facebookresearch/fairseq/tree/main/examples/xglm}{\cmark} }&\cgreen{ \xmark }\\
      2022.01 & LaMDA \citep{lamda} & Google & Eng. & 137B & 768B  & \dec & NTP & BPE & T5 &\cblue{ \xmark }&\corange{ \xmark }&\cyellow{ \xmark }&\cpurple{ \xmark }&\cgreen{ \xmark }\\ 
      2022.01 & MT-NLG \citep{smith2022using} & Microsoft & Eng. & 530B & 270B  & \dec & NTP & BPE & Sinus. &\cblue{ \xmark  }&\corange{ \xmark }&\cyellow{ \xmark }&\cpurple{ \xmark}&\cgreen{ \xmark }\\ 
      2022.02 & ST-MoE \citep{stmoe} & Google & Eng. & 269B & 1.5T  & \encdec & SC & SP & Sinus. &\cblue{ \xmark }&\corange{ \cmark }&\cyellow{ \href{https://github.com/tensorflow/mesh/blob/master/mesh\_tensorflow/transformer/moe.py}{\cmark} }&\cpurple{ \xmark}&\cgreen{ \xmark }\\ 
      2022.03 & InstructGPT \citep{ouyang2022gptinstruct} & OpenAI & Eng. & 175B & \missing{N/A}  & \dec & RLHF & BPE & Learned &\cblue{ \cmark }&\corange{ \xmark }&\cyellow{ \xmark }&\cpurple{ \xmark }&\cgreen{ \cmark }\\ 
      2022.03 & GopherCite \citep{gophercite} & DeepMind & Eng. & 280B & \missing{N/A}  & \dec & RLHF & BPE & Rel. &\cblue{ \cmark }&\corange{ \xmark }&\cyellow{ \xmark }&\cpurple{ \xmark }&\cgreen{ \cmark }\\ 
      2022.03 & sMLP \citep{smlp} & Meta & Eng. & 9.4B & \missing{N/A} & \encdec & NTP & BPE & Sinus. &\cblue{ \xmark }&\corange{ \cmark }&\cyellow{ \xmark }&\cpurple{ \xmark}&\cgreen{ \xmark }\\ 
      2022.03 & Chinchilla \citep{chinchilla} & DeepMind & Eng. & 70B & 1.4T  & \dec & NTP & SP & Rel. &\cblue{ \xmark }&\corange{ \xmark }&\cyellow{ \xmark }&\cpurple{ \xmark}&\cgreen{ \xmark }\\ 
      2022.04 & PaLM \citep{chowdhery2022palm} & Google & Multil. & 540B & 780B  & \dec & NTP & SP & RoPE &\cblue{ \xmark }&\corange{ \cmark }&\cyellow{ \xmark }&\cpurple{ \xmark }&\cgreen{ \xmark }\\ 
      2022.04 & GPT-NeoX \citep{black2022gptneox} & EleutherAI & Eng. & 20B & 472B  & \dec & NTP & BPE & RoPE &\cblue{ \xmark  }&\corange{ \xmark }&\cyellow{ \href{https://github.com/EleutherAI/gpt-neox}{\cmark} }&\cpurple{ \href{https://github.com/EleutherAI/gpt-neox\#pretrained-models}{\cmark} }&\cgreen{ \xmark }\\
      2022.04 & Tk-Instruct \citep{wang2022super} & AI2 & Eng. & 11B & 1B & \encdec & NTP & SP & T5 &\cblue{ \cmark }&\corange{ \xmark }&\cyellow{ \href{https://github.com/yizhongw/Tk-Instruct}{\cmark} }&\cpurple{ \href{https://huggingface.co/models?search=tk-instruct-}{\cmark} }&\cgreen{ \xmark }\\ 
      2022.04 & METRO-LM \citep{bajaj2022metro} & Microsoft & Eng. & 5.4B & 2T  & \enc & METRO & SP & T5 &\cblue{ \xmark }&\corange{ \xmark }&\cyellow{ \xmark }&\cpurple{ \xmark}&\cgreen{ \xmark }\\
      2022.04 & mGPT \citep{shliazhko2022mgpt} & Sber & Multi. & 13B & 440B  & \dec & NTP & BPE & Learned &\cblue{ \xmark  }&\corange{ \xmark }&\cyellow{ \href{https://github.com/ai-forever/mgpt}{\cmark} }&\cpurple{ \href{https://huggingface.co/sberbank-ai/mGPT}{\cmark} }&\cgreen{ \xmark }\\
      2022.05 & OPT \citep{zhang2022opt} & Meta & Eng. & 175B & 300B  & \dec & NTP & BPE & Learned &\cblue{ \xmark }&\corange{ \xmark }&\cyellow{ \href{https://github.com/facebookresearch/metaseq}{\cmark} }&\cpurple{ \href{https://huggingface.co/facebook/}{\cmark}}&\cgreen{ \xmark }\\
      2022.05 & UL2 \citep{tay2022ul2} & Google & Eng. & 20B & 1T& \encdec & MoD & Unigram & T5 &\cblue{ \xmark }&\corange{ \xmark }&\cyellow{ \xmark }&\cpurple{ \href{https://github.com/google-research/google-research/tree/master/ul2}{\cmark} }&\cgreen{ \xmark }\\ 
      2022.05 & DeepStruct \citep{wang-etal-2022-deepstruct} & UC Berkeley & Eng. & 10B & \missing{N/A}  & \encdec & Struc. & BPE & Sinus. &\cblue{ \xmark }&\corange{ \xmark }&\cyellow{ \xmark }&\cpurple{ \xmark }&\cgreen{ \xmark }\\ 
      2022.07 & Minerva \citep{minerva} & Google & Eng. & 540B & 26B & \dec & NTP & SP & RoPE &\cblue{ \xmark }&\corange{ \xmark }&\cyellow{ \xmark }&\cpurple{ \xmark}&\cgreen{ \xmark }\\ 
      2022.08 & PEER \citep{peer} & Meta & Eng. & 11B & 5B & \encdec & NTP & SP & T5 &\cblue{ \xmark }&\corange{ \xmark }&\cyellow{ \xmark }&\cpurple{ \xmark }&\cgreen{ \cmark }\\
      2022.08 & AlexaTM \citep{soltan2022alexatm} & Amazon & Multil. & 20B & 1T & \encdec & MoD, NTP & SP & Sinus. &\cblue{ \xmark }&\corange{ \xmark }&\cyellow{ \xmark }&\cpurple{ \href{https://github.com/amazon-science/alexa-teacher-models}{\cmark} }&\cgreen{ \cmark }\\
      2022.10 & GLM-130B \citep{zeng2022glm130b} & Tsinghua Uni. & Multil. & 130B & 400B & Uni. \encdec & ARBF & SP & RoPE &\cblue{ \xmark }&\corange{ \xmark }&\cyellow{ \href{https://github.com/THUDM/GLM-130B}{\cmark} }&\cpurple{ \href{https://docs.google.com/forms/d/e/1FAIpQLSehr5Dh_i3TwACmFFi8QEgIVNYGmSPwV0GueIcsUev0NEfUug/viewform}{\cmark}}&\cgreen{ \xmark }\\ 
      2022.10 & U-PaLM \citep{tay2022upalm} & Google & Eng. & 540B & 1.3B & \dec & MoD & SP & RoPE &\cblue{ \xmark }&\corange{ \cmark }&\cyellow{ \xmark }&\cpurple{ \xmark }&\cgreen{ \cmark }\\ 
      2022.10 & FLAN-PaLM \citep{flan} & Google & Eng. & 540B & 1.4B & \dec & NTP & SP & RoPE &\cblue{ \cmark }&\corange{ \cmark }&\cyellow{ \xmark }&\cpurple{ \xmark }&\cgreen{ \cmark }\\ 
      2022.11 & BLOOM \citep{scao2021bloom} & BigScience & Multil. & 176B & 366B  & \dec & NTP & BPE & ALiBi &\cblue{ \xmark }&\corange{ \xmark }&\cyellow{ \href{https://github.com/bigscience-workshop/Megatron-DeepSpeed}{\cmark} }&\cpurple{ \href{https://huggingface.co/bigscience/bloom}{\cmark}}&\cgreen{ \xmark }\\ 
      2022.11 & Galactica \citep{taylor2022galactica} & Meta & Eng. & 120B & 450B  & \dec & NTP & BPE & Learned &\cblue{ \xmark }&\corange{ \xmark }&\cyellow{ \href{https://github.com/paperswithcode/galai}{\cmark} }&\cpurple{ \href{https://huggingface.co/facebook/galactica-120b}{\cmark}}&\cgreen{ \xmark }\\ 
      2022.11 & Atlas \citep{atlas} & Meta & Eng. & 11B & \missing{N/A} & \encdec & MLM & BPE & T5 &\cblue{ \xmark }&\corange{ \xmark }&\cyellow{ \href{https://github.com/facebookresearch/atlas}{\cmark} }&\cpurple{ \href{https://github.com/facebookresearch/atlas\#models}{\cmark} }&\cgreen{ \cmark }\\  
      2022.11 & BLOOMZ \citep{muennighoff2022crosslingual} & BigScience & Multil. & 176B & 13B & \dec & NTP & BPE & ALiBi &\cblue{ \cmark }&\corange{ \xmark }&\cyellow{ \href{https://github.com/bigscience-workshop/Megatron-DeepSpeed}{\cmark} }&\cpurple{ \href{https://huggingface.co/bigscience/bloomz}{\cmark}}&\cgreen{ \cmark }\\ 
      2022.11 & mT0 \citep{muennighoff2022crosslingual} & BigScience & Multil. & 13B & 13B & \encdec & NTP & SP & T5 &\cblue{ \cmark }&\corange{ \xmark }&\cyellow{ \href{https://github.com/google-research/t5x/blob/main/docs/usage/finetune.md}{\cmark} }&\cpurple{ \href{https://huggingface.co/bigscience/mt0-xxl}{\cmark} }&\cgreen{ \cmark }\\
      2022.12 & OPT-IML \citep{opt_iml} & Meta & Eng. & 175B & 2B  & \dec & NTP & BPE & Sinus. &\cblue{ \cmark }&\corange{ \xmark }&\cyellow{ \href{https://github.com/facebookresearch/metaseq}{\cmark} }&\cpurple{ \href{https://docs.google.com/forms/d/19jE4WVSMcUy1YcVFGZcU2Q4KvDWGgwFy1tYqGZ02x1k/}{\cmark} }&\cgreen{ \cmark }\\ 
      2022.12 & Med-PaLM \citep{medpalm} & Google & Eng. & 540B & 0B  & \dec & NTP & SP & RoPE &\cblue{ \xmark }&\corange{ \xmark }&\cyellow{ \xmark }&\cpurple{ \xmark }&\cgreen{ \cmark }\\ 
    2023.02 & LLaMA\{-I\} \citep{touvronLLaMAOpenEfficient2023} & Meta & Eng. & 65B & 1.4T & \dec & NTP & BPE & RoPE &\cblue{ \cmark }&\corange{ \xmark }&\cyellow{ \href{https://github.com/facebookresearch/llama}{\cmark} }&\cpurple{ \href{https://docs.google.com/forms/d/e/1FAIpQLSfqNECQnMkycAp2jP4Z9TFX0cGR4uf7b_fBxjY_OjhJILlKGA/viewform}{\cmark} }&\cgreen{ \xmark }\\ 
    2023.03 & PanGu-{$\Sigma$} \citep{ren2023pangusigma} & Huawei & Multil. & 1T & 329B  & \dec & NTP & BPE & Learned &\cblue{ \xmark }&\corange{ \cmark }&\cyellow{ \xmark }&\cpurple{ \xmark }&\cgreen{ \cmark  }\\
    2023.03 & CoLT5 \citep{ainslie2023colt5} & Google & Eng. & 5.3B & 1T & \encdec & MoD & \missing{N/A} & T5 &\cblue{ \xmark }&\corange{ \xmark }&\cyellow{ \xmark }&\cpurple{ \xmark }&\cgreen{ \xmark }\\
    2023.03 & BloombergGPT \citep{wu2023bloomberggpt} & Bloomberg & Eng. & 50B & 569B & \dec & NTP & Unigram & ALiBi &\cblue{ \xmark }&\corange{ \xmark }&\cyellow{ \xmark }&\cpurple{ \xmark }&\cgreen{ \xmark }\\
    2023.04 & Cerebras-GPT \citep{dey2023cerebrasgpt} & Cerebras & Eng. & 13B & 257B  & \dec & NTP & BPE & RoPE &\cblue{ \xmark }&\corange{ \xmark }&\cyellow{ \xmark }&\cpurple{ \href{https://huggingface.co/cerebras}{\cmark} }&\cgreen{ \xmark }\\
    2023.04 & Pythia \citep{biderman2023pythia} & EleutherAI & Eng. & 12B & 300B  & \dec & NTP & BPE & RoPE &\cblue{ \xmark }&\corange{ \xmark }&\cyellow{ \href{https://github.com/EleutherAI/pythia}{\cmark} }&\cpurple{ \href{https://github.com/EleutherAI/pythia}{\cmark}  }&\cgreen{ \xmark }\\
    2023.04 & WizardLM \citep{xu2023wizardlm} & Microsoft & Eng. & 30B & \missing{N/A}  & \dec & NTP & BPE & RoPE &\cblue{ \cmark }&\corange{ \xmark }&\cyellow{ \href{https://github.com/nlpxucan/WizardLM}{\cmark} }&\cpurple{ \href{https://huggingface.co/WizardLM/WizardLM-30B-V1.0}{\cmark}  }&\cgreen{ \cmark }\\
    2023.05 & Guanaco \citep{dettmers_qlora_2023} & Univ. of Washington & Multil. & 65B & 82M  & \dec & NTP & BPE & RoPE &\cblue{ \cmark }&\corange{ \xmark }&\cyellow{ \xmark }&\cpurple{ \href{https://huggingface.co/timdettmers/guanaco-65b-merged}{\cmark}  }&\cgreen{ \cmark }\\
    2023.04 & RWKV \citep{peng2023rwkv} & RWKV & Eng. & 14B & \missing{N/A} & \dec & NTP & BPE & RoPE &\cblue{ \cmark }&\corange{ \xmark }&\cyellow{ \href{https://github.com/nlpxucan/WizardLM}{\cmark} }&\cpurple{ \href{https://huggingface.co/WizardLM/WizardLM-30B-V1.0}{\cmark}  }&\cgreen{ \cmark }\\
    2023.06 & Orca \citep{mukherjee2023orca} & Microsoft & Eng. & 13B & \missing{N/A} & \dec & NTP & BPE & RoPE &\cblue{ \cmark }&\corange{ \xmark }&\cyellow{ \xmark }&\cpurple{ \xmark  }&\cgreen{ \cmark }\\
    2023.07 & LLaMA 2 \citep{touvronLlamaOpenFoundationa} & Meta & Eng. & 70B & 2T & \dec & NTP & BPE & RoPE &\cblue{ \cmark }&\corange{ \xmark }&\cyellow{ \href{https://github.com/facebookresearch/llama}{\cmark} }&\cpurple{ \href{https://huggingface.co/meta-llama}{\cmark}  }&\cgreen{ \cmark }\\
      \bottomrule
        \end{tabular}}
\end{table*}

%% file: tables/applications.tex
\tikzstyle{mybox2}=[
    rectangle,
    draw=hidden-grey,
    rounded corners,
    text opacity=1,
    minimum height=1.5em,
    minimum width=5em,
    inner sep=2pt,
    align=center,
    fill opacity=.5,
    ]

\tikzstyle{leaf}=[mybox2,minimum height=1em,
fill=hiddendraw!40, text width=16em,  text=black,align=left,font=\tiny,
inner xsep=2pt,
inner ysep=1pt,
]
\tikzstyle{leaf_prompt}=[mybox2,minimum height=1em,
fill=hidden-blue!40, text width=16em,  text=black,align=left,font=\tiny,
inner xsep=2pt,
inner ysep=1pt,
]

\tikzstyle{leaf_fine}=[mybox2,minimum height=1em,
fill=hidden-orange!40, text width=16em,  text=black,align=left,font=\tiny,
inner xsep=2pt,
inner ysep=1pt,
]

\tikzstyle{leaf_evaluation}=[mybox2,minimum height=1em,
fill=hidden-grey!40, text width=16em,  text=black,align=left,font=\tiny,
inner xsep=2pt,
inner ysep=1pt,
]

\begin{figure*}[tp]
  \centering
\begin{forest}
  forked edges,
  for tree={
  grow=east,
  reversed=true,
  anchor=base west,
  parent anchor=east,
  child anchor=west,
  base=left,
  font=\small,
  rectangle,
  draw=hidden-grey,
  rounded corners,align=left,
  minimum width=2.5em,
s sep=3pt,
inner xsep=1.5pt,
inner ysep=1pt,
ver/.style={rotate=90, child anchor=north, parent anchor=south, anchor=center},
  },
  where level=1{text width=8em,font=\scriptsize,}{},
  where level=2{text width=9em,font=\tiny}{},
[Applications, ver
[Chatbots \ref{sec:chatbots}
[BlenderBot3 (OPT-175) \cite{blenderbot3}{,} Bard (LaMDA{,} PaLM2) \cite{lamda}{,} \\ Sparrow (Chinchilla) \cite{sparrow}{,} ChatGPT (GPT-3.5{,} GPT-4) \cite{chatgpt}{,} \\ OpenAssistant (LLaMA) \cite{kopf2023openassistant}, leaf_fine]
[GPT-4 Technical Report \cite{openai2023gpt4}{,} Sparks of AGI (GPT-4) \cite{bubeck2023sparks}{,} \\ Capabilities of ChatGPT \cite{chatgpt_jack_of_all_trades}, leaf_evaluation]
]
[Computational Biology \ref{sec:computational_biology}
[Proteins
[ESM-2 \cite{lin2022language}{,} 
ProtT5 \cite{elnaggar2020prottrans}{,}
ProtST \cite{xu2023protst}{,}
CaLM \cite{outeiral2022codon}{,} 
ProGen \cite{madani2023large}{,}\\
IgLM \cite{shuai2021generative}{,} xTrimoPGLM~\cite{xTrimoPGLM}
, leaf]
]
[Genomics
[GenSLM \cite{zvyagin2022genslms}{,} Nucleotide Transformers \cite{dalla2023nucleotide}, leaf]
]
]
[Computer Programming \ref{sec:computer_programming}
[InCoder \cite{incoder}{,}
CodeGen \cite{codegen}{,} AlphaCode~\cite{li2022competition} {,} SantaCoder \cite{santacoder}{,} \\ Polycoder \cite{code_evaluation}{,} phi-1 \cite{gunasekar2023textbooks},leaf]
[Codex (GPT-3) \cite{codex}, leaf_fine]
[Self-Debugging (Codex) \cite{chen2023teaching}{,} ViperGPT (Codex) \cite{suris2023vipergpt}{,} \\ RepoCoder \cite{zhang2023repocoder}{,} Repo-Level Prompt Generator \cite{shrivastava2022repository} , leaf_prompt]
]
[Creative Work \ref{sec:creative_work}
[Long Form
[Dramatron (Chinchilla) \cite{mirowski2022co}{,} Re3 (GPT-3) \cite{yang2022re3}{,} \\ Detailed Outline Control (GPT-3) \cite{yang2022doc}, leaf_prompt]
]
[Short Form
[CoPoet (T5{,} T0) \cite{chakrabarty2022help}{,} Spindle - Interactive Fiction (GPT-3) \cite{calderwoodspinning}, leaf_fine]
[Cross-lingual Short Stories (PaLM) \cite{razumovskaia2022little}{,} ReelFramer (GPT-4) \cite{wang2023reelframer}, leaf_prompt]
[Idea Generation \cite{haase2023artificial}, leaf_evaluation]
]
[Visual
[LayoutGPT \cite{fengLayoutGPTCompositionalVisual2023}{,} LLM Grounded Diffusion~\cite{lian2023llmgrounded}, leaf_prompt]
]
]
[Knowledge Work \ref{sec:knowledge_work}
[Galactica \cite{taylor2022galactica}{,} BloombergGPT \cite{wu2023bloomberggpt}, leaf]
[Scientific NERRE (GPT-3) \cite{dunn2022structured}, leaf_fine]
[Data Analysis (GPT-4)~\cite{luo2021synthesizing}, leaf_prompt]
[Professional Exams \cite{bommarito2023gpt}{,} News Summarization \cite{news_summary}{,} \\ Email Management \cite{thiergart2021understanding}{,} Academic Paper Review (GPT-4) \cite{liuReviewerGPTExploratoryStudy2023}, leaf_evaluation]
]
[Law \ref{sec:law}
[Legal Question Answering
[Legal Entailment (GPT-3.5) \cite{yu2022legal}{,} Bar Examination (GPT-3.5) \cite{bommarito2022gpt}
,leaf_fine]
[Explaining Legal Concepts (GPT-4 + Retrieval) \cite{savelka2023explaining}, leaf_prompt]
[Law School (ChatGPT) \cite{choi2023chatgpt}{,} Bar Examination (GPT-4) \cite{katz2023gpt} \\ Statutory Reasoning (GPT-3.5) \cite{blair2023can}{,} Law Professor (ChatGPT) \cite{pettinato2023chatgpt}{,}\\  Summarizing Judgments (GPT-3.5) \cite{deroy2023ready}{,} Litigation (ChatGPT) \cite{iu2023chatgpt}, leaf_evaluation]
]
[Case Prediction
[US Supreme Court (GPT-2 + GPT-3) \cite{hamilton2023blind}, leaf_fine]
]
]
[Medicine \ref{sec:medicine}
[Medical Question Answering
[PubMedGPT \cite{pubmedgpt}{,} GatorTronGPT~\cite{peng2023study}, leaf]
[MedPaLM(2) (PaLM)~\citep{medpalm, medpalm2}{,} ChatDoctor (LLaMA) \cite{yunxiang2023chatdoctor}, leaf_fine]
[GPT-3.5 + Retrieval \cite{lievin2022can}, leaf_prompt]
[Medical Challenge Problems (GPT-4) \cite{nori2023capabilities}{,} \\ Triage and Diagnosis (GPT-3)~\cite{levine2023diagnostic}{,} \\ Surgical Knowledge QA (GPT-4) \cite{chatgptoperating}{,} \\ Social Media - Genetics Questions (ChatGPT) \cite{duong2023analysis}{,} \\
Social Media - General Questions (ChatGPT) \cite{ayers2023comparing}{,} \\
Ophthalmology QA (ChatGPT) \cite{ChatGPTOphthalmology}{,}\\ Medical Summarization (GPT-3.5{,} ChatGPT)~\cite{tang2023evaluating}, leaf_evaluation]
]
[Medical Information Retrieval
[Medical Acronym Disambiguation (T5) \cite{rajkomar2022deciphering}{,} \\ Adverse Drug Event Extraction \cite{gu2023distilling}, leaf_fine]
[Clinical Information Extraction (InstructGPT) \cite{agrawal2022large}, leaf_prompt]
]
]
[Reasoning \ref{sec:reasoning}
[Self Improvement (PaLM) \cite{self_improvement}{,} Processed Based Fine-Tuning \cite{solving_math_problems} 
,leaf_fine]
[DIVERSE (GPT-3.5) \cite{better_reasoners}{,} Socratic Sub-Questions (GPT-3) \cite{Shridhar2022AutomaticGO}{,} \\ Mathematical Formalization (Codex) \cite{gadgiltowards}, leaf_prompt]
[Causal Factors in Performance \cite{robustness_math_lm}{,} Analogical Reasoning \cite{analogical_reasoning}{,}\\ Causal Reasoning~\citep{kiciman2023causal, gao2023chatgpt, srivastava2022beyond, jin2023large, lampinen2023passive}{,} \\ Common-Sense Reasoning~\cite{valmeekam2023large}, leaf_evaluation]
]
[Robotics \ref{sec:robotics}
[PaLM-E \cite{palm-e}, leaf_fine]
[SayCan (PaLM + Scoring) \cite{saycan}{,} 
ChatGPT for Robotics \cite{chatgpt_robo}{,} \\
REFLECT (GPT-4) \cite{liu2023reflect}{,} Code as Policies (Codex) \cite{liang2023code}{,} \\
PROGPROMPT (Codex) \cite{singh2022progprompt}{,} Inner Monologue \cite{huang2022inner}{,} \\Statler (GPT-3.5)~\cite{yoneda2023statler}, leaf_prompt]
]
[Social Sciences \ref{sec:social_sciences}
[Using LLMs to Model Human Behavior \cite{aher2022using, griffin2023susceptibility}{,} \\ Analyzing Behavioral Characteristics of LLMs \cite{miotto2022gpt, pellert2023ai}{,} \\ Simulating Social Relationships with LLMs \cite{park2023generative}, leaf_prompt]
]
[Synthetic Training Data \ref{sec:synthetic_data}
[Automated Labeling (GPT-3) \cite{label_cost}{,} AugGPT (ChatGPT) \cite{dai2023chataug}{,} \\ Labeling + Generation (GPT-3) \cite{ding2022gpt}{,} \\ Information Retrieval (GPT-3) \cite{bonifacio2022inpars}{,} \\ Decompositional Distillation (GPT-3) \cite{shridhar2022distilling}{,} \\ Code `Textbooks' (GPT-3.5)~\cite{gunasekar2023textbooks}{,} GPT3Mix \cite{yoo-etal-2021-gpt3mix-leveraging} , leaf_prompt]
]
]
\end{forest}
\caption{Overview of LLM Applications. Color = Level of Model Adaption (\textcolor{red}{Pre-Trained}, \textcolor{orange}{Fine-Tuned}, \textcolor{blue}{Prompting Strategy}, \textcolor{gray}{Evaluation}).}
\label{fig:applications}
\end{figure*}

%% file: main.bbl
\begin{thebibliography}{688}
\expandafter\ifx\csname natexlab\endcsname\relax\def\natexlab#1{#1}\fi

\bibitem[{Chi()}]{ChipsShortage}

\newblock \href
  {https://fortune.com/2023/06/05/openai-ceo-sam-altman-chatgpt-chips-shortage-gpus-london-leak/}
  {A blog post detailed a {Sam} {Altman} freakout about a huge chips shortage
  threatening {OpenAI}. {Then} it was taken down}.

\bibitem[{Ope()}]{OpenLLMLeaderboard}

\newblock \href
  {https://huggingface.co/spaces/HuggingFaceH4/open_llm_leaderboard} {Open
  {LLM} {Leaderboard} - a {Hugging} {Face} {Space} by {HuggingFaceH4}}.

\bibitem[{Rep()}]{ReproducibilityPyTorchDocumentation}

\newblock \href
  {https://pytorch.org/docs/stable/notes/randomness.html#avoiding-nondeterministic-algorithms}
  {Reproducibility — {PyTorch} 2.0 documentation}.

\bibitem[{Neg(2023)}]{NegativePromptsText2023}
 2023.
\newblock \href
  {https://community.openai.com/t/negative-prompts-for-text-generation/203346}
  {Negative prompts for text generation}.
\newblock Section: Prompting.

\bibitem[{Rep(2023)}]{Reproducibility2023}
 2023.
\newblock \href
  {https://en.wikipedia.org/w/index.php?title=Reproducibility&oldid=1163331755#Terminology}
  {Reproducibility}.
\newblock Page Version ID: 1163331755.

\bibitem[{Abbas et~al.(2023)Abbas, Tirumala, Simig, Ganguli, and
  Morcos}]{abbas2023semdedup}
A.~Abbas, K.~Tirumala, D.~Simig, S.~Ganguli and A.~S. Morcos. 2023.
\newblock Semdedup: Data-efficient learning at web-scale through semantic
  deduplication.
\newblock \emph{arXiv preprint arXiv:2303.09540}.

\bibitem[{Abernethy et~al.(2023)Abernethy, Agarwal, Marinov, and
  Warmuth}]{Abernethy2023AMF}
J.~D. Abernethy, A.~Agarwal, T.~V. Marinov and M.~K. Warmuth. 2023.
\newblock A mechanism for sample-efficient in-context learning for sparse
  retrieval tasks.
\newblock \emph{ArXiv}, abs/2305.17040.

\bibitem[{Adiwardana et~al.(2020)Adiwardana, Luong, So, Hall, Fiedel,
  Thoppilan, Yang, Kulshreshtha, Nemade, Lu et~al.}]{adiwardana2020towards}
D.~Adiwardana, M.-T. Luong, D.~R. So, J.~Hall, N.~Fiedel, R.~Thoppilan,
  Z.~Yang, A.~Kulshreshtha et~al. 2020.
\newblock Towards a human-like open-domain chatbot.
\newblock \emph{arXiv preprint arXiv:2001.09977}.

\bibitem[{Agarwal et~al.(2021)Agarwal, Schwarzer, Castro, Courville, and
  Bellemare}]{agarwalDeepReinforcementLearning2021}
R.~Agarwal, M.~Schwarzer, P.~S. Castro, A.~C. Courville and M.~Bellemare. 2021.
\newblock \href
  {https://proceedings.neurips.cc/paper_files/paper/2021/hash/f514cec81cb148559cf475e7426eed5e-Abstract.html}
  {Deep {Reinforcement} {Learning} at the {Edge} of the {Statistical}
  {Precipice}}.
\newblock In \emph{Advances in {Neural} {Information} {Processing} {Systems}},
  volume~34, pages 29304--29320. Curran Associates, Inc.

\bibitem[{Agrawal et~al.(2022{\natexlab{a}})Agrawal, Hegselmann, Lang, Kim, and
  Sontag}]{agrawal2022large}
M.~Agrawal, S.~Hegselmann, H.~Lang, Y.~Kim and D.~Sontag. 2022{\natexlab{a}}.
\newblock Large language models are zero-shot clinical information extractors.
\newblock \emph{arXiv preprint arXiv:2205.12689}.

\bibitem[{Agrawal et~al.(2022{\natexlab{b}})Agrawal, Alberti, Huot, Maynez, Ma,
  Ruder, Ganchev, Das, and Lapata}]{agrawal2022qameleon}
P.~Agrawal, C.~Alberti, F.~Huot, J.~Maynez, J.~Ma, S.~Ruder, K.~Ganchev, D.~Das
  et~al. 2022{\natexlab{b}}.
\newblock Qameleon: Multilingual qa with only 5 examples.
\newblock \emph{arXiv preprint arXiv:2211.08264}.

\bibitem[{Aher et~al.(2022)Aher, Arriaga, and Kalai}]{aher2022using}
G.~Aher, R.~I. Arriaga and A.~T. Kalai. 2022.
\newblock Using large language models to simulate multiple humans.
\newblock \emph{arXiv preprint arXiv:2208.10264}.

\bibitem[{Ahia et~al.(2023)Ahia, Kumar, Gonen, Kasai, Mortensen, Smith, and
  Tsvetkov}]{ahia2023all}
O.~Ahia, S.~Kumar, H.~Gonen, J.~Kasai, D.~R. Mortensen, N.~A. Smith and
  Y.~Tsvetkov. 2023.
\newblock Do all languages cost the same? tokenization in the era of commercial
  language models.
\newblock \emph{arXiv preprint arXiv:2305.13707}.

\bibitem[{Ahn et~al.(2022)Ahn, Brohan, Brown, Chebotar, Cortes, David, Finn,
  Gopalakrishnan, Hausman, Herzog et~al.}]{saycan}
M.~Ahn, A.~Brohan, N.~Brown, Y.~Chebotar, O.~Cortes, B.~David, C.~Finn,
  K.~Gopalakrishnan et~al. 2022.
\newblock Do as i can, not as i say: Grounding language in robotic affordances.
\newblock \emph{arXiv preprint arXiv:2204.01691}.

\bibitem[{Ainslie et~al.(2023)Ainslie, Lei, de~Jong, Onta{\~n}{\'o}n, Brahma,
  Zemlyanskiy, Uthus, Guo, Lee-Thorp, Tay et~al.}]{ainslie2023colt5}
J.~Ainslie, T.~Lei, M.~de~Jong, S.~Onta{\~n}{\'o}n, S.~Brahma, Y.~Zemlyanskiy,
  D.~Uthus, M.~Guo et~al. 2023.
\newblock Colt5: Faster long-range transformers with conditional computation.
\newblock \emph{arXiv preprint arXiv:2303.09752}.

\bibitem[{Aky{\"u}rek et~al.(2023)Aky{\"u}rek, Schuurmans, Andreas, Ma, and
  Zhou}]{rek2023what}
E.~Aky{\"u}rek, D.~Schuurmans, J.~Andreas, T.~Ma and D.~Zhou. 2023.
\newblock \href {https://openreview.net/forum?id=0g0X4H8yN4I} {What learning
  algorithm is in-context learning? investigations with linear models}.
\newblock In \emph{The Eleventh International Conference on Learning
  Representations}.

\bibitem[{Allal et~al.(2023)Allal, Li, Kocetkov, Mou, Akiki, Ferrandis,
  Muennighoff, Mishra, Gu, Dey, Umapathi, Anderson, Zi, Poirier, Schoelkopf,
  Troshin, Abulkhanov, Romero, Lappert, De~Toni, del Río, Liu, Bose,
  Bhattacharyya, Zhuo, Yu, Villegas, Zocca, Mangrulkar, Lansky, Nguyen,
  Contractor, Villa, Li, Bahdanau, Jernite, Hughes, Fried, Guha, de~Vries, and
  von Werra}]{santacoder}
L.~B. Allal, R.~Li, D.~Kocetkov, C.~Mou, C.~Akiki, C.~M. Ferrandis,
  N.~Muennighoff, M.~Mishra et~al. 2023.
\newblock \href {https://doi.org/10.48550/ARXIV.2301.03988} {Santacoder: don't
  reach for the stars!}

\bibitem[{Andreas(2022)}]{agent_models}
J.~Andreas. 2022.
\newblock \href {https://doi.org/10.48550/ARXIV.2212.01681} {Language models as
  agent models}.

\bibitem[{Anil et~al.(2022)Anil, Wu, Andreassen, Lewkowycz, Misra, Ramasesh,
  Slone, Gur-Ari, Dyer, and Neyshabur}]{anilExploringLengthGeneralization2022}
C.~Anil, Y.~Wu, A.~Andreassen, A.~Lewkowycz, V.~Misra, V.~Ramasesh, A.~Slone,
  G.~Gur-Ari et~al. 2022.
\newblock \href {http://arxiv.org/abs/2207.04901} {Exploring {Length}
  {Generalization} in {Large} {Language} {Models}}.
\newblock ArXiv:2207.04901 [cs].

\bibitem[{Anil et~al.(2023)Anil, Dai, Firat, Johnson, Lepikhin, Passos,
  Shakeri, Taropa, Bailey, Chen et~al.}]{anil2023palm}
R.~Anil, A.~M. Dai, O.~Firat, M.~Johnson, D.~Lepikhin, A.~Passos, S.~Shakeri,
  E.~Taropa et~al. 2023.
\newblock Palm 2 technical report.
\newblock \emph{arXiv preprint arXiv:2305.10403}.

\bibitem[{Antaki et~al.(2023)Antaki, Touma, Milad, El-Khoury, and
  Duval}]{ChatGPTOphthalmology}
F.~Antaki, S.~Touma, D.~Milad, J.~El-Khoury and R.~Duval. 2023.
\newblock \href {https://doi.org/10.1101/2023.01.22.23284882} {Evaluating the
  performance of chatgpt in ophthalmology: An analysis of its successes and
  shortcomings}.
\newblock \emph{medRxiv}.

\bibitem[{Argyle et~al.(2022)Argyle, Busby, Fulda, Gubler, Rytting, and
  Wingate}]{argyle2022out}
L.~P. Argyle, E.~C. Busby, N.~Fulda, J.~Gubler, C.~Rytting and D.~Wingate.
  2022.
\newblock Out of one, many: Using language models to simulate human samples.
\newblock \emph{arXiv preprint arXiv:2209.06899}.

\bibitem[{Aribandi et~al.(2022)Aribandi, Tay, Schuster, Rao, Zheng, Mehta,
  Zhuang, Tran, Bahri, Ni, Gupta, Hui, Ruder, and Metzler}]{aribandi2022ext}
V.~Aribandi, Y.~Tay, T.~Schuster, J.~Rao, H.~S. Zheng, S.~V. Mehta, H.~Zhuang,
  V.~Q. Tran et~al. 2022.
\newblock \href {https://openreview.net/forum?id=Vzh1BFUCiIX} {Ext5: Towards
  extreme multi-task scaling for transfer learning}.
\newblock In \emph{International Conference on Learning Representations}.

\bibitem[{Arora et~al.(2022)Arora, Narayan, Chen, Orr, Guha, Bhatia, Chami,
  Sala, and Ré}]{arora2022ama}
S.~Arora, A.~Narayan, M.~F. Chen, L.~Orr, N.~Guha, K.~Bhatia, I.~Chami, F.~Sala
  et~al. 2022.
\newblock \href {https://doi.org/10.48550/ARXIV.2210.02441} {Ask me anything: A
  simple strategy for prompting language models}.

\bibitem[{Asai et~al.(2022)Asai, Schick, Lewis, Chen, Izacard, Riedel,
  Hajishirzi, and Yih}]{taskaware_retrieval}
A.~Asai, T.~Schick, P.~Lewis, X.~Chen, G.~Izacard, S.~Riedel, H.~Hajishirzi and
  W.-t. Yih. 2022.
\newblock \href {https://doi.org/10.48550/ARXIV.2211.09260} {Task-aware
  retrieval with instructions}.

\bibitem[{Asher et~al.(2023)Asher, Bhar, Chaturvedi, Hunter, and
  Paul}]{asherLimitsLearningLanguage2023}
N.~Asher, S.~Bhar, A.~Chaturvedi, J.~Hunter and S.~Paul. 2023.
\newblock \href {http://arxiv.org/abs/2306.12213} {Limits for {Learning} with
  {Language} {Models}}.
\newblock ArXiv:2306.12213 [cs].

\bibitem[{Ashton and Lee(2009)}]{ashton2009hexaco}
M.~C. Ashton and K.~Lee. 2009.
\newblock The hexaco--60: A short measure of the major dimensions of
  personality.
\newblock \emph{Journal of personality assessment}, 91(4):340--345.

\bibitem[{Austin et~al.(2021)Austin, Odena, Nye, Bosma, Michalewski, Dohan,
  Jiang, Cai, Terry, Le et~al.}]{austin2021program}
J.~Austin, A.~Odena, M.~Nye, M.~Bosma, H.~Michalewski, D.~Dohan, E.~Jiang,
  C.~Cai et~al. 2021.
\newblock Program synthesis with large language models.
\newblock \emph{arXiv preprint arXiv:2108.07732}.

\bibitem[{AUTOMATIC1111(2023)}]{automatic1111StableDiffusionWeb2023}
AUTOMATIC1111. 2023.
\newblock \href {https://github.com/AUTOMATIC1111/stable-diffusion-webui}
  {Stable {Diffusion} web {UI}}.
\newblock Original-date: 2022-08-22T14:05:26Z.

\bibitem[{Ayers et~al.(2023)Ayers, Poliak, Dredze, Leas, Zhu, Kelley, Faix,
  Goodman, Longhurst, Hogarth et~al.}]{ayers2023comparing}
J.~W. Ayers, A.~Poliak, M.~Dredze, E.~C. Leas, Z.~Zhu, J.~B. Kelley, D.~J.
  Faix, A.~M. Goodman et~al. 2023.
\newblock Comparing physician and artificial intelligence chatbot responses to
  patient questions posted to a public social media forum.
\newblock \emph{JAMA internal medicine}.

\bibitem[{Bai et~al.(2022)Bai, Kadavath, Kundu, Askell, Kernion, Jones, Chen,
  Goldie, Mirhoseini, McKinnon et~al.}]{bai2022constitutional}
Y.~Bai, S.~Kadavath, S.~Kundu, A.~Askell, J.~Kernion, A.~Jones, A.~Chen,
  A.~Goldie et~al. 2022.
\newblock Constitutional ai: Harmlessness from ai feedback.
\newblock \emph{arXiv preprint arXiv:2212.08073}.

\bibitem[{Bajaj et~al.(2018)Bajaj, Campos, Craswell, Deng, Gao, Liu, Majumder,
  McNamara, Mitra, Nguyen, Rosenberg, Song, Stoica, Tiwary, and
  Wang}]{bajaj2018ms}
P.~Bajaj, D.~Campos, N.~Craswell, L.~Deng, J.~Gao, X.~Liu, R.~Majumder,
  A.~McNamara et~al. 2018.
\newblock \href {http://arxiv.org/abs/1611.09268} {Ms marco: A human generated
  machine reading comprehension dataset}.

\bibitem[{Bajaj et~al.(2022)Bajaj, Xiong, Ke, Liu, He, Tiwary, Liu, Bennett,
  Song, and Gao}]{bajaj2022metro}
P.~Bajaj, C.~Xiong, G.~Ke, X.~Liu, D.~He, S.~Tiwary, T.-Y. Liu, P.~Bennett
  et~al. 2022.
\newblock Metro: Efficient denoising pretraining of large scale autoencoding
  language models with model generated signals.
\newblock \emph{arXiv preprint arXiv:2204.06644}.

\bibitem[{Bakhtin et~al.(2019)Bakhtin, Gross, Ott, Deng, Ranzato, and
  Szlam}]{bakhtinRealFakeLearning2019}
A.~Bakhtin, S.~Gross, M.~Ott, Y.~Deng, M.~Ranzato and A.~Szlam. 2019.
\newblock \href {http://arxiv.org/abs/1906.03351} {Real or {Fake}? {Learning}
  to {Discriminate} {Machine} from {Human} {Generated} {Text}}.
\newblock ArXiv:1906.03351 [cs, stat].

\bibitem[{Balestriero et~al.(2021)Balestriero, Pesenti, and
  LeCun}]{balestriero2021learning}
R.~Balestriero, J.~Pesenti and Y.~LeCun. 2021.
\newblock Learning in high dimension always amounts to extrapolation.
\newblock \emph{arXiv preprint arXiv:2110.09485}.

\bibitem[{Bandy and Vincent(2021)}]{bookcorpus_2}
J.~Bandy and N.~Vincent. 2021.
\newblock \href {https://doi.org/10.48550/ARXIV.2105.05241} {Addressing
  "documentation debt" in machine learning research: A retrospective datasheet
  for bookcorpus}.

\bibitem[{Barham et~al.(2022)Barham, Chowdhery, Dean, Ghemawat, Hand, Hurt,
  Isard, Lim, Pang, Roy et~al.}]{barham2022pathways}
P.~Barham, A.~Chowdhery, J.~Dean, S.~Ghemawat, S.~Hand, D.~Hurt, M.~Isard,
  H.~Lim et~al. 2022.
\newblock Pathways: Asynchronous distributed dataflow for ml.
\newblock \emph{Proceedings of Machine Learning and Systems}, 4:430--449.

\bibitem[{Bavarian et~al.(2022)Bavarian, Jun, Tezak, Schulman, McLeavey,
  Tworek, and Chen}]{bavarian2022efficient}
M.~Bavarian, H.~Jun, N.~Tezak, J.~Schulman, C.~McLeavey, J.~Tworek and M.~Chen.
  2022.
\newblock Efficient training of language models to fill in the middle.
\newblock \emph{arXiv preprint arXiv:2207.14255}.

\bibitem[{Belrose et~al.(2023)Belrose, Furman, Smith, Halawi, Ostrovsky,
  McKinney, Biderman, and Steinhardt}]{belrose2023eliciting}
N.~Belrose, Z.~Furman, L.~Smith, D.~Halawi, I.~Ostrovsky, L.~McKinney,
  S.~Biderman and J.~Steinhardt. 2023.
\newblock \href {http://arxiv.org/abs/2303.08112} {Eliciting latent predictions
  from transformers with the tuned lens}.

\bibitem[{Ben~Zaken et~al.(2022)Ben~Zaken, Goldberg, and
  Ravfogel}]{ben-zaken-etal-2022-bitfit}
E.~Ben~Zaken, Y.~Goldberg and S.~Ravfogel. 2022.
\newblock \href {https://doi.org/10.18653/v1/2022.acl-short.1} {{B}it{F}it:
  Simple parameter-efficient fine-tuning for transformer-based masked
  language-models}.
\newblock In \emph{Proceedings of the 60th Annual Meeting of the Association
  for Computational Linguistics (Volume 2: Short Papers)}, pages 1--9, Dublin,
  Ireland. Association for Computational Linguistics.

\bibitem[{Biderman et~al.(2022)Biderman, Bicheno, and
  Gao}]{biderman2022datasheet}
S.~Biderman, K.~Bicheno and L.~Gao. 2022.
\newblock Datasheet for the pile.
\newblock \emph{arXiv preprint arXiv:2201.07311}.

\bibitem[{Biderman et~al.(2023{\natexlab{a}})Biderman, Prashanth, Sutawika,
  Schoelkopf, Anthony, Purohit, and
  Raff}]{bidermanEmergentPredictableMemorization2023a}
S.~Biderman, U.~S. Prashanth, L.~Sutawika, H.~Schoelkopf, Q.~Anthony,
  S.~Purohit and E.~Raff. 2023{\natexlab{a}}.
\newblock \href {http://arxiv.org/abs/2304.11158} {Emergent and {Predictable}
  {Memorization} in {Large} {Language} {Models}}.
\newblock ArXiv:2304.11158 [cs].

\bibitem[{Biderman and Scheirer(2021)}]{biderman2021pitfalls}
S.~Biderman and W.~J. Scheirer. 2021.
\newblock \href {http://arxiv.org/abs/2011.02832} {Pitfalls in machine learning
  research: Reexamining the development cycle}.

\bibitem[{Biderman et~al.(2023{\natexlab{b}})Biderman, Schoelkopf, Anthony,
  Bradley, O'Brien, Hallahan, Khan, Purohit, Prashanth, Raff, Skowron,
  Sutawika, and Van Der~Wal}]{biderman2023pythia}
S.~Biderman, H.~Schoelkopf, Q.~G. Anthony, H.~Bradley, K.~O'Brien, E.~Hallahan,
  M.~A. Khan, S.~Purohit et~al. 2023{\natexlab{b}}.
\newblock \href {https://proceedings.mlr.press/v202/biderman23a.html} {Pythia:
  A suite for analyzing large language models across training and scaling}.
\newblock In \emph{Proceedings of the 40th International Conference on Machine
  Learning}, volume 202 of \emph{Proceedings of Machine Learning Research},
  pages 2397--2430. PMLR.

\bibitem[{Biderman(2023)}]{BlancheMinerva_2023}
S.~R. Biderman. 2023.
\newblock [...] we aren’t running out of text data any time soon. ml
  researchers massively underestimate how much text is out there.
\newblock
  \url{https://twitter.com/BlancheMinerva/status/1644154144431677442?s=20}.
\newblock Accessed: 2023-05-28.

\bibitem[{Birhane et~al.(2021)Birhane, Prabhu, and
  Kahembwe}]{birhane2021multimodal}
A.~Birhane, V.~U. Prabhu and E.~Kahembwe. 2021.
\newblock Multimodal datasets: misogyny, pornography, and malignant
  stereotypes.
\newblock \emph{arXiv preprint arXiv:2110.01963}.

\bibitem[{Black et~al.(2022)Black, Biderman, Hallahan, Anthony, Gao, Golding,
  He, Leahy, McDonell, Phang, Pieler, Prashanth, Purohit, Reynolds, Tow, Wang,
  and Weinbach}]{black2022gptneox}
S.~Black, S.~Biderman, E.~Hallahan, Q.~Anthony, L.~Gao, L.~Golding, H.~He,
  C.~Leahy et~al. 2022.
\newblock \href {https://doi.org/10.48550/ARXIV.2204.06745} {Gpt-neox-20b: An
  open-source autoregressive language model}.

\bibitem[{Blair-Stanek et~al.(2023)Blair-Stanek, Holzenberger, and
  Van~Durme}]{blair2023can}
A.~Blair-Stanek, N.~Holzenberger and B.~Van~Durme. 2023.
\newblock Can gpt-3 perform statutory reasoning?
\newblock \emph{arXiv preprint arXiv:2302.06100}.

\bibitem[{Bommarito et~al.(2023)Bommarito, Bommarito, Katz, and
  Katz}]{bommarito2023gpt}
J.~Bommarito, M.~Bommarito, D.~M. Katz and J.~Katz. 2023.
\newblock Gpt as knowledge worker: A zero-shot evaluation of (ai) cpa
  capabilities.
\newblock \emph{arXiv preprint arXiv:2301.04408}.

\bibitem[{Bommarito~II and Katz(2022)}]{bommarito2022gpt}
M.~Bommarito~II and D.~M. Katz. 2022.
\newblock Gpt takes the bar exam.
\newblock \emph{arXiv preprint arXiv:2212.14402}.

\bibitem[{Bonifacio et~al.(2022)Bonifacio, Abonizio, Fadaee, and
  Nogueira}]{bonifacio2022inpars}
L.~Bonifacio, H.~Abonizio, M.~Fadaee and R.~Nogueira. 2022.
\newblock \href {https://doi.org/10.1145/3477495.3531863} {Inpars: Unsupervised
  dataset generation for information retrieval}.
\newblock In \emph{Proceedings of the 45th International ACM SIGIR Conference
  on Research and Development in Information Retrieval}, SIGIR '22, page
  2387–2392, New York, NY, USA. Association for Computing Machinery.

\bibitem[{Borgeaud et~al.(2021)Borgeaud, Mensch, Hoffmann, Cai, Rutherford,
  Millican, Driessche, Lespiau, Damoc, Clark et~al.}]{retro}
S.~Borgeaud, A.~Mensch, J.~Hoffmann, T.~Cai, E.~Rutherford, K.~Millican,
  G.~v.~d. Driessche, J.-B. Lespiau et~al. 2021.
\newblock Improving language models by retrieving from trillions of tokens.
\newblock \emph{arXiv preprint arXiv:2112.04426}.

\bibitem[{Borji(2023)}]{borjiCategoricalArchiveChatGPT2023}
A.~Borji. 2023.
\newblock \href {http://arxiv.org/abs/2302.03494} {A {Categorical} {Archive} of
  {ChatGPT} {Failures}}.
\newblock ArXiv:2302.03494 [cs].

\bibitem[{Borzunov et~al.(2022)Borzunov, Baranchuk, Dettmers, Ryabinin,
  Belkada, Chumachenko, Samygin, and Raffel}]{borzunov2022petals}
A.~Borzunov, D.~Baranchuk, T.~Dettmers, M.~Ryabinin, Y.~Belkada,
  A.~Chumachenko, P.~Samygin and C.~Raffel. 2022.
\newblock Petals: Collaborative inference and fine-tuning of large models.
\newblock \emph{arXiv preprint arXiv:2209.01188}.

\bibitem[{Bos et~al.(2020)Bos, Schemer, Corbu, Hameleers, Andreadis, Schulz,
  Schmuck, Reinemann, and Fawzi}]{bos2020effects}
L.~Bos, C.~Schemer, N.~Corbu, M.~Hameleers, I.~Andreadis, A.~Schulz,
  D.~Schmuck, C.~Reinemann et~al. 2020.
\newblock The effects of populism as a social identity frame on persuasion and
  mobilisation: Evidence from a 15-country experiment.
\newblock \emph{European Journal of Political Research}, 59(1):3--24.

\bibitem[{Britz et~al.(2017)Britz, Guan, and Luong}]{britz2017efficient}
D.~Britz, M.~Y. Guan and M.-T. Luong. 2017.
\newblock Efficient attention using a fixed-size memory representation.
\newblock \emph{arXiv preprint arXiv:1707.00110}.

\bibitem[{Broder et~al.(1998)Broder, Charikar, Frieze, and
  Mitzenmacher}]{broder1998min}
A.~Z. Broder, M.~Charikar, A.~M. Frieze and M.~Mitzenmacher. 1998.
\newblock Min-wise independent permutations.
\newblock In \emph{Proceedings of the thirtieth annual ACM symposium on Theory
  of computing}, pages 327--336.

\bibitem[{Brown et~al.(2021)Brown, Bun, Feldman, Smith, and
  Talwar}]{brown2021memorization}
G.~Brown, M.~Bun, V.~Feldman, A.~Smith and K.~Talwar. 2021.
\newblock When is memorization of irrelevant training data necessary for
  high-accuracy learning?
\newblock In \emph{Proceedings of the 53rd annual ACM SIGACT symposium on
  theory of computing}, pages 123--132.

\bibitem[{Brown et~al.(2020)Brown, Mann, Ryder, Subbiah, Kaplan, Dhariwal,
  Neelakantan, Shyam, Sastry, Askell, Agarwal, Herbert-Voss, Krueger, Henighan,
  Child, Ramesh, Ziegler, Wu, Winter, Hesse, Chen, Sigler, Litwin, Gray, Chess,
  Clark, Berner, McCandlish, Radford, Sutskever, and Amodei}]{brown2020gpt3}
T.~Brown, B.~Mann, N.~Ryder, M.~Subbiah, J.~D. Kaplan, P.~Dhariwal,
  A.~Neelakantan, P.~Shyam et~al. 2020.
\newblock \href
  {https://proceedings.neurips.cc/paper/2020/file/1457c0d6bfcb4967418bfb8ac142f64a-Paper.pdf}
  {Language models are few-shot learners}.
\newblock In \emph{Advances in Neural Information Processing Systems},
  volume~33, pages 1877--1901. Curran Associates, Inc.

\bibitem[{Brundage et~al.(2018)Brundage, Avin, Clark, Toner, Eckersley,
  Garfinkel, Dafoe, Scharre, Zeitzoff, Filar et~al.}]{brundage2018malicious}
M.~Brundage, S.~Avin, J.~Clark, H.~Toner, P.~Eckersley, B.~Garfinkel, A.~Dafoe,
  P.~Scharre et~al. 2018.
\newblock The malicious use of artificial intelligence: Forecasting,
  prevention, and mitigation.
\newblock \emph{arXiv preprint arXiv:1802.07228}.

\bibitem[{Bubeck et~al.(2023)Bubeck, Chandrasekaran, Eldan, Gehrke, Horvitz,
  Kamar, Lee, Lee, Li, Lundberg, Nori, Palangi, Ribeiro, and
  Zhang}]{bubeck2023sparks}
S.~Bubeck, V.~Chandrasekaran, R.~Eldan, J.~Gehrke, E.~Horvitz, E.~Kamar,
  P.~Lee, Y.~T. Lee et~al. 2023.
\newblock \href {http://arxiv.org/abs/2303.12712} {Sparks of artificial general
  intelligence: Early experiments with gpt-4}.

\bibitem[{Burns et~al.(2022)Burns, Ye, Klein, and
  Steinhardt}]{burns2022discovering}
C.~Burns, H.~Ye, D.~Klein and J.~Steinhardt. 2022.
\newblock \href {http://arxiv.org/abs/2212.03827} {Discovering latent knowledge
  in language models without supervision}.

\bibitem[{Calderwood et~al.(2022)Calderwood, Wardrip-Fruin, and
  Mateas}]{calderwoodspinning}
A.~Calderwood, N.~Wardrip-Fruin and M.~Mateas. 2022.
\newblock Spinning coherent interactive fiction through foundation model
  prompts.
\newblock \emph{International Conference of Computation and Creativity}.

\bibitem[{Carlini et~al.(2023{\natexlab{a}})Carlini, Jagielski, Choquette-Choo,
  Paleka, Pearce, Anderson, Terzis, Thomas, and
  Tramèr}]{carliniPoisoningWebScaleTraining2023}
N.~Carlini, M.~Jagielski, C.~A. Choquette-Choo, D.~Paleka, W.~Pearce,
  H.~Anderson, A.~Terzis, K.~Thomas et~al. 2023{\natexlab{a}}.
\newblock \href {http://arxiv.org/abs/2302.10149} {Poisoning {Web}-{Scale}
  {Training} {Datasets} is {Practical}}.
\newblock ArXiv:2302.10149 [cs].

\bibitem[{Carlini et~al.(2019)Carlini, Liu, Erlingsson, Kos, and
  Song}]{carlini2019secret}
N.~Carlini, C.~Liu, {\'U}.~Erlingsson, J.~Kos and D.~Song. 2019.
\newblock The secret sharer: Evaluating and testing unintended memorization in
  neural networks.
\newblock In \emph{USENIX Security Symposium}, volume 267.

\bibitem[{Carlini et~al.(2023{\natexlab{b}})Carlini, Nasr, Choquette-Choo,
  Jagielski, Gao, Awadalla, Koh, Ippolito, Lee, Tramer, and
  Schmidt}]{carlini2023aligned}
N.~Carlini, M.~Nasr, C.~A. Choquette-Choo, M.~Jagielski, I.~Gao, A.~Awadalla,
  P.~W. Koh, D.~Ippolito et~al. 2023{\natexlab{b}}.
\newblock \href {http://arxiv.org/abs/2306.15447} {Are aligned neural networks
  adversarially aligned?}

\bibitem[{Carlini et~al.(2020)Carlini, Tramer, Wallace, Jagielski,
  Herbert-Voss, Lee, Roberts, Brown, Song, Erlingsson, Oprea, and
  Raffel}]{extracting_data}
N.~Carlini, F.~Tramer, E.~Wallace, M.~Jagielski, A.~Herbert-Voss, K.~Lee,
  A.~Roberts, T.~Brown et~al. 2020.
\newblock \href {https://doi.org/10.48550/ARXIV.2012.07805} {Extracting
  training data from large language models}.

\bibitem[{Casper et~al.(2023)Casper, Lin, Kwon, Culp, and
  Hadfield-Menell}]{casper2023explore}
S.~Casper, J.~Lin, J.~Kwon, G.~Culp and D.~Hadfield-Menell. 2023.
\newblock Explore, establish, exploit: Red teaming language models from
  scratch.
\newblock \emph{arXiv preprint arXiv:2306.09442}.

\bibitem[{Chakrabarty et~al.(2022)Chakrabarty, Padmakumar, and
  He}]{chakrabarty2022help}
T.~Chakrabarty, V.~Padmakumar and H.~He. 2022.
\newblock Help me write a poem: Instruction tuning as a vehicle for
  collaborative poetry writing.
\newblock \emph{arXiv preprint arXiv:2210.13669}.

\bibitem[{Chalkidis et~al.(2019)Chalkidis, Androutsopoulos, and
  Aletras}]{chalkidis2019neural}
I.~Chalkidis, I.~Androutsopoulos and N.~Aletras. 2019.
\newblock Neural legal judgment prediction in english.
\newblock \emph{arXiv preprint arXiv:1906.02059}.

\bibitem[{Chalkidis et~al.(2020)Chalkidis, Fergadiotis, Malakasiotis, Aletras,
  and Androutsopoulos}]{chalkidis2020legal}
I.~Chalkidis, M.~Fergadiotis, P.~Malakasiotis, N.~Aletras and
  I.~Androutsopoulos. 2020.
\newblock Legal-bert: The muppets straight out of law school.
\newblock \emph{arXiv preprint arXiv:2010.02559}.

\bibitem[{Chang et~al.(2023)Chang, Wang, Wang, Wu, Zhu, Chen, Yang, Yi, Wang,
  Wang, Ye, Zhang, Chang, Yu, Yang, and Xie}]{changSurveyEvaluationLarge2023a}
Y.~Chang, X.~Wang, J.~Wang, Y.~Wu, K.~Zhu, H.~Chen, L.~Yang, X.~Yi et~al. 2023.
\newblock \href {https://doi.org/10.48550/arXiv.2307.03109} {A {Survey} on
  {Evaluation} of {Large} {Language} {Models}}.
\newblock ArXiv:2307.03109 [cs].

\bibitem[{Chen et~al.(2023{\natexlab{a}})Chen, Cheng, ao~Gengyang, Li, Zeng,
  Wang, Jing, Liu, Zeng, Dong, Tang, and Song}]{xTrimoPGLM}
B.~Chen, X.~Cheng, L.~ao~Gengyang, S.~Li, X.~Zeng, B.~Wang, G.~Jing, C.~Liu
  et~al. 2023{\natexlab{a}}.
\newblock \href {https://doi.org/10.1101/2023.07.05.547496} {xtrimopglm:
  Unified 100b-scale pre-trained transformer for deciphering the language of
  protein}.
\newblock \emph{bioRxiv}.

\bibitem[{Chen et~al.(2023{\natexlab{b}})Chen, Borgeaud, Irving, Lespiau,
  Sifre, and Jumper}]{chen2023accelerating}
C.~Chen, S.~Borgeaud, G.~Irving, J.-B. Lespiau, L.~Sifre and J.~Jumper.
  2023{\natexlab{b}}.
\newblock Accelerating large language model decoding with speculative sampling.
\newblock \emph{arXiv preprint arXiv:2302.01318}.

\bibitem[{Chen et~al.(2023{\natexlab{c}})Chen, Zaharia, and
  Zou}]{chenFrugalGPTHowUse2023}
L.~Chen, M.~Zaharia and J.~Zou. 2023{\natexlab{c}}.
\newblock \href {https://doi.org/10.48550/arXiv.2305.05176} {{FrugalGPT}: {How}
  to {Use} {Large} {Language} {Models} {While} {Reducing} {Cost} and
  {Improving} {Performance}}.
\newblock ArXiv:2305.05176 [cs].

\bibitem[{Chen et~al.(2023{\natexlab{d}})Chen, Zaharia, and
  Zou}]{chenHowChatGPTBehavior2023}
L.~Chen, M.~Zaharia and J.~Zou. 2023{\natexlab{d}}.
\newblock \href {http://arxiv.org/abs/2307.09009} {How is {ChatGPT}'s behavior
  changing over time?}
\newblock ArXiv:2307.09009 [cs].

\bibitem[{Chen et~al.(2021)Chen, Tworek, Jun, Yuan, Pinto, Kaplan, Edwards,
  Burda, Joseph, Brockman, Ray, Puri, Krueger, Petrov, Khlaaf, Sastry, Mishkin,
  Chan, Gray, Ryder, Pavlov, Power, Kaiser, Bavarian, Winter, Tillet, Such,
  Cummings, Plappert, Chantzis, Barnes, Herbert-Voss, Guss, Nichol, Paino,
  Tezak, Tang, Babuschkin, Balaji, Jain, Saunders, Hesse, Carr, Leike, Achiam,
  Misra, Morikawa, Radford, Knight, Brundage, Murati, Mayer, Welinder, McGrew,
  Amodei, McCandlish, Sutskever, and Zaremba}]{codex}
M.~Chen, J.~Tworek, H.~Jun, Q.~Yuan, H.~P. d.~O. Pinto, J.~Kaplan, H.~Edwards,
  Y.~Burda et~al. 2021.
\newblock \href {https://doi.org/10.48550/ARXIV.2107.03374} {Evaluating large
  language models trained on code}.

\bibitem[{Chen et~al.(2023{\natexlab{e}})Chen, Papangelis, Tao, Kim, Rosenbaum,
  Liu, Yu, and Hakkani-Tur}]{chen2023places}
M.~Chen, A.~Papangelis, C.~Tao, S.~Kim, A.~Rosenbaum, Y.~Liu, Z.~Yu and
  D.~Hakkani-Tur. 2023{\natexlab{e}}.
\newblock Places: Prompting language models for social conversation synthesis.
\newblock \emph{arXiv preprint arXiv:2302.03269}.

\bibitem[{Chen et~al.(2023{\natexlab{f}})Chen, Wong, Chen, and
  Tian}]{chen2023extending}
S.~Chen, S.~Wong, L.~Chen and Y.~Tian. 2023{\natexlab{f}}.
\newblock \href {http://arxiv.org/abs/2306.15595} {Extending context window of
  large language models via positional interpolation}.

\bibitem[{Chen et~al.(2023{\natexlab{g}})Chen, Zhang, Jaiswal, Liu, and
  Wang}]{chen2023sparse}
T.~Chen, Z.~Zhang, A.~Jaiswal, S.~Liu and Z.~Wang. 2023{\natexlab{g}}.
\newblock \href {http://arxiv.org/abs/2303.01610} {Sparse moe as the new
  dropout: Scaling dense and self-slimmable transformers}.

\bibitem[{Chen et~al.(2023{\natexlab{h}})Chen, Lin, Sch{\"a}rli, and
  Zhou}]{chen2023teaching}
X.~Chen, M.~Lin, N.~Sch{\"a}rli and D.~Zhou. 2023{\natexlab{h}}.
\newblock Teaching large language models to self-debug.
\newblock \emph{arXiv preprint arXiv:2304.05128}.

\bibitem[{Cheng et~al.(2023)Cheng, Li, and Bing}]{cheng2023gpt4}
L.~Cheng, X.~Li and L.~Bing. 2023.
\newblock \href {http://arxiv.org/abs/2305.15038} {Is gpt-4 a good data
  analyst?}

\bibitem[{Choe et~al.(2019)Choe, Al-Rfou, Guo, Lee, and
  Constant}]{choeBridgingGapTokenizerFree2019}
D.~Choe, R.~Al-Rfou, M.~Guo, H.~Lee and N.~Constant. 2019.
\newblock \href {http://arxiv.org/abs/1908.10322} {Bridging the {Gap} for
  {Tokenizer}-{Free} {Language} {Models}}.
\newblock ArXiv:1908.10322 [cs].

\bibitem[{Choi et~al.(2023)Choi, Hickman, Monahan, and
  Schwarcz}]{choi2023chatgpt}
J.~H. Choi, K.~E. Hickman, A.~Monahan and D.~Schwarcz. 2023.
\newblock Chatgpt goes to law school.
\newblock \emph{Available at SSRN}.

\bibitem[{Choromanski et~al.(2020)Choromanski, Likhosherstov, Dohan, Song,
  Gane, Sarlos, Hawkins, Davis, Mohiuddin, Kaiser
  et~al.}]{choromanski2020rethinking}
K.~Choromanski, V.~Likhosherstov, D.~Dohan, X.~Song, A.~Gane, T.~Sarlos,
  P.~Hawkins, J.~Davis et~al. 2020.
\newblock Rethinking attention with performers.
\newblock \emph{arXiv preprint arXiv:2009.14794}.

\bibitem[{Chowdhery et~al.(2022)Chowdhery, Narang, Devlin, Bosma, Mishra,
  Roberts, Barham, Chung, Sutton, Gehrmann et~al.}]{chowdhery2022palm}
A.~Chowdhery, S.~Narang, J.~Devlin, M.~Bosma, G.~Mishra, A.~Roberts, P.~Barham,
  H.~W. Chung et~al. 2022.
\newblock Palm: Scaling language modeling with pathways.
\newblock \emph{arXiv preprint arXiv:2204.02311}.

\bibitem[{Christ et~al.(2023)Christ, Gunn, and
  Zamir}]{christUndetectableWatermarksLanguage}
M.~Christ, S.~Gunn and O.~Zamir. 2023.
\newblock \href {https://eprint.iacr.org/2023/763} {Undetectable {Watermarks}
  for {Language} {Models}}.

\bibitem[{Christiano et~al.(2017)Christiano, Leike, Brown, Martic, Legg, and
  Amodei}]{christiano2017rlhf}
P.~Christiano, J.~Leike, T.~B. Brown, M.~Martic, S.~Legg and D.~Amodei. 2017.
\newblock \href {https://doi.org/10.48550/ARXIV.1706.03741} {Deep reinforcement
  learning from human preferences}.

\bibitem[{Christianson et~al.(2001)Christianson, Hollingworth, Halliwell, and
  Ferreira}]{christianson2001thematic}
K.~Christianson, A.~Hollingworth, J.~F. Halliwell and F.~Ferreira. 2001.
\newblock Thematic roles assigned along the garden path linger.
\newblock \emph{Cognitive psychology}, 42(4):368--407.

\bibitem[{Chung(2023)}]{chung2023details}
H.~W. Chung. 2023.
\newblock \href {https://twitter.com/hwchung27/status/1668729544701001729}
  {Missing model details (tweet)}.

\bibitem[{Chung et~al.(2023)Chung, Garcia, Roberts, Tay, Firat, Narang, and
  Constant}]{chung2023unimax}
H.~W. Chung, X.~Garcia, A.~Roberts, Y.~Tay, O.~Firat, S.~Narang and
  N.~Constant. 2023.
\newblock \href {https://openreview.net/forum?id=kXwdL1cWOAi} {Unimax: Fairer
  and more effective language sampling for large-scale multilingual
  pretraining}.
\newblock In \emph{The Eleventh International Conference on Learning
  Representations}.

\bibitem[{Chung et~al.(2020)Chung, Garrette, Tan, and
  Riesa}]{chung-etal-2020-improving}
H.~W. Chung, D.~Garrette, K.~C. Tan and J.~Riesa. 2020.
\newblock \href {https://doi.org/10.18653/v1/2020.emnlp-main.367} {Improving
  multilingual models with language-clustered vocabularies}.
\newblock In \emph{Proceedings of the 2020 Conference on Empirical Methods in
  Natural Language Processing (EMNLP)}, pages 4536--4546, Online. Association
  for Computational Linguistics.

\bibitem[{Chung et~al.(2022)Chung, Hou, Longpre, Zoph, Tay, Fedus, Li, Wang,
  Dehghani, Brahma, Webson, Gu, Dai, Suzgun, Chen, Chowdhery, Castro-Ros,
  Pellat, Robinson, Valter, Narang, Mishra, Yu, Zhao, Huang, Dai, Yu, Petrov,
  Chi, Dean, Devlin, Roberts, Zhou, Le, and Wei}]{flan}
H.~W. Chung, L.~Hou, S.~Longpre, B.~Zoph, Y.~Tay, W.~Fedus, Y.~Li, X.~Wang
  et~al. 2022.
\newblock \href {https://doi.org/10.48550/ARXIV.2210.11416} {Scaling
  instruction-finetuned language models}.

\bibitem[{Clark et~al.(2022)Clark, Garrette, Turc, and
  Wieting}]{clark-etal-2022-canine}
J.~H. Clark, D.~Garrette, I.~Turc and J.~Wieting. 2022.
\newblock \href {https://doi.org/10.1162/tacl_a_00448} {Canine: Pre-training an
  efficient tokenization-free encoder for language representation}.
\newblock \emph{Transactions of the Association for Computational Linguistics},
  10:73--91.

\bibitem[{Cobbe et~al.(2021)Cobbe, Kosaraju, Bavarian, Chen, Jun, Kaiser,
  Plappert, Tworek, Hilton, Nakano, Hesse, and Schulman}]{gsm8k}
K.~Cobbe, V.~Kosaraju, M.~Bavarian, M.~Chen, H.~Jun, L.~Kaiser, M.~Plappert,
  J.~Tworek et~al. 2021.
\newblock \href {https://doi.org/10.48550/ARXIV.2110.14168} {Training verifiers
  to solve math word problems}.

\bibitem[{Cohen et~al.(2022)Cohen, Ryu, Chow, Keller, Greenberg, Hassidim,
  Fink, Matias, Szpektor, Boutilier et~al.}]{cohen2022dynamic}
D.~Cohen, M.~Ryu, Y.~Chow, O.~Keller, I.~Greenberg, A.~Hassidim, M.~Fink,
  Y.~Matias et~al. 2022.
\newblock Dynamic planning in open-ended dialogue using reinforcement learning.
\newblock \emph{arXiv preprint arXiv:2208.02294}.

\bibitem[{Cohen et~al.(2023)Cohen, Hamri, Geva, and
  Globerson}]{cohenLMVsLM2023}
R.~Cohen, M.~Hamri, M.~Geva and A.~Globerson. 2023.
\newblock \href {http://arxiv.org/abs/2305.13281} {{LM} vs {LM}: {Detecting}
  {Factual} {Errors} via {Cross} {Examination}}.
\newblock ArXiv:2305.13281 [cs].

\bibitem[{Computer(2023)}]{together2023redpajama}
T.~Computer. 2023.
\newblock \href {https://github.com/togethercomputer/RedPajama-Data}
  {Redpajama: An open source recipe to reproduce llama training dataset}.

\bibitem[{Conmy et~al.(2023)Conmy, Mavor-Parker, Lynch, Heimersheim, and
  Garriga-Alonso}]{conmy2023towards}
A.~Conmy, A.~N. Mavor-Parker, A.~Lynch, S.~Heimersheim and A.~Garriga-Alonso.
  2023.
\newblock Towards automated circuit discovery for mechanistic interpretability.
\newblock \emph{arXiv preprint arXiv:2304.14997}.

\bibitem[{Conneau et~al.(2020)Conneau, Khandelwal, Goyal, Chaudhary, Wenzek,
  Guzm{\'a}n, Grave, Ott, Zettlemoyer, and
  Stoyanov}]{conneau-etal-2020-unsupervised}
A.~Conneau, K.~Khandelwal, N.~Goyal, V.~Chaudhary, G.~Wenzek, F.~Guzm{\'a}n,
  E.~Grave, M.~Ott et~al. 2020.
\newblock \href {https://doi.org/10.18653/v1/2020.acl-main.747} {Unsupervised
  cross-lingual representation learning at scale}.
\newblock In \emph{Proceedings of the 58th Annual Meeting of the Association
  for Computational Linguistics}, pages 8440--8451, Online. Association for
  Computational Linguistics.

\bibitem[{Cyphert(2021)}]{cyphert2021human}
A.~B. Cyphert. 2021.
\newblock A human being wrote this law review article: Gpt-3 and the practice
  of law.
\newblock \emph{UC Davis L. Rev.}, 55:401.

\bibitem[{Dai et~al.(2022{\natexlab{a}})Dai, Dong, Hao, Sui, Chang, and
  Wei}]{dai-etal-2022-knowledge}
D.~Dai, L.~Dong, Y.~Hao, Z.~Sui, B.~Chang and F.~Wei. 2022{\natexlab{a}}.
\newblock \href {https://doi.org/10.18653/v1/2022.acl-long.581} {Knowledge
  neurons in pretrained transformers}.
\newblock In \emph{Proceedings of the 60th Annual Meeting of the Association
  for Computational Linguistics (Volume 1: Long Papers)}, pages 8493--8502,
  Dublin, Ireland. Association for Computational Linguistics.

\bibitem[{Dai et~al.(2022{\natexlab{b}})Dai, Sun, Dong, Hao, Sui, and
  Wei}]{dai2022can}
D.~Dai, Y.~Sun, L.~Dong, Y.~Hao, Z.~Sui and F.~Wei. 2022{\natexlab{b}}.
\newblock Why can gpt learn in-context? language models secretly perform
  gradient descent as meta optimizers.
\newblock \emph{arXiv preprint arXiv:2212.10559}.

\bibitem[{Dai et~al.(2023)Dai, Liu, Liao, Huang, Wu, Zhao, Liu, Liu, Li, Zhu,
  Cai, Li, Shen, Liu, and Li}]{dai2023chataug}
H.~Dai, Z.~Liu, W.~Liao, X.~Huang, Z.~Wu, L.~Zhao, W.~Liu, N.~Liu et~al. 2023.
\newblock \href {http://arxiv.org/abs/2302.13007} {Chataug: Leveraging chatgpt
  for text data augmentation}.

\bibitem[{Dai et~al.(2019)Dai, Yang, Yang, Carbonell, Le, and
  Salakhutdinov}]{dai-etal-2019-transformer}
Z.~Dai, Z.~Yang, Y.~Yang, J.~Carbonell, Q.~Le and R.~Salakhutdinov. 2019.
\newblock \href {https://doi.org/10.18653/v1/P19-1285} {Transformer-{XL}:
  Attentive language models beyond a fixed-length context}.
\newblock In \emph{Proceedings of the 57th Annual Meeting of the Association
  for Computational Linguistics}, pages 2978--2988, Florence, Italy.
  Association for Computational Linguistics.

\bibitem[{Dalla-Torre et~al.(2023)Dalla-Torre, Gonzalez, Mendoza~Revilla,
  Lopez~Carranza, Henryk~Grywaczewski, Oteri, Dallago, Trop, Sirelkhatim,
  Richard et~al.}]{dalla2023nucleotide}
H.~Dalla-Torre, L.~Gonzalez, J.~Mendoza~Revilla, N.~Lopez~Carranza,
  A.~Henryk~Grywaczewski, F.~Oteri, C.~Dallago, E.~Trop et~al. 2023.
\newblock The nucleotide transformer: Building and evaluating robust foundation
  models for human genomics.
\newblock \emph{bioRxiv}, pages 2023--01.

\bibitem[{Dao et~al.(2022)Dao, Fu, Ermon, Rudra, and
  R{\'e}}]{dao2022flashattention}
T.~Dao, D.~Y. Fu, S.~Ermon, A.~Rudra and C.~R{\'e}. 2022.
\newblock Flashattention: Fast and memory-efficient exact attention with
  io-awareness.
\newblock \emph{arXiv preprint arXiv:2205.14135}.

\bibitem[{Dao et~al.(2023)Dao, Fu, Saab, Thomas, Rudra, and
  Ré}]{daoHungryHungryHippos2023}
T.~Dao, D.~Y. Fu, K.~K. Saab, A.~W. Thomas, A.~Rudra and C.~Ré. 2023.
\newblock \href {http://arxiv.org/abs/2212.14052} {Hungry {Hungry} {Hippos}:
  {Towards} {Language} {Modeling} with {State} {Space} {Models}}.
\newblock ArXiv:2212.14052 [cs].

\bibitem[{Dathathri et~al.(2020)Dathathri, Madotto, Lan, Hung, Frank, Molino,
  Yosinski, and Liu}]{dathathri2020plug}
S.~Dathathri, A.~Madotto, J.~Lan, J.~Hung, E.~Frank, P.~Molino, J.~Yosinski and
  R.~Liu. 2020.
\newblock \href {http://arxiv.org/abs/1912.02164} {Plug and play language
  models: A simple approach to controlled text generation}.

\bibitem[{Dauparas et~al.(2022)Dauparas, Anishchenko, Bennett, Bai, Ragotte,
  Milles, Wicky, Courbet, de~Haas, Bethel, Leung, Huddy, Pellock, Tischer,
  Chan, Koepnick, Nguyen, Kang, Sankaran, Bera, King, and
  Baker}]{dauparas2022proteinmpnn}
J.~Dauparas, I.~Anishchenko, N.~Bennett, H.~Bai, R.~J. Ragotte, L.~F. Milles,
  B.~I.~M. Wicky, A.~Courbet et~al. 2022.
\newblock \href {https://doi.org/10.1126/science.add2187} {Robust deep
  learning\&\#x2013;based protein sequence design using proteinmpnn}.
\newblock \emph{Science}, 378(6615):49--56.

\bibitem[{De~Cao et~al.(2021)De~Cao, Aziz, and
  Titov}]{de-cao-etal-2021-editing}
N.~De~Cao, W.~Aziz and I.~Titov. 2021.
\newblock \href {https://doi.org/10.18653/v1/2021.emnlp-main.522} {Editing
  factual knowledge in language models}.
\newblock In \emph{Proceedings of the 2021 Conference on Empirical Methods in
  Natural Language Processing}, pages 6491--6506, Online and Punta Cana,
  Dominican Republic. Association for Computational Linguistics.

\bibitem[{Dehghani et~al.(2022)Dehghani, Arnab, Beyer, Vaswani, and
  Tay}]{dehghani_efficiency_2022}
M.~Dehghani, A.~Arnab, L.~Beyer, A.~Vaswani and Y.~Tay. 2022.
\newblock \href {http://arxiv.org/abs/2110.12894} {The {Efficiency}
  {Misnomer}}.
\newblock ArXiv:2110.12894 [cs, stat].

\bibitem[{Dehghani et~al.(2021)Dehghani, Tay, Gritsenko, Zhao, Houlsby, Diaz,
  Metzler, and Vinyals}]{dehghani2021benchmark}
M.~Dehghani, Y.~Tay, A.~A. Gritsenko, Z.~Zhao, N.~Houlsby, F.~Diaz, D.~Metzler
  and O.~Vinyals. 2021.
\newblock The benchmark lottery.
\newblock \emph{arXiv preprint arXiv:2107.07002}.

\bibitem[{Del~Corro et~al.(2023)Del~Corro, Del~Giorno, Agarwal, Yu, Awadallah,
  and Mukherjee}]{delcorroSkipDecodeAutoregressiveSkip2023}
L.~Del~Corro, A.~Del~Giorno, S.~Agarwal, B.~Yu, A.~Awadallah and S.~Mukherjee.
  2023.
\newblock \href {https://doi.org/10.48550/arXiv.2307.02628} {{SkipDecode}:
  {Autoregressive} {Skip} {Decoding} with {Batching} and {Caching} for
  {Efficient} {LLM} {Inference}}.
\newblock ArXiv:2307.02628 [cs].

\bibitem[{Deroy et~al.(2023)Deroy, Ghosh, and Ghosh}]{deroy2023ready}
A.~Deroy, K.~Ghosh and S.~Ghosh. 2023.
\newblock \href {http://arxiv.org/abs/2306.01248} {How ready are pre-trained
  abstractive models and llms for legal case judgement summarization?}

\bibitem[{Deshpande et~al.(2023)Deshpande, Murahari, Rajpurohit, Kalyan, and
  Narasimhan}]{deshpande2023toxicity}
A.~Deshpande, V.~Murahari, T.~Rajpurohit, A.~Kalyan and K.~Narasimhan. 2023.
\newblock Toxicity in chatgpt: Analyzing persona-assigned language models.
\newblock \emph{arXiv preprint arXiv:2304.05335}.

\bibitem[{Dettmers et~al.(2022)Dettmers, Lewis, Belkada, and
  Zettlemoyer}]{llm8bit}
T.~Dettmers, M.~Lewis, Y.~Belkada and L.~Zettlemoyer. 2022.
\newblock \href {https://doi.org/10.48550/ARXIV.2208.07339} {Llm.int8(): 8-bit
  matrix multiplication for transformers at scale}.

\bibitem[{Dettmers et~al.(2023{\natexlab{a}})Dettmers, Pagnoni, Holtzman, and
  Zettlemoyer}]{dettmers_qlora_2023}
T.~Dettmers, A.~Pagnoni, A.~Holtzman and L.~Zettlemoyer. 2023{\natexlab{a}}.
\newblock \href {http://arxiv.org/abs/2305.14314} {{QLoRA}: {Efficient}
  {Finetuning} of {Quantized} {LLMs}}.
\newblock ArXiv:2305.14314 [cs].

\bibitem[{Dettmers et~al.(2023{\natexlab{b}})Dettmers, Svirschevski,
  Egiazarian, Kuznedelev, Frantar, Ashkboos, Borzunov, Hoefler, and
  Alistarh}]{dettmers2023spqr}
T.~Dettmers, R.~Svirschevski, V.~Egiazarian, D.~Kuznedelev, E.~Frantar,
  S.~Ashkboos, A.~Borzunov, T.~Hoefler et~al. 2023{\natexlab{b}}.
\newblock Spqr: A sparse-quantized representation for near-lossless llm weight
  compression.
\newblock \emph{arXiv preprint arXiv:2306.03078}.

\bibitem[{Devlin et~al.(2019)Devlin, Chang, Lee, and
  Toutanova}]{devlin-etal-2019-bert}
J.~Devlin, M.-W. Chang, K.~Lee and K.~Toutanova. 2019.
\newblock \href {https://doi.org/10.18653/v1/N19-1423} {{BERT}: Pre-training of
  deep bidirectional transformers for language understanding}.
\newblock In \emph{Proceedings of the 2019 Conference of the North {A}merican
  Chapter of the Association for Computational Linguistics: Human Language
  Technologies, Volume 1 (Long and Short Papers)}, pages 4171--4186,
  Minneapolis, Minnesota. Association for Computational Linguistics.

\bibitem[{Dey et~al.(2023)Dey, Gosal, Zhiming, Chen, Khachane, Marshall,
  Pathria, Tom, and Hestness}]{dey2023cerebrasgpt}
N.~Dey, G.~Gosal, Zhiming, Chen, H.~Khachane, W.~Marshall, R.~Pathria, M.~Tom
  et~al. 2023.
\newblock \href {http://arxiv.org/abs/2304.03208} {Cerebras-gpt: Open
  compute-optimal language models trained on the cerebras wafer-scale cluster}.

\bibitem[{Diao et~al.(2022)Diao, Li, Lin, Huang, and Zhang}]{diao2022black}
S.~Diao, X.~Li, Y.~Lin, Z.~Huang and T.~Zhang. 2022.
\newblock Black-box prompt learning for pre-trained language models.
\newblock \emph{arXiv preprint arXiv:2201.08531}.

\bibitem[{Ding et~al.(2022)Ding, Qin, Liu, Bing, Joty, and Li}]{ding2022gpt}
B.~Ding, C.~Qin, L.~Liu, L.~Bing, S.~Joty and B.~Li. 2022.
\newblock Is gpt-3 a good data annotator?
\newblock \emph{arXiv preprint arXiv:2212.10450}.

\bibitem[{Ding et~al.(2023)Ding, Ma, Dong, Zhang, Huang, Wang, and
  Wei}]{ding2023longnet}
J.~Ding, S.~Ma, L.~Dong, X.~Zhang, S.~Huang, W.~Wang and F.~Wei. 2023.
\newblock \href {http://arxiv.org/abs/2307.02486} {Longnet: Scaling
  transformers to 1,000,000,000 tokens}.

\bibitem[{Dodge et~al.(2021)Dodge, Sap, Marasovi{\'c}, Agnew, Ilharco,
  Groeneveld, Mitchell, and Gardner}]{dodge2021documenting}
J.~Dodge, M.~Sap, A.~Marasovi{\'c}, W.~Agnew, G.~Ilharco, D.~Groeneveld,
  M.~Mitchell and M.~Gardner. 2021.
\newblock Documenting large webtext corpora: A case study on the colossal clean
  crawled corpus.
\newblock \emph{arXiv preprint arXiv:2104.08758}.

\bibitem[{Dominguez-Olmedo et~al.(2023)Dominguez-Olmedo, Hardt, and
  Mendler-D{\"u}nner}]{dominguez2023questioning}
R.~Dominguez-Olmedo, M.~Hardt and C.~Mendler-D{\"u}nner. 2023.
\newblock Questioning the survey responses of large language models.
\newblock \emph{arXiv preprint arXiv:2306.07951}.

\bibitem[{Dong et~al.(2022)Dong, Dai, Song, Xu, Sui, and
  Li}]{dong-etal-2022-calibrating}
Q.~Dong, D.~Dai, Y.~Song, J.~Xu, Z.~Sui and L.~Li. 2022.
\newblock \href {https://aclanthology.org/2022.findings-emnlp.438} {Calibrating
  factual knowledge in pretrained language models}.
\newblock In \emph{Findings of the Association for Computational Linguistics:
  EMNLP 2022}, pages 5937--5947, Abu Dhabi, United Arab Emirates. Association
  for Computational Linguistics.

\bibitem[{Dowty et~al.(2012)Dowty, Wall, and Peters}]{dowty2012introduction}
D.~R. Dowty, R.~Wall and S.~Peters. 2012.
\newblock \emph{Introduction to Montague semantics}, volume~11.
\newblock Springer Science \& Business Media.

\bibitem[{Driess et~al.(2023)Driess, Xia, Sajjadi, Lynch, Chowdhery, Ichter,
  Wahid, Tompson, Vuong, Yu et~al.}]{palm-e}
D.~Driess, F.~Xia, M.~S. Sajjadi, C.~Lynch, A.~Chowdhery, B.~Ichter, A.~Wahid,
  J.~Tompson et~al. 2023.
\newblock Palm-e: An embodied multimodal language model.
\newblock \emph{arXiv preprint arXiv:2303.03378}.

\bibitem[{Du et~al.(2022{\natexlab{a}})Du, Huang, Dai, Tong, Lepikhin, Xu,
  Krikun, Zhou, Yu, Firat et~al.}]{glam}
N.~Du, Y.~Huang, A.~M. Dai, S.~Tong, D.~Lepikhin, Y.~Xu, M.~Krikun, Y.~Zhou
  et~al. 2022{\natexlab{a}}.
\newblock Glam: Efficient scaling of language models with mixture-of-experts.
\newblock In \emph{International Conference on Machine Learning}, pages
  5547--5569. PMLR.

\bibitem[{Du et~al.(2023)Du, Li, Torralba, Tenenbaum, and
  Mordatch}]{du2023improving}
Y.~Du, S.~Li, A.~Torralba, J.~B. Tenenbaum and I.~Mordatch. 2023.
\newblock \href {http://arxiv.org/abs/2305.14325} {Improving {Factuality} and
  {Reasoning} in {Language} {Models} through {Multiagent} {Debate}}.
\newblock ArXiv:2305.14325 [cs].

\bibitem[{Du et~al.(2022{\natexlab{b}})Du, Qian, Liu, Ding, Qiu, Yang, and
  Tang}]{du-etal-2022-glm}
Z.~Du, Y.~Qian, X.~Liu, M.~Ding, J.~Qiu, Z.~Yang and J.~Tang.
  2022{\natexlab{b}}.
\newblock \href {https://doi.org/10.18653/v1/2022.acl-long.26} {{GLM}: General
  language model pretraining with autoregressive blank infilling}.
\newblock In \emph{Proceedings of the 60th Annual Meeting of the Association
  for Computational Linguistics (Volume 1: Long Papers)}, pages 320--335,
  Dublin, Ireland. Association for Computational Linguistics.

\bibitem[{Dunn et~al.(2022)Dunn, Dagdelen, Walker, Lee, Rosen, Ceder, Persson,
  and Jain}]{dunn2022structured}
A.~Dunn, J.~Dagdelen, N.~Walker, S.~Lee, A.~S. Rosen, G.~Ceder, K.~Persson and
  A.~Jain. 2022.
\newblock Structured information extraction from complex scientific text with
  fine-tuned large language models.
\newblock \emph{arXiv preprint arXiv:2212.05238}.

\bibitem[{Duong and Solomon(2023)}]{duong2023analysis}
D.~Duong and B.~D. Solomon. 2023.
\newblock Analysis of large-language model versus human performance for
  genetics questions.
\newblock \emph{European Journal of Human Genetics}, pages 1--3.

\bibitem[{Dziri et~al.(2023)Dziri, Lu, Sclar, Li, Jiang, Lin, West,
  Bhagavatula, Bras, Hwang, Sanyal, Welleck, Ren, Ettinger, Harchaoui, and
  Choi}]{dziriFaithFateLimits2023}
N.~Dziri, X.~Lu, M.~Sclar, X.~L. Li, L.~Jiang, B.~Y. Lin, P.~West,
  C.~Bhagavatula et~al. 2023.
\newblock \href {http://arxiv.org/abs/2305.18654} {Faith and {Fate}: {Limits}
  of {Transformers} on {Compositionality}}.
\newblock ArXiv:2305.18654 [cs].

\bibitem[{Dziri et~al.(2021)Dziri, Madotto, Zaiane, and
  Bose}]{dziriNeuralPathHunter2021}
N.~Dziri, A.~Madotto, O.~Zaiane and A.~J. Bose. 2021.
\newblock \href {http://arxiv.org/abs/2104.08455} {Neural {Path} {Hunter}:
  {Reducing} {Hallucination} in {Dialogue} {Systems} via {Path} {Grounding}}.
\newblock ArXiv:2104.08455 [cs].

\bibitem[{El-Mhamdi et~al.(2023)El-Mhamdi, Farhadkhani, Guerraoui, Gupta,
  Hoang, Pinot, Rouault, and Stephan}]{el-mhamdiImpossibleSafetyLarge2023}
E.-M. El-Mhamdi, S.~Farhadkhani, R.~Guerraoui, N.~Gupta, L.-N. Hoang, R.~Pinot,
  S.~Rouault and J.~Stephan. 2023.
\newblock \href {http://arxiv.org/abs/2209.15259} {On the {Impossible} {Safety}
  of {Large} {AI} {Models}}.
\newblock ArXiv:2209.15259 [cs].

\bibitem[{Elhage et~al.(2021)Elhage, Nanda, Olsson, Henighan, Joseph, Mann,
  Askell, Bai, Chen, Conerly et~al.}]{elhage2021mathematical}
N.~Elhage, N.~Nanda, C.~Olsson, T.~Henighan, N.~Joseph, B.~Mann, A.~Askell,
  Y.~Bai et~al. 2021.
\newblock A mathematical framework for transformer circuits.
\newblock \emph{Transformer Circuits Thread}.

\bibitem[{Elnaggar et~al.(2020)Elnaggar, Heinzinger, Dallago, Rihawi, Wang,
  Jones, Gibbs, Feher, Angerer, Steinegger et~al.}]{elnaggar2020prottrans}
A.~Elnaggar, M.~Heinzinger, C.~Dallago, G.~Rihawi, Y.~Wang, L.~Jones, T.~Gibbs,
  T.~Feher et~al. 2020.
\newblock Prottrans: towards cracking the language of life's code through
  self-supervised deep learning and high performance computing.
\newblock \emph{arXiv preprint arXiv:2007.06225}.

\bibitem[{Eloundou et~al.(2023)Eloundou, Manning, Mishkin, and
  Rock}]{eloundou2023gpts}
T.~Eloundou, S.~Manning, P.~Mishkin and D.~Rock. 2023.
\newblock \href {http://arxiv.org/abs/2303.10130} {Gpts are gpts: An early look
  at the labor market impact potential of large language models}.

\bibitem[{Faal et~al.(2023)Faal, Schmitt, and Yu}]{faal2023reward}
F.~Faal, K.~Schmitt and J.~Y. Yu. 2023.
\newblock Reward modeling for mitigating toxicity in transformer-based language
  models.
\newblock \emph{Applied Intelligence}, 53(7):8421--8435.

\bibitem[{Fan et~al.(2021)Fan, Gardent, Braud, and
  Bordes}]{fan-etal-2021-augmenting}
A.~Fan, C.~Gardent, C.~Braud and A.~Bordes. 2021.
\newblock \href {https://doi.org/10.1162/tacl_a_00356} {Augmenting transformers
  with {KNN}-based composite memory for dialog}.
\newblock \emph{Transactions of the Association for Computational Linguistics},
  9:82--99.

\bibitem[{Fan et~al.(2020)Fan, Grave, and Joulin}]{Fan2020Reducing}
A.~Fan, E.~Grave and A.~Joulin. 2020.
\newblock \href {https://openreview.net/forum?id=SylO2yStDr} {Reducing
  transformer depth on demand with structured dropout}.
\newblock In \emph{International Conference on Learning Representations}.

\bibitem[{Fathi et~al.(2023)Fathi, Pilault, Bacon, Pal, Firat, and
  Goroshin}]{fathiBlockStateTransformer2023}
M.~Fathi, J.~Pilault, P.-L. Bacon, C.~Pal, O.~Firat and R.~Goroshin. 2023.
\newblock \href {http://arxiv.org/abs/2306.09539} {Block-{State}
  {Transformer}}.
\newblock ArXiv:2306.09539 [cs].

\bibitem[{Fedus et~al.(2021)Fedus, Zoph, and Shazeer}]{switch_transformer}
W.~Fedus, B.~Zoph and N.~Shazeer. 2021.
\newblock \href {https://doi.org/10.48550/ARXIV.2101.03961} {Switch
  transformers: Scaling to trillion parameter models with simple and efficient
  sparsity}.

\bibitem[{Feldman(2020)}]{feldman2020does}
V.~Feldman. 2020.
\newblock Does learning require memorization? a short tale about a long tail.
\newblock In \emph{Proceedings of the 52nd Annual ACM SIGACT Symposium on
  Theory of Computing}, pages 954--959.

\bibitem[{Feng et~al.(2023{\natexlab{a}})Feng, Park, Liu, and
  Tsvetkov}]{fengpretrainingDataLanguage2023}
S.~Feng, C.~Y. Park, Y.~Liu and Y.~Tsvetkov. 2023{\natexlab{a}}.
\newblock \href {http://arxiv.org/abs/2305.08283} {From {Pretraining} {Data} to
  {Language} {Models} to {Downstream} {Tasks}: {Tracking} the {Trails} of
  {Political} {Biases} {Leading} to {Unfair} {NLP} {Models}}.
\newblock ArXiv:2305.08283 [cs].

\bibitem[{Feng et~al.(2023{\natexlab{b}})Feng, Zhu, Fu, Jampani, Akula, He,
  Basu, Wang, and Wang}]{fengLayoutGPTCompositionalVisual2023}
W.~Feng, W.~Zhu, T.-j. Fu, V.~Jampani, A.~Akula, X.~He, S.~Basu, X.~E. Wang
  et~al. 2023{\natexlab{b}}.
\newblock \href {https://doi.org/10.48550/arXiv.2305.15393} {{LayoutGPT}:
  {Compositional} {Visual} {Planning} and {Generation} with {Large} {Language}
  {Models}}.
\newblock ArXiv:2305.15393 [cs].

\bibitem[{Ferrara(2023)}]{ferrara2023should}
E.~Ferrara. 2023.
\newblock Should chatgpt be biased? challenges and risks of bias in large
  language models.
\newblock \emph{arXiv preprint arXiv:2304.03738}.

\bibitem[{Ficek et~al.(2022)Ficek, Liu, and Collier}]{ficek-etal-2022-tackle}
A.~Ficek, F.~Liu and N.~Collier. 2022.
\newblock \href {https://aclanthology.org/2022.aacl-short.60} {How to tackle an
  emerging topic? combining strong and weak labels for covid news {NER}}.
\newblock In \emph{Proceedings of the 2nd Conference of the Asia-Pacific
  Chapter of the Association for Computational Linguistics and the 12th
  International Joint Conference on Natural Language Processing (Volume 2:
  Short Papers)}, pages 488--496, Online only. Association for Computational
  Linguistics.

\bibitem[{Fourrier et~al.(2023)Fourrier, Habib, Launay, and Wolf}]{hfmmlu2023}
C.~Fourrier, N.~Habib, J.~Launay and T.~Wolf. 2023.
\newblock What's going on with the open llm leaderboard?
\newblock Available from:
  \url{https://huggingface.co/blog/evaluating-mmlu-leaderboard}.
\newblock Accessed: 27/06/2023.

\bibitem[{Frantar and Alistarh(2023)}]{frantar2023massive}
E.~Frantar and D.~Alistarh. 2023.
\newblock Massive language models can be accurately pruned in one-shot.
\newblock \emph{arXiv preprint arXiv:2301.00774}.

\bibitem[{Frantar et~al.(2022)Frantar, Ashkboos, Hoefler, and
  Alistarh}]{frantar2022gptq}
E.~Frantar, S.~Ashkboos, T.~Hoefler and D.~Alistarh. 2022.
\newblock Gptq: Accurate post-training quantization for generative pre-trained
  transformers.
\newblock \emph{arXiv preprint arXiv:2210.17323}.

\bibitem[{Fried et~al.(2022)Fried, Aghajanyan, Lin, Wang, Wallace, Shi, Zhong,
  Yih, Zettlemoyer, and Lewis}]{incoder}
D.~Fried, A.~Aghajanyan, J.~Lin, S.~Wang, E.~Wallace, F.~Shi, R.~Zhong, W.-t.
  Yih et~al. 2022.
\newblock \href {https://doi.org/10.48550/ARXIV.2204.05999} {Incoder: A
  generative model for code infilling and synthesis}.

\bibitem[{Frömmgen and Kharatyan(2023)}]{code_review_ml}
A.~Frömmgen and L.~Kharatyan. 2023.
\newblock Resolving code review comments with ml.
\newblock Available from:
  \url{https://ai.googleblog.com/2023/05/resolving-code-review-comments-with-ml.html}.
\newblock Accessed: 26/06/2023.

\bibitem[{Fu et~al.(2023)Fu, Ng, Jiang, and Liu}]{fu2023gptscore}
J.~Fu, S.-K. Ng, Z.~Jiang and P.~Liu. 2023.
\newblock Gptscore: Evaluate as you desire.
\newblock \emph{arXiv preprint arXiv:2302.04166}.

\bibitem[{Fujii et~al.(2023)Fujii, Shibata, Yamaguchi, Morishita, and
  Sogawa}]{fujii2023different}
T.~Fujii, K.~Shibata, A.~Yamaguchi, T.~Morishita and Y.~Sogawa. 2023.
\newblock How do different tokenizers perform on downstream tasks in scriptio
  continua languages?: A case study in japanese.
\newblock \emph{arXiv preprint arXiv:2306.09572}.

\bibitem[{Gabriel(2020)}]{gabriel2020artificial}
I.~Gabriel. 2020.
\newblock Artificial intelligence, values, and alignment.
\newblock \emph{Minds and machines}, 30(3):411--437.

\bibitem[{Gadgil et~al.(2022)Gadgil, Tadipatri, Agrawal, Narayanan, and
  Goyal}]{gadgiltowards}
S.~Gadgil, A.~R. Tadipatri, A.~Agrawal, A.~Narayanan and N.~Goyal. 2022.
\newblock Towards automating formalisation of theorem statements using large
  language models.
\newblock \emph{36th Conference on Neural Information Processing Systems
  (NeurIPS 2022) Workshop on MATH-AI}.

\bibitem[{Gale et~al.(2022)Gale, Narayanan, Young, and
  Zaharia}]{gale2022megablocks}
T.~Gale, D.~Narayanan, C.~Young and M.~Zaharia. 2022.
\newblock Megablocks: Efficient sparse training with mixture-of-experts.
\newblock \emph{arXiv preprint arXiv:2211.15841}.

\bibitem[{Gale et~al.(2020)Gale, Zaharia, Young, and Elsen}]{gale_sparse_2020}
T.~Gale, M.~Zaharia, C.~Young and E.~Elsen. 2020.
\newblock \href {http://arxiv.org/abs/2006.10901} {Sparse {GPU} {Kernels} for
  {Deep} {Learning}}.
\newblock ArXiv:2006.10901 [cs, stat].

\bibitem[{Gandhi et~al.(2023)Gandhi, Fr{\"a}nken, Gerstenbrg, and
  Goodman}]{gandhi2023understanding}
K.~Gandhi, J.-P. Fr{\"a}nken, T.~Gerstenbrg and N.~D. Goodman. 2023.
\newblock Understanding social reasoning in language models with language
  models.
\newblock \emph{arXiv preprint arXiv:2306.15448}.

\bibitem[{Ganguli et~al.(2022)Ganguli, Lovitt, Kernion, Askell, Bai, Kadavath,
  Mann, Perez, Schiefer, Ndousse et~al.}]{ganguli2022red}
D.~Ganguli, L.~Lovitt, J.~Kernion, A.~Askell, Y.~Bai, S.~Kadavath, B.~Mann,
  E.~Perez et~al. 2022.
\newblock Red teaming language models to reduce harms: Methods, scaling
  behaviors, and lessons learned.
\newblock \emph{arXiv preprint arXiv:2209.07858}.

\bibitem[{Gao et~al.(2023)Gao, Ding, Qin, and Liu}]{gao2023chatgpt}
J.~Gao, X.~Ding, B.~Qin and T.~Liu. 2023.
\newblock Is chatgpt a good causal reasoner? a comprehensive evaluation.
\newblock \emph{arXiv preprint arXiv:2305.07375}.

\bibitem[{Gao et~al.(2020)Gao, Biderman, Black, Golding, Hoppe, Foster, Phang,
  He, Thite, Nabeshima et~al.}]{gao2020pile}
L.~Gao, S.~Biderman, S.~Black, L.~Golding, T.~Hoppe, C.~Foster, J.~Phang, H.~He
  et~al. 2020.
\newblock The pile: An 800gb dataset of diverse text for language modeling.
\newblock \emph{arXiv preprint arXiv:2101.00027}.

\bibitem[{Gao et~al.(2021)Gao, Tow, Biderman, Black, DiPofi, Foster, Golding,
  Hsu, McDonell, Muennighoff, Phang, Reynolds, Tang, Thite, Wang, Wang, and
  Zou}]{eval-harness}
L.~Gao, J.~Tow, S.~Biderman, S.~Black, A.~DiPofi, C.~Foster, L.~Golding, J.~Hsu
  et~al. 2021.
\newblock \href {https://doi.org/10.5281/zenodo.5371628} {A framework for
  few-shot language model evaluation}.

\bibitem[{Gehman et~al.(2020)Gehman, Gururangan, Sap, Choi, and
  Smith}]{gehman2020realtoxicityprompts}
S.~Gehman, S.~Gururangan, M.~Sap, Y.~Choi and N.~A. Smith. 2020.
\newblock Realtoxicityprompts: Evaluating neural toxic degeneration in language
  models.
\newblock \emph{arXiv preprint arXiv:2009.11462}.

\bibitem[{Gehrmann et~al.(2019)Gehrmann, Strobelt, and
  Rush}]{gehrmannGLTRStatisticalDetection2019}
S.~Gehrmann, H.~Strobelt and A.~M. Rush. 2019.
\newblock \href {http://arxiv.org/abs/1906.04043} {{GLTR}: {Statistical}
  {Detection} and {Visualization} of {Generated} {Text}}.
\newblock ArXiv:1906.04043 [cs].

\bibitem[{Geirhos et~al.(2020)Geirhos, Jacobsen, Michaelis, Zemel, Brendel,
  Bethge, and Wichmann}]{geirhos2020shortcut}
R.~Geirhos, J.-H. Jacobsen, C.~Michaelis, R.~Zemel, W.~Brendel, M.~Bethge and
  F.~A. Wichmann. 2020.
\newblock Shortcut learning in deep neural networks.
\newblock \emph{Nature Machine Intelligence}, 2(11):665--673.

\bibitem[{Glaese et~al.(2022)Glaese, McAleese, Trębacz, Aslanides, Firoiu,
  Ewalds, Rauh, Weidinger, Chadwick, Thacker, Campbell-Gillingham, Uesato,
  Huang, Comanescu, Yang, See, Dathathri, Greig, Chen, Fritz, Elias, Green,
  Mokrá, Fernando, Wu, Foley, Young, Gabriel, Isaac, Mellor, Hassabis,
  Kavukcuoglu, Hendricks, and Irving}]{sparrow}
A.~Glaese, N.~McAleese, M.~Trębacz, J.~Aslanides, V.~Firoiu, T.~Ewalds,
  M.~Rauh, L.~Weidinger et~al. 2022.
\newblock \href {https://doi.org/10.48550/ARXIV.2209.14375} {Improving
  alignment of dialogue agents via targeted human judgements}.

\bibitem[{Goldberg(1991)}]{goldbergWhatEveryComputer1991}
D.~Goldberg. 1991.
\newblock \href {https://doi.org/10.1145/103162.103163} {What every computer
  scientist should know about floating-point arithmetic}.
\newblock \emph{ACM Computing Surveys}, 23(1):5--48.

\bibitem[{Gomez et~al.(2022)Gomez, Key, Perlin, Gou, Frosst, Dean, and
  Gal}]{gomez2022interlocking}
A.~N. Gomez, O.~Key, K.~Perlin, S.~Gou, N.~Frosst, J.~Dean and Y.~Gal. 2022.
\newblock Interlocking backpropagation: Improving depthwise model-parallelism.
\newblock \emph{The Journal of Machine Learning Research}, 23(1):7714--7741.

\bibitem[{Gong et~al.(2019)Gong, He, Li, Qin, Wang, and Liu}]{pmlr-v97-gong19a}
L.~Gong, D.~He, Z.~Li, T.~Qin, L.~Wang and T.~Liu. 2019.
\newblock \href {https://proceedings.mlr.press/v97/gong19a.html} {Efficient
  training of {BERT} by progressively stacking}.
\newblock In \emph{Proceedings of the 36th International Conference on Machine
  Learning}, volume~97 of \emph{Proceedings of Machine Learning Research},
  pages 2337--2346. PMLR.

\bibitem[{Gou et~al.(2023)Gou, Shao, Gong, Shen, Yang, Duan, and
  Chen}]{gou2023critic}
Z.~Gou, Z.~Shao, Y.~Gong, Y.~Shen, Y.~Yang, N.~Duan and W.~Chen. 2023.
\newblock Critic: Large language models can self-correct with tool-interactive
  critiquing.
\newblock \emph{arXiv preprint arXiv:2305.11738}.

\bibitem[{Greshake et~al.(2023)Greshake, Abdelnabi, Mishra, Endres, Holz, and
  Fritz}]{greshake2023more}
K.~Greshake, S.~Abdelnabi, S.~Mishra, C.~Endres, T.~Holz and M.~Fritz. 2023.
\newblock More than you've asked for: A comprehensive analysis of novel prompt
  injection threats to application-integrated large language models.
\newblock \emph{arXiv preprint arXiv:2302.12173}.

\bibitem[{Griffin et~al.(2023)Griffin, Kleinberg, Mozes, Mai, Vau, Caldwell,
  and Marvor-Parker}]{griffin2023susceptibility}
L.~D. Griffin, B.~Kleinberg, M.~Mozes, K.~T. Mai, M.~Vau, M.~Caldwell and
  A.~Marvor-Parker. 2023.
\newblock Susceptibility to influence of large language models.
\newblock \emph{arXiv preprint arXiv:2303.06074}.

\bibitem[{Gu et~al.(2021)Gu, Tinn, Cheng, Lucas, Usuyama, Liu, Naumann, Gao,
  and Poon}]{gu2021domain}
Y.~Gu, R.~Tinn, H.~Cheng, M.~Lucas, N.~Usuyama, X.~Liu, T.~Naumann, J.~Gao
  et~al. 2021.
\newblock Domain-specific language model pretraining for biomedical natural
  language processing.
\newblock \emph{ACM Transactions on Computing for Healthcare (HEALTH)},
  3(1):1--23.

\bibitem[{Gu et~al.(2023)Gu, Zhang, Usuyama, Woldesenbet, Wong, Sanapathi, Wei,
  Valluri, Strandberg, Naumann, and Poon}]{gu2023distilling}
Y.~Gu, S.~Zhang, N.~Usuyama, Y.~Woldesenbet, C.~Wong, P.~Sanapathi, M.~Wei,
  N.~Valluri et~al. 2023.
\newblock \href {http://arxiv.org/abs/2307.06439} {Distilling large language
  models for biomedical knowledge extraction: A case study on adverse drug
  events}.

\bibitem[{Gu et~al.(2022)Gu, Han, Liu, and Huang}]{gu-etal-2022-ppt}
Y.~Gu, X.~Han, Z.~Liu and M.~Huang. 2022.
\newblock \href {https://doi.org/10.18653/v1/2022.acl-long.576} {{PPT}:
  Pre-trained prompt tuning for few-shot learning}.
\newblock In \emph{Proceedings of the 60th Annual Meeting of the Association
  for Computational Linguistics (Volume 1: Long Papers)}, pages 8410--8423,
  Dublin, Ireland. Association for Computational Linguistics.

\bibitem[{Gudibande et~al.(2023)Gudibande, Wallace, Snell, Geng, Liu, Abbeel,
  Levine, and Song}]{gudibande2023false}
A.~Gudibande, E.~Wallace, C.~Snell, X.~Geng, H.~Liu, P.~Abbeel, S.~Levine and
  D.~Song. 2023.
\newblock The false promise of imitating proprietary llms.
\newblock \emph{arXiv preprint arXiv:2305.15717}.

\bibitem[{Guilford(1967)}]{guilford1967creativity}
J.~P. Guilford. 1967.
\newblock Creativity: Yesterday, today and tomorrow.
\newblock \emph{The Journal of Creative Behavior}, 1(1):3--14.

\bibitem[{Gunasekar et~al.(2023)Gunasekar, Zhang, Aneja, Mendes, Giorno, Gopi,
  Javaheripi, Kauffmann, de~Rosa, Saarikivi, Salim, Shah, Behl, Wang, Bubeck,
  Eldan, Kalai, Lee, and Li}]{gunasekar2023textbooks}
S.~Gunasekar, Y.~Zhang, J.~Aneja, C.~C.~T. Mendes, A.~D. Giorno, S.~Gopi,
  M.~Javaheripi, P.~Kauffmann et~al. 2023.
\newblock \href {http://arxiv.org/abs/2306.11644} {Textbooks are all you need}.

\bibitem[{Guo et~al.(2022)Guo, Ainslie, Uthus, Ontanon, Ni, Sung, and
  Yang}]{guo-etal-2022-longt5}
M.~Guo, J.~Ainslie, D.~Uthus, S.~Ontanon, J.~Ni, Y.-H. Sung and Y.~Yang. 2022.
\newblock \href {https://doi.org/10.18653/v1/2022.findings-naacl.55}
  {{L}ong{T}5: {E}fficient text-to-text transformer for long sequences}.
\newblock In \emph{Findings of the Association for Computational Linguistics:
  NAACL 2022}, pages 724--736, Seattle, United States. Association for
  Computational Linguistics.

\bibitem[{Gupta(2023)}]{guptaProbingQuantifierComprehension2023}
A.~Gupta. 2023.
\newblock \href {http://arxiv.org/abs/2306.07384} {Probing {Quantifier}
  {Comprehension} in {Large} {Language} {Models}}.
\newblock ArXiv:2306.07384 [cs].

\bibitem[{Gupta and Kembhavi(2022)}]{gupta2022visual}
T.~Gupta and A.~Kembhavi. 2022.
\newblock \href {http://arxiv.org/abs/2211.11559} {Visual programming:
  Compositional visual reasoning without training}.

\bibitem[{Guu et~al.(2020)Guu, Lee, Tung, Pasupat, and
  Chang}]{guu2020retrieval}
K.~Guu, K.~Lee, Z.~Tung, P.~Pasupat and M.~Chang. 2020.
\newblock Retrieval augmented language model pre-training.
\newblock In \emph{International Conference on Machine Learning}, pages
  3929--3938. PMLR.

\bibitem[{Haase and Hanel(2023)}]{haase2023artificial}
J.~Haase and P.~H.~P. Hanel. 2023.
\newblock \href {http://arxiv.org/abs/2303.12003} {Artificial muses: Generative
  artificial intelligence chatbots have risen to human-level creativity}.

\bibitem[{Hahn and Goyal(2023)}]{Hahn2023ATO}
M.~Hahn and N.~Goyal. 2023.
\newblock A theory of emergent in-context learning as implicit structure
  induction.
\newblock \emph{ArXiv}, abs/2303.07971.

\bibitem[{Hamilton(2023)}]{hamilton2023blind}
S.~Hamilton. 2023.
\newblock Blind judgement: Agent-based supreme court modelling with gpt.
\newblock \emph{arXiv preprint arXiv:2301.05327}.

\bibitem[{Han et~al.(2023)Han, Wang, Zhao, and Ji}]{Han2023InContextLO}
C.~Han, Z.~Wang, H.~Zhao and H.~Ji. 2023.
\newblock In-context learning of large language models explained as kernel
  regression.
\newblock \emph{ArXiv}, abs/2305.12766.

\bibitem[{Hartvigsen et~al.(2022)Hartvigsen, Sankaranarayanan, Palangi, Kim,
  and Ghassemi}]{hartvigsen2022aging}
T.~Hartvigsen, S.~Sankaranarayanan, H.~Palangi, Y.~Kim and M.~Ghassemi. 2022.
\newblock Aging with grace: Lifelong model editing with discrete key-value
  adaptors.
\newblock \emph{arXiv preprint arXiv:2211.11031}.

\bibitem[{Haviv et~al.(2022)Haviv, Ram, Press, Izsak, and
  Levy}]{haviv-etal-2022-transformer}
A.~Haviv, O.~Ram, O.~Press, P.~Izsak and O.~Levy. 2022.
\newblock \href {https://aclanthology.org/2022.findings-emnlp.99} {Transformer
  language models without positional encodings still learn positional
  information}.
\newblock In \emph{Findings of the Association for Computational Linguistics:
  EMNLP 2022}, pages 1382--1390, Abu Dhabi, United Arab Emirates. Association
  for Computational Linguistics.

\bibitem[{Hazell(2023)}]{hazell2023large}
J.~Hazell. 2023.
\newblock Large language models can be used to effectively scale spear phishing
  campaigns.
\newblock \emph{arXiv preprint arXiv:2305.06972}.

\bibitem[{Henderson et~al.(2022{\natexlab{a}})Henderson, Westwood, and
  Simons}]{henderson2022reproducible}
E.~L. Henderson, S.~J. Westwood and D.~J. Simons. 2022{\natexlab{a}}.
\newblock A reproducible systematic map of research on the illusory truth
  effect.
\newblock \emph{Psychonomic Bulletin \& Review}, pages 1--24.

\bibitem[{Henderson et~al.(2022{\natexlab{b}})Henderson, Krass, Zheng, Guha,
  Manning, Jurafsky, and Ho}]{henderson2022pile}
P.~Henderson, M.~S. Krass, L.~Zheng, N.~Guha, C.~D. Manning, D.~Jurafsky and
  D.~E. Ho. 2022{\natexlab{b}}.
\newblock \href {https://openreview.net/forum?id=3HCT3xfNm9r} {Pile of law:
  Learning responsible data filtering from the law and a 256{GB} open-source
  legal dataset}.
\newblock In \emph{Thirty-sixth Conference on Neural Information Processing
  Systems Datasets and Benchmarks Track}.

\bibitem[{Hendrycks et~al.(2020)Hendrycks, Burns, Basart, Critch, Li, Song, and
  Steinhardt}]{hendrycks2020aligning}
D.~Hendrycks, C.~Burns, S.~Basart, A.~Critch, J.~Li, D.~Song and J.~Steinhardt.
  2020.
\newblock Aligning ai with shared human values.
\newblock \emph{arXiv preprint arXiv:2008.02275}.

\bibitem[{Hendrycks et~al.(2021{\natexlab{a}})Hendrycks, Burns, Basart, Zou,
  Mazeika, Song, and Steinhardt}]{mmlu}
D.~Hendrycks, C.~Burns, S.~Basart, A.~Zou, M.~Mazeika, D.~Song and
  J.~Steinhardt. 2021{\natexlab{a}}.
\newblock \href {http://arxiv.org/abs/2009.03300} {Measuring massive multitask
  language understanding}.

\bibitem[{Hendrycks et~al.(2021{\natexlab{b}})Hendrycks, Carlini, Schulman, and
  Steinhardt}]{hendrycks2021unsolved}
D.~Hendrycks, N.~Carlini, J.~Schulman and J.~Steinhardt. 2021{\natexlab{b}}.
\newblock Unsolved problems in ml safety.
\newblock \emph{arXiv preprint arXiv:2109.13916}.

\bibitem[{Hendrycks and Mazeika(2022)}]{hendrycks2022x}
D.~Hendrycks and M.~Mazeika. 2022.
\newblock X-risk analysis for ai research.
\newblock \emph{arXiv preprint arXiv:2206.05862}.

\bibitem[{Hernandez et~al.(2022)Hernandez, Brown, Conerly, DasSarma, Drain,
  El-Showk, Elhage, Hatfield-Dodds, Henighan, Hume
  et~al.}]{hernandez2022scaling}
D.~Hernandez, T.~Brown, T.~Conerly, N.~DasSarma, D.~Drain, S.~El-Showk,
  N.~Elhage, Z.~Hatfield-Dodds et~al. 2022.
\newblock Scaling laws and interpretability of learning from repeated data.
\newblock \emph{arXiv preprint arXiv:2205.10487}.

\bibitem[{Hestness et~al.(2017)Hestness, Narang, Ardalani, Diamos, Jun,
  Kianinejad, Patwary, Ali, Yang, and Zhou}]{hestness2017deep}
J.~Hestness, S.~Narang, N.~Ardalani, G.~Diamos, H.~Jun, H.~Kianinejad,
  M.~Patwary, M.~Ali et~al. 2017.
\newblock Deep learning scaling is predictable, empirically.
\newblock \emph{arXiv preprint arXiv:1712.00409}.

\bibitem[{Hie et~al.(2023)Hie, Shanker, Xu, Bruun, Weidenbacher, Tang, Wu, Pak,
  and Kim}]{hie2023efficient}
B.~L. Hie, V.~R. Shanker, D.~Xu, T.~U. Bruun, P.~A. Weidenbacher, S.~Tang,
  W.~Wu, J.~E. Pak et~al. 2023.
\newblock Efficient evolution of human antibodies from general protein language
  models.
\newblock \emph{Nature Biotechnology}.

\bibitem[{Hingston and Preuss(2011)}]{hingston2011red}
P.~Hingston and M.~Preuss. 2011.
\newblock Red teaming with coevolution.
\newblock In \emph{2011 IEEE Congress of Evolutionary Computation (CEC)}, pages
  1155--1163. IEEE.

\bibitem[{Ho and Salimans(2022)}]{ho2022classifierfree}
J.~Ho and T.~Salimans. 2022.
\newblock \href {http://arxiv.org/abs/2207.12598} {Classifier-free diffusion
  guidance}.

\bibitem[{Hoelscher-Obermaier et~al.(2023)Hoelscher-Obermaier, Persson, Kran,
  Konstas, and Barez}]{hoelscherObermaierDetectingEditFailures2023}
J.~Hoelscher-Obermaier, J.~Persson, E.~Kran, I.~Konstas and F.~Barez. 2023.
\newblock \href {https://doi.org/10.48550/arXiv.2305.17553} {Detecting {Edit}
  {Failures} {In} {Large} {Language} {Models}: {An} {Improved} {Specificity}
  {Benchmark}}.
\newblock ArXiv:2305.17553 [cs].

\bibitem[{Hoffmann et~al.(2022)Hoffmann, Borgeaud, Mensch, Buchatskaya, Cai,
  Rutherford, de~las Casas, Hendricks, Welbl, Clark, Hennigan, Noland,
  Millican, van~den Driessche, Damoc, Guy, Osindero, Simonyan, Elsen, Vinyals,
  Rae, and Sifre}]{chinchilla}
J.~Hoffmann, S.~Borgeaud, A.~Mensch, E.~Buchatskaya, T.~Cai, E.~Rutherford,
  D.~de~las Casas, L.~A. Hendricks et~al. 2022.
\newblock \href {https://openreview.net/forum?id=iBBcRUlOAPR} {An empirical
  analysis of compute-optimal large language model training}.
\newblock In \emph{Advances in Neural Information Processing Systems}.

\bibitem[{Holtzman et~al.(2020)Holtzman, Buys, Du, Forbes, and
  Choi}]{holtzman2020topp}
A.~Holtzman, J.~Buys, L.~Du, M.~Forbes and Y.~Choi. 2020.
\newblock \href {https://openreview.net/forum?id=rygGQyrFvH} {The curious case
  of neural text degeneration}.
\newblock In \emph{International Conference on Learning Representations}.

\bibitem[{Holzenberger et~al.(2020)Holzenberger, Blair-Stanek, and
  Van~Durme}]{holzenberger2020dataset}
N.~Holzenberger, A.~Blair-Stanek and B.~Van~Durme. 2020.
\newblock A dataset for statutory reasoning in tax law entailment and question
  answering.
\newblock \emph{arXiv preprint arXiv:2005.05257}.

\bibitem[{Honovich et~al.(2022)Honovich, Scialom, Levy, and
  Schick}]{honovich2022unnatural}
O.~Honovich, T.~Scialom, O.~Levy and T.~Schick. 2022.
\newblock Unnatural instructions: Tuning language models with (almost) no human
  labor.
\newblock \emph{arXiv preprint arXiv:2212.09689}.

\bibitem[{Hooker(2021)}]{hooker2021hardware}
S.~Hooker. 2021.
\newblock The hardware lottery.
\newblock \emph{Communications of the ACM}, 64(12):58--65.

\bibitem[{Horton(2023)}]{horton2023large}
J.~J. Horton. 2023.
\newblock Large language models as simulated economic agents: What can we learn
  from homo silicus?
\newblock \emph{arXiv preprint arXiv:2301.07543}.

\bibitem[{Horton et~al.(2023)Horton, Mehta, Farhadi, and
  Rastegari}]{hortonBytesAreAll2023}
M.~Horton, S.~Mehta, A.~Farhadi and M.~Rastegari. 2023.
\newblock \href {http://arxiv.org/abs/2306.00238} {Bytes {Are} {All} {You}
  {Need}: {Transformers} {Operating} {Directly} {On} {File} {Bytes}}.
\newblock ArXiv:2306.00238 [cs].

\bibitem[{Houlsby et~al.(2019)Houlsby, Giurgiu, Jastrzebski, Morrone,
  De~Laroussilhe, Gesmundo, Attariyan, and Gelly}]{houlsby2019parameter}
N.~Houlsby, A.~Giurgiu, S.~Jastrzebski, B.~Morrone, Q.~De~Laroussilhe,
  A.~Gesmundo, M.~Attariyan and S.~Gelly. 2019.
\newblock Parameter-efficient transfer learning for nlp.
\newblock In \emph{International Conference on Machine Learning}, pages
  2790--2799. PMLR.

\bibitem[{Houser and McCabe(2014)}]{houser2014experimental}
D.~Houser and K.~McCabe. 2014.
\newblock Experimental economics and experimental game theory.
\newblock In \emph{Neuroeconomics}, pages 19--34. Elsevier.

\bibitem[{Howard and Ruder(2018)}]{howard-ruder-2018-universal}
J.~Howard and S.~Ruder. 2018.
\newblock \href {https://doi.org/10.18653/v1/P18-1031} {Universal language
  model fine-tuning for text classification}.
\newblock In \emph{Proceedings of the 56th Annual Meeting of the Association
  for Computational Linguistics (Volume 1: Long Papers)}, pages 328--339,
  Melbourne, Australia. Association for Computational Linguistics.

\bibitem[{Hsiao(2023)}]{bard_palm_2}
S.~Hsiao. 2023.
\newblock What’s ahead for bard: More global, more visual, more integrated.
\newblock Available from:
  \url{https://blog.google/technology/ai/google-bard-updates-io-2023/}.
\newblock Accessed: 28/06/2023.

\bibitem[{Hu et~al.(2022)Hu, Xia, Zheng, Tan, Huang, Xu, and
  Li}]{hu2022protein}
B.~Hu, J.~Xia, J.~Zheng, C.~Tan, Y.~Huang, Y.~Xu and S.~Z. Li. 2022.
\newblock \href {http://arxiv.org/abs/2211.16742} {Protein language models and
  structure prediction: Connection and progression}.

\bibitem[{Hu et~al.(2021)Hu, Shen, Wallis, Allen-Zhu, Li, Wang, Wang, and
  Chen}]{hu2021lora}
E.~J. Hu, Y.~Shen, P.~Wallis, Z.~Allen-Zhu, Y.~Li, S.~Wang, L.~Wang and
  W.~Chen. 2021.
\newblock \href {http://arxiv.org/abs/2106.09685} {Lora: Low-rank adaptation of
  large language models}.

\bibitem[{Hu et~al.(2023)Hu, Lan, Wang, Xu, Lim, Lee, Bing, and
  Poria}]{hu2023llm}
Z.~Hu, Y.~Lan, L.~Wang, W.~Xu, E.-P. Lim, R.~K.-W. Lee, L.~Bing and S.~Poria.
  2023.
\newblock Llm-adapters: An adapter family for parameter-efficient fine-tuning
  of large language models.
\newblock \emph{arXiv preprint arXiv:2304.01933}.

\bibitem[{Hua et~al.(2022)Hua, Dai, Liu, and
  Le}]{huaTransformerQualityLinear2022}
W.~Hua, Z.~Dai, H.~Liu and Q.~Le. 2022.
\newblock \href {https://proceedings.mlr.press/v162/hua22a.html} {Transformer
  {Quality} in {Linear} {Time}}.
\newblock In \emph{Proceedings of the 39th {International} {Conference} on
  {Machine} {Learning}}, pages 9099--9117. PMLR.
\newblock ISSN: 2640-3498.

\bibitem[{Huang et~al.(2019)Huang, Vaswani, Uszkoreit, Simon, Hawthorne,
  Shazeer, Dai, Hoffman, Dinculescu, and Eck}]{huang2018music}
C.-Z.~A. Huang, A.~Vaswani, J.~Uszkoreit, I.~Simon, C.~Hawthorne, N.~Shazeer,
  A.~M. Dai, M.~D. Hoffman et~al. 2019.
\newblock Music transformer.
\newblock In \emph{International Conference on Learning Representations}.

\bibitem[{Huang et~al.(2022{\natexlab{a}})Huang, Gu, Hou, Wu, Wang, Yu, and
  Han}]{self_improvement}
J.~Huang, S.~S. Gu, L.~Hou, Y.~Wu, X.~Wang, H.~Yu and J.~Han.
  2022{\natexlab{a}}.
\newblock \href {https://doi.org/10.48550/ARXIV.2210.11610} {Large language
  models can self-improve}.

\bibitem[{Huang and Chang(2023)}]{huangReasoningLargeLanguage2023}
J.~Huang and K.~C.-C. Chang. 2023.
\newblock \href {http://arxiv.org/abs/2212.10403} {Towards {Reasoning} in
  {Large} {Language} {Models}: {A} {Survey}}.
\newblock ArXiv:2212.10403 [cs].

\bibitem[{Huang et~al.(2022{\natexlab{b}})Huang, Abbeel, Pathak, and
  Mordatch}]{huang2022language}
W.~Huang, P.~Abbeel, D.~Pathak and I.~Mordatch. 2022{\natexlab{b}}.
\newblock Language models as zero-shot planners: Extracting actionable
  knowledge for embodied agents.
\newblock In \emph{International Conference on Machine Learning}, pages
  9118--9147. PMLR.

\bibitem[{Huang et~al.(2022{\natexlab{c}})Huang, Xia, Xiao, Chan, Liang,
  Florence, Zeng, Tompson, Mordatch, Chebotar et~al.}]{huang2022inner}
W.~Huang, F.~Xia, T.~Xiao, H.~Chan, J.~Liang, P.~Florence, A.~Zeng, J.~Tompson
  et~al. 2022{\natexlab{c}}.
\newblock Inner monologue: Embodied reasoning through planning with language
  models.
\newblock \emph{arXiv preprint arXiv:2207.05608}.

\bibitem[{Huang et~al.(2018)Huang, Cheng, Bapna, Firat, Chen, Chen, Lee, Ngiam,
  Le, Wu, and Chen}]{gpipe}
Y.~Huang, Y.~Cheng, A.~Bapna, O.~Firat, M.~X. Chen, D.~Chen, H.~Lee, J.~Ngiam
  et~al. 2018.
\newblock \href {https://doi.org/10.48550/ARXIV.1811.06965} {Gpipe: Efficient
  training of giant neural networks using pipeline parallelism}.

\bibitem[{Huang et~al.(2023)Huang, Shen, Zhang, Zhou, Rong, and
  Xiong}]{huang2023transformerpatcher}
Z.~Huang, Y.~Shen, X.~Zhang, J.~Zhou, W.~Rong and Z.~Xiong. 2023.
\newblock \href {https://openreview.net/forum?id=4oYUGeGBPm}
  {Transformer-patcher: One mistake worth one neuron}.
\newblock In \emph{The Eleventh International Conference on Learning
  Representations}.

\bibitem[{Hubara et~al.(2021)Hubara, Chmiel, Island, Banner, Naor, and
  Soudry}]{hubara2021sparsity}
I.~Hubara, B.~Chmiel, M.~Island, R.~Banner, J.~Naor and D.~Soudry. 2021.
\newblock \href
  {https://proceedings.neurips.cc/paper_files/paper/2021/file/b0490b85e92b64dbb5db76bf8fca6a82-Paper.pdf}
  {Accelerated sparse neural training: A provable and efficient method to find
  n:m transposable masks}.
\newblock In \emph{Advances in Neural Information Processing Systems},
  volume~34, pages 21099--21111. Curran Associates, Inc.

\bibitem[{HuggingFace(2023)}]{hugging_chat}
HuggingFace. 2023.
\newblock Huggingchat v0.3.0.
\newblock Available from: \url{https://huggingface.co/chat}.
\newblock Accessed: 28/06/2023.

\bibitem[{Hwang et~al.(2022)Hwang, Cui, Xiong, Yang, Liu, Hu, Wang, Salas,
  Jose, Ram et~al.}]{hwang2022tutel}
C.~Hwang, W.~Cui, Y.~Xiong, Z.~Yang, Z.~Liu, H.~Hu, Z.~Wang, R.~Salas et~al.
  2022.
\newblock Tutel: Adaptive mixture-of-experts at scale.
\newblock \emph{arXiv preprint arXiv:2206.03382}.

\bibitem[{Ioannidis(2005)}]{ioannidisWhyMostPublished2005}
J.~P.~A. Ioannidis. 2005.
\newblock \href {https://doi.org/10.1371/journal.pmed.0020124} {Why {Most}
  {Published} {Research} {Findings} {Are} {False}}.
\newblock \emph{PLoS Medicine}, 2(8):e124.

\bibitem[{Ippolito et~al.(2022)Ippolito, Yuan, Coenen, and
  Burnam}]{ippolito2022creative}
D.~Ippolito, A.~Yuan, A.~Coenen and S.~Burnam. 2022.
\newblock Creative writing with an ai-powered writing assistant: Perspectives
  from professional writers.
\newblock \emph{arXiv preprint arXiv:2211.05030}.

\bibitem[{Irving et~al.(2018)Irving, Christiano, and Amodei}]{irving2018ai}
G.~Irving, P.~Christiano and D.~Amodei. 2018.
\newblock Ai safety via debate.
\newblock \emph{arXiv preprint arXiv:1805.00899}.

\bibitem[{Iu and Wong(2023)}]{iu2023chatgpt}
K.~Y. Iu and V.~M.-Y. Wong. 2023.
\newblock Chatgpt by openai: The end of litigation lawyers?
\newblock \emph{Available at SSRN}.

\bibitem[{Iyer et~al.(2022)Iyer, Lin, Pasunuru, Mihaylov, Simig, Yu, Shuster,
  Wang, Liu, Koura, Li, O'Horo, Pereyra, Wang, Dewan, Celikyilmaz, Zettlemoyer,
  and Stoyanov}]{opt_iml}
S.~Iyer, X.~V. Lin, R.~Pasunuru, T.~Mihaylov, D.~Simig, P.~Yu, K.~Shuster,
  T.~Wang et~al. 2022.
\newblock \href {https://doi.org/10.48550/ARXIV.2212.12017} {Opt-iml: Scaling
  language model instruction meta learning through the lens of generalization}.

\bibitem[{Izacard et~al.(2022)Izacard, Lewis, Lomeli, Hosseini, Petroni,
  Schick, Dwivedi-Yu, Joulin, Riedel, and Grave}]{atlas}
G.~Izacard, P.~Lewis, M.~Lomeli, L.~Hosseini, F.~Petroni, T.~Schick,
  J.~Dwivedi-Yu, A.~Joulin et~al. 2022.
\newblock Few-shot learning with retrieval augmented language models.
\newblock \emph{arXiv preprint arXiv:2208.03299}.

\bibitem[{Jacovi et~al.(2023)Jacovi, Caciularu, Goldman, and
  Goldberg}]{jacoviStopUploadingTest2023}
A.~Jacovi, A.~Caciularu, O.~Goldman and Y.~Goldberg. 2023.
\newblock \href {http://arxiv.org/abs/2305.10160} {Stop {Uploading} {Test}
  {Data} in {Plain} {Text}: {Practical} {Strategies} for {Mitigating} {Data}
  {Contamination} by {Evaluation} {Benchmarks}}.
\newblock ArXiv:2305.10160 [cs].

\bibitem[{Jain et~al.(2023)Jain, Saifullah, Wen, Kirchenbauer, Shu, Saha,
  Goldblum, Geiping, and Goldstein}]{jain2023bring}
N.~Jain, K.~Saifullah, Y.~Wen, J.~Kirchenbauer, M.~Shu, A.~Saha, M.~Goldblum,
  J.~Geiping et~al. 2023.
\newblock Bring your own data! self-supervised evaluation for large language
  models.
\newblock \emph{arXiv preprint arXiv:23062.13651}.

\bibitem[{Jang et~al.(2023)Jang, Kim, Ye, Kim, Logeswaran, Lee, Lee, and
  Seo}]{jangExploringBenefitsTraining2023a}
J.~Jang, S.~Kim, S.~Ye, D.~Kim, L.~Logeswaran, M.~Lee, K.~Lee and M.~Seo. 2023.
\newblock \href {http://arxiv.org/abs/2302.03202} {Exploring the {Benefits} of
  {Training} {Expert} {Language} {Models} over {Instruction} {Tuning}}.
\newblock ArXiv:2302.03202 [cs].

\bibitem[{Jeliazkov et~al.(2023)Jeliazkov, del Alamo, and
  Karpiak}]{jeliazkov2023hallucinate}
J.~R. Jeliazkov, D.~del Alamo and J.~D. Karpiak. 2023.
\newblock \href {https://doi.org/10.1101/2023.05.23.541774} {Esmfold
  hallucinates native-like protein sequences}.
\newblock \emph{bioRxiv}.

\bibitem[{Ji et~al.(2023)Ji, Lee, Frieske, Yu, Su, Xu, Ishii, Bang, Madotto,
  and Fung}]{jiSurveyHallucinationNatural2023}
Z.~Ji, N.~Lee, R.~Frieske, T.~Yu, D.~Su, Y.~Xu, E.~Ishii, Y.~J. Bang et~al.
  2023.
\newblock \href {https://doi.org/10.1145/3571730} {Survey of {Hallucination} in
  {Natural} {Language} {Generation}}.
\newblock \emph{ACM Computing Surveys}, 55(12):1--38.

\bibitem[{Jiang et~al.(2022)Jiang, Xu, Zhu, Han, Zhang, and Zhu}]{jiang2022mpi}
G.~Jiang, M.~Xu, S.-C. Zhu, W.~Han, C.~Zhang and Y.~Zhu. 2022.
\newblock Mpi: Evaluating and inducing personality in pre-trained language
  models.
\newblock \emph{arXiv preprint arXiv:2206.07550}.

\bibitem[{Jiao et~al.(2020)Jiao, Yin, Shang, Jiang, Chen, Li, Wang, and
  Liu}]{jiao-etal-2020-tinybert}
X.~Jiao, Y.~Yin, L.~Shang, X.~Jiang, X.~Chen, L.~Li, F.~Wang and Q.~Liu. 2020.
\newblock \href {https://doi.org/10.18653/v1/2020.findings-emnlp.372}
  {{T}iny{BERT}: Distilling {BERT} for natural language understanding}.
\newblock In \emph{Findings of the Association for Computational Linguistics:
  EMNLP 2020}, pages 4163--4174, Online. Association for Computational
  Linguistics.

\bibitem[{Jin et~al.(2023)Jin, Liu, Lyu, Poff, Sachan, Mihalcea, Diab, and
  Schölkopf}]{jin2023large}
Z.~Jin, J.~Liu, Z.~Lyu, S.~Poff, M.~Sachan, R.~Mihalcea, M.~Diab and
  B.~Schölkopf. 2023.
\newblock \href {http://arxiv.org/abs/2306.05836} {Can large language models
  infer causation from correlation?}

\bibitem[{Jinich et~al.(2022)Jinich, Nazia, Tellez, Rappoport, AlQuraishi, and
  Rhee}]{jinich2022predicting}
A.~Jinich, S.~Z. Nazia, A.~V. Tellez, D.~Rappoport, M.~AlQuraishi and K.~Rhee.
  2022.
\newblock Predicting enzyme substrate chemical structure with protein language
  models.
\newblock \emph{bioRxiv}, pages 2022--09.

\bibitem[{{Jonathan Frankle
  [@jefrankle]}(2022)}]{jonathanfrankle[@jefrankle]LouderPeopleBack2022}
{Jonathan Frankle [@jefrankle]}. 2022.
\newblock \href {https://twitter.com/jefrankle/status/1577313906250465282}
  {Louder for the people in the back: {LARGE} {MODELS} ({GPT}, {DALLE}) =
  {DATABASES} {PROMPTS} = {QUERIES} {OUTPUTS} = {RESPONSES} {NNs} find new
  relations w/in data. {Anyone}, no matter the resources, can study better
  querying langs and possibly beat a big model they could never afford to
  train.}

\bibitem[{Jones(2022)}]{jones-2022-development}
D.~Jones. 2022.
\newblock \href {https://aclanthology.org/2022.cltw-1.8} {Development and
  evaluation of speech recognition for the {W}elsh language}.
\newblock In \emph{Proceedings of the 4th Celtic Language Technology Workshop
  within LREC2022}, pages 52--59, Marseille, France. European Language
  Resources Association.

\bibitem[{Jumper et~al.(2021)Jumper, Evans, Pritzel, Green, Figurnov,
  Ronneberger, Tunyasuvunakool, Bates, {\v{Z}}{\'\i}dek, Potapenko
  et~al.}]{alphafold2}
J.~Jumper, R.~Evans, A.~Pritzel, T.~Green, M.~Figurnov, O.~Ronneberger,
  K.~Tunyasuvunakool, R.~Bates et~al. 2021.
\newblock Highly accurate protein structure prediction with alphafold.
\newblock \emph{Nature}, 596(7873):583--589.

\bibitem[{Kaddour(2022)}]{kaddour2022stop}
J.~Kaddour. 2022.
\newblock Stop wasting my time! saving days of imagenet and bert training with
  latest weight averaging.
\newblock \emph{arXiv preprint arXiv:2209.14981}.

\bibitem[{Kaddour(2023)}]{minipile}
J.~Kaddour. 2023.
\newblock \href {https://doi.org/10.48550/arXiv.2304.08442} {The {MiniPile}
  {Challenge} for {Data}-{Efficient} {Language} {Models}}.
\newblock ArXiv:2304.08442 [cs].

\bibitem[{Kaddour et~al.(2023)Kaddour, Key, Nawrot, Minervini, and
  Kusner}]{ntng}
J.~Kaddour, O.~Key, P.~Nawrot, P.~Minervini and M.~J. Kusner. 2023.
\newblock \href {https://doi.org/10.48550/arXiv.2307.06440} {No {Train} {No}
  {Gain}: {Revisiting} {Efficient} {Training} {Algorithms} {For}
  {Transformer}-based {Language} {Models}}.
\newblock ArXiv:2307.06440 [cs].

\bibitem[{Kaddour et~al.(2022{\natexlab{a}})Kaddour, Liu, Silva, and
  Kusner}]{kaddour2022when}
J.~Kaddour, L.~Liu, R.~Silva and M.~Kusner. 2022{\natexlab{a}}.
\newblock \href {https://openreview.net/forum?id=vDeh2yxTvuh} {When do flat
  minima optimizers work?}
\newblock In \emph{Advances in Neural Information Processing Systems}.

\bibitem[{Kaddour et~al.(2022{\natexlab{b}})Kaddour, Lynch, Liu, Kusner, and
  Silva}]{cml}
J.~Kaddour, A.~Lynch, Q.~Liu, M.~J. Kusner and R.~Silva. 2022{\natexlab{b}}.
\newblock Causal machine learning: A survey and open problems.
\newblock \emph{arXiv preprint arXiv:2206.15475}.

\bibitem[{Kaddour et~al.(2021)Kaddour, Zhu, Liu, Kusner, and Silva}]{sin}
J.~Kaddour, Y.~Zhu, Q.~Liu, M.~J. Kusner and R.~Silva. 2021.
\newblock \href
  {https://proceedings.neurips.cc/paper/2021/hash/d02e9bdc27a894e882fa0c9055c99722-Abstract.html}
  {Causal {Effect} {Inference} for {Structured} {Treatments}}.
\newblock In \emph{Advances in {Neural} {Information} {Processing} {Systems}},
  volume~34, pages 24841--24854. Curran Associates, Inc.

\bibitem[{Kale et~al.(2021)Kale, Siddhant, Al-Rfou, Xue, Constant, and
  Johnson}]{kale-etal-2021-nmt5}
M.~Kale, A.~Siddhant, R.~Al-Rfou, L.~Xue, N.~Constant and M.~Johnson. 2021.
\newblock \href {https://doi.org/10.18653/v1/2021.acl-short.87} {nm{T}5 - is
  parallel data still relevant for pre-training massively multilingual language
  models?}
\newblock In \emph{Proceedings of the 59th Annual Meeting of the Association
  for Computational Linguistics and the 11th International Joint Conference on
  Natural Language Processing (Volume 2: Short Papers)}, pages 683--691,
  Online. Association for Computational Linguistics.

\bibitem[{Kaplan et~al.(2020)Kaplan, McCandlish, Henighan, Brown, Chess, Child,
  Gray, Radford, Wu, and Amodei}]{kaplan2020scaling}
J.~Kaplan, S.~McCandlish, T.~Henighan, T.~B. Brown, B.~Chess, R.~Child,
  S.~Gray, A.~Radford et~al. 2020.
\newblock Scaling laws for neural language models.
\newblock \emph{arXiv preprint arXiv:2001.08361}.

\bibitem[{Karpathy(2023)}]{karpathy2023tokenization}
A.~Karpathy. 2023.
\newblock \href {https://twitter.com/karpathy/status/1657949234535211009?s=20}
  {Tokenization issues (tweet)}.

\bibitem[{Katz et~al.(2023)Katz, Bommarito, Gao, and Arredondo}]{katz2023gpt}
D.~M. Katz, M.~J. Bommarito, S.~Gao and P.~Arredondo. 2023.
\newblock Gpt-4 passes the bar exam.
\newblock \emph{Available at SSRN 4389233}.

\bibitem[{Kazemnejad et~al.(2023)Kazemnejad, Padhi, Ramamurthy, Das, and
  Reddy}]{kazemnejad2023impact}
A.~Kazemnejad, I.~Padhi, K.~N. Ramamurthy, P.~Das and S.~Reddy. 2023.
\newblock The impact of positional encoding on length generalization in
  transformers.
\newblock \emph{arXiv preprint arXiv:2305.19466}.

\bibitem[{Kenton et~al.(2021)Kenton, Everitt, Weidinger, Gabriel, Mikulik, and
  Irving}]{kenton2021alignment}
Z.~Kenton, T.~Everitt, L.~Weidinger, I.~Gabriel, V.~Mikulik and G.~Irving.
  2021.
\newblock Alignment of language agents.
\newblock \emph{arXiv preprint arXiv:2103.14659}.

\bibitem[{Keskar et~al.(2019)Keskar, McCann, Varshney, Xiong, and
  Socher}]{keskar2019ctrl}
N.~S. Keskar, B.~McCann, L.~R. Varshney, C.~Xiong and R.~Socher. 2019.
\newblock Ctrl: A conditional transformer language model for controllable
  generation.
\newblock \emph{arXiv preprint arXiv:1909.05858}.

\bibitem[{Khattab et~al.(2023)Khattab, Santhanam, Li, Hall, Liang, Potts, and
  Zaharia}]{khattabDemonstrateSearchPredictComposingRetrieval2023}
O.~Khattab, K.~Santhanam, X.~L. Li, D.~Hall, P.~Liang, C.~Potts and M.~Zaharia.
  2023.
\newblock \href {https://doi.org/10.48550/arXiv.2212.14024}
  {Demonstrate-{Search}-{Predict}: {Composing} retrieval and language models
  for knowledge-intensive {NLP}}.
\newblock ArXiv:2212.14024 [cs].

\bibitem[{Kiela et~al.(2021)Kiela, Bartolo, Nie, Kaushik, Geiger, Wu, Vidgen,
  Prasad, Singh, Ringshia et~al.}]{kiela2021dynabench}
D.~Kiela, M.~Bartolo, Y.~Nie, D.~Kaushik, A.~Geiger, Z.~Wu, B.~Vidgen,
  G.~Prasad et~al. 2021.
\newblock Dynabench: Rethinking benchmarking in nlp.
\newblock \emph{arXiv preprint arXiv:2104.14337}.

\bibitem[{Kim et~al.(2022)Kim, Kim, and Mozafari}]{kim2022provable}
J.~Kim, M.~Kim and B.~Mozafari. 2022.
\newblock Provable memorization capacity of transformers.
\newblock In \emph{The Eleventh International Conference on Learning
  Representations}.

\bibitem[{Kim et~al.(2023)Kim, Mangalam, Malik, Mahoney, Gholami, and
  Keutzer}]{kim2023big}
S.~Kim, K.~Mangalam, J.~Malik, M.~W. Mahoney, A.~Gholami and K.~Keutzer. 2023.
\newblock Big little transformer decoder.
\newblock \emph{arXiv preprint arXiv:2302.07863}.

\bibitem[{Kim(2022)}]{kim-2022-revisiting}
T.~Kim. 2022.
\newblock \href {https://aclanthology.org/2022.coling-1.479} {Revisiting the
  practical effectiveness of constituency parse extraction from pre-trained
  language models}.
\newblock In \emph{Proceedings of the 29th International Conference on
  Computational Linguistics}, pages 5398--5408, Gyeongju, Republic of Korea.
  International Committee on Computational Linguistics.

\bibitem[{Kinch et~al.(2021)Kinch, Schaeffer, Kryshtafovych, and
  Grishin}]{kinch2021casp}
L.~N. Kinch, R.~D. Schaeffer, A.~Kryshtafovych and N.~V. Grishin. 2021.
\newblock \href {https://doi.org/https://doi.org/10.1002/prot.26202} {Target
  classification in the 14th round of the critical assessment of protein
  structure prediction (casp14)}.
\newblock \emph{Proteins: Structure, Function, and Bioinformatics},
  89(12):1618--1632.

\bibitem[{Kirchenbauer et~al.(2023{\natexlab{a}})Kirchenbauer, Geiping, Wen,
  Katz, Miers, and Goldstein}]{kirchenbauerWatermarkLargeLanguage2023}
J.~Kirchenbauer, J.~Geiping, Y.~Wen, J.~Katz, I.~Miers and T.~Goldstein.
  2023{\natexlab{a}}.
\newblock \href {http://arxiv.org/abs/2301.10226} {A {Watermark} for {Large}
  {Language} {Models}}.
\newblock ArXiv:2301.10226 [cs].

\bibitem[{Kirchenbauer et~al.(2023{\natexlab{b}})Kirchenbauer, Geiping, Wen,
  Shu, Saifullah, Kong, Fernando, Saha, Goldblum, and
  Goldstein}]{kirchenbauerReliabilityWatermarksLarge2023}
J.~Kirchenbauer, J.~Geiping, Y.~Wen, M.~Shu, K.~Saifullah, K.~Kong,
  K.~Fernando, A.~Saha et~al. 2023{\natexlab{b}}.
\newblock \href {http://arxiv.org/abs/2306.04634} {On the {Reliability} of
  {Watermarks} for {Large} {Language} {Models}}.
\newblock ArXiv:2306.04634 [cs].

\bibitem[{Klein et~al.(2018)Klein, Vianello, Hasselman, Adams, Adams~Jr, Alper,
  Aveyard, Axt, Babalola, Bahn{\'\i}k et~al.}]{klein2018many}
R.~A. Klein, M.~Vianello, F.~Hasselman, B.~G. Adams, R.~B. Adams~Jr, S.~Alper,
  M.~Aveyard, J.~R. Axt et~al. 2018.
\newblock Many labs 2: Investigating variation in replicability across samples
  and settings.
\newblock \emph{Advances in Methods and Practices in Psychological Science},
  1(4):443--490.

\bibitem[{Kocetkov et~al.(2022)Kocetkov, Li, Allal, Li, Mou, Ferrandis,
  Jernite, Mitchell, Hughes, Wolf, Bahdanau, von Werra, and
  de~Vries}]{the_stack}
D.~Kocetkov, R.~Li, L.~B. Allal, J.~Li, C.~Mou, C.~M. Ferrandis, Y.~Jernite,
  M.~Mitchell et~al. 2022.
\newblock \href {https://doi.org/10.48550/ARXIV.2211.15533} {The stack: 3 tb of
  permissively licensed source code}.

\bibitem[{Kocoń et~al.(2023)Kocoń, Cichecki, Kaszyca, Kochanek, Szydło,
  Baran, Bielaniewicz, Gruza, Janz, Kanclerz, Kocoń, Koptyra,
  Mieleszczenko-Kowszewicz, Miłkowski, Oleksy, Piasecki, Radliński, Wojtasik,
  Woźniak, and Kazienko}]{chatgpt_jack_of_all_trades}
J.~Kocoń, I.~Cichecki, O.~Kaszyca, M.~Kochanek, D.~Szydło, J.~Baran,
  J.~Bielaniewicz, M.~Gruza et~al. 2023.
\newblock \href {https://doi.org/10.48550/ARXIV.2302.10724} {Chatgpt: Jack of
  all trades, master of none}.

\bibitem[{Kojima et~al.(2022)Kojima, Gu, Reid, Matsuo, and
  Iwasawa}]{kojima2022large}
T.~Kojima, S.~S. Gu, M.~Reid, Y.~Matsuo and Y.~Iwasawa. 2022.
\newblock \href {https://openreview.net/forum?id=e2TBb5y0yFf} {Large language
  models are zero-shot reasoners}.
\newblock In \emph{Advances in Neural Information Processing Systems}.

\bibitem[{K{\"o}pf et~al.(2023)K{\"o}pf, Kilcher, von R{\"u}tte, Anagnostidis,
  Tam, Stevens, Barhoum, Duc, Stanley, Nagyfi et~al.}]{kopf2023openassistant}
A.~K{\"o}pf, Y.~Kilcher, D.~von R{\"u}tte, S.~Anagnostidis, Z.-R. Tam,
  K.~Stevens, A.~Barhoum, N.~M. Duc et~al. 2023.
\newblock Openassistant conversations--democratizing large language model
  alignment.
\newblock \emph{arXiv preprint arXiv:2304.07327}.

\bibitem[{Korbak et~al.(2023)Korbak, Shi, Chen, Bhalerao, Buckley, Phang,
  Bowman, and Perez}]{korbak2023rlhfpre}
T.~Korbak, K.~Shi, A.~Chen, R.~Bhalerao, C.~L. Buckley, J.~Phang, S.~R. Bowman
  and E.~Perez. 2023.
\newblock Pretraining language models with human preferences.
\newblock \emph{arXiv preprint arXiv:2302.08582}.

\bibitem[{Korngiebel and Mooney(2021)}]{korngiebel2021considering}
D.~M. Korngiebel and S.~D. Mooney. 2021.
\newblock Considering the possibilities and pitfalls of generative pre-trained
  transformer 3 (gpt-3) in healthcare delivery.
\newblock \emph{NPJ Digital Medicine}, 4(1):1--3.

\bibitem[{Kosinski(2023)}]{kosinski2023theory}
M.~Kosinski. 2023.
\newblock \href {http://arxiv.org/abs/2302.02083} {Theory of mind may have
  spontaneously emerged in large language models}.

\bibitem[{Krause et~al.(2021)Krause, Gotmare, McCann, Keskar, Joty, Socher, and
  Rajani}]{krause-etal-2021-gedi-generative}
B.~Krause, A.~D. Gotmare, B.~McCann, N.~S. Keskar, S.~Joty, R.~Socher and N.~F.
  Rajani. 2021.
\newblock \href {https://doi.org/10.18653/v1/2021.findings-emnlp.424}
  {{G}e{D}i: Generative discriminator guided sequence generation}.
\newblock In \emph{Findings of the Association for Computational Linguistics:
  EMNLP 2021}, pages 4929--4952, Punta Cana, Dominican Republic. Association
  for Computational Linguistics.

\bibitem[{Krawczyk(2018)}]{krawczyk2018introduction}
D.~C. Krawczyk. 2018.
\newblock Introduction to reasoning.
\newblock \emph{Reasoning—The Neuroscience of How We Think; Academic Press:
  Cambridge, MA, USA}, pages 1--11.

\bibitem[{Krishna et~al.(2023)Krishna, Song, Karpinska, Wieting, and
  Iyyer}]{krishnaParaphrasingEvadesDetectors2023}
K.~Krishna, Y.~Song, M.~Karpinska, J.~Wieting and M.~Iyyer. 2023.
\newblock \href {http://arxiv.org/abs/2303.13408} {Paraphrasing evades
  detectors of {AI}-generated text, but retrieval is an effective defense}.
\newblock ArXiv:2303.13408 [cs].

\bibitem[{Kudo(2018)}]{kudo-2018-subword}
T.~Kudo. 2018.
\newblock \href {https://doi.org/10.18653/v1/P18-1007} {Subword regularization:
  Improving neural network translation models with multiple subword
  candidates}.
\newblock In \emph{Proceedings of the 56th Annual Meeting of the Association
  for Computational Linguistics (Volume 1: Long Papers)}, pages 66--75,
  Melbourne, Australia. Association for Computational Linguistics.

\bibitem[{Kudo and Richardson(2018)}]{kudo2018sentencepiece}
T.~Kudo and J.~Richardson. 2018.
\newblock Sentencepiece: A simple and language independent subword tokenizer
  and detokenizer for neural text processing.
\newblock \emph{arXiv preprint arXiv:1808.06226}.

\bibitem[{Kulkarni(2021)}]{kulkarniGitHubCopilotAI2021}
A.~Kulkarni. 2021.
\newblock \href
  {https://analyticsdrift.com/github-copilot-ai-is-leaking-functional-api-keys/}
  {{GitHub} {Copilot} {AI} {Is} {Leaking} {Functional} {API} {Keys}}.

\bibitem[{K{\"u}nzel et~al.(2019)K{\"u}nzel, Sekhon, Bickel, and
  Yu}]{kunzel2019metalearners}
S.~R. K{\"u}nzel, J.~S. Sekhon, P.~J. Bickel and B.~Yu. 2019.
\newblock Metalearners for estimating heterogeneous treatment effects using
  machine learning.
\newblock \emph{Proceedings of the national academy of sciences},
  116(10):4156--4165.

\bibitem[{Kwon et~al.(2023)Kwon, Li, Zhuang, Sheng, Zheng, Yu, Gonzalez, Zhang,
  and Stoica}]{kwon2023vllm}
W.~Kwon, Z.~Li, S.~Zhuang, Y.~Sheng, L.~Zheng, C.~Yu, J.~Gonzalez, H.~Zhang
  et~al. 2023.
\newblock \href {https://vllm.ai/} {vllm: Easy, fast, and cheap llm serving
  with pagedattention}.

\bibitem[{Kıcıman et~al.(2023)Kıcıman, Ness, Sharma, and
  Tan}]{kiciman2023causal}
E.~Kıcıman, R.~Ness, A.~Sharma and C.~Tan. 2023.
\newblock \href {http://arxiv.org/abs/2305.00050} {Causal reasoning and large
  language models: Opening a new frontier for causality}.

\bibitem[{Lab(2023)}]{promptslabAwesomePromptEngineering2023}
P.~Lab. 2023.
\newblock \href {https://github.com/promptslab/Awesome-Prompt-Engineering}
  {Awesome-{Prompt}-{Engineering}}.
\newblock Original-date: 2023-02-09T18:22:52Z.

\bibitem[{Lampinen et~al.(2023)Lampinen, Chan, Dasgupta, Nam, and
  Wang}]{lampinen2023passive}
A.~K. Lampinen, S.~C. Chan, I.~Dasgupta, A.~J. Nam and J.~X. Wang. 2023.
\newblock Passive learning of active causal strategies in agents and language
  models.
\newblock \emph{arXiv preprint arXiv:2305.16183}.

\bibitem[{Lauren{\c{c}}on et~al.(2022)Lauren{\c{c}}on, Saulnier, Wang, Akiki,
  del Moral, Scao, Werra, Mou, Ponferrada, Nguyen, Frohberg, {\v{S}}a{\v{s}}ko,
  Lhoest, McMillan-Major, Dupont, Biderman, Rogers, allal, Toni, Pistilli,
  Nguyen, Nikpoor, Masoud, Colombo, de~la Rosa, Villegas, Thrush, Longpre,
  Nagel, Weber, Mu{\~n}oz, Zhu, Strien, Alyafeai, Almubarak, Chien,
  Gonzalez-Dios, Soroa, Lo, Dey, Suarez, Gokaslan, Bose, Adelani, Phan, Tran,
  Yu, Pai, Chim, Lepercq, Ilic, Mitchell, Luccioni, and Jernite}]{roots}
H.~Lauren{\c{c}}on, L.~Saulnier, T.~Wang, C.~Akiki, A.~V. del Moral, T.~L.
  Scao, L.~V. Werra, C.~Mou et~al. 2022.
\newblock \href {https://openreview.net/forum?id=UoEw6KigkUn} {The bigscience
  {ROOTS} corpus: A 1.6{TB} composite multilingual dataset}.
\newblock In \emph{Thirty-sixth Conference on Neural Information Processing
  Systems Datasets and Benchmarks Track}.

\bibitem[{Lazaridou et~al.(2022)Lazaridou, Gribovskaya, Stokowiec, and
  Grigorev}]{lazaridou2022internetaugmented}
A.~Lazaridou, E.~Gribovskaya, W.~Stokowiec and N.~Grigorev. 2022.
\newblock \href {http://arxiv.org/abs/2203.05115} {Internet-augmented language
  models through few-shot prompting for open-domain question answering}.

\bibitem[{Lee et~al.(2023{\natexlab{a}})Lee, Miranda, and
  Koyejo}]{leeScaleDiversityCoefficient2023}
A.~Lee, B.~Miranda and S.~Koyejo. 2023{\natexlab{a}}.
\newblock \href {http://arxiv.org/abs/2306.13840} {Beyond {Scale}: the
  {Diversity} {Coefficient} as a {Data} {Quality} {Metric} {Demonstrates}
  {LLMs} are {Pre}-trained on {Formally} {Diverse} {Data}}.
\newblock ArXiv:2306.13840 [cs].

\bibitem[{Lee et~al.(2023{\natexlab{b}})Lee, Lee, Ha, Kim, Lee, Lee, and
  Song}]{lee2023query}
D.~Lee, J.~Lee, J.-W. Ha, J.-H. Kim, S.-W. Lee, H.~Lee and H.~O. Song.
  2023{\natexlab{b}}.
\newblock Query-efficient black-box red teaming via bayesian optimization.
\newblock \emph{arXiv preprint arXiv:2305.17444}.

\bibitem[{Lee et~al.(2018)Lee, Firat, Agarwal, Fannjiang, and
  Sussillo}]{lee2018hallucinations}
K.~Lee, O.~Firat, A.~Agarwal, C.~Fannjiang and D.~Sussillo. 2018.
\newblock Hallucinations in neural machine translation.

\bibitem[{Lee et~al.(2021)Lee, Ippolito, Nystrom, Zhang, Eck, Callison-Burch,
  and Carlini}]{lee2021deduplicating}
K.~Lee, D.~Ippolito, A.~Nystrom, C.~Zhang, D.~Eck, C.~Callison-Burch and
  N.~Carlini. 2021.
\newblock Deduplicating training data makes language models better.
\newblock \emph{arXiv preprint arXiv:2107.06499}.

\bibitem[{Lee et~al.()Lee, Ping, Xu, Patwary, Fung, Shoeybi, and
  Catanzaro}]{leeFactualityEnhancedLanguage}
N.~Lee, W.~Ping, P.~Xu, M.~Patwary, P.~Fung, M.~Shoeybi and B.~Catanzaro.
\newblock Factuality {Enhanced} {Language} {Models} for {Open}-{Ended} {Text}
  {Generation}.

\bibitem[{Lee et~al.(2023{\natexlab{c}})Lee, Bubeck, and
  Petro}]{lee2023benefits}
P.~Lee, S.~Bubeck and J.~Petro. 2023{\natexlab{c}}.
\newblock Benefits, limits, and risks of gpt-4 as an ai chatbot for medicine.
\newblock \emph{New England Journal of Medicine}, 388(13):1233--1239.

\bibitem[{Lehman et~al.(2023)Lehman, Hernandez, Mahajan, Wulff, Smith, Ziegler,
  Nadler, Szolovits, Johnson, and
  Alsentzer}]{do_we_need_clinical_language_models}
E.~Lehman, E.~Hernandez, D.~Mahajan, J.~Wulff, M.~J. Smith, Z.~Ziegler,
  D.~Nadler, P.~Szolovits et~al. 2023.
\newblock \href {https://doi.org/10.48550/ARXIV.2302.08091} {Do we still need
  clinical language models?}

\bibitem[{Lepikhin et~al.(2020)Lepikhin, Lee, Xu, Chen, Firat, Huang, Krikun,
  Shazeer, and Chen}]{gshard}
D.~Lepikhin, H.~Lee, Y.~Xu, D.~Chen, O.~Firat, Y.~Huang, M.~Krikun, N.~Shazeer
  et~al. 2020.
\newblock \href {https://doi.org/10.48550/ARXIV.2006.16668} {Gshard: Scaling
  giant models with conditional computation and automatic sharding}.

\bibitem[{Lester et~al.(2021)Lester, Al-Rfou, and
  Constant}]{lester-etal-2021-power}
B.~Lester, R.~Al-Rfou and N.~Constant. 2021.
\newblock \href {https://doi.org/10.18653/v1/2021.emnlp-main.243} {The power of
  scale for parameter-efficient prompt tuning}.
\newblock In \emph{Proceedings of the 2021 Conference on Empirical Methods in
  Natural Language Processing}, pages 3045--3059, Online and Punta Cana,
  Dominican Republic. Association for Computational Linguistics.

\bibitem[{Leviathan et~al.(2022)Leviathan, Kalman, and
  Matias}]{leviathan2022fast}
Y.~Leviathan, M.~Kalman and Y.~Matias. 2022.
\newblock Fast inference from transformers via speculative decoding.
\newblock \emph{arXiv preprint arXiv:2211.17192}.

\bibitem[{Levine et~al.(2023)Levine, Tuwani, Kompa, Varma, Finlayson, Mehrotra,
  and Beam}]{levine2023diagnostic}
D.~M. Levine, R.~Tuwani, B.~Kompa, A.~Varma, S.~G. Finlayson, A.~Mehrotra and
  A.~Beam. 2023.
\newblock The diagnostic and triage accuracy of the gpt-3 artificial
  intelligence model.
\newblock \emph{medRxiv}, pages 2023--01.

\bibitem[{Lewis et~al.(2021)Lewis, Bhosale, Dettmers, Goyal, and
  Zettlemoyer}]{base}
M.~Lewis, S.~Bhosale, T.~Dettmers, N.~Goyal and L.~Zettlemoyer. 2021.
\newblock \href {https://doi.org/10.48550/ARXIV.2103.16716} {Base layers:
  Simplifying training of large, sparse models}.

\bibitem[{Lewis et~al.(2020{\natexlab{a}})Lewis, Liu, Goyal, Ghazvininejad,
  Mohamed, Levy, Stoyanov, and Zettlemoyer}]{lewis-etal-2020-bart}
M.~Lewis, Y.~Liu, N.~Goyal, M.~Ghazvininejad, A.~Mohamed, O.~Levy, V.~Stoyanov
  and L.~Zettlemoyer. 2020{\natexlab{a}}.
\newblock \href {https://doi.org/10.18653/v1/2020.acl-main.703} {{BART}:
  Denoising sequence-to-sequence pre-training for natural language generation,
  translation, and comprehension}.
\newblock In \emph{Proceedings of the 58th Annual Meeting of the Association
  for Computational Linguistics}, pages 7871--7880, Online. Association for
  Computational Linguistics.

\bibitem[{Lewis et~al.(2020{\natexlab{b}})Lewis, Perez, Piktus, Petroni,
  Karpukhin, Goyal, K{\"u}ttler, Lewis, Yih, Rockt{\"a}schel
  et~al.}]{lewis2020retrieval}
P.~Lewis, E.~Perez, A.~Piktus, F.~Petroni, V.~Karpukhin, N.~Goyal,
  H.~K{\"u}ttler, M.~Lewis et~al. 2020{\natexlab{b}}.
\newblock Retrieval-augmented generation for knowledge-intensive nlp tasks.
\newblock \emph{Advances in Neural Information Processing Systems},
  33:9459--9474.

\bibitem[{Lewkowycz et~al.(2022)Lewkowycz, Andreassen, Dohan, Dyer,
  Michalewski, Ramasesh, Slone, Anil, Schlag, Gutman-Solo, Wu, Neyshabur,
  Gur-Ari, and Misra}]{minerva}
A.~Lewkowycz, A.~Andreassen, D.~Dohan, E.~Dyer, H.~Michalewski, V.~Ramasesh,
  A.~Slone, C.~Anil et~al. 2022.
\newblock \href {https://doi.org/10.48550/ARXIV.2206.14858} {Solving
  quantitative reasoning problems with language models}.

\bibitem[{Li et~al.(2021{\natexlab{a}})Li, Nye, and Andreas}]{li2021implicit}
B.~Z. Li, M.~Nye and J.~Andreas. 2021{\natexlab{a}}.
\newblock Implicit representations of meaning in neural language models.
\newblock \emph{arXiv preprint arXiv:2106.00737}.

\bibitem[{Li et~al.(2021{\natexlab{b}})Li, Awan, Tang, Rajbhandari, and
  He}]{li20211communication}
C.~Li, A.~A. Awan, H.~Tang, S.~Rajbhandari and Y.~He. 2021{\natexlab{b}}.
\newblock 1-bit lamb: Communication efficient large-scale large-batch training
  with lamb's convergence speed.
\newblock \emph{arXiv preprint arXiv:2104.06069}.

\bibitem[{Li et~al.(2023{\natexlab{a}})Li, Shao, Xie, Sheng, Zheng, Gonzalez,
  Stoica, Ma, and Zhang}]{longchat2023}
D.~Li, R.~Shao, A.~Xie, Y.~Sheng, L.~Zheng, J.~E. Gonzalez, I.~Stoica, X.~Ma
  et~al. 2023{\natexlab{a}}.
\newblock \href {https://lmsys.org/blog/2023-06-29-longchat} {How long can
  open-source llms truly promise on context length?}

\bibitem[{Li et~al.(2023{\natexlab{b}})Li, Guo, Fan, Xu, and
  Song}]{li2023multi}
H.~Li, D.~Guo, W.~Fan, M.~Xu and Y.~Song. 2023{\natexlab{b}}.
\newblock Multi-step jailbreaking privacy attacks on chatgpt.
\newblock \emph{arXiv preprint arXiv:2304.05197}.

\bibitem[{Li et~al.(2020)Li, Su, Duan, and Zheng}]{li2020linear}
R.~Li, J.~Su, C.~Duan and S.~Zheng. 2020.
\newblock Linear attention mechanism: An efficient attention for semantic
  segmentation.
\newblock \emph{arXiv preprint arXiv:2007.14902}.

\bibitem[{Li and Liang(2021)}]{li-liang-2021-prefix}
X.~L. Li and P.~Liang. 2021.
\newblock \href {https://doi.org/10.18653/v1/2021.acl-long.353} {Prefix-tuning:
  Optimizing continuous prompts for generation}.
\newblock In \emph{Proceedings of the 59th Annual Meeting of the Association
  for Computational Linguistics and the 11th International Joint Conference on
  Natural Language Processing (Volume 1: Long Papers)}, pages 4582--4597,
  Online. Association for Computational Linguistics.

\bibitem[{Li et~al.(2022{\natexlab{a}})Li, Lin, Zhang, Fu, Chen, Lou, and
  Chen}]{better_reasoners}
Y.~Li, Z.~Lin, S.~Zhang, Q.~Fu, B.~Chen, J.-G. Lou and W.~Chen.
  2022{\natexlab{a}}.
\newblock \href {https://doi.org/10.48550/ARXIV.2206.02336} {On the advance of
  making language models better reasoners}.

\bibitem[{Li et~al.(2022{\natexlab{b}})Li, Choi, Chung, Kushman, Schrittwieser,
  Leblond, Eccles, Keeling, Gimeno, Dal~Lago et~al.}]{li2022competition}
Y.~Li, D.~Choi, J.~Chung, N.~Kushman, J.~Schrittwieser, R.~Leblond, T.~Eccles,
  J.~Keeling et~al. 2022{\natexlab{b}}.
\newblock Competition-level code generation with alphacode.
\newblock \emph{Science}, 378(6624):1092--1097.

\bibitem[{Li et~al.(2023{\natexlab{c}})Li, You, Bhojanapalli, Li, Rawat, Reddi,
  Ye, Chern, Yu, Guo, and Kumar}]{liLazyNeuronPhenomenon2023a}
Z.~Li, C.~You, S.~Bhojanapalli, D.~Li, A.~S. Rawat, S.~J. Reddi, K.~Ye,
  F.~Chern et~al. 2023{\natexlab{c}}.
\newblock \href {http://arxiv.org/abs/2210.06313} {The {Lazy} {Neuron}
  {Phenomenon}: {On} {Emergence} of {Activation} {Sparsity} in {Transformers}}.
\newblock ArXiv:2210.06313 [cs, stat].

\bibitem[{Lian et~al.(2023)Lian, Li, Yala, and Darrell}]{lian2023llmgrounded}
L.~Lian, B.~Li, A.~Yala and T.~Darrell. 2023.
\newblock \href {http://arxiv.org/abs/2305.13655} {Llm-grounded diffusion:
  Enhancing prompt understanding of text-to-image diffusion models with large
  language models}.

\bibitem[{Liang et~al.(2023)Liang, Huang, Xia, Xu, Hausman, Ichter, Florence,
  and Zeng}]{liang2023code}
J.~Liang, W.~Huang, F.~Xia, P.~Xu, K.~Hausman, B.~Ichter, P.~Florence and
  A.~Zeng. 2023.
\newblock \href {http://arxiv.org/abs/2209.07753} {Code as policies: Language
  model programs for embodied control}.

\bibitem[{Liang et~al.(2021)Liang, Wu, Morency, and
  Salakhutdinov}]{liang2021towards}
P.~P. Liang, C.~Wu, L.-P. Morency and R.~Salakhutdinov. 2021.
\newblock Towards understanding and mitigating social biases in language
  models.
\newblock In \emph{International Conference on Machine Learning}, pages
  6565--6576. PMLR.

\bibitem[{Liang et~al.(2022)Liang, Bommasani, Lee, Tsipras, Soylu, Yasunaga,
  Zhang, Narayanan, Wu, Kumar et~al.}]{liang2022holistic}
P.~Liang, R.~Bommasani, T.~Lee, D.~Tsipras, D.~Soylu, M.~Yasunaga, Y.~Zhang,
  D.~Narayanan et~al. 2022.
\newblock Holistic evaluation of language models.
\newblock \emph{arXiv preprint arXiv:2211.09110}.

\bibitem[{Lieber et~al.(2021)Lieber, Sharir, Lenz, and
  Shoham}]{lieber2021jurassic}
O.~Lieber, O.~Sharir, B.~Lenz and Y.~Shoham. 2021.
\newblock Jurassic-1: Technical details and evaluation.
\newblock \emph{White Paper. AI21 Labs}, 1.

\bibitem[{Li{\'e}vin et~al.(2022)Li{\'e}vin, Hother, and
  Winther}]{lievin2022can}
V.~Li{\'e}vin, C.~E. Hother and O.~Winther. 2022.
\newblock Can large language models reason about medical questions?
\newblock \emph{arXiv preprint arXiv:2207.08143}.

\bibitem[{Lin et~al.(2020)Lin, Jaech, Li, Gormley, and
  Eisner}]{lin2020limitations}
C.-C. Lin, A.~Jaech, X.~Li, M.~R. Gormley and J.~Eisner. 2020.
\newblock Limitations of autoregressive models and their alternatives.
\newblock \emph{arXiv preprint arXiv:2010.11939}.

\bibitem[{Lin et~al.(2021{\natexlab{a}})Lin, Yang, Bai, Zhou, Jiang, Jia, Wang,
  Zhang, Li, Lin et~al.}]{lin2021m6}
J.~Lin, A.~Yang, J.~Bai, C.~Zhou, L.~Jiang, X.~Jia, A.~Wang, J.~Zhang et~al.
  2021{\natexlab{a}}.
\newblock M6-10t: A sharing-delinking paradigm for efficient multi-trillion
  parameter pretraining.
\newblock \emph{arXiv preprint arXiv:2110.03888}.

\bibitem[{Lin et~al.(2021{\natexlab{b}})Lin, Hilton, and
  Evans}]{lin2021truthfulqa}
S.~Lin, J.~Hilton and O.~Evans. 2021{\natexlab{b}}.
\newblock Truthfulqa: Measuring how models mimic human falsehoods.
\newblock \emph{arXiv preprint arXiv:2109.07958}.

\bibitem[{Lin et~al.(2022{\natexlab{a}})Lin, Mihaylov, Artetxe, Wang, Chen,
  Simig, Ott, Goyal, Bhosale, Du, Pasunuru, Shleifer, Koura, Chaudhary,
  O{'}Horo, Wang, Zettlemoyer, Kozareva, Diab, Stoyanov, and
  Li}]{lin-etal-2022-shot}
X.~V. Lin, T.~Mihaylov, M.~Artetxe, T.~Wang, S.~Chen, D.~Simig, M.~Ott,
  N.~Goyal et~al. 2022{\natexlab{a}}.
\newblock \href {https://aclanthology.org/2022.emnlp-main.616} {Few-shot
  learning with multilingual generative language models}.
\newblock In \emph{Proceedings of the 2022 Conference on Empirical Methods in
  Natural Language Processing}, pages 9019--9052, Abu Dhabi, United Arab
  Emirates. Association for Computational Linguistics.

\bibitem[{Lin and Chen(2023)}]{lin2023llm}
Y.-T. Lin and Y.-N. Chen. 2023.
\newblock Llm-eval: Unified multi-dimensional automatic evaluation for
  open-domain conversations with large language models.
\newblock \emph{arXiv preprint arXiv:2305.13711}.

\bibitem[{Lin et~al.(2022{\natexlab{b}})Lin, Akin, Rao, Hie, Zhu, Lu, dos
  Santos~Costa, Fazel-Zarandi, Sercu, Candido et~al.}]{lin2022language}
Z.~Lin, H.~Akin, R.~Rao, B.~Hie, Z.~Zhu, W.~Lu, A.~dos Santos~Costa,
  M.~Fazel-Zarandi et~al. 2022{\natexlab{b}}.
\newblock Language models of protein sequences at the scale of evolution enable
  accurate structure prediction.
\newblock \emph{BioRxiv}.

\bibitem[{Ling et~al.(2017)Ling, Yogatama, Dyer, and
  Blunsom}]{ling-etal-2017-program}
W.~Ling, D.~Yogatama, C.~Dyer and P.~Blunsom. 2017.
\newblock \href {https://doi.org/10.18653/v1/P17-1015} {Program induction by
  rationale generation: Learning to solve and explain algebraic word problems}.
\newblock In \emph{Proceedings of the 55th Annual Meeting of the Association
  for Computational Linguistics (Volume 1: Long Papers)}, pages 158--167,
  Vancouver, Canada. Association for Computational Linguistics.

\bibitem[{Liu et~al.(2023{\natexlab{a}})Liu, Ash, Goel, Krishnamurthy, and
  Zhang}]{liuExposingAttentionGlitches2023}
B.~Liu, J.~T. Ash, S.~Goel, A.~Krishnamurthy and C.~Zhang. 2023{\natexlab{a}}.
\newblock \href {http://arxiv.org/abs/2306.00946} {Exposing {Attention}
  {Glitches} with {Flip}-{Flop} {Language} {Modeling}}.
\newblock ArXiv:2306.00946 [cs].

\bibitem[{Liu et~al.(2022{\natexlab{a}})Liu, Eisenschlos, Piccinno, Krichene,
  Pang, Lee, Joshi, Chen, Collier, and Altun}]{liu2022deplot}
F.~Liu, J.~M. Eisenschlos, F.~Piccinno, S.~Krichene, C.~Pang, K.~Lee, M.~Joshi,
  W.~Chen et~al. 2022{\natexlab{a}}.
\newblock Deplot: One-shot visual language reasoning by plot-to-table
  translation.
\newblock \emph{arXiv preprint arXiv:2212.10505}.

\bibitem[{Liu et~al.(2023{\natexlab{b}})Liu, Sferrazza, and
  Abbeel}]{liu2023languages}
H.~Liu, C.~Sferrazza and P.~Abbeel. 2023{\natexlab{b}}.
\newblock Languages are rewards: Hindsight finetuning using human feedback.
\newblock \emph{arXiv preprint arXiv:2302.02676}.

\bibitem[{Liu et~al.(2022{\natexlab{b}})Liu, Tam, Muqeeth, Mohta, Huang,
  Bansal, and Raffel}]{liu2022few}
H.~Liu, D.~Tam, M.~Muqeeth, J.~Mohta, T.~Huang, M.~Bansal and C.~A. Raffel.
  2022{\natexlab{b}}.
\newblock Few-shot parameter-efficient fine-tuning is better and cheaper than
  in-context learning.
\newblock \emph{Advances in Neural Information Processing Systems},
  35:1950--1965.

\bibitem[{Liu et~al.(2022{\natexlab{c}})Liu, Xie, Li, and Ma}]{Liu2022SamePL}
H.~Liu, S.~M. Xie, Z.~Li and T.~Ma. 2022{\natexlab{c}}.
\newblock Same pre-training loss, better downstream: Implicit bias matters for
  language models.
\newblock \emph{ArXiv}, abs/2210.14199.

\bibitem[{Liu et~al.(2023{\natexlab{c}})Liu, Lin, Hewitt, Paranjape,
  Bevilacqua, Petroni, and Liang}]{liuLostMiddleHow2023}
N.~F. Liu, K.~Lin, J.~Hewitt, A.~Paranjape, M.~Bevilacqua, F.~Petroni and
  P.~Liang. 2023{\natexlab{c}}.
\newblock \href {https://doi.org/10.48550/arXiv.2307.03172} {Lost in the
  {Middle}: {How} {Language} {Models} {Use} {Long} {Contexts}}.
\newblock ArXiv:2307.03172 [cs].

\bibitem[{Liu et~al.(2022{\natexlab{d}})Liu, Jia, Wei, Xu, and
  Vosoughi}]{liu2022quantifying}
R.~Liu, C.~Jia, J.~Wei, G.~Xu and S.~Vosoughi. 2022{\natexlab{d}}.
\newblock Quantifying and alleviating political bias in language models.
\newblock \emph{Artificial Intelligence}, 304:103654.

\bibitem[{Liu and Shah(2023)}]{liuReviewerGPTExploratoryStudy2023}
R.~Liu and N.~B. Shah. 2023.
\newblock \href {http://arxiv.org/abs/2306.00622} {{ReviewerGPT}? {An}
  {Exploratory} {Study} on {Using} {Large} {Language} {Models} for {Paper}
  {Reviewing}}.
\newblock ArXiv:2306.00622 [cs].

\bibitem[{Liu and Wang(2023)}]{liu2023ten}
S.~Liu and Z.~Wang. 2023.
\newblock Ten lessons we have learned in the new" sparseland": A short handbook
  for sparse neural network researchers.
\newblock \emph{arXiv preprint arXiv:2302.02596}.

\bibitem[{Liu et~al.(2022{\natexlab{e}})Liu, Yang, Ouyang, Guo, Su, Xi, Yuan,
  and Yuan}]{liu2022protein}
X.~Liu, X.~Yang, L.~Ouyang, G.~Guo, J.~Su, R.~Xi, K.~Yuan and F.~Yuan.
  2022{\natexlab{e}}.
\newblock Protein language model predicts mutation pathogenicity and clinical
  prognosis.
\newblock \emph{bioRxiv}, pages 2022--09.

\bibitem[{Liu et~al.(2023{\natexlab{d}})Liu, Bahety, and Song}]{liu2023reflect}
Z.~Liu, A.~Bahety and S.~Song. 2023{\natexlab{d}}.
\newblock \href {http://arxiv.org/abs/2306.15724} {Reflect: Summarizing robot
  experiences for failure explanation and correction}.

\bibitem[{Liu et~al.(2023{\natexlab{e}})Liu, Gan, and Tegmark}]{liu2023seeing}
Z.~Liu, E.~Gan and M.~Tegmark. 2023{\natexlab{e}}.
\newblock Seeing is believing: Brain-inspired modular training for mechanistic
  interpretability.
\newblock \emph{arXiv preprint arXiv:2305.08746}.

\bibitem[{Longpre et~al.(2023{\natexlab{a}})Longpre, Hou, Vu, Webson, Chung,
  Tay, Zhou, Le, Zoph, Wei, and Roberts}]{flan_collection}
S.~Longpre, L.~Hou, T.~Vu, A.~Webson, H.~W. Chung, Y.~Tay, D.~Zhou, Q.~V. Le
  et~al. 2023{\natexlab{a}}.
\newblock \href {https://doi.org/10.48550/ARXIV.2301.13688} {The flan
  collection: Designing data and methods for effective instruction tuning}.

\bibitem[{Longpre et~al.(2023{\natexlab{b}})Longpre, Yauney, Reif, Lee,
  Roberts, Zoph, Zhou, Wei, Robinson, Mimno, and
  Ippolito}]{longprepretrainerGuideTraining2023}
S.~Longpre, G.~Yauney, E.~Reif, K.~Lee, A.~Roberts, B.~Zoph, D.~Zhou, J.~Wei
  et~al. 2023{\natexlab{b}}.
\newblock \href {http://arxiv.org/abs/2305.13169} {A {Pretrainer}'s {Guide} to
  {Training} {Data}: {Measuring} the {Effects} of {Data} {Age}, {Domain}
  {Coverage}, {Quality}, \& {Toxicity}}.
\newblock ArXiv:2305.13169 [cs].

\bibitem[{Lu et~al.(2022{\natexlab{a}})Lu, Bartolo, Moore, Riedel, and
  Stenetorp}]{lu-etal-2022-fantastically}
Y.~Lu, M.~Bartolo, A.~Moore, S.~Riedel and P.~Stenetorp. 2022{\natexlab{a}}.
\newblock \href {https://doi.org/10.18653/v1/2022.acl-long.556} {Fantastically
  ordered prompts and where to find them: Overcoming few-shot prompt order
  sensitivity}.
\newblock In \emph{Proceedings of the 60th Annual Meeting of the Association
  for Computational Linguistics (Volume 1: Long Papers)}, pages 8086--8098,
  Dublin, Ireland. Association for Computational Linguistics.

\bibitem[{Lu et~al.(2022{\natexlab{b}})Lu, Li, Zhang, De~Sa, and
  He}]{lu2022maximizing}
Y.~Lu, C.~Li, M.~Zhang, C.~De~Sa and Y.~He. 2022{\natexlab{b}}.
\newblock Maximizing communication efficiency for large-scale training via 0/1
  adam.
\newblock \emph{arXiv preprint arXiv:2202.06009}.

\bibitem[{Lukas et~al.(2023)Lukas, Salem, Sim, Tople, Wutschitz, and
  Zanella-Béguelin}]{lukasAnalyzingLeakagePersonally2023}
N.~Lukas, A.~Salem, R.~Sim, S.~Tople, L.~Wutschitz and S.~Zanella-Béguelin.
  2023.
\newblock \href {http://arxiv.org/abs/2302.00539} {Analyzing {Leakage} of
  {Personally} {Identifiable} {Information} in {Language} {Models}}.
\newblock ArXiv:2302.00539 [cs].

\bibitem[{Luo et~al.(2022)Luo, Lau, Li, and Si}]{luo2022critical}
B.~Luo, R.~Y. Lau, C.~Li and Y.-W. Si. 2022.
\newblock A critical review of state-of-the-art chatbot designs and
  applications.
\newblock \emph{Wiley Interdisciplinary Reviews: Data Mining and Knowledge
  Discovery}, 12(1):e1434.

\bibitem[{Luo et~al.(2021)Luo, Tang, Li, Chai, Li, and
  Qin}]{luo2021synthesizing}
Y.~Luo, N.~Tang, G.~Li, C.~Chai, W.~Li and X.~Qin. 2021.
\newblock Synthesizing natural language to visualization (nl2vis) benchmarks
  from nl2sql benchmarks.
\newblock In \emph{Proceedings of the 2021 International Conference on
  Management of Data}, pages 1235--1247.

\bibitem[{Lynch et~al.(2023)Lynch, Dovonon, Kaddour, and
  Silva}]{lynch2023spawrious}
A.~Lynch, G.~J. Dovonon, J.~Kaddour and R.~Silva. 2023.
\newblock Spawrious: A benchmark for fine control of spurious correlation
  biases.
\newblock \emph{arXiv preprint arXiv:2303.05470}.

\bibitem[{Ma et~al.(2023{\natexlab{a}})Ma, Li, Sun, and Wang}]{ma2023oops}
P.~Ma, Z.~Li, A.~Sun and S.~Wang. 2023{\natexlab{a}}.
\newblock "oops, did i just say that?" testing and repairing unethical
  suggestions of large language models with suggest-critique-reflect process.
\newblock \emph{arXiv preprint arXiv:2305.02626}.

\bibitem[{Ma et~al.(2023{\natexlab{b}})Ma, Fang, and Wang}]{ma2023llm}
X.~Ma, G.~Fang and X.~Wang. 2023{\natexlab{b}}.
\newblock Llm-pruner: On the structural pruning of large language models.
\newblock \emph{arXiv preprint arXiv:2305.11627}.

\bibitem[{Ma et~al.(2021)Ma, Kong, Wang, Zhou, May, Ma, and
  Zettlemoyer}]{ma2021luna}
X.~Ma, X.~Kong, S.~Wang, C.~Zhou, J.~May, H.~Ma and L.~Zettlemoyer. 2021.
\newblock Luna: Linear unified nested attention.
\newblock \emph{Advances in Neural Information Processing Systems},
  34:2441--2453.

\bibitem[{Madaan et~al.(2023)Madaan, Tandon, Gupta, Hallinan, Gao, Wiegreffe,
  Alon, Dziri, Prabhumoye, Yang, Welleck, Majumder, Gupta, Yazdanbakhsh, and
  Clark}]{madaan2023selfrefine}
A.~Madaan, N.~Tandon, P.~Gupta, S.~Hallinan, L.~Gao, S.~Wiegreffe, U.~Alon,
  N.~Dziri et~al. 2023.
\newblock \href {http://arxiv.org/abs/2303.17651} {Self-refine: Iterative
  refinement with self-feedback}.

\bibitem[{Madani et~al.(2023)Madani, Krause, Greene, Subramanian, Mohr, Holton,
  Olmos~Jr, Xiong, Sun, Socher et~al.}]{madani2023large}
A.~Madani, B.~Krause, E.~R. Greene, S.~Subramanian, B.~P. Mohr, J.~M. Holton,
  J.~L. Olmos~Jr, C.~Xiong et~al. 2023.
\newblock Large language models generate functional protein sequences across
  diverse families.
\newblock \emph{Nature Biotechnology}, pages 1--8.

\bibitem[{Maddela et~al.(2023)Maddela, Ung, Xu, Madotto, Foran, and
  Boureau}]{maddelaTrainingModelsGenerate2023}
M.~Maddela, M.~Ung, J.~Xu, A.~Madotto, H.~Foran and Y.-L. Boureau. 2023.
\newblock \href {http://arxiv.org/abs/2307.02768} {Training {Models} to
  {Generate}, {Recognize}, and {Reframe} {Unhelpful} {Thoughts}}.
\newblock ArXiv:2307.02768 [cs].

\bibitem[{Mahdavi et~al.(2023)Mahdavi, Liao, and
  Thrampoulidis}]{mahdaviMemorizationCapacityMultiHead2023}
S.~Mahdavi, R.~Liao and C.~Thrampoulidis. 2023.
\newblock \href {http://arxiv.org/abs/2306.02010} {Memorization {Capacity} of
  {Multi}-{Head} {Attention} in {Transformers}}.
\newblock ArXiv:2306.02010 [cs].

\bibitem[{Malladi et~al.(2023)Malladi, Gao, Nichani, Damian, Lee, Chen, and
  Arora}]{malladiFineTuningLanguageModels2023}
S.~Malladi, T.~Gao, E.~Nichani, A.~Damian, J.~D. Lee, D.~Chen and S.~Arora.
  2023.
\newblock \href {http://arxiv.org/abs/2305.17333} {Fine-{Tuning} {Language}
  {Models} with {Just} {Forward} {Passes}}.
\newblock ArXiv:2305.17333 [cs].

\bibitem[{Mangrulkar et~al.(2022)Mangrulkar, Gugger, Debut, Belkada, and
  Paul}]{hfpeft}
S.~Mangrulkar, S.~Gugger, L.~Debut, Y.~Belkada and S.~Paul. 2022.
\newblock Peft: State-of-the-art parameter-efficient fine-tuning methods.
\newblock \url{https://github.com/huggingface/peft}.

\bibitem[{Maniatis and Tarlow(2023)}]{didact}
P.~Maniatis and D.~Tarlow. 2023.
\newblock Large sequence models for software development activities.
\newblock Available from:
  \url{https://ai.googleblog.com/2023/05/large-sequence-models-for-software.html}.
\newblock Accessed: 26/06/2023.

\bibitem[{McCrae and Costa~Jr(1997)}]{mccrae1997personality}
R.~R. McCrae and P.~T. Costa~Jr. 1997.
\newblock Personality trait structure as a human universal.
\newblock \emph{American psychologist}, 52(5):509.

\bibitem[{McKenzie et~al.(2023)McKenzie, Lyzhov, Pieler, Parrish, Mueller,
  Prabhu, McLean, Kirtland, Ross, Liu, Gritsevskiy, Wurgaft, Kauffman, Recchia,
  Liu, Cavanagh, Weiss, Huang, Droid, Tseng, Korbak, Shen, Zhang, Zhou, Kim,
  Bowman, and Perez}]{mckenzieInverseScalingWhen2023}
I.~R. McKenzie, A.~Lyzhov, M.~Pieler, A.~Parrish, A.~Mueller, A.~Prabhu,
  E.~McLean, A.~Kirtland et~al. 2023.
\newblock \href {https://doi.org/10.48550/arXiv.2306.09479} {Inverse {Scaling}:
  {When} {Bigger} {Isn}'t {Better}}.
\newblock ArXiv:2306.09479 [cs].

\bibitem[{Meng et~al.(2022)Meng, Bau, Andonian, and
  Belinkov}]{meng2022locating}
K.~Meng, D.~Bau, A.~J. Andonian and Y.~Belinkov. 2022.
\newblock \href {https://openreview.net/forum?id=-h6WAS6eE4} {Locating and
  editing factual associations in {GPT}}.
\newblock In \emph{Advances in Neural Information Processing Systems}.

\bibitem[{Meng et~al.(2023)Meng, Sharma, Andonian, Belinkov, and
  Bau}]{meng2023massediting}
K.~Meng, A.~S. Sharma, A.~J. Andonian, Y.~Belinkov and D.~Bau. 2023.
\newblock \href {https://openreview.net/forum?id=MkbcAHIYgyS} {Mass-editing
  memory in a transformer}.
\newblock In \emph{The Eleventh International Conference on Learning
  Representations}.

\bibitem[{Menick et~al.(2022)Menick, Trebacz, Mikulik, Aslanides, Song,
  Chadwick, Glaese, Young, Campbell-Gillingham, Irving, and
  McAleese}]{gophercite}
J.~Menick, M.~Trebacz, V.~Mikulik, J.~Aslanides, F.~Song, M.~Chadwick,
  M.~Glaese, S.~Young et~al. 2022.
\newblock \href {https://doi.org/10.48550/ARXIV.2203.11147} {Teaching language
  models to support answers with verified quotes}.

\bibitem[{Mialon et~al.(2023)Mialon, Dess{\`\i}, Lomeli, Nalmpantis, Pasunuru,
  Raileanu, Rozi{\`e}re, Schick, Dwivedi-Yu, Celikyilmaz
  et~al.}]{mialon2023augmented}
G.~Mialon, R.~Dess{\`\i}, M.~Lomeli, C.~Nalmpantis, R.~Pasunuru, R.~Raileanu,
  B.~Rozi{\`e}re, T.~Schick et~al. 2023.
\newblock Augmented language models: a survey.
\newblock \emph{arXiv preprint arXiv:2302.07842}.

\bibitem[{Milgram(1963)}]{milgram1963behavioral}
S.~Milgram. 1963.
\newblock Behavioral study of obedience.
\newblock \emph{The Journal of abnormal and social psychology}, 67(4):371.

\bibitem[{Min et~al.(2023)Min, Krishna, Lyu, Lewis, Yih, Koh, Iyyer,
  Zettlemoyer, and Hajishirzi}]{minFActScoreFinegrainedAtomic2023}
S.~Min, K.~Krishna, X.~Lyu, M.~Lewis, W.-t. Yih, P.~W. Koh, M.~Iyyer,
  L.~Zettlemoyer et~al. 2023.
\newblock \href {http://arxiv.org/abs/2305.14251} {{FActScore}: {Fine}-grained
  {Atomic} {Evaluation} of {Factual} {Precision} in {Long} {Form} {Text}
  {Generation}}.
\newblock ArXiv:2305.14251 [cs].

\bibitem[{Min et~al.(2022)Min, Lyu, Holtzman, Artetxe, Lewis, Hajishirzi, and
  Zettlemoyer}]{min2022rethinking}
S.~Min, X.~Lyu, A.~Holtzman, M.~Artetxe, M.~Lewis, H.~Hajishirzi and
  L.~Zettlemoyer. 2022.
\newblock \href {http://arxiv.org/abs/2202.12837} {Rethinking the role of
  demonstrations: What makes in-context learning work?}

\bibitem[{Miotto et~al.(2022)Miotto, Rossberg, and Kleinberg}]{miotto2022gpt}
M.~Miotto, N.~Rossberg and B.~Kleinberg. 2022.
\newblock Who is gpt-3? an exploration of personality, values and demographics.
\newblock \emph{arXiv preprint arXiv:2209.14338}.

\bibitem[{Mirowski et~al.(2022)Mirowski, Mathewson, Pittman, and
  Evans}]{mirowski2022co}
P.~Mirowski, K.~W. Mathewson, J.~Pittman and R.~Evans. 2022.
\newblock Co-writing screenplays and theatre scripts with language models: An
  evaluation by industry professionals.
\newblock \emph{arXiv preprint arXiv:2209.14958}.

\bibitem[{Mishra et~al.(2021)Mishra, Latorre, Pool, Stosic, Stosic, Venkatesh,
  Yu, and Micikevicius}]{mishra2021accelerating}
A.~Mishra, J.~A. Latorre, J.~Pool, D.~Stosic, D.~Stosic, G.~Venkatesh, C.~Yu
  and P.~Micikevicius. 2021.
\newblock Accelerating sparse deep neural networks.
\newblock \emph{arXiv preprint arXiv:2104.08378}.

\bibitem[{Mishra et~al.(2022)Mishra, Khashabi, Baral, and
  Hajishirzi}]{mishra-etal-2022-cross}
S.~Mishra, D.~Khashabi, C.~Baral and H.~Hajishirzi. 2022.
\newblock \href {https://doi.org/10.18653/v1/2022.acl-long.244} {Cross-task
  generalization via natural language crowdsourcing instructions}.
\newblock In \emph{Proceedings of the 60th Annual Meeting of the Association
  for Computational Linguistics (Volume 1: Long Papers)}, pages 3470--3487,
  Dublin, Ireland. Association for Computational Linguistics.

\bibitem[{Mitchell et~al.(2023)Mitchell, Lee, Khazatsky, Manning, and
  Finn}]{mitchellDetectGPTZeroShotMachineGenerated2023}
E.~Mitchell, Y.~Lee, A.~Khazatsky, C.~D. Manning and C.~Finn. 2023.
\newblock \href {https://doi.org/10.48550/arXiv.2301.11305} {{DetectGPT}:
  {Zero}-{Shot} {Machine}-{Generated} {Text} {Detection} using {Probability}
  {Curvature}}.
\newblock ArXiv:2301.11305 [cs].

\bibitem[{Mitchell et~al.(2022{\natexlab{a}})Mitchell, Lin, Bosselut, Finn, and
  Manning}]{mitchell2022fast}
E.~Mitchell, C.~Lin, A.~Bosselut, C.~Finn and C.~D. Manning.
  2022{\natexlab{a}}.
\newblock \href {https://openreview.net/forum?id=0DcZxeWfOPt} {Fast model
  editing at scale}.
\newblock In \emph{International Conference on Learning Representations}.

\bibitem[{Mitchell et~al.(2022{\natexlab{b}})Mitchell, Lin, Bosselut, Manning,
  and Finn}]{mitchell2022serac}
E.~Mitchell, C.~Lin, A.~Bosselut, C.~D. Manning and C.~Finn.
  2022{\natexlab{b}}.
\newblock \href {https://proceedings.mlr.press/v162/mitchell22a.html}
  {Memory-based model editing at scale}.
\newblock In \emph{Proceedings of the 39th International Conference on Machine
  Learning}, volume 162 of \emph{Proceedings of Machine Learning Research},
  pages 15817--15831. PMLR.

\bibitem[{Moriconi et~al.(2020)Moriconi, Deisenroth, and
  Sesh~Kumar}]{moriconi2020high}
R.~Moriconi, M.~P. Deisenroth and K.~Sesh~Kumar. 2020.
\newblock High-dimensional bayesian optimization using low-dimensional feature
  spaces.
\newblock \emph{Machine Learning}, 109:1925--1943.

\bibitem[{Moussa{\"\i}d et~al.(2013)Moussa{\"\i}d, K{\"a}mmer, Analytis, and
  Neth}]{moussaid2013social}
M.~Moussa{\"\i}d, J.~E. K{\"a}mmer, P.~P. Analytis and H.~Neth. 2013.
\newblock Social influence and the collective dynamics of opinion formation.
\newblock \emph{PloS one}, 8(11):e78433.

\bibitem[{Mozes et~al.(2023)Mozes, Hoffmann, Tomanek, Kouate, Thain, Yuan,
  Bolukbasi, and Dixon}]{mozes2023towards}
M.~Mozes, J.~Hoffmann, K.~Tomanek, M.~Kouate, N.~Thain, A.~Yuan, T.~Bolukbasi
  and L.~Dixon. 2023.
\newblock Towards agile text classifiers for everyone.
\newblock \emph{arXiv preprint arXiv:2302.06541}.

\bibitem[{Muennighoff et~al.(2022)Muennighoff, Wang, Sutawika, Roberts,
  Biderman, Scao, Bari, Shen, Yong, Schoelkopf
  et~al.}]{muennighoff2022crosslingual}
N.~Muennighoff, T.~Wang, L.~Sutawika, A.~Roberts, S.~Biderman, T.~L. Scao,
  M.~S. Bari, S.~Shen et~al. 2022.
\newblock Crosslingual generalization through multitask finetuning.
\newblock \emph{arXiv preprint arXiv:2211.01786}.

\bibitem[{Mukherjee et~al.(2023)Mukherjee, Mitra, Jawahar, Agarwal, Palangi,
  and Awadallah}]{mukherjee2023orca}
S.~Mukherjee, A.~Mitra, G.~Jawahar, S.~Agarwal, H.~Palangi and A.~Awadallah.
  2023.
\newblock Orca: Progressive learning from complex explanation traces of gpt-4.
\newblock \emph{arXiv preprint arXiv:2306.02707}.

\bibitem[{Nakano et~al.(2021)Nakano, Hilton, Balaji, Wu, Ouyang, Kim, Hesse,
  Jain, Kosaraju, Saunders et~al.}]{reiichiro2021webgpt}
R.~Nakano, J.~Hilton, S.~Balaji, J.~Wu, L.~Ouyang, C.~Kim, C.~Hesse, S.~Jain
  et~al. 2021.
\newblock Webgpt: Browser-assisted question-answering with human feedback.
\newblock \emph{arXiv preprint arXiv:2112.09332}.

\bibitem[{Nanda et~al.(2023)Nanda, Chan, Lieberum, Smith, and
  Steinhardt}]{nanda2023progress}
N.~Nanda, L.~Chan, T.~Lieberum, J.~Smith and J.~Steinhardt. 2023.
\newblock \href {https://openreview.net/forum?id=9XFSbDPmdW} {Progress measures
  for grokking via mechanistic interpretability}.
\newblock In \emph{The Eleventh International Conference on Learning
  Representations}.

\bibitem[{Nerella et~al.(2023)Nerella, Bandyopadhyay, Zhang, Contreras, Siegel,
  Bumin, Silva, Sena, Shickel, Bihorac, Khezeli, and
  Rashidi}]{nerella2023transformers}
S.~Nerella, S.~Bandyopadhyay, J.~Zhang, M.~Contreras, S.~Siegel, A.~Bumin,
  B.~Silva, J.~Sena et~al. 2023.
\newblock \href {http://arxiv.org/abs/2307.00067} {Transformers in healthcare:
  A survey}.

\bibitem[{Nguyen et~al.(2023{\natexlab{a}})Nguyen, Karampatziakis, and
  Chen}]{nguyen2023meet}
A.~Nguyen, N.~Karampatziakis and W.~Chen. 2023{\natexlab{a}}.
\newblock Meet in the middle: A new pre-training paradigm.
\newblock \emph{arXiv preprint arXiv:2303.07295}.

\bibitem[{Nguyen et~al.(2023{\natexlab{b}})Nguyen, Poli, Faizi, Thomas,
  Birch-Sykes, Wornow, Patel, Rabideau, Massaroli, Bengio
  et~al.}]{nguyen2023hyenadna}
E.~Nguyen, M.~Poli, M.~Faizi, A.~Thomas, C.~Birch-Sykes, M.~Wornow, A.~Patel,
  C.~Rabideau et~al. 2023{\natexlab{b}}.
\newblock Hyenadna: Long-range genomic sequence modeling at single nucleotide
  resolution.
\newblock \emph{arXiv preprint arXiv:2306.15794}.

\bibitem[{Nichol et~al.(2022)Nichol, Dhariwal, Ramesh, Shyam, Mishkin, McGrew,
  Sutskever, and Chen}]{nichol2022glide}
A.~Nichol, P.~Dhariwal, A.~Ramesh, P.~Shyam, P.~Mishkin, B.~McGrew,
  I.~Sutskever and M.~Chen. 2022.
\newblock \href {http://arxiv.org/abs/2112.10741} {Glide: Towards
  photorealistic image generation and editing with text-guided diffusion
  models}.

\bibitem[{Nie and Wager(2021)}]{nie2021quasi}
X.~Nie and S.~Wager. 2021.
\newblock Quasi-oracle estimation of heterogeneous treatment effects.
\newblock \emph{Biometrika}, 108(2):299--319.

\bibitem[{Nijkamp et~al.(2022)Nijkamp, Pang, Hayashi, Tu, Wang, Zhou, Savarese,
  and Xiong}]{codegen}
E.~Nijkamp, B.~Pang, H.~Hayashi, L.~Tu, H.~Wang, Y.~Zhou, S.~Savarese and
  C.~Xiong. 2022.
\newblock \href {https://doi.org/10.48550/ARXIV.2203.13474} {Codegen: An open
  large language model for code with multi-turn program synthesis}.

\bibitem[{Niu et~al.()Niu, Recht, Re, Wright, and
  St}]{niuHogwildLockFreeApproach}
F.~Niu, B.~Recht, C.~Re, S.~J. Wright and W.~D. St.
\newblock Hogwild!: {A} {Lock}-{Free} {Approach} to {Parallelizing}
  {Stochastic} {Gradient} {Descent}.

\bibitem[{Nori et~al.(2023)Nori, King, McKinney, Carignan, and
  Horvitz}]{nori2023capabilities}
H.~Nori, N.~King, S.~M. McKinney, D.~Carignan and E.~Horvitz. 2023.
\newblock \href {http://arxiv.org/abs/2303.13375} {Capabilities of gpt-4 on
  medical challenge problems}.

\bibitem[{Nottingham et~al.(2023)Nottingham, Ammanabrolu, Suhr, Choi,
  Hajishirzi, Singh, and Fox}]{nottingham2023embodied}
K.~Nottingham, P.~Ammanabrolu, A.~Suhr, Y.~Choi, H.~Hajishirzi, S.~Singh and
  R.~Fox. 2023.
\newblock Do embodied agents dream of pixelated sheep?: Embodied decision
  making using language guided world modelling.
\newblock \emph{arXiv preprint arXiv:2301.12050}.

\bibitem[{Nurk et~al.(2022)Nurk, Koren, Rhie, Rautiainen, Bzikadze, Mikheenko,
  Vollger, Altemose, Uralsky, Gershman, Aganezov, Hoyt, Diekhans, Logsdon,
  Alonge, Antonarakis, Borchers, Bouffard, Brooks, Caldas, Chen, Cheng, Chin,
  Chow, de~Lima, Dishuck, Durbin, Dvorkina, Fiddes, Formenti, Fulton,
  Fungtammasan, Garrison, Grady, Graves-Lindsay, Hall, Hansen, Hartley,
  Haukness, Howe, Hunkapiller, Jain, Jain, Jarvis, Kerpedjiev, Kirsche,
  Kolmogorov, Korlach, Kremitzki, Li, Maduro, Marschall, McCartney, McDaniel,
  Miller, Mullikin, Myers, Olson, Paten, Peluso, Pevzner, Porubsky, Potapova,
  Rogaev, Rosenfeld, Salzberg, Schneider, Sedlazeck, Shafin, Shew, Shumate,
  Sims, Smit, Soto, Sović, Storer, Streets, Sullivan, Thibaud-Nissen,
  Torrance, Wagner, Walenz, Wenger, Wood, Xiao, Yan, Young, Zarate, Surti,
  McCoy, Dennis, Alexandrov, Gerton, O’Neill, Timp, Zook, Schatz, Eichler,
  Miga, and Phillippy}]{nurk2022genome}
S.~Nurk, S.~Koren, A.~Rhie, M.~Rautiainen, A.~V. Bzikadze, A.~Mikheenko, M.~R.
  Vollger, N.~Altemose et~al. 2022.
\newblock \href {https://doi.org/10.1126/science.abj6987} {The complete
  sequence of a human genome}.
\newblock \emph{Science}, 376(6588):44--53.

\bibitem[{Nye et~al.(2021)Nye, Andreassen, Gur-Ari, Michalewski, Austin,
  Bieber, Dohan, Lewkowycz, Bosma, Luan, Sutton, and Odena}]{nye2021scratchpad}
M.~Nye, A.~J. Andreassen, G.~Gur-Ari, H.~Michalewski, J.~Austin, D.~Bieber,
  D.~Dohan, A.~Lewkowycz et~al. 2021.
\newblock \href {https://doi.org/10.48550/ARXIV.2112.00114} {Show your work:
  Scratchpads for intermediate computation with language models}.

\bibitem[{{Ofir Press
  [@OfirPress]}(2022)}]{ofirpress[@ofirpress]GPT3SeemsBe2022}
{Ofir Press [@OfirPress]}. 2022.
\newblock \href {https://twitter.com/OfirPress/status/1542610741668093952}
  {{GPT}-3 seems to be nondeterministic even when it should be (i.e.
  temperature == 0). {Has} anyone else noticed this? {Is} there a known fix?
  {Video} by my collaborator {Muru} {Zhang}. https://t.co/{dOWYWPBYyP}}.

\bibitem[{Oh et~al.(2023)Oh, Choi, and Lee}]{chatgptoperating}
N.~Oh, G.-S. Choi and W.~Y. Lee. 2023.
\newblock \href {https://doi.org/10.1101/2023.03.16.23287340} {Chatgpt goes to
  operating room: Evaluating gpt-4 performance and its potential in surgical
  education and training in the era of large language models}.
\newblock \emph{medRxiv}.

\bibitem[{Olah()}]{MechanisticInterpretabilityVariables}
C.~Olah.
\newblock \href
  {https://transformer-circuits.pub/2022/mech-interp-essay/index.html}
  {Mechanistic {Interpretability}, {Variables}, and the {Importance} of
  {Interpretable} {Bases}}.

\bibitem[{Olsson et~al.(2022)Olsson, Elhage, Nanda, Joseph, DasSarma, Henighan,
  Mann, Askell, Bai, Chen et~al.}]{olsson2022context}
C.~Olsson, N.~Elhage, N.~Nanda, N.~Joseph, N.~DasSarma, T.~Henighan, B.~Mann,
  A.~Askell et~al. 2022.
\newblock In-context learning and induction heads.
\newblock \emph{arXiv preprint arXiv:2209.11895}.

\bibitem[{OpenAI(2022)}]{chatgpt}
OpenAI. 2022.
\newblock Chatgpt: Optimizing language models for dialogue.
\newblock \url{https://openai.com/blog/chatgpt/}.
\newblock Accessed: 2023-02-18.

\bibitem[{OpenAI(2023{\natexlab{a}})}]{openaiforumgpt4}
OpenAI. 2023{\natexlab{a}}.
\newblock Chat gpt 4 painfully slow.
\newblock
  \url{https://community.openai.com/t/chat-gpt-4-painfully-slow/117996}.

\bibitem[{OpenAI(2023{\natexlab{b}})}]{openai2023gpt4}
OpenAI. 2023{\natexlab{b}}.
\newblock \href {http://arxiv.org/abs/2303.08774} {Gpt-4 technical report}.

\bibitem[{{Ortiz Su{'a}rez} et~al.(2019){Ortiz Su{'a}rez}, Sagot, and
  Romary}]{oscar}
P.~J. {Ortiz Su{'a}rez}, B.~Sagot and L.~Romary. 2019.
\newblock \href {https://doi.org/10.14618/ids-pub-9021} {Asynchronous pipelines
  for processing huge corpora on medium to low resource infrastructures}.
\newblock In \emph{Proceedings of the Workshop on Challenges in the Management
  of Large Corpora (CMLC-7) 2019. Cardiff, 22nd July 2019}, pages 9 -- 16,
  Mannheim. Leibniz-Institut f{"u}r Deutsche Sprache.

\bibitem[{Ott et~al.(2019)Ott, Edunov, Baevski, Fan, Gross, Ng, Grangier, and
  Auli}]{ott2019fairseq}
M.~Ott, S.~Edunov, A.~Baevski, A.~Fan, S.~Gross, N.~Ng, D.~Grangier and
  M.~Auli. 2019.
\newblock fairseq: A fast, extensible toolkit for sequence modeling.
\newblock \emph{arXiv preprint arXiv:1904.01038}.

\bibitem[{Ousidhoum et~al.(2021)Ousidhoum, Zhao, Fang, Song, and
  Yeung}]{ousidhoum2021probing}
N.~Ousidhoum, X.~Zhao, T.~Fang, Y.~Song and D.-Y. Yeung. 2021.
\newblock Probing toxic content in large pre-trained language models.
\newblock In \emph{Proceedings of the 59th Annual Meeting of the Association
  for Computational Linguistics and the 11th International Joint Conference on
  Natural Language Processing (Volume 1: Long Papers)}, pages 4262--4274.

\bibitem[{Outeiral and Deane(2022)}]{outeiral2022codon}
C.~Outeiral and C.~Deane. 2022.
\newblock Codon language embeddings provide strong signals for protein
  engineering.
\newblock \emph{bioRxiv}, pages 2022--12.

\bibitem[{Ouyang et~al.(2022)Ouyang, Wu, Jiang, Almeida, Wainwright, Mishkin,
  Zhang, Agarwal, Slama, Gray, Schulman, Hilton, Kelton, Miller, Simens,
  Askell, Welinder, Christiano, Leike, and Lowe}]{ouyang2022gptinstruct}
L.~Ouyang, J.~Wu, X.~Jiang, D.~Almeida, C.~Wainwright, P.~Mishkin, C.~Zhang,
  S.~Agarwal et~al. 2022.
\newblock \href {https://openreview.net/forum?id=TG8KACxEON} {Training language
  models to follow instructions with human feedback}.
\newblock In \emph{Advances in Neural Information Processing Systems}.

\bibitem[{Pagliardini et~al.(2023)Pagliardini, Paliotta, Jaggi, and
  Fleuret}]{pagliardini2023faster}
M.~Pagliardini, D.~Paliotta, M.~Jaggi and F.~Fleuret. 2023.
\newblock \href {http://arxiv.org/abs/2306.01160} {Faster causal attention over
  large sequences through sparse flash attention}.

\bibitem[{Pan et~al.(2023)Pan, Gao, Chen, and Chen}]{pan2023incontext}
J.~Pan, T.~Gao, H.~Chen and D.~Chen. 2023.
\newblock \href {http://arxiv.org/abs/2305.09731} {What in-context learning
  "learns" in-context: Disentangling task recognition and task learning}.

\bibitem[{Paranjape et~al.(2023)Paranjape, Lundberg, Singh, Hajishirzi,
  Zettlemoyer, and Ribeiro}]{paranjape2023art}
B.~Paranjape, S.~Lundberg, S.~Singh, H.~Hajishirzi, L.~Zettlemoyer and M.~T.
  Ribeiro. 2023.
\newblock \href {http://arxiv.org/abs/2303.09014} {Art: Automatic multi-step
  reasoning and tool-use for large language models}.

\bibitem[{Park et~al.(2022)Park, Park, Kwon, Kim, Lee, and Lee}]{park2022nuqmm}
G.~Park, B.~Park, S.~J. Kwon, B.~Kim, Y.~Lee and D.~Lee. 2022.
\newblock nuqmm: Quantized matmul for efficient inference of large-scale
  generative language models.
\newblock \emph{arXiv preprint arXiv:2206.09557}.

\bibitem[{Park et~al.(2023{\natexlab{a}})Park, O'Brien, Cai, Morris, Liang, and
  Bernstein}]{park2023generative}
J.~S. Park, J.~C. O'Brien, C.~J. Cai, M.~R. Morris, P.~Liang and M.~S.
  Bernstein. 2023{\natexlab{a}}.
\newblock \href {http://arxiv.org/abs/2304.03442} {Generative agents:
  Interactive simulacra of human behavior}.

\bibitem[{Park et~al.(2023{\natexlab{b}})Park, Schoenegger, and
  Zhu}]{park2023artificial}
P.~S. Park, P.~Schoenegger and C.~Zhu. 2023{\natexlab{b}}.
\newblock Artificial intelligence in psychology research.
\newblock \emph{arXiv preprint arXiv:2302.07267}.

\bibitem[{Patel et~al.(2023)Patel, Li, Rasooli, Constant, Raffel, and
  Callison-Burch}]{patel2023bidirectional}
A.~Patel, B.~Li, M.~S. Rasooli, N.~Constant, C.~Raffel and C.~Callison-Burch.
  2023.
\newblock \href {http://arxiv.org/abs/2209.14500} {Bidirectional language
  models are also few-shot learners}.

\bibitem[{Patson et~al.(2009)Patson, Darowski, Moon, and
  Ferreira}]{patson2009lingering}
N.~D. Patson, E.~S. Darowski, N.~Moon and F.~Ferreira. 2009.
\newblock Lingering misinterpretations in garden-path sentences: evidence from
  a paraphrasing task.
\newblock \emph{Journal of Experimental Psychology: Learning, Memory, and
  Cognition}, 35(1):280.

\bibitem[{Patterson et~al.(2022)Patterson, Gonzalez, Hölzle, Le, Liang,
  Munguia, Rothchild, So, Texier, and Dean}]{9810097}
D.~Patterson, J.~Gonzalez, U.~Hölzle, Q.~Le, C.~Liang, L.-M. Munguia,
  D.~Rothchild, D.~R. So et~al. 2022.
\newblock \href {https://doi.org/10.1109/MC.2022.3148714} {The carbon footprint
  of machine learning training will plateau, then shrink}.
\newblock \emph{Computer}, 55(7):18--28.

\bibitem[{Paullada et~al.(2021)Paullada, Raji, Bender, Denton, and
  Hanna}]{paullada2021data}
A.~Paullada, I.~D. Raji, E.~M. Bender, E.~Denton and A.~Hanna. 2021.
\newblock Data and its (dis) contents: A survey of dataset development and use
  in machine learning research.
\newblock \emph{Patterns}, 2(11):100336.

\bibitem[{Pellert et~al.(2023)Pellert, Lechner, Wagner, Rammstedt, and
  Strohmaier}]{pellert2023ai}
M.~Pellert, C.~M. Lechner, C.~Wagner, B.~Rammstedt and M.~Strohmaier. 2023.
\newblock Ai psychometrics: Using psychometric inventories to obtain
  psychological profiles of large language models.

\bibitem[{Penedo et~al.(2023)Penedo, Malartic, Hesslow, Cojocaru, Cappelli,
  Alobeidli, Pannier, Almazrouei, and
  Launay}]{penedoRefinedWebDatasetFalcon2023}
G.~Penedo, Q.~Malartic, D.~Hesslow, R.~Cojocaru, A.~Cappelli, H.~Alobeidli,
  B.~Pannier, E.~Almazrouei et~al. 2023.
\newblock \href {http://arxiv.org/abs/2306.01116} {The {RefinedWeb} {Dataset}
  for {Falcon} {LLM}: {Outperforming} {Curated} {Corpora} with {Web} {Data},
  and {Web} {Data} {Only}}.
\newblock ArXiv:2306.01116 [cs].

\bibitem[{Peng et~al.(2023{\natexlab{a}})Peng, Alcaide, Anthony, Albalak,
  Arcadinho, Cao, Cheng, Chung, Grella, GV, He, Hou, Kazienko, Kocon, Kong,
  Koptyra, Lau, Mantri, Mom, Saito, Tang, Wang, Wind, Wozniak, Zhang, Zhang,
  Zhao, Zhou, Zhu, and Zhu}]{pengRWKVReinventingRNNs2023}
B.~Peng, E.~Alcaide, Q.~Anthony, A.~Albalak, S.~Arcadinho, H.~Cao, X.~Cheng,
  M.~Chung et~al. 2023{\natexlab{a}}.
\newblock \href {https://doi.org/10.48550/arXiv.2305.13048} {{RWKV}:
  {Reinventing} {RNNs} for the {Transformer} {Era}}.
\newblock ArXiv:2305.13048 [cs].

\bibitem[{Peng et~al.(2023{\natexlab{b}})Peng, Alcaide, Anthony, Albalak,
  Arcadinho, Cao, Cheng, Chung, Grella, GV et~al.}]{peng2023rwkv}
B.~Peng, E.~Alcaide, Q.~Anthony, A.~Albalak, S.~Arcadinho, H.~Cao, X.~Cheng,
  M.~Chung et~al. 2023{\natexlab{b}}.
\newblock Rwkv: Reinventing rnns for the transformer era.
\newblock \emph{arXiv preprint arXiv:2305.13048}.

\bibitem[{Peng et~al.(2023{\natexlab{c}})Peng, Yang, Chen, Smith, PourNejatian,
  Costa, Martin, Flores, Zhang, Magoc, Lipori, Mitchell, Ospina, Ahmed, Hogan,
  Shenkman, Guo, Bian, and Wu}]{peng2023study}
C.~Peng, X.~Yang, A.~Chen, K.~E. Smith, N.~PourNejatian, A.~B. Costa,
  C.~Martin, M.~G. Flores et~al. 2023{\natexlab{c}}.
\newblock \href {http://arxiv.org/abs/2305.13523} {A study of generative large
  language model for medical research and healthcare}.

\bibitem[{Peng(2021)}]{peng-2021-marvs}
Y.~Peng. 2021.
\newblock \href {https://aclanthology.org/2021.paclic-1.51} {A {MARVS} analysis
  of two {C}hinese near-synonymous verbs of jumping based on {C}hinese
  corpora}.
\newblock In \emph{Proceedings of the 35th Pacific Asia Conference on Language,
  Information and Computation}, pages 483--492, Shanghai, China. Association
  for Computational Lingustics.

\bibitem[{Perez et~al.(2022{\natexlab{a}})Perez, Huang, Song, Cai, Ring,
  Aslanides, Glaese, McAleese, and Irving}]{perez2022red}
E.~Perez, S.~Huang, F.~Song, T.~Cai, R.~Ring, J.~Aslanides, A.~Glaese,
  N.~McAleese et~al. 2022{\natexlab{a}}.
\newblock Red teaming language models with language models.
\newblock \emph{arXiv preprint arXiv:2202.03286}.

\bibitem[{Perez et~al.(2022{\natexlab{b}})Perez, Ringer, Lukošiūtė, Nguyen,
  Chen, Heiner, Pettit, Olsson, Kundu, Kadavath, Jones, Chen, Mann, Israel,
  Seethor, McKinnon, Olah, Yan, Amodei, Amodei, Drain, Li, Tran-Johnson,
  Khundadze, Kernion, Landis, Kerr, Mueller, Hyun, Landau, Ndousse, Goldberg,
  Lovitt, Lucas, Sellitto, Zhang, Kingsland, Elhage, Joseph, Mercado, DasSarma,
  Rausch, Larson, McCandlish, Johnston, Kravec, Showk, Lanham, Telleen-Lawton,
  Brown, Henighan, Hume, Bai, Hatfield-Dodds, Clark, Bowman, Askell, Grosse,
  Hernandez, Ganguli, Hubinger, Schiefer, and Kaplan}]{perez2022discovering}
E.~Perez, S.~Ringer, K.~Lukošiūtė, K.~Nguyen, E.~Chen, S.~Heiner, C.~Pettit,
  C.~Olsson et~al. 2022{\natexlab{b}}.
\newblock \href {http://arxiv.org/abs/2212.09251} {Discovering language model
  behaviors with model-written evaluations}.

\bibitem[{Perez and Ribeiro(2022)}]{perez2022ignore}
F.~Perez and I.~Ribeiro. 2022.
\newblock Ignore previous prompt: Attack techniques for language models.
\newblock \emph{arXiv preprint arXiv:2211.09527}.

\bibitem[{Peric et~al.(2020)Peric, Mijic, Stammbach, and Ash}]{peric2020legal}
L.~Peric, S.~Mijic, D.~Stammbach and E.~Ash. 2020.
\newblock Legal language modeling with transformers.
\newblock In \emph{Proceedings of the Fourth Workshop on Automated Semantic
  Analysis of Information in Legal Text (ASAIL 2020) held online in conjunction
  with te 33rd International Conference on Legal Knowledge and Information
  Systems (JURIX 2020) December 9, 2020}, volume 2764. CEUR-WS.

\bibitem[{Peters and Martins(2021)}]{peters-martins-2021-smoothing}
B.~Peters and A.~F.~T. Martins. 2021.
\newblock \href {https://doi.org/10.18653/v1/2021.naacl-main.210} {Smoothing
  and shrinking the sparse {S}eq2{S}eq search space}.
\newblock In \emph{Proceedings of the 2021 Conference of the North American
  Chapter of the Association for Computational Linguistics: Human Language
  Technologies}, pages 2642--2654, Online. Association for Computational
  Linguistics.

\bibitem[{Peters et~al.(2017)Peters, Janzing, and
  Sch{\"o}lkopf}]{peters2017elements}
J.~Peters, D.~Janzing and B.~Sch{\"o}lkopf. 2017.
\newblock \emph{Elements of causal inference: foundations and learning
  algorithms}.
\newblock The MIT Press.

\bibitem[{Petrov et~al.(2023)Petrov, La~Malfa, Torr, and
  Bibi}]{petrov2023language}
A.~Petrov, E.~La~Malfa, P.~H. Torr and A.~Bibi. 2023.
\newblock Language model tokenizers introduce unfairness between languages.
\newblock \emph{arXiv preprint arXiv:2305.15425}.

\bibitem[{Pettinato~Oltz(2023)}]{pettinato2023chatgpt}
T.~Pettinato~Oltz. 2023.
\newblock Chatgpt, professor of law.
\newblock \emph{Professor of Law (February 4, 2023)}.

\bibitem[{Pfeiffer et~al.(2020)Pfeiffer, R{\"u}ckl{\'e}, Poth, Kamath,
  Vuli{\'c}, Ruder, Cho, and Gurevych}]{pfeiffer-etal-2020-adapterhub}
J.~Pfeiffer, A.~R{\"u}ckl{\'e}, C.~Poth, A.~Kamath, I.~Vuli{\'c}, S.~Ruder,
  K.~Cho and I.~Gurevych. 2020.
\newblock \href {https://doi.org/10.18653/v1/2020.emnlp-demos.7}
  {{A}dapter{H}ub: A framework for adapting transformers}.
\newblock In \emph{Proceedings of the 2020 Conference on Empirical Methods in
  Natural Language Processing: System Demonstrations}, pages 46--54, Online.
  Association for Computational Linguistics.

\bibitem[{Pichai(2023)}]{bard}
S.~Pichai. 2023.
\newblock An important next step on our ai journey.
\newblock
  \url{https://blog.google/technology/ai/bard-google-ai-search-updates/}.
\newblock Accessed: 2023-02-18.

\bibitem[{Poli et~al.(2023)Poli, Massaroli, Nguyen, Fu, Dao, Baccus, Bengio,
  Ermon, and Ré}]{poliHyenaHierarchyLarger2023a}
M.~Poli, S.~Massaroli, E.~Nguyen, D.~Y. Fu, T.~Dao, S.~Baccus, Y.~Bengio,
  S.~Ermon et~al. 2023.
\newblock \href {https://doi.org/10.48550/arXiv.2302.10866} {Hyena {Hierarchy}:
  {Towards} {Larger} {Convolutional} {Language} {Models}}.
\newblock ArXiv:2302.10866 [cs].

\bibitem[{Pope et~al.(2022{\natexlab{a}})Pope, Douglas, Chowdhery, Devlin,
  Bradbury, Levskaya, Heek, Xiao, Agrawal, and
  Dean}]{popeEfficientlyScalingTransformer2022}
R.~Pope, S.~Douglas, A.~Chowdhery, J.~Devlin, J.~Bradbury, A.~Levskaya,
  J.~Heek, K.~Xiao et~al. 2022{\natexlab{a}}.
\newblock \href {http://arxiv.org/abs/2211.05102} {Efficiently {Scaling}
  {Transformer} {Inference}}.
\newblock ArXiv:2211.05102 [cs].

\bibitem[{Pope et~al.(2022{\natexlab{b}})Pope, Douglas, Chowdhery, Devlin,
  Bradbury, Levskaya, Heek, Xiao, Agrawal, and Dean}]{pope2022efficiently}
R.~Pope, S.~Douglas, A.~Chowdhery, J.~Devlin, J.~Bradbury, A.~Levskaya,
  J.~Heek, K.~Xiao et~al. 2022{\natexlab{b}}.
\newblock Efficiently scaling transformer inference.
\newblock \emph{arXiv preprint arXiv:2211.05102}.

\bibitem[{Prabhakaran et~al.(2021)Prabhakaran, Mostafazadeh~Davani, and
  Diaz}]{prabhakaran-etal-2021-releasing}
V.~Prabhakaran, A.~Mostafazadeh~Davani and M.~Diaz. 2021.
\newblock \href {https://doi.org/10.18653/v1/2021.law-1.14} {On releasing
  annotator-level labels and information in datasets}.
\newblock In \emph{Proceedings of the Joint 15th Linguistic Annotation Workshop
  (LAW) and 3rd Designing Meaning Representations (DMR) Workshop}, pages
  133--138, Punta Cana, Dominican Republic. Association for Computational
  Linguistics.

\bibitem[{Press et~al.(2021)Press, Smith, and Lewis}]{press2021alibi}
O.~Press, N.~A. Smith and M.~Lewis. 2021.
\newblock \href {https://doi.org/10.48550/ARXIV.2108.12409} {Train short, test
  long: Attention with linear biases enables input length extrapolation}.

\bibitem[{Press et~al.(2023)Press, Zhang, Min, Schmidt, Smith, and
  Lewis}]{pressMeasuringNarrowingCompositionality2023}
O.~Press, M.~Zhang, S.~Min, L.~Schmidt, N.~A. Smith and M.~Lewis. 2023.
\newblock \href {http://arxiv.org/abs/2210.03350} {Measuring and {Narrowing}
  the {Compositionality} {Gap} in {Language} {Models}}.
\newblock ArXiv:2210.03350 [cs].

\bibitem[{Qian et~al.(2022)Qian, Wang, Li, Li, and Yan}]{qian2022limitations}
J.~Qian, H.~Wang, Z.~Li, S.~Li and X.~Yan. 2022.
\newblock Limitations of language models in arithmetic and symbolic induction.
\newblock \emph{arXiv preprint arXiv:2208.05051}.

\bibitem[{Rabelo et~al.(2022)Rabelo, Goebel, Kim, Kano, Yoshioka, and
  Satoh}]{Rabelo2022-xe}
J.~Rabelo, R.~Goebel, M.-Y. Kim, Y.~Kano, M.~Yoshioka and K.~Satoh. 2022.
\newblock Overview and discussion of the competition on legal information
  {Extraction/Entailment} ({COLIEE}) 2021.
\newblock \emph{The Review of Socionetwork Strategies}, 16(1):111--133.

\bibitem[{Radford et~al.(2017)Radford, Jozefowicz, and
  Sutskever}]{radford2017learning}
A.~Radford, R.~Jozefowicz and I.~Sutskever. 2017.
\newblock Learning to generate reviews and discovering sentiment.
\newblock \emph{arXiv preprint arXiv:1704.01444}.

\bibitem[{Radford et~al.(2022)Radford, Kim, Xu, Brockman, McLeavey, and
  Sutskever}]{radfordRobustSpeechRecognition2022}
A.~Radford, J.~W. Kim, T.~Xu, G.~Brockman, C.~McLeavey and I.~Sutskever. 2022.
\newblock \href {http://arxiv.org/abs/2212.04356} {Robust {Speech}
  {Recognition} via {Large}-{Scale} {Weak} {Supervision}}.
\newblock ArXiv:2212.04356 [cs, eess].

\bibitem[{Radford et~al.(2019)Radford, Wu, Child, Luan, Amodei, and
  Sutskever}]{gpt2}
A.~Radford, J.~Wu, R.~Child, D.~Luan, D.~Amodei and I.~Sutskever. 2019.
\newblock Language models are unsupervised multitask learners.

\bibitem[{Rae et~al.(2021)Rae, Borgeaud, Cai, Millican, Hoffmann, Song,
  Aslanides, Henderson, Ring, Young et~al.}]{rae2021gopher}
J.~W. Rae, S.~Borgeaud, T.~Cai, K.~Millican, J.~Hoffmann, F.~Song,
  J.~Aslanides, S.~Henderson et~al. 2021.
\newblock Scaling language models: Methods, analysis \& insights from training
  gopher.
\newblock \emph{arXiv preprint arXiv:2112.11446}.

\bibitem[{Rafailov et~al.(2023)Rafailov, Sharma, Mitchell, Ermon, Manning, and
  Finn}]{rafailov2023direct}
R.~Rafailov, A.~Sharma, E.~Mitchell, S.~Ermon, C.~D. Manning and C.~Finn. 2023.
\newblock Direct preference optimization: Your language model is secretly a
  reward model.
\newblock \emph{arXiv preprint arXiv:2305.18290}.

\bibitem[{Raffel et~al.(2022)Raffel, Shazeer, Roberts, Lee, Narang, Matena,
  Zhou, Li, and Liu}]{raffel2022t5}
C.~Raffel, N.~Shazeer, A.~Roberts, K.~Lee, S.~Narang, M.~Matena, Y.~Zhou, W.~Li
  et~al. 2022.
\newblock Exploring the limits of transfer learning with a unified text-to-text
  transformer.
\newblock \emph{J. Mach. Learn. Res.}, 21(1).

\bibitem[{Rajbhandari et~al.(2022)Rajbhandari, Li, Yao, Zhang, Aminabadi, Awan,
  Rasley, and He}]{rajbhandari2022moe}
S.~Rajbhandari, C.~Li, Z.~Yao, M.~Zhang, R.~Y. Aminabadi, A.~A. Awan, J.~Rasley
  and Y.~He. 2022.
\newblock \href {https://proceedings.mlr.press/v162/rajbhandari22a.html}
  {{D}eep{S}peed-{M}o{E}: Advancing mixture-of-experts inference and training
  to power next-generation {AI} scale}.
\newblock In \emph{Proceedings of the 39th International Conference on Machine
  Learning}, volume 162 of \emph{Proceedings of Machine Learning Research},
  pages 18332--18346. PMLR.

\bibitem[{Rajbhandari et~al.(2020)Rajbhandari, Rasley, Ruwase, and
  He}]{rajbhandari2020zero}
S.~Rajbhandari, J.~Rasley, O.~Ruwase and Y.~He. 2020.
\newblock Zero: Memory optimizations toward training trillion parameter models.
\newblock In \emph{Proceedings of the International Conference for High
  Performance Computing, Networking, Storage and Analysis}, SC '20. IEEE Press.

\bibitem[{Rajbhandari et~al.(2021)Rajbhandari, Ruwase, Rasley, Smith, and
  He}]{rajbhandari2021zeroinf}
S.~Rajbhandari, O.~Ruwase, J.~Rasley, S.~Smith and Y.~He. 2021.
\newblock \href {https://doi.org/10.1145/3458817.3476205} {Zero-infinity:
  Breaking the gpu memory wall for extreme scale deep learning}.
\newblock In \emph{Proceedings of the International Conference for High
  Performance Computing, Networking, Storage and Analysis}, SC '21, New York,
  NY, USA. Association for Computing Machinery.

\bibitem[{Raji et~al.(2021)Raji, Bender, Paullada, Denton, and
  Hanna}]{raji2021aibenchmark}
I.~D. Raji, E.~M. Bender, A.~Paullada, E.~Denton and A.~Hanna. 2021.
\newblock Ai and the everything in the whole wide world benchmark.
\newblock \emph{arXiv preprint arXiv:2111.15366}.

\bibitem[{Rajkomar et~al.(2022)Rajkomar, Loreaux, Liu, Kemp, Li, Chen, Zhang,
  Mohiuddin, and Gottweis}]{rajkomar2022deciphering}
A.~Rajkomar, E.~Loreaux, Y.~Liu, J.~Kemp, B.~Li, M.-J. Chen, Y.~Zhang,
  A.~Mohiuddin et~al. 2022.
\newblock Deciphering clinical abbreviations with a privacy protecting machine
  learning system.
\newblock \emph{Nature Communications}, 13(1):7456.

\bibitem[{Ramamurthy et~al.(2022)Ramamurthy, Ammanabrolu, Brantley, Hessel,
  Sifa, Bauckhage, Hajishirzi, and Choi}]{ramamurthy2022reinforcement}
R.~Ramamurthy, P.~Ammanabrolu, K.~Brantley, J.~Hessel, R.~Sifa, C.~Bauckhage,
  H.~Hajishirzi and Y.~Choi. 2022.
\newblock Is reinforcement learning (not) for natural language processing?:
  Benchmarks, baselines, and building blocks for natural language policy
  optimization.
\newblock \emph{arXiv preprint arXiv:2210.01241}.

\bibitem[{Rasley et~al.(2020)Rasley, Rajbhandari, Ruwase, and
  He}]{rasley2020deepspeed}
J.~Rasley, S.~Rajbhandari, O.~Ruwase and Y.~He. 2020.
\newblock \href {https://doi.org/10.1145/3394486.3406703} {Deepspeed: System
  optimizations enable training deep learning models with over 100 billion
  parameters}.
\newblock In \emph{Proceedings of the 26th ACM SIGKDD International Conference
  on Knowledge Discovery \& Data Mining}, KDD '20, page 3505–3506, New York,
  NY, USA. Association for Computing Machinery.

\bibitem[{Ray(2023)}]{rayChatGPTComprehensiveReview2023}
P.~P. Ray. 2023.
\newblock \href {https://doi.org/10.1016/j.iotcps.2023.04.003} {{ChatGPT}: {A}
  comprehensive review on background, applications, key challenges, bias,
  ethics, limitations and future scope}.
\newblock \emph{Internet of Things and Cyber-Physical Systems}, 3:121--154.

\bibitem[{Razumovskaia et~al.(2022)Razumovskaia, Maynez, Louis, Lapata, and
  Narayan}]{razumovskaia2022little}
E.~Razumovskaia, J.~Maynez, A.~Louis, M.~Lapata and S.~Narayan. 2022.
\newblock Little red riding hood goes around the globe: Crosslingual story
  planning and generation with large language models.
\newblock \emph{arXiv preprint arXiv:2212.10471}.

\bibitem[{Recht et~al.(2011)Recht, Re, Wright, and Niu}]{recht2011hogwild}
B.~Recht, C.~Re, S.~Wright and F.~Niu. 2011.
\newblock Hogwild!: A lock-free approach to parallelizing stochastic gradient
  descent.
\newblock \emph{Advances in neural information processing systems}, 24.

\bibitem[{Ren et~al.(2021)Ren, Rajbhandari, Aminabadi, Ruwase, Yang, Zhang, Li,
  and He}]{ren2021zero}
J.~Ren, S.~Rajbhandari, R.~Y. Aminabadi, O.~Ruwase, S.~Yang, M.~Zhang, D.~Li
  and Y.~He. 2021.
\newblock $\{$ZeRO-Offload$\}$: Democratizing $\{$Billion-Scale$\}$ model
  training.
\newblock In \emph{2021 USENIX Annual Technical Conference (USENIX ATC 21)},
  pages 551--564.

\bibitem[{Ren et~al.(2023)Ren, Zhou, Meng, Huang, Wang, Wang, Li, Zhang,
  Podolskiy, Arshinov, Bout, Piontkovskaya, Wei, Jiang, Su, Liu, and
  Yao}]{ren2023pangusigma}
X.~Ren, P.~Zhou, X.~Meng, X.~Huang, Y.~Wang, W.~Wang, P.~Li, X.~Zhang et~al.
  2023.
\newblock \href {http://arxiv.org/abs/2303.10845} {Pangu-{\\Sigma}: Towards
  trillion parameter language model with sparse heterogeneous computing}.

\bibitem[{{Riley Goodside
  [@goodside]}(2022)}]{rileygoodside[@goodside]EdgecaseGPT3Big2022}
{Riley Goodside [@goodside]}. 2022.
\newblock \href {https://twitter.com/goodside/status/1608525976702525440} {An
  edge-case in {GPT}-3 with big implications: {Inference} is non-deterministic
  (even at temperature=0) when top-2 token probabilities are {\textless}1\%
  different. {So} temperature=0 output is *very close* to deterministic, but
  actually isn't. {Worth} remembering.}

\bibitem[{Robin et~al.(2021)Robin, Haas, Gumienny, Smolinski, Tauriello, and
  Schwede}]{robin2021cameo}
X.~Robin, J.~Haas, R.~Gumienny, A.~Smolinski, G.~Tauriello and T.~Schwede.
  2021.
\newblock \href {https://doi.org/https://doi.org/10.1002/prot.26213}
  {Continuous automated model evaluation (cameo)—perspectives on the future
  of fully automated evaluation of structure prediction methods}.
\newblock \emph{Proteins: Structure, Function, and Bioinformatics},
  89(12):1977--1986.

\bibitem[{Rohrbach et~al.(2018)Rohrbach, Hendricks, Burns, Darrell, and
  Saenko}]{rohrbach2018object}
A.~Rohrbach, L.~A. Hendricks, K.~Burns, T.~Darrell and K.~Saenko. 2018.
\newblock Object hallucination in image captioning.
\newblock \emph{arXiv preprint arXiv:1809.02156}.

\bibitem[{Roller et~al.(2021)Roller, Sukhbaatar, Szlam, and
  Weston}]{roller2021hash}
S.~Roller, S.~Sukhbaatar, A.~Szlam and J.~Weston. 2021.
\newblock \href {http://arxiv.org/abs/2106.04426} {Hash layers for large sparse
  models}.

\bibitem[{Rosa et~al.(2022)Rosa, Bonifacio, Jeronymo, Abonizio, Lotufo, and
  Nogueira}]{rosa2022billions}
G.~M. Rosa, L.~Bonifacio, V.~Jeronymo, H.~Abonizio, R.~Lotufo and R.~Nogueira.
  2022.
\newblock Billions of parameters are worth more than in-domain training data: A
  case study in the legal case entailment task.
\newblock \emph{arXiv preprint arXiv:2205.15172}.

\bibitem[{Ross et~al.(1977)Ross, Greene, and House}]{ross1977false}
L.~Ross, D.~Greene and P.~House. 1977.
\newblock The “false consensus effect”: An egocentric bias in social
  perception and attribution processes.
\newblock \emph{Journal of experimental social psychology}, 13(3):279--301.

\bibitem[{Rottenstreich and Hsee(2001)}]{rottenstreich2001money}
Y.~Rottenstreich and C.~K. Hsee. 2001.
\newblock Money, kisses, and electric shocks: On the affective psychology of
  risk.
\newblock \emph{Psychological science}, 12(3):185--190.

\bibitem[{Roush()}]{roushallenYouProbablyDon}
A.~Roush.
\newblock \href
  {https://gist.github.com/Hellisotherpeople/45c619ee22aac6865ca4bb328eb58faf}
  {You probably don't know how to do {Prompt} {Engineering}, let me educate
  you.}

\bibitem[{Ruis et~al.(2022)Ruis, Khan, Biderman, Hooker, Rocktäschel, and
  Grefenstette}]{ruis2022large}
L.~Ruis, A.~Khan, S.~Biderman, S.~Hooker, T.~Rocktäschel and E.~Grefenstette.
  2022.
\newblock \href {http://arxiv.org/abs/2210.14986} {Large language models are
  not zero-shot communicators}.

\bibitem[{Rumbelow and mwatkins()}]{rumbelowSolidGoldMagikarpPromptGeneration}
J.~Rumbelow and mwatkins.
\newblock \href
  {https://www.alignmentforum.org/posts/aPeJE8bSo6rAFoLqg/solidgoldmagikarp-plus-prompt-generation}
  {{SolidGoldMagikarp} (plus, prompt generation)}.

\bibitem[{Russell(2021)}]{russell2021human}
S.~Russell. 2021.
\newblock Human-compatible artificial intelligence.
\newblock \emph{Human-like machine intelligence}, pages 3--23.

\bibitem[{Rust et~al.(2023)Rust, Lotz, Bugliarello, Salesky, de~Lhoneux, and
  Elliott}]{rustLanguageModellingPixels2023}
P.~Rust, J.~F. Lotz, E.~Bugliarello, E.~Salesky, M.~de~Lhoneux and D.~Elliott.
  2023.
\newblock \href {http://arxiv.org/abs/2207.06991} {Language {Modelling} with
  {Pixels}}.
\newblock ArXiv:2207.06991 [cs].

\bibitem[{Sabne(2020)}]{50530}
A.~Sabne. 2020.
\newblock Xla : Compiling machine learning for peak performance.

\bibitem[{Sadasivan et~al.(2023)Sadasivan, Kumar, Balasubramanian, Wang, and
  Feizi}]{sadasivanCanAIGeneratedText2023}
V.~S. Sadasivan, A.~Kumar, S.~Balasubramanian, W.~Wang and S.~Feizi. 2023.
\newblock \href {http://arxiv.org/abs/2303.11156} {Can {AI}-{Generated} {Text}
  be {Reliably} {Detected}?}
\newblock ArXiv:2303.11156 [cs].

\bibitem[{Safdari et~al.(2023)Safdari, Serapio-García, Crepy, Fitz, Romero,
  Sun, Abdulhai, Faust, and Matarić}]{safdari2023personality}
M.~Safdari, G.~Serapio-García, C.~Crepy, S.~Fitz, P.~Romero, L.~Sun,
  M.~Abdulhai, A.~Faust et~al. 2023.
\newblock \href {http://arxiv.org/abs/2307.00184} {Personality traits in large
  language models}.

\bibitem[{Sagawa et~al.(2020)Sagawa, Koh, Hashimoto, and
  Liang}]{sagawa2020distributionally}
S.~Sagawa, P.~W. Koh, T.~B. Hashimoto and P.~Liang. 2020.
\newblock \href {http://arxiv.org/abs/1911.08731} {Distributionally robust
  neural networks for group shifts: On the importance of regularization for
  worst-case generalization}.

\bibitem[{Sainz et~al.()Sainz, Campos, García-Ferrero, Etxaniz, and
  Agirre}]{Lmcontamination}
O.~Sainz, J.~C. Campos, I.~García-Ferrero, J.~Etxaniz and E.~Agirre.
\newblock \href {https://hitz-zentroa.github.io/lm-contamination/blog/}
  {lm-contamination}.

\bibitem[{Salewski et~al.(2023)Salewski, Alaniz, Rio-Torto, Schulz, and
  Akata}]{salewski2023context}
L.~Salewski, S.~Alaniz, I.~Rio-Torto, E.~Schulz and Z.~Akata. 2023.
\newblock In-context impersonation reveals large language models' strengths and
  biases.
\newblock \emph{arXiv preprint arXiv:2305.14930}.

\bibitem[{Sanchez et~al.(2023)Sanchez, Fan, Spangher, Levi, Ammanamanchi, and
  Biderman}]{sanchezStayTopicClassifierFree2023}
G.~Sanchez, H.~Fan, A.~Spangher, E.~Levi, P.~S. Ammanamanchi and S.~Biderman.
  2023.
\newblock \href {http://arxiv.org/abs/2306.17806} {Stay on topic with
  {Classifier}-{Free} {Guidance}}.
\newblock ArXiv:2306.17806 [cs].

\bibitem[{Sanh et~al.(2022)Sanh, Webson, Raffel, Bach, Sutawika, Alyafeai,
  Chaffin, Stiegler, Raja, Dey, Bari, Xu, Thakker, Sharma, Szczechla, Kim,
  Chhablani, Nayak, Datta, Chang, Jiang, Wang, Manica, Shen, Yong, Pandey,
  Bawden, Wang, Neeraj, Rozen, Sharma, Santilli, Fevry, Fries, Teehan, Scao,
  Biderman, Gao, Wolf, and Rush}]{t0}
V.~Sanh, A.~Webson, C.~Raffel, S.~Bach, L.~Sutawika, Z.~Alyafeai, A.~Chaffin,
  A.~Stiegler et~al. 2022.
\newblock \href {https://openreview.net/forum?id=9Vrb9D0WI4} {Multitask
  prompted training enables zero-shot task generalization}.
\newblock In \emph{International Conference on Learning Representations}.

\bibitem[{Sanyal et~al.(2023)Sanyal, Kaddour, Kumar, and
  Sanghavi}]{sanyal2023understanding}
S.~Sanyal, J.~Kaddour, A.~Kumar and S.~Sanghavi. 2023.
\newblock \href {http://arxiv.org/abs/2306.03241} {Understanding the
  effectiveness of early weight averaging for training large language models}.

\bibitem[{Saravia(2022)}]{saraviaPromptEngineeringGuide2022}
E.~Saravia. 2022.
\newblock \href {https://github.com/dair-ai/Prompt-Engineering-Guide} {Prompt
  {Engineering} {Guide}}.
\newblock Publication Title:
  https://github.com/dair-ai/Prompt-Engineering-Guide original-date:
  2022-12-16T16:04:50Z.

\bibitem[{Savelka et~al.(2023)Savelka, Ashley, Gray, Westermann, and
  Xu}]{savelka2023explaining}
J.~Savelka, K.~D. Ashley, M.~A. Gray, H.~Westermann and H.~Xu. 2023.
\newblock \href {http://arxiv.org/abs/2306.09525} {Explaining legal concepts
  with augmented large language models (gpt-4)}.

\bibitem[{Scao et~al.(2022)Scao, Fan, Akiki, Pavlick, Ilić, Hesslow,
  Castagné, Luccioni, Yvon, Gallé, Tow, Rush, Biderman, Webson, Ammanamanchi,
  Wang, Sagot, Muennighoff, del Moral, Ruwase, Bawden, Bekman, McMillan-Major,
  Beltagy, Nguyen, Saulnier, Tan, Suarez, Sanh, Laurençon, Jernite, Launay,
  Mitchell, Raffel, Gokaslan, Simhi, Soroa, Aji, Alfassy, Rogers, Nitzav, Xu,
  Mou, Emezue, Klamm, Leong, van Strien, Adelani, Radev, Ponferrada, Levkovizh,
  Kim, Natan, De~Toni, Dupont, Kruszewski, Pistilli, Elsahar, Benyamina, Tran,
  Yu, Abdulmumin, Johnson, Gonzalez-Dios, de~la Rosa, Chim, Dodge, Zhu, Chang,
  Frohberg, Tobing, Bhattacharjee, Almubarak, Chen, Lo, Von~Werra, Weber, Phan,
  allal, Tanguy, Dey, Muñoz, Masoud, Grandury, Šaško, Huang, Coavoux, Singh,
  Jiang, Vu, Jauhar, Ghaleb, Subramani, Kassner, Khamis, Nguyen, Espejel,
  de~Gibert, Villegas, Henderson, Colombo, Amuok, Lhoest, Harliman, Bommasani,
  López, Ribeiro, Osei, Pyysalo, Nagel, Bose, Muhammad, Sharma, Longpre,
  Nikpoor, Silberberg, Pai, Zink, Torrent, Schick, Thrush, Danchev, Nikoulina,
  Laippala, Lepercq, Prabhu, Alyafeai, Talat, Raja, Heinzerling, Si, Salesky,
  Mielke, Lee, Sharma, Santilli, Chaffin, Stiegler, Datta, Szczechla,
  Chhablani, Wang, Pandey, Strobelt, Fries, Rozen, Gao, Sutawika, Bari,
  Al-shaibani, Manica, Nayak, Teehan, Albanie, Shen, Ben-David, Bach, Kim,
  Bers, Fevry, Neeraj, Thakker, Raunak, Tang, Yong, Sun, Brody, Uri, Tojarieh,
  Roberts, Chung, Tae, Phang, Press, Li, Narayanan, Bourfoune, Casper, Rasley,
  Ryabinin, Mishra, Zhang, Shoeybi, Peyrounette, Patry, Tazi, Sanseviero, von
  Platen, Cornette, Lavallée, Lacroix, Rajbhandari, Gandhi, Smith, Requena,
  Patil, Dettmers, Baruwa, Singh, Cheveleva, Ligozat, Subramonian, Névéol,
  Lovering, Garrette, Tunuguntla, Reiter, Taktasheva, Voloshina, Bogdanov,
  Winata, Schoelkopf, Kalo, Novikova, Forde, Clive, Kasai, Kawamura, Hazan,
  Carpuat, Clinciu, Kim, Cheng, Serikov, Antverg, van~der Wal, Zhang, Zhang,
  Gehrmann, Pais, Shavrina, Scialom, Yun, Limisiewicz, Rieser, Protasov,
  Mikhailov, Pruksachatkun, Belinkov, Bamberger, Kasner, Rueda, Pestana,
  Feizpour, Khan, Faranak, Santos, Hevia, Unldreaj, Aghagol, Abdollahi,
  Tammour, HajiHosseini, Behroozi, Ajibade, Saxena, Ferrandis, Contractor,
  Lansky, David, Kiela, Nguyen, Tan, Baylor, Ozoani, Mirza, Ononiwu, Rezanejad,
  Jones, Bhattacharya, Solaiman, Sedenko, Nejadgholi, Passmore, Seltzer, Sanz,
  Fort, Dutra, Samagaio, Elbadri, Mieskes, Gerchick, Akinlolu, McKenna, Qiu,
  Ghauri, Burynok, Abrar, Rajani, Elkott, Fahmy, Samuel, An, Kromann, Hao,
  Alizadeh, Shubber, Wang, Roy, Viguier, Le, Oyebade, Le, Yang, Nguyen,
  Kashyap, Palasciano, Callahan, Shukla, Miranda-Escalada, Singh, Beilharz,
  Wang, Brito, Zhou, Jain, Xu, Fourrier, Periñán, Molano, Yu, Manjavacas,
  Barth, Fuhrimann, Altay, Bayrak, Burns, Vrabec, Bello, Dash, Kang, Giorgi,
  Golde, Posada, Sivaraman, Bulchandani, Liu, Shinzato, de~Bykhovetz, Takeuchi,
  Pàmies, Castillo, Nezhurina, Sänger, Samwald, Cullan, Weinberg, De~Wolf,
  Mihaljcic, Liu, Freidank, Kang, Seelam, Dahlberg, Broad, Muellner, Fung,
  Haller, Chandrasekhar, Eisenberg, Martin, Canalli, Su, Su, Cahyawijaya,
  Garda, Deshmukh, Mishra, Kiblawi, Ott, Sang-aroonsiri, Kumar, Schweter,
  Bharati, Laud, Gigant, Kainuma, Kusa, Labrak, Bajaj, Venkatraman, Xu, Xu, Xu,
  Tan, Xie, Ye, Bras, Belkada, and Wolf}]{scao2021bloom}
T.~L. Scao, A.~Fan, C.~Akiki, E.~Pavlick, S.~Ilić, D.~Hesslow, R.~Castagné,
  A.~S. Luccioni et~al. 2022.
\newblock \href {https://doi.org/10.48550/ARXIV.2211.05100} {Bloom: A
  176b-parameter open-access multilingual language model}.

\bibitem[{Schaeffer et~al.(2023)Schaeffer, Miranda, and
  Koyejo}]{schaeffer2023emergent}
R.~Schaeffer, B.~Miranda and S.~Koyejo. 2023.
\newblock \href {http://arxiv.org/abs/2304.15004} {Are emergent abilities of
  large language models a mirage?}

\bibitem[{Schick et~al.(2023)Schick, Dwivedi-Yu, Dessì, Raileanu, Lomeli,
  Zettlemoyer, Cancedda, and Scialom}]{schick2023toolformer}
T.~Schick, J.~Dwivedi-Yu, R.~Dessì, R.~Raileanu, M.~Lomeli, L.~Zettlemoyer,
  N.~Cancedda and T.~Scialom. 2023.
\newblock Toolformer: Language models can teach themselves to use tools.
\newblock \emph{arXiv preprint arXiv:2302.04761}.

\bibitem[{Schick et~al.(2022)Schick, Dwivedi-Yu, Jiang, Petroni, Lewis,
  Izacard, You, Nalmpantis, Grave, and Riedel}]{peer}
T.~Schick, J.~Dwivedi-Yu, Z.~Jiang, F.~Petroni, P.~Lewis, G.~Izacard, Q.~You,
  C.~Nalmpantis et~al. 2022.
\newblock \href {https://doi.org/10.48550/ARXIV.2208.11663} {Peer: A
  collaborative language model}.

\bibitem[{Schick and Sch{\"u}tze(2021)}]{schick2021s}
T.~Schick and H.~Sch{\"u}tze. 2021.
\newblock It’s not just size that matters: Small language models are also
  few-shot learners.
\newblock In \emph{Proceedings of the 2021 Conference of the North American
  Chapter of the Association for Computational Linguistics: Human Language
  Technologies}, pages 2339--2352.

\bibitem[{Schulman et~al.(2017)Schulman, Wolski, Dhariwal, Radford, and
  Klimov}]{schulman2017proximal}
J.~Schulman, F.~Wolski, P.~Dhariwal, A.~Radford and O.~Klimov. 2017.
\newblock Proximal policy optimization algorithms.
\newblock \emph{arXiv preprint arXiv:1707.06347}.

\bibitem[{Schuster and Nakajima(2012)}]{schuster2012wordpiece}
M.~Schuster and K.~Nakajima. 2012.
\newblock \href {https://doi.org/10.1109/ICASSP.2012.6289079} {Japanese and
  korean voice search}.
\newblock In \emph{2012 IEEE International Conference on Acoustics, Speech and
  Signal Processing (ICASSP)}, pages 5149--5152.

\bibitem[{Schuster et~al.(2020)Schuster, Schuster, Shah, and
  Barzilay}]{schuster2020limitations}
T.~Schuster, R.~Schuster, D.~J. Shah and R.~Barzilay. 2020.
\newblock The limitations of stylometry for detecting machine-generated fake
  news.
\newblock \emph{Computational Linguistics}, 46(2):499--510.

\bibitem[{Schwartz et~al.(2019)Schwartz, Dodge, Smith, and
  Etzioni}]{schwartzGreenAI2019}
R.~Schwartz, J.~Dodge, N.~A. Smith and O.~Etzioni. 2019.
\newblock \href {http://arxiv.org/abs/1907.10597} {Green {AI}}.
\newblock ArXiv:1907.10597 [cs, stat].

\bibitem[{Schwartz et~al.(2015)Schwartz, Breyer, and Danner}]{humvalscale}
S.~H. Schwartz, B.~Breyer and D.~Danner. 2015.
\newblock \href {https://doi.org/10.6102/zis234} {Human values scale (ess)}.
\newblock \emph{Zusammenstellung sozialwissenschaftlicher Items und Skalen
  (ZIS)}.

\bibitem[{See et~al.(2019)See, Pappu, Saxena, Yerukola, and
  Manning}]{see-etal-2019-massively}
A.~See, A.~Pappu, R.~Saxena, A.~Yerukola and C.~D. Manning. 2019.
\newblock \href {https://doi.org/10.18653/v1/K19-1079} {Do massively pretrained
  language models make better storytellers?}
\newblock In \emph{Proceedings of the 23rd Conference on Computational Natural
  Language Learning (CoNLL)}, pages 843--861, Hong Kong, China. Association for
  Computational Linguistics.

\bibitem[{Sennrich et~al.(2015)Sennrich, Haddow, and Birch}]{BPE}
R.~Sennrich, B.~Haddow and A.~Birch. 2015.
\newblock Neural machine translation of rare words with subword units.
\newblock \emph{arXiv preprint arXiv:1508.07909}.

\bibitem[{Sezgin et~al.(2022)Sezgin, Sirrianni, Linwood
  et~al.}]{sezgin2022operationalizing}
E.~Sezgin, J.~Sirrianni, S.~L. Linwood et~al. 2022.
\newblock Operationalizing and implementing pretrained, large artificial
  intelligence linguistic models in the us health care system: Outlook of
  generative pretrained transformer 3 (gpt-3) as a service model.
\newblock \emph{JMIR Medical Informatics}, 10(2):e32875.

\bibitem[{Shaw et~al.(2018)Shaw, Uszkoreit, and Vaswani}]{shaw-etal-2018-self}
P.~Shaw, J.~Uszkoreit and A.~Vaswani. 2018.
\newblock \href {https://doi.org/10.18653/v1/N18-2074} {Self-attention with
  relative position representations}.
\newblock In \emph{Proceedings of the 2018 Conference of the North {A}merican
  Chapter of the Association for Computational Linguistics: Human Language
  Technologies, Volume 2 (Short Papers)}, pages 464--468, New Orleans,
  Louisiana. Association for Computational Linguistics.

\bibitem[{Shazeer(2019{\natexlab{a}})}]{fast_transformer_decoding}
N.~Shazeer. 2019{\natexlab{a}}.
\newblock \href {https://doi.org/10.48550/ARXIV.1911.02150} {Fast transformer
  decoding: One write-head is all you need}.

\bibitem[{Shazeer(2019{\natexlab{b}})}]{shazeer2019fast}
N.~Shazeer. 2019{\natexlab{b}}.
\newblock \href {http://arxiv.org/abs/1911.02150} {Fast transformer decoding:
  One write-head is all you need}.

\bibitem[{Shazeer et~al.(2017)Shazeer, Mirhoseini, Maziarz, Davis, Le, Hinton,
  and Dean}]{shazeer2017outrageously}
N.~Shazeer, A.~Mirhoseini, K.~Maziarz, A.~Davis, Q.~Le, G.~Hinton and J.~Dean.
  2017.
\newblock Outrageously large neural networks: The sparsely-gated
  mixture-of-experts layer.
\newblock \emph{arXiv preprint arXiv:1701.06538}.

\bibitem[{Shen et~al.(2021)Shen, Zhang, Zhao, Yi, and Li}]{shen2021efficient}
Z.~Shen, M.~Zhang, H.~Zhao, S.~Yi and H.~Li. 2021.
\newblock Efficient attention: Attention with linear complexities.
\newblock In \emph{Proceedings of the IEEE/CVF winter conference on
  applications of computer vision}, pages 3531--3539.

\bibitem[{Sheng et~al.(2023)Sheng, Zheng, Yuan, Li, Ryabinin, Chen, Liang, Ré,
  Stoica, and Zhang}]{sheng2023flexgen}
Y.~Sheng, L.~Zheng, B.~Yuan, Z.~Li, M.~Ryabinin, B.~Chen, P.~Liang, C.~Ré
  et~al. 2023.
\newblock High-throughput generative inference of large language models with a
  single gpu.

\bibitem[{Shevlane et~al.(2023)Shevlane, Farquhar, Garfinkel, Phuong,
  Whittlestone, Leung, Kokotajlo, Marchal, Anderljung, Kolt
  et~al.}]{shevlane2023model}
T.~Shevlane, S.~Farquhar, B.~Garfinkel, M.~Phuong, J.~Whittlestone, J.~Leung,
  D.~Kokotajlo, N.~Marchal et~al. 2023.
\newblock Model evaluation for extreme risks.
\newblock \emph{arXiv preprint arXiv:2305.15324}.

\bibitem[{Shirafuji et~al.(2023)Shirafuji, Watanobe, Ito, Morishita, Nakamura,
  Oda, and Suzuki}]{shirafuji2023exploring}
A.~Shirafuji, Y.~Watanobe, T.~Ito, M.~Morishita, Y.~Nakamura, Y.~Oda and
  J.~Suzuki. 2023.
\newblock \href {http://arxiv.org/abs/2306.14583} {Exploring the robustness of
  large language models for solving programming problems}.

\bibitem[{Shliazhko et~al.(2022)Shliazhko, Fenogenova, Tikhonova, Mikhailov,
  Kozlova, and Shavrina}]{shliazhko2022mgpt}
O.~Shliazhko, A.~Fenogenova, M.~Tikhonova, V.~Mikhailov, A.~Kozlova and
  T.~Shavrina. 2022.
\newblock mgpt: Few-shot learners go multilingual.
\newblock \emph{arXiv preprint arXiv:2204.07580}.

\bibitem[{Shoeybi et~al.(2019)Shoeybi, Patwary, Puri, LeGresley, Casper, and
  Catanzaro}]{shoeybi2019megatron}
M.~Shoeybi, M.~Patwary, R.~Puri, P.~LeGresley, J.~Casper and B.~Catanzaro.
  2019.
\newblock Megatron-lm: Training multi-billion parameter language models using
  model parallelism.
\newblock \emph{arXiv preprint arXiv:1909.08053}.

\bibitem[{Shridhar et~al.(2022{\natexlab{a}})Shridhar, Macina, El-Assady,
  Sinha, Kapur, and Sachan}]{Shridhar2022AutomaticGO}
K.~Shridhar, J.~Macina, M.~El-Assady, T.~Sinha, M.~Kapur and M.~Sachan.
  2022{\natexlab{a}}.
\newblock Automatic generation of socratic subquestions for teaching math word
  problems.
\newblock \emph{ArXiv}, abs/2211.12835.

\bibitem[{Shridhar et~al.(2022{\natexlab{b}})Shridhar, Stolfo, and
  Sachan}]{shridhar2022distilling}
K.~Shridhar, A.~Stolfo and M.~Sachan. 2022{\natexlab{b}}.
\newblock Distilling multi-step reasoning capabilities of large language models
  into smaller models via semantic decompositions.
\newblock \emph{arXiv preprint arXiv:2212.00193}.

\bibitem[{Shrivastava et~al.(2022)Shrivastava, Larochelle, and
  Tarlow}]{shrivastava2022repository}
D.~Shrivastava, H.~Larochelle and D.~Tarlow. 2022.
\newblock Repository-level prompt generation for large language models of code.
\newblock \emph{arXiv preprint arXiv:2206.12839}.

\bibitem[{Shuai et~al.(2021)Shuai, Ruffolo, and Gray}]{shuai2021generative}
R.~W. Shuai, J.~A. Ruffolo and J.~J. Gray. 2021.
\newblock Generative language modeling for antibody design.
\newblock \emph{bioRxiv}, pages 2021--12.

\bibitem[{Shumailov et~al.(2023)Shumailov, Shumaylov, Zhao, Gal, Papernot, and
  Anderson}]{shumailov2023curse}
I.~Shumailov, Z.~Shumaylov, Y.~Zhao, Y.~Gal, N.~Papernot and R.~Anderson. 2023.
\newblock \href {http://arxiv.org/abs/2305.17493} {The curse of recursion:
  Training on generated data makes models forget}.

\bibitem[{Shuster et~al.(2021)Shuster, Poff, Chen, Kiela, and
  Weston}]{shuster2021retrieval}
K.~Shuster, S.~Poff, M.~Chen, D.~Kiela and J.~Weston. 2021.
\newblock Retrieval augmentation reduces hallucination in conversation.
\newblock \emph{arXiv preprint arXiv:2104.07567}.

\bibitem[{Shuster et~al.(2022)Shuster, Xu, Komeili, Ju, Smith, Roller, Ung,
  Chen, Arora, Lane, Behrooz, Ngan, Poff, Goyal, Szlam, Boureau, Kambadur, and
  Weston}]{blenderbot3}
K.~Shuster, J.~Xu, M.~Komeili, D.~Ju, E.~M. Smith, S.~Roller, M.~Ung, M.~Chen
  et~al. 2022.
\newblock \href {https://doi.org/10.48550/ARXIV.2208.03188} {Blenderbot 3: a
  deployed conversational agent that continually learns to responsibly engage}.

\bibitem[{Sia and Duh(2023)}]{Sia2023IncontextLA}
S.~Sia and K.~Duh. 2023.
\newblock In-context learning as maintaining coherency: A study of on-the-fly
  machine translation using large language models.
\newblock \emph{ArXiv}, abs/2305.03573.

\bibitem[{Singh et~al.(2022)Singh, Blukis, Mousavian, Goyal, Xu, Tremblay, Fox,
  Thomason, and Garg}]{singh2022progprompt}
I.~Singh, V.~Blukis, A.~Mousavian, A.~Goyal, D.~Xu, J.~Tremblay, D.~Fox,
  J.~Thomason et~al. 2022.
\newblock \href {http://arxiv.org/abs/2209.11302} {Progprompt: Generating
  situated robot task plans using large language models}.

\bibitem[{Singhal et~al.(2022)Singhal, Azizi, Tu, Mahdavi, Wei, Chung, Scales,
  Tanwani, Cole-Lewis, Pfohl, Payne, Seneviratne, Gamble, Kelly, Scharli,
  Chowdhery, Mansfield, Arcas, Webster, Corrado, Matias, Chou, Gottweis,
  Tomasev, Liu, Rajkomar, Barral, Semturs, Karthikesalingam, and
  Natarajan}]{medpalm}
K.~Singhal, S.~Azizi, T.~Tu, S.~S. Mahdavi, J.~Wei, H.~W. Chung, N.~Scales,
  A.~Tanwani et~al. 2022.
\newblock \href {https://doi.org/10.48550/ARXIV.2212.13138} {Large language
  models encode clinical knowledge}.

\bibitem[{Singhal et~al.(2023)Singhal, Tu, Gottweis, Sayres, Wulczyn, Hou,
  Clark, Pfohl, Cole-Lewis, Neal et~al.}]{medpalm2}
K.~Singhal, T.~Tu, J.~Gottweis, R.~Sayres, E.~Wulczyn, L.~Hou, K.~Clark,
  S.~Pfohl et~al. 2023.
\newblock Towards expert-level medical question answering with large language
  models.
\newblock \emph{arXiv preprint arXiv:2305.09617}.

\bibitem[{Sinitsin et~al.(2020)Sinitsin, Pyrkin, Babenko, Plokhotnyuk, and
  Popov}]{sinitsinEDITABLENEURALNETWORKS2020}
A.~Sinitsin, D.~Pyrkin, A.~Babenko, V.~Plokhotnyuk and S.~Popov. 2020.
\newblock {EDITABLE} {NEURAL} {NETWORKS}.

\bibitem[{Smith et~al.(2017)Smith, Kindermans, Ying, and Le}]{smith2017don}
S.~L. Smith, P.-J. Kindermans, C.~Ying and Q.~V. Le. 2017.
\newblock Don't decay the learning rate, increase the batch size.
\newblock \emph{arXiv preprint arXiv:1711.00489}.

\bibitem[{Smith et~al.(2022)Smith, Patwary, Norick, LeGresley, Rajbhandari,
  Casper, Liu, Prabhumoye, Zerveas, Korthikanti et~al.}]{smith2022using}
S.~Smith, M.~Patwary, B.~Norick, P.~LeGresley, S.~Rajbhandari, J.~Casper,
  Z.~Liu, S.~Prabhumoye et~al. 2022.
\newblock Using deepspeed and megatron to train megatron-turing nlg 530b, a
  large-scale generative language model.
\newblock \emph{arXiv preprint arXiv:2201.11990}.

\bibitem[{Solaiman and Dennison(2021)}]{solaiman2021process}
I.~Solaiman and C.~Dennison. 2021.
\newblock Process for adapting language models to society (palms) with
  values-targeted datasets.
\newblock \emph{Advances in Neural Information Processing Systems},
  34:5861--5873.

\bibitem[{Soltan et~al.(2022)Soltan, Ananthakrishnan, FitzGerald, Gupta, Hamza,
  Khan, Peris, Rawls, Rosenbaum, Rumshisky et~al.}]{soltan2022alexatm}
S.~Soltan, S.~Ananthakrishnan, J.~FitzGerald, R.~Gupta, W.~Hamza, H.~Khan,
  C.~Peris, S.~Rawls et~al. 2022.
\newblock Alexatm 20b: Few-shot learning using a large-scale multilingual
  seq2seq model.
\newblock \emph{arXiv preprint arXiv:2208.01448}.

\bibitem[{Sorscher et~al.(2022)Sorscher, Geirhos, Shekhar, Ganguli, and
  Morcos}]{sorscher2022beyond}
B.~Sorscher, R.~Geirhos, S.~Shekhar, S.~Ganguli and A.~S. Morcos. 2022.
\newblock Beyond neural scaling laws: beating power law scaling via data
  pruning.
\newblock \emph{arXiv preprint arXiv:2206.14486}.

\bibitem[{Srivastava et~al.(2022)Srivastava, Rastogi, Rao, Shoeb, Abid, Fisch,
  Brown, Santoro, Gupta, Garriga-Alonso et~al.}]{srivastava2022beyond}
A.~Srivastava, A.~Rastogi, A.~Rao, A.~A.~M. Shoeb, A.~Abid, A.~Fisch, A.~R.
  Brown, A.~Santoro et~al. 2022.
\newblock Beyond the imitation game: Quantifying and extrapolating the
  capabilities of language models.
\newblock \emph{arXiv preprint arXiv:2206.04615}.

\bibitem[{Steinhardt(2022)}]{steinhardt2022future}
J.~Steinhardt. 2022.
\newblock Future ml systems will be qualitatively different.
\newblock \emph{Accessed May}, 20:2022.

\bibitem[{Steinhardt(2023)}]{emergent_deception}
J.~Steinhardt. 2023.
\newblock Emergent deception and emergent optimization.
\newblock Available from:
  \url{https://bounded-regret.ghost.io/emergent-deception-optimization/}.
\newblock Accessed: 29/04/2023.

\bibitem[{Stern et~al.(2018)Stern, Shazeer, and Uszkoreit}]{stern2018blockwise}
M.~Stern, N.~Shazeer and J.~Uszkoreit. 2018.
\newblock Blockwise parallel decoding for deep autoregressive models.
\newblock In \emph{Proceedings of the 32nd International Conference on Neural
  Information Processing Systems}, NIPS'18, page 10107–10116, Red Hook, NY,
  USA. Curran Associates Inc.

\bibitem[{Stevenson et~al.(2022)Stevenson, Smal, Baas, Grasman, and van~der
  Maas}]{stevenson2022putting}
C.~Stevenson, I.~Smal, M.~Baas, R.~Grasman and H.~van~der Maas. 2022.
\newblock Putting gpt-3's creativity to the (alternative uses) test.
\newblock \emph{arXiv preprint arXiv:2206.08932}.

\bibitem[{Stiennon et~al.(2020)Stiennon, Ouyang, Wu, Ziegler, Lowe, Voss,
  Radford, Amodei, and Christiano}]{stiennon2020learning}
N.~Stiennon, L.~Ouyang, J.~Wu, D.~Ziegler, R.~Lowe, C.~Voss, A.~Radford,
  D.~Amodei et~al. 2020.
\newblock Learning to summarize with human feedback.
\newblock In \emph{Conference on Neural Information Processing Systems}.

\bibitem[{Stolfo et~al.(2022)Stolfo, Jin, Shridhar, Schölkopf, and
  Sachan}]{robustness_math_lm}
A.~Stolfo, Z.~Jin, K.~Shridhar, B.~Schölkopf and M.~Sachan. 2022.
\newblock \href {https://doi.org/10.48550/ARXIV.2210.12023} {A causal framework
  to quantify the robustness of mathematical reasoning with language models}.

\bibitem[{Su et~al.(2021)Su, Lu, Pan, Wen, and Liu}]{su2021roformer}
J.~Su, Y.~Lu, S.~Pan, B.~Wen and Y.~Liu. 2021.
\newblock \href {http://arxiv.org/abs/2104.09864} {Roformer: Enhanced
  transformer with rotary position embedding}.

\bibitem[{Sun et~al.(2023)Sun, Liu, Bair, and Kolter}]{sun2023simple}
M.~Sun, Z.~Liu, A.~Bair and J.~Z. Kolter. 2023.
\newblock \href {http://arxiv.org/abs/2306.11695} {A simple and effective
  pruning approach for large language models}.

\bibitem[{Sun et~al.(2022)Sun, Shao, Qian, Huang, and Qiu}]{pmlr-v162-sun22e}
T.~Sun, Y.~Shao, H.~Qian, X.~Huang and X.~Qiu. 2022.
\newblock \href {https://proceedings.mlr.press/v162/sun22e.html} {Black-box
  tuning for language-model-as-a-service}.
\newblock In \emph{Proceedings of the 39th International Conference on Machine
  Learning}, volume 162 of \emph{Proceedings of Machine Learning Research},
  pages 20841--20855. PMLR.

\bibitem[{Sun et~al.(2021{\natexlab{a}})Sun, Ge, Wei, and
  Wang}]{sun-etal-2021-instantaneous}
X.~Sun, T.~Ge, F.~Wei and H.~Wang. 2021{\natexlab{a}}.
\newblock \href {https://doi.org/10.18653/v1/2021.acl-long.462} {Instantaneous
  grammatical error correction with shallow aggressive decoding}.
\newblock In \emph{Proceedings of the 59th Annual Meeting of the Association
  for Computational Linguistics and the 11th International Joint Conference on
  Natural Language Processing (Volume 1: Long Papers)}, pages 5937--5947,
  Online. Association for Computational Linguistics.

\bibitem[{Sun et~al.(2021{\natexlab{b}})Sun, Wang, Feng, Ding, Pang, Shang,
  Liu, Chen, Zhao, Lu et~al.}]{sun2021ernie}
Y.~Sun, S.~Wang, S.~Feng, S.~Ding, C.~Pang, J.~Shang, J.~Liu, X.~Chen et~al.
  2021{\natexlab{b}}.
\newblock Ernie 3.0: Large-scale knowledge enhanced pre-training for language
  understanding and generation.
\newblock \emph{arXiv preprint arXiv:2107.02137}.

\bibitem[{Sun(2023)}]{sun2023short}
Z.~Sun. 2023.
\newblock \href {http://arxiv.org/abs/2303.09136} {A short survey of viewing
  large language models in legal aspect}.

\bibitem[{Sur{\'\i}s et~al.(2023)Sur{\'\i}s, Menon, and
  Vondrick}]{suris2023vipergpt}
D.~Sur{\'\i}s, S.~Menon and C.~Vondrick. 2023.
\newblock Vipergpt: Visual inference via python execution for reasoning.
\newblock \emph{arXiv preprint arXiv:2303.08128}.

\bibitem[{{Susan Zhang [@suchenzang]}(2023)}]{susanzhangPilingPileonSorry2023}
{Susan Zhang [@suchenzang]}. 2023.
\newblock \href {https://twitter.com/suchenzang/status/1617093563061522432}
  {Piling on to the pile-on (sorry - it's always easy to criticize), here's a
  rant about benchmarks for {LLMs} that are used to back claims of "stronger"
  or "better" models. {Let}'s start with a tour through {GPT}-3's {Appendix}
  {G}... 1/8}.

\bibitem[{Suzgun et~al.(2022)Suzgun, Scales, Sch{\"a}rli, Gehrmann, Tay, Chung,
  Chowdhery, Le, Chi, Zhou et~al.}]{suzgun2022challenging}
M.~Suzgun, N.~Scales, N.~Sch{\"a}rli, S.~Gehrmann, Y.~Tay, H.~W. Chung,
  A.~Chowdhery, Q.~V. Le et~al. 2022.
\newblock Challenging big-bench tasks and whether chain-of-thought can solve
  them.
\newblock \emph{arXiv preprint arXiv:2210.09261}.

\bibitem[{Swaminathan et~al.(2023)Swaminathan, Dedieu, Raju, Shanahan,
  Lazaro-Gredilla, and
  George}]{swaminathanSchemalearningRebindingMechanisms2023}
S.~Swaminathan, A.~Dedieu, R.~V. Raju, M.~Shanahan, M.~Lazaro-Gredilla and
  D.~George. 2023.
\newblock \href {https://doi.org/10.48550/arXiv.2307.01201} {Schema-learning
  and rebinding as mechanisms of in-context learning and emergence}.
\newblock ArXiv:2307.01201 [cs].

\bibitem[{Tang et~al.(2021)Tang, Gan, Awan, Rajbhandari, Li, Lian, Liu, Zhang,
  and He}]{tang2021communication}
H.~Tang, S.~Gan, A.~A. Awan, S.~Rajbhandari, C.~Li, X.~Lian, J.~Liu, C.~Zhang
  et~al. 2021.
\newblock \href {https://proceedings.mlr.press/v139/tang21a.html} {1-bit adam:
  Communication efficient large-scale training with adam’s convergence
  speed}.
\newblock In \emph{Proceedings of the 38th International Conference on Machine
  Learning}, volume 139 of \emph{Proceedings of Machine Learning Research},
  pages 10118--10129. PMLR.

\bibitem[{Tang et~al.(2023{\natexlab{a}})Tang, Uberti, and
  Shlomi}]{tangBaselinesIdentifyingWatermarked2023}
L.~Tang, G.~Uberti and T.~Shlomi. 2023{\natexlab{a}}.
\newblock \href {http://arxiv.org/abs/2305.18456} {Baselines for {Identifying}
  {Watermarked} {Large} {Language} {Models}}.
\newblock ArXiv:2305.18456 [cs].

\bibitem[{Tang et~al.(2023{\natexlab{b}})Tang, Sun, Idnay, Nestor, Soroush,
  Elias, Xu, Ding, Durrett, Rousseau et~al.}]{tang2023evaluating}
L.~Tang, Z.~Sun, B.~Idnay, J.~G. Nestor, A.~Soroush, P.~A. Elias, Z.~Xu,
  Y.~Ding et~al. 2023{\natexlab{b}}.
\newblock Evaluating large language models on medical evidence summarization.
\newblock \emph{medRxiv}, pages 2023--04.

\bibitem[{Tang et~al.(2023{\natexlab{c}})Tang, Chuang, and
  Hu}]{tangScienceDetectingLLMGenerated2023}
R.~Tang, Y.-N. Chuang and X.~Hu. 2023{\natexlab{c}}.
\newblock \href {http://arxiv.org/abs/2303.07205} {The {Science} of {Detecting}
  {LLM}-{Generated} {Texts}}.
\newblock ArXiv:2303.07205 [cs].

\bibitem[{Taori et~al.(2023)Taori, Gulrajani, Zhang, Dubois, Li, Guestrin,
  Liang, and Hashimoto}]{taori2023alpaca}
R.~Taori, I.~Gulrajani, T.~Zhang, Y.~Dubois, X.~Li, C.~Guestrin, P.~Liang and
  T.~B. Hashimoto. 2023.
\newblock \href {https://crfm.stanford.edu/2023/03/13/alpaca.html} {Alpaca: A
  strong, replicable instruction-following model}.

\bibitem[{Tay et~al.(2021)Tay, Bahri, Metzler, Juan, Zhao, and
  Zheng}]{tay2021synthesizer}
Y.~Tay, D.~Bahri, D.~Metzler, D.-C. Juan, Z.~Zhao and C.~Zheng. 2021.
\newblock Synthesizer: Rethinking self-attention for transformer models.
\newblock In \emph{International conference on machine learning}, pages
  10183--10192. PMLR.

\bibitem[{Tay et~al.(2022{\natexlab{a}})Tay, Dehghani, Abnar, Chung, Fedus,
  Rao, Narang, Tran, Yogatama, and Metzler}]{tay2022scaling}
Y.~Tay, M.~Dehghani, S.~Abnar, H.~W. Chung, W.~Fedus, J.~Rao, S.~Narang, V.~Q.
  Tran et~al. 2022{\natexlab{a}}.
\newblock \href {http://arxiv.org/abs/2207.10551} {Scaling laws vs model
  architectures: How does inductive bias influence scaling?}

\bibitem[{Tay et~al.(2022{\natexlab{b}})Tay, Dehghani, Bahri, and
  Metzler}]{tay2022efficient}
Y.~Tay, M.~Dehghani, D.~Bahri and D.~Metzler. 2022{\natexlab{b}}.
\newblock Efficient transformers: A survey.
\newblock \emph{ACM Computing Surveys}, 55(6):1--28.

\bibitem[{Tay et~al.(2022{\natexlab{c}})Tay, Dehghani, Rao, Fedus, Abnar,
  Chung, Narang, Yogatama, Vaswani, and
  Metzler}]{tayScaleEfficientlyInsights2022}
Y.~Tay, M.~Dehghani, J.~Rao, W.~Fedus, S.~Abnar, H.~W. Chung, S.~Narang,
  D.~Yogatama et~al. 2022{\natexlab{c}}.
\newblock \href {http://arxiv.org/abs/2109.10686} {Scale {Efficiently}:
  {Insights} from {Pre}-training and {Fine}-tuning {Transformers}}.
\newblock ArXiv:2109.10686 [cs].

\bibitem[{Tay et~al.(2022{\natexlab{d}})Tay, Dehghani, Tran, Garcia, Wei, Wang,
  Chung, Bahri, Schuster, Zheng, Zhou, Houlsby, and Metzler}]{tay2022ul2}
Y.~Tay, M.~Dehghani, V.~Q. Tran, X.~Garcia, J.~Wei, X.~Wang, H.~W. Chung,
  D.~Bahri et~al. 2022{\natexlab{d}}.
\newblock \href {https://doi.org/10.48550/ARXIV.2205.05131} {Ul2: Unifying
  language learning paradigms}.

\bibitem[{Tay et~al.(2022{\natexlab{e}})Tay, Tran, Ruder, Gupta, Chung, Bahri,
  Qin, Baumgartner, Yu, and Metzler}]{tay2022charformer}
Y.~Tay, V.~Q. Tran, S.~Ruder, J.~Gupta, H.~W. Chung, D.~Bahri, Z.~Qin,
  S.~Baumgartner et~al. 2022{\natexlab{e}}.
\newblock \href {http://arxiv.org/abs/2106.12672} {Charformer: Fast character
  transformers via gradient-based subword tokenization}.

\bibitem[{Tay et~al.(2022{\natexlab{f}})Tay, Wei, Chung, Tran, So, Shakeri,
  Garcia, Zheng, Rao, Chowdhery, Zhou, Metzler, Petrov, Houlsby, Le, and
  Dehghani}]{tay2022upalm}
Y.~Tay, J.~Wei, H.~W. Chung, V.~Q. Tran, D.~R. So, S.~Shakeri, X.~Garcia, H.~S.
  Zheng et~al. 2022{\natexlab{f}}.
\newblock \href {https://doi.org/10.48550/ARXIV.2210.11399} {Transcending
  scaling laws with 0.1\% extra compute}.

\bibitem[{Taylor et~al.(2022)Taylor, Kardas, Cucurull, Scialom, Hartshorn,
  Saravia, Poulton, Kerkez, and Stojnic}]{taylor2022galactica}
R.~Taylor, M.~Kardas, G.~Cucurull, T.~Scialom, A.~Hartshorn, E.~Saravia,
  A.~Poulton, V.~Kerkez et~al. 2022.
\newblock Galactica: A large language model for science.
\newblock \emph{arXiv preprint arXiv:2211.09085}.

\bibitem[{Taylor(1953)}]{taylor1953cloze}
W.~L. Taylor. 1953.
\newblock “cloze procedure”: A new tool for measuring readability.
\newblock \emph{Journalism quarterly}, 30(4):415--433.

\bibitem[{Thiergart et~al.(2021)Thiergart, Huber, and
  {\"U}bellacker}]{thiergart2021understanding}
J.~Thiergart, S.~Huber and T.~{\"U}bellacker. 2021.
\newblock Understanding emails and drafting responses--an approach using gpt-3.
\newblock \emph{arXiv preprint arXiv:2102.03062}.

\bibitem[{Thoppilan et~al.(2022)Thoppilan, De~Freitas, Hall, Shazeer,
  Kulshreshtha, Cheng, Jin, Bos, Baker, Du et~al.}]{lamda}
R.~Thoppilan, D.~De~Freitas, J.~Hall, N.~Shazeer, A.~Kulshreshtha, H.-T. Cheng,
  A.~Jin, T.~Bos et~al. 2022.
\newblock Lamda: Language models for dialog applications.
\newblock \emph{arXiv preprint arXiv:2201.08239}.

\bibitem[{Tian et~al.(2020)Tian, Narayan, Sellam, and
  Parikh}]{tianStickingFactsConfident2020}
R.~Tian, S.~Narayan, T.~Sellam and A.~P. Parikh. 2020.
\newblock \href {http://arxiv.org/abs/1910.08684} {Sticking to the {Facts}:
  {Confident} {Decoding} for {Faithful} {Data}-to-{Text} {Generation}}.
\newblock ArXiv:1910.08684 [cs].

\bibitem[{Tirumala et~al.()Tirumala, Markosyan, Zettlemoyer, and
  Aghajanyan}]{tirumalaMemorizationOverfittingAnalyzing}
K.~Tirumala, A.~H. Markosyan, L.~Zettlemoyer and A.~Aghajanyan.
\newblock Memorization {Without} {Overﬁtting}: {Analyzing} the {Training}
  {Dynamics} of {Large} {Language} {Models}.

\bibitem[{To et~al.(2023)To, Bui, Guo, and Nguyen}]{to2023better}
H.~Q. To, N.~D. Bui, J.~Guo and T.~N. Nguyen. 2023.
\newblock Better language models of code through self-improvement.
\newblock \emph{arXiv preprint arXiv:2304.01228}.

\bibitem[{Tornede et~al.(2023)Tornede, Deng, Eimer, Giovanelli, Mohan, Ruhkopf,
  Segel, Theodorakopoulos, Tornede, Wachsmuth, and
  Lindauer}]{tornedeAutoMLAgeLarge2023}
A.~Tornede, D.~Deng, T.~Eimer, J.~Giovanelli, A.~Mohan, T.~Ruhkopf, S.~Segel,
  D.~Theodorakopoulos et~al. 2023.
\newblock \href {http://arxiv.org/abs/2306.08107} {{AutoML} in the {Age} of
  {Large} {Language} {Models}: {Current} {Challenges}, {Future} {Opportunities}
  and {Risks}}.
\newblock ArXiv:2306.08107 [cs].

\bibitem[{Touvron et~al.(2023)Touvron, Lavril, Izacard, Martinet, Lachaux,
  Lacroix, Rozière, Goyal, Hambro, Azhar, Rodriguez, Joulin, Grave, and
  Lample}]{touvronLLaMAOpenEfficient2023}
H.~Touvron, T.~Lavril, G.~Izacard, X.~Martinet, M.-A. Lachaux, T.~Lacroix,
  B.~Rozière, N.~Goyal et~al. 2023.
\newblock \href {http://arxiv.org/abs/2302.13971} {{LLaMA}: {Open} and
  {Efficient} {Foundation} {Language} {Models}}.
\newblock ArXiv:2302.13971 [cs].

\bibitem[{Touvron et~al.()Touvron, Martin, and
  Stone}]{touvronLlamaOpenFoundationa}
H.~Touvron, L.~Martin and K.~Stone.
\newblock Llama 2: {Open} {Foundation} and {Fine}-{Tuned} {Chat} {Models}.

\bibitem[{Tran et~al.(2023)Tran, Khadkikar, and Porollo}]{tran2023survey}
C.~Tran, S.~Khadkikar and A.~Porollo. 2023.
\newblock Survey of protein sequence embedding models.
\newblock \emph{International Journal of Molecular Sciences}, 24(4):3775.

\bibitem[{Uchendu et~al.(2020)Uchendu, Le, Shu, and
  Lee}]{uchenduAuthorshipAttributionNeural2020}
A.~Uchendu, T.~Le, K.~Shu and D.~Lee. 2020.
\newblock \href {https://doi.org/10.18653/v1/2020.emnlp-main.673} {Authorship
  {Attribution} for {Neural} {Text} {Generation}}.
\newblock In \emph{Proceedings of the 2020 {Conference} on {Empirical}
  {Methods} in {Natural} {Language} {Processing} ({EMNLP})}, pages 8384--8395,
  Online. Association for Computational Linguistics.

\bibitem[{Uesato et~al.(2022)Uesato, Kushman, Kumar, Song, Siegel, Wang,
  Creswell, Irving, and Higgins}]{solving_math_problems}
J.~Uesato, N.~Kushman, R.~Kumar, F.~Song, N.~Siegel, L.~Wang, A.~Creswell,
  G.~Irving et~al. 2022.
\newblock \href {https://doi.org/10.48550/ARXIV.2211.14275} {Solving math word
  problems with process- and outcome-based feedback}.

\bibitem[{University(2023)}]{helm_website}
S.~University. 2023.
\newblock Holistic evaluation of langauge models results page.
\newblock Available from:
  \url{https://crfm.stanford.edu/helm/latest/?groups=1}.
\newblock Accessed: 23/03/2023.

\bibitem[{Valmeekam et~al.(2023)Valmeekam, Olmo, Sreedharan, and
  Kambhampati}]{valmeekam2023large}
K.~Valmeekam, A.~Olmo, S.~Sreedharan and S.~Kambhampati. 2023.
\newblock \href {http://arxiv.org/abs/2206.10498} {Large language models still
  can't plan (a benchmark for llms on planning and reasoning about change)}.

\bibitem[{Vaswani et~al.(2017)Vaswani, Shazeer, Parmar, Uszkoreit, Jones,
  Gomez, Kaiser, and Polosukhin}]{transformers}
A.~Vaswani, N.~Shazeer, N.~Parmar, J.~Uszkoreit, L.~Jones, A.~N. Gomez, L.~u.
  Kaiser and I.~Polosukhin. 2017.
\newblock \href
  {https://proceedings.neurips.cc/paper/2017/file/3f5ee243547dee91fbd053c1c4a845aa-Paper.pdf}
  {Attention is all you need}.
\newblock In \emph{Advances in Neural Information Processing Systems},
  volume~30. Curran Associates, Inc.

\bibitem[{Vemprala et~al.(2023)Vemprala, Bonatti, Bucker, and
  Kapoor}]{chatgpt_robo}
S.~Vemprala, R.~Bonatti, A.~Bucker and A.~Kapoor. 2023.
\newblock Chatgpt for robotics: Design principles and model abilities.

\bibitem[{Venigalla et~al.(2022)Venigalla, Frankle, and Carbin}]{pubmedgpt}
A.~Venigalla, J.~Frankle and M.~Carbin. 2022.
\newblock Pubmed gpt: A domain- specific large language model for biomedical
  text.
\newblock \url{https://www.mosaicml.com/blog/introducing-pubmed-gpt}.
\newblock Accessed: 2023-01-24.

\bibitem[{Verkuil et~al.(2022)Verkuil, Kabeli, Du, Wicky, Milles, Dauparas,
  Baker, Ovchinnikov, Sercu, and Rives}]{verkuil2022language}
R.~Verkuil, O.~Kabeli, Y.~Du, B.~I. Wicky, L.~F. Milles, J.~Dauparas, D.~Baker,
  S.~Ovchinnikov et~al. 2022.
\newblock Language models generalize beyond natural proteins.
\newblock \emph{bioRxiv}, pages 2022--12.

\bibitem[{Vijayakumar et~al.(2018)Vijayakumar, Cogswell, Selvaraju, Sun, Lee,
  Crandall, and Batra}]{vijayakumar2018dbs}
A.~Vijayakumar, M.~Cogswell, R.~Selvaraju, Q.~Sun, S.~Lee, D.~Crandall and
  D.~Batra. 2018.
\newblock \href {https://doi.org/10.1609/aaai.v32i1.12340} {Diverse beam search
  for improved description of complex scenes}.
\newblock \emph{Proceedings of the AAAI Conference on Artificial Intelligence},
  32(1).

\bibitem[{Villalobos et~al.(2022)Villalobos, Sevilla, Heim, Besiroglu,
  Hobbhahn, and Ho}]{villalobos2022will}
P.~Villalobos, J.~Sevilla, L.~Heim, T.~Besiroglu, M.~Hobbhahn and A.~Ho. 2022.
\newblock Will we run out of data? an analysis of the limits of scaling
  datasets in machine learning.
\newblock \emph{arXiv preprint arXiv:2211.04325}.

\bibitem[{Viswanath and Zhang(2023)}]{viswanath2023fairpy}
H.~Viswanath and T.~Zhang. 2023.
\newblock Fairpy: A toolkit for evaluation of social biases and their
  mitigation in large language models.
\newblock \emph{arXiv preprint arXiv:2302.05508}.

\bibitem[{von Oswald et~al.(2022)von Oswald, Niklasson, Randazzo, Sacramento,
  Mordvintsev, Zhmoginov, and Vladymyrov}]{von2022transformers}
J.~von Oswald, E.~Niklasson, E.~Randazzo, J.~Sacramento, A.~Mordvintsev,
  A.~Zhmoginov and M.~Vladymyrov. 2022.
\newblock Transformers learn in-context by gradient descent.
\newblock \emph{arXiv preprint arXiv:2212.07677}.

\bibitem[{Vries(2023)}]{vriesGoSmolGo2023}
H.~d. Vries. 2023.
\newblock \href
  {https://www.harmdevries.com/post/model-size-vs-compute-overhead/} {Go smol
  or go home}.

\bibitem[{Vu et~al.(2022)Vu, Lester, Constant, Al-Rfou{'}, and
  Cer}]{vu-etal-2022-spot}
T.~Vu, B.~Lester, N.~Constant, R.~Al-Rfou{'} and D.~Cer. 2022.
\newblock \href {https://doi.org/10.18653/v1/2022.acl-long.346} {{SP}o{T}:
  Better frozen model adaptation through soft prompt transfer}.
\newblock In \emph{Proceedings of the 60th Annual Meeting of the Association
  for Computational Linguistics (Volume 1: Long Papers)}, pages 5039--5059,
  Dublin, Ireland. Association for Computational Linguistics.

\bibitem[{Wahle et~al.(2022{\natexlab{a}})Wahle, Ruas, Folt{\`y}nek, Meuschke,
  and Gipp}]{wahle2022identifying}
J.~P. Wahle, T.~Ruas, T.~Folt{\`y}nek, N.~Meuschke and B.~Gipp.
  2022{\natexlab{a}}.
\newblock Identifying machine-paraphrased plagiarism.
\newblock In \emph{International Conference on Information}, pages 393--413.
  Springer.

\bibitem[{Wahle et~al.(2022{\natexlab{b}})Wahle, Ruas, Kirstein, and
  Gipp}]{wahle2022large}
J.~P. Wahle, T.~Ruas, F.~Kirstein and B.~Gipp. 2022{\natexlab{b}}.
\newblock How large language models are transforming machine-paraphrased
  plagiarism.
\newblock \emph{arXiv preprint arXiv:2210.03568}.

\bibitem[{Wang et~al.(2018)Wang, Singh, Michael, Hill, Levy, and
  Bowman}]{wang-etal-2018-glue}
A.~Wang, A.~Singh, J.~Michael, F.~Hill, O.~Levy and S.~Bowman. 2018.
\newblock \href {https://doi.org/10.18653/v1/W18-5446} {{GLUE}: A multi-task
  benchmark and analysis platform for natural language understanding}.
\newblock In \emph{Proceedings of the 2018 {EMNLP} Workshop {B}lackbox{NLP}:
  Analyzing and Interpreting Neural Networks for {NLP}}, pages 353--355,
  Brussels, Belgium. Association for Computational Linguistics.

\bibitem[{Wang and Komatsuzaki(2021)}]{gpt-j}
B.~Wang and A.~Komatsuzaki. 2021.
\newblock {GPT-J-6B: A 6 Billion Parameter Autoregressive Language Model}.
\newblock \url{https://github.com/kingoflolz/mesh-transformer-jax}.

\bibitem[{Wang et~al.(2020)Wang, Cho, and Gu}]{wang2020neural}
C.~Wang, K.~Cho and J.~Gu. 2020.
\newblock \href {https://doi.org/10.1609/aaai.v34i05.6451} {Neural machine
  translation with byte-level subwords}.
\newblock \emph{Proceedings of the AAAI Conference on Artificial Intelligence},
  34(05):9154--9160.

\bibitem[{Wang et~al.(2022{\natexlab{a}})Wang, Liu, Chen, Hong, Tang, and
  Song}]{wang-etal-2022-deepstruct}
C.~Wang, X.~Liu, Z.~Chen, H.~Hong, J.~Tang and D.~Song. 2022{\natexlab{a}}.
\newblock \href {https://doi.org/10.18653/v1/2022.findings-acl.67}
  {{D}eep{S}truct: Pretraining of language models for structure prediction}.
\newblock In \emph{Findings of the Association for Computational Linguistics:
  ACL 2022}, pages 803--823, Dublin, Ireland. Association for Computational
  Linguistics.

\bibitem[{Wang et~al.(2023{\natexlab{a}})Wang, Xie, Jiang, Mandlekar, Xiao,
  Zhu, Fan, and Anandkumar}]{wang2023voyager}
G.~Wang, Y.~Xie, Y.~Jiang, A.~Mandlekar, C.~Xiao, Y.~Zhu, L.~Fan and
  A.~Anandkumar. 2023{\natexlab{a}}.
\newblock Voyager: An open-ended embodied agent with large language models.
\newblock \emph{arXiv preprint arXiv:2305.16291}.

\bibitem[{Wang et~al.(2022{\natexlab{b}})Wang, Kaddour, Liu, Tang, Kusner,
  Lasenby, and Liu}]{wang2022evaluating}
H.~Wang, J.~Kaddour, S.~Liu, J.~Tang, M.~Kusner, J.~Lasenby and Q.~Liu.
  2022{\natexlab{b}}.
\newblock Evaluating self-supervised learning for molecular graph embeddings.
\newblock \emph{arXiv preprint arXiv:2206.08005}.

\bibitem[{Wang et~al.(2023{\natexlab{b}})Wang, Li, Chen, Zhu, Lin, Cao, Liu,
  Liu, and Sui}]{wangLargeLanguageModels2023}
P.~Wang, L.~Li, L.~Chen, D.~Zhu, B.~Lin, Y.~Cao, Q.~Liu, T.~Liu et~al.
  2023{\natexlab{b}}.
\newblock \href {http://arxiv.org/abs/2305.17926} {Large {Language} {Models}
  are not {Fair} {Evaluators}}.
\newblock ArXiv:2305.17926 [cs].

\bibitem[{Wang et~al.(2023{\natexlab{c}})Wang, Wang, Mi, Chen, Xu, and
  Wong}]{wang2023self}
R.~Wang, H.~Wang, F.~Mi, Y.~Chen, R.~Xu and K.-F. Wong. 2023{\natexlab{c}}.
\newblock Self-critique prompting with large language models for inductive
  instructions.
\newblock \emph{arXiv preprint arXiv:2305.13733}.

\bibitem[{Wang et~al.(2021)Wang, Liu, Xu, Zhu, and Zeng}]{label_cost}
S.~Wang, Y.~Liu, Y.~Xu, C.~Zhu and M.~Zeng. 2021.
\newblock \href {https://doi.org/10.48550/ARXIV.2108.13487} {Want to reduce
  labeling cost? gpt-3 can help}.

\bibitem[{Wang et~al.(2023{\natexlab{d}})Wang, Menon, Long, Henderson, Li,
  Crowston, Hansen, Nickerson, and Chilton}]{wang2023reelframer}
S.~Wang, S.~Menon, T.~Long, K.~Henderson, D.~Li, K.~Crowston, M.~Hansen, J.~V.
  Nickerson et~al. 2023{\natexlab{d}}.
\newblock Reelframer: Co-creating news reels on social media with generative
  ai.
\newblock \emph{arXiv preprint arXiv:2304.09653}.

\bibitem[{Wang et~al.(2022{\natexlab{c}})Wang, Wei, Schuurmans, Le, Chi,
  Narang, Chowdhery, and Zhou}]{wang2022selfconsistency}
X.~Wang, J.~Wei, D.~Schuurmans, Q.~Le, E.~Chi, S.~Narang, A.~Chowdhery and
  D.~Zhou. 2022{\natexlab{c}}.
\newblock \href {https://doi.org/10.48550/ARXIV.2203.11171} {Self-consistency
  improves chain of thought reasoning in language models}.

\bibitem[{Wang et~al.(2023{\natexlab{e}})Wang, Yu, Zeng, Yang, Wang, Chen,
  Jiang, Xie, Wang, Xie et~al.}]{wang2023pandalm}
Y.~Wang, Z.~Yu, Z.~Zeng, L.~Yang, C.~Wang, H.~Chen, C.~Jiang, R.~Xie et~al.
  2023{\natexlab{e}}.
\newblock Pandalm: An automatic evaluation benchmark for llm instruction tuning
  optimization.
\newblock \emph{arXiv preprint arXiv:2306.05087}.

\bibitem[{Wang(2021)}]{wang-2021-comment}
Y.~Wang. 2021.
\newblock \href {https://aclanthology.org/2021.hackashop-1.12} {Comment section
  personalization: Algorithmic, interface, and interaction design}.
\newblock In \emph{Proceedings of the EACL Hackashop on News Media Content
  Analysis and Automated Report Generation}, pages 84--88, Online. Association
  for Computational Linguistics.

\bibitem[{Wang et~al.(2022{\natexlab{d}})Wang, Kordi, Mishra, Liu, Smith,
  Khashabi, and Hajishirzi}]{wang2022selfinstruct}
Y.~Wang, Y.~Kordi, S.~Mishra, A.~Liu, N.~A. Smith, D.~Khashabi and
  H.~Hajishirzi. 2022{\natexlab{d}}.
\newblock \href {https://doi.org/10.48550/ARXIV.2212.10560} {Self-instruct:
  Aligning language model with self generated instructions}.

\bibitem[{Wang et~al.(2022{\natexlab{e}})Wang, Mishra, Alipoormolabashi, Kordi,
  Mirzaei, Naik, Ashok, Dhanasekaran, Arunkumar, Stap et~al.}]{wang2022super}
Y.~Wang, S.~Mishra, P.~Alipoormolabashi, Y.~Kordi, A.~Mirzaei, A.~Naik,
  A.~Ashok, A.~S. Dhanasekaran et~al. 2022{\natexlab{e}}.
\newblock Super-naturalinstructions: Generalization via declarative
  instructions on 1600+ nlp tasks.
\newblock In \emph{Proceedings of the 2022 Conference on Empirical Methods in
  Natural Language Processing}, pages 5085--5109.

\bibitem[{Wang et~al.(2023{\natexlab{f}})Wang, Zhao, and
  Petzold}]{wang2023large}
Y.~Wang, Y.~Zhao and L.~Petzold. 2023{\natexlab{f}}.
\newblock \href {http://arxiv.org/abs/2304.05368} {Are large language models
  ready for healthcare? a comparative study on clinical language
  understanding}.

\bibitem[{Wang et~al.(2023{\natexlab{g}})Wang, Cai, Liu, Ma, and
  Liang}]{wang2023describe}
Z.~Wang, S.~Cai, A.~Liu, X.~Ma and Y.~Liang. 2023{\natexlab{g}}.
\newblock Describe, explain, plan and select: Interactive planning with large
  language models enables open-world multi-task agents.
\newblock \emph{arXiv preprint arXiv:2302.01560}.

\bibitem[{Wang et~al.(2019{\natexlab{a}})Wang, Wohlwend, and
  Lei}]{wang2019structured}
Z.~Wang, J.~Wohlwend and T.~Lei. 2019{\natexlab{a}}.
\newblock Structured pruning of large language models.
\newblock \emph{arXiv preprint arXiv:1910.04732}.

\bibitem[{Wang et~al.(2019{\natexlab{b}})Wang, Dai, P{\'o}czos, and
  Carbonell}]{wang2019characterizing}
Z.~Wang, Z.~Dai, B.~P{\'o}czos and J.~Carbonell. 2019{\natexlab{b}}.
\newblock Characterizing and avoiding negative transfer.
\newblock In \emph{Proceedings of the IEEE/CVF conference on computer vision
  and pattern recognition}, pages 11293--11302.

\bibitem[{Wang et~al.(2013)Wang, Zoghi, Hutter, Matheson, De~Freitas
  et~al.}]{wang2013bayesian}
Z.~Wang, M.~Zoghi, F.~Hutter, D.~Matheson, N.~De~Freitas et~al. 2013.
\newblock Bayesian optimization in high dimensions via random embeddings.
\newblock In \emph{IJCAI}, volume~13, pages 1778--1784.

\bibitem[{Webb et~al.(2022)Webb, Holyoak, and Lu}]{analogical_reasoning}
T.~Webb, K.~J. Holyoak and H.~Lu. 2022.
\newblock \href {https://doi.org/10.48550/ARXIV.2212.09196} {Emergent
  analogical reasoning in large language models}.

\bibitem[{Webson and Pavlick(2022)}]{webson-pavlick-2022-prompt}
A.~Webson and E.~Pavlick. 2022.
\newblock \href {https://doi.org/10.18653/v1/2022.naacl-main.167} {Do
  prompt-based models really understand the meaning of their prompts?}
\newblock In \emph{Proceedings of the 2022 Conference of the North American
  Chapter of the Association for Computational Linguistics: Human Language
  Technologies}, pages 2300--2344, Seattle, United States. Association for
  Computational Linguistics.

\bibitem[{Wei et~al.(2023)Wei, Haghtalab, and
  Steinhardt}]{weiJailbrokenHowDoes2023}
A.~Wei, N.~Haghtalab and J.~Steinhardt. 2023.
\newblock \href {http://arxiv.org/abs/2307.02483} {Jailbroken: {How} {Does}
  {LLM} {Safety} {Training} {Fail}?}
\newblock ArXiv:2307.02483 [cs].

\bibitem[{Wei et~al.(2022{\natexlab{a}})Wei, Bosma, Zhao, Guu, Yu, Lester, Du,
  Dai, and Le}]{wei2022finetuned}
J.~Wei, M.~Bosma, V.~Zhao, K.~Guu, A.~W. Yu, B.~Lester, N.~Du, A.~M. Dai et~al.
  2022{\natexlab{a}}.
\newblock \href {https://openreview.net/forum?id=gEZrGCozdqR} {Finetuned
  language models are zero-shot learners}.
\newblock In \emph{International Conference on Learning Representations}.

\bibitem[{Wei et~al.(2022{\natexlab{b}})Wei, Tay, Bommasani, Raffel, Zoph,
  Borgeaud, Yogatama, Bosma, Zhou, Metzler, Chi, Hashimoto, Vinyals, Liang,
  Dean, and Fedus}]{wei2022emergent}
J.~Wei, Y.~Tay, R.~Bommasani, C.~Raffel, B.~Zoph, S.~Borgeaud, D.~Yogatama,
  M.~Bosma et~al. 2022{\natexlab{b}}.
\newblock \href {http://arxiv.org/abs/2206.07682} {Emergent abilities of large
  language models}.

\bibitem[{Wei et~al.(2022{\natexlab{c}})Wei, Tay, and Le}]{wei2022inverse}
J.~Wei, Y.~Tay and Q.~V. Le. 2022{\natexlab{c}}.
\newblock Inverse scaling can become u-shaped.
\newblock \emph{arXiv preprint arXiv:2211.02011}.

\bibitem[{Wei et~al.(2022{\natexlab{d}})Wei, Wang, Schuurmans, Bosma, brian
  ichter, Xia, Chi, Le, and Zhou}]{wei2022chain}
J.~Wei, X.~Wang, D.~Schuurmans, M.~Bosma, brian ichter, F.~Xia, E.~H. Chi,
  Q.~V. Le et~al. 2022{\natexlab{d}}.
\newblock \href {https://openreview.net/forum?id=_VjQlMeSB_J} {Chain of thought
  prompting elicits reasoning in large language models}.
\newblock In \emph{Advances in Neural Information Processing Systems}.

\bibitem[{Weidinger et~al.(2021)Weidinger, Mellor, Rauh, Griffin, Uesato,
  Huang, Cheng, Glaese, Balle, Kasirzadeh et~al.}]{weidinger2021ethical}
L.~Weidinger, J.~Mellor, M.~Rauh, C.~Griffin, J.~Uesato, P.-S. Huang, M.~Cheng,
  M.~Glaese et~al. 2021.
\newblock Ethical and social risks of harm from language models.
\newblock \emph{arXiv preprint arXiv:2112.04359}.

\bibitem[{Weiss(2019)}]{weiss2019deepfake}
M.~Weiss. 2019.
\newblock Deepfake bot submissions to federal public comment websites cannot be
  distinguished from human submissions.
\newblock \emph{Technology Science}, 2019121801.

\bibitem[{Welleck et~al.(2019)Welleck, Kulikov, Roller, Dinan, Cho, and
  Weston}]{welleck2019neural}
S.~Welleck, I.~Kulikov, S.~Roller, E.~Dinan, K.~Cho and J.~Weston. 2019.
\newblock Neural text generation with unlikelihood training.
\newblock \emph{arXiv preprint arXiv:1908.04319}.

\bibitem[{Weng(2023{\natexlab{a}})}]{weng2023inference}
L.~Weng. 2023{\natexlab{a}}.
\newblock \href
  {https://lilianweng.github.io/posts/2023-01-10-inference-optimization/}
  {Large transformer model inference optimization}.
\newblock \emph{Lil'Log}.

\bibitem[{Weng(2023{\natexlab{b}})}]{weng2023prompt}
L.~Weng. 2023{\natexlab{b}}.
\newblock \href
  {https://lilianweng.github.io/posts/2023-03-15-prompt-engineering/} {Prompt
  engineering}.
\newblock \emph{lilianweng.github.io}.

\bibitem[{Willig et~al.(2023)Willig, ZE{\v{C}}EVI{\'C}, Dhami, and
  Kersting}]{willig2023causal}
M.~Willig, M.~ZE{\v{C}}EVI{\'C}, D.~S. Dhami and K.~Kersting. 2023.
\newblock Causal parrots: Large language models may talk causality but are not
  causal.
\newblock \emph{preprint}.

\bibitem[{Winkelmolen et~al.(2020)Winkelmolen, Ivkin, Bozkurt, and
  Karnin}]{winkelmolen2020practical}
F.~Winkelmolen, N.~Ivkin, H.~F. Bozkurt and Z.~Karnin. 2020.
\newblock Practical and sample efficient zero-shot hpo.
\newblock \emph{arXiv preprint arXiv:2007.13382}.

\bibitem[{Wolf et~al.(2023)Wolf, Wies, Levine, and
  Shashua}]{wolf2023fundamental}
Y.~Wolf, N.~Wies, Y.~Levine and A.~Shashua. 2023.
\newblock Fundamental limitations of alignment in large language models.
\newblock \emph{arXiv preprint arXiv:2304.11082}.

\bibitem[{Wornow et~al.(2023)Wornow, Xu, Thapa, Patel, Steinberg, Fleming,
  Pfeffer, Fries, and Shah}]{wornow2023shaky}
M.~Wornow, Y.~Xu, R.~Thapa, B.~Patel, E.~Steinberg, S.~Fleming, M.~A. Pfeffer,
  J.~Fries et~al. 2023.
\newblock \href {http://arxiv.org/abs/2303.12961} {The shaky foundations of
  clinical foundation models: A survey of large language models and foundation
  models for emrs}.

\bibitem[{Wu et~al.(2023{\natexlab{a}})Wu, Radev, and Xu}]{wu2023geometric}
F.~Wu, D.~Radev and J.~Xu. 2023{\natexlab{a}}.
\newblock When geometric deep learning meets pretrained protein language
  models.
\newblock \emph{bioRxiv}, pages 2023--01.

\bibitem[{Wu et~al.(2021{\natexlab{a}})Wu, Ouyang, Ziegler, Stiennon, Lowe,
  Leike, and Christiano}]{wu2021recursively}
J.~Wu, L.~Ouyang, D.~M. Ziegler, N.~Stiennon, R.~Lowe, J.~Leike and
  P.~Christiano. 2021{\natexlab{a}}.
\newblock Recursively summarizing books with human feedback.
\newblock \emph{arXiv preprint arXiv:2109.10862}.

\bibitem[{Wu et~al.(2022{\natexlab{a}})Wu, Wu, Jiang, Liu, and
  Zhao}]{wu2022tfold}
J.~Wu, F.~Wu, B.~Jiang, W.~Liu and P.~Zhao. 2022{\natexlab{a}}.
\newblock tfold-ab: Fast and accurate antibody structure prediction without
  sequence homologs.
\newblock \emph{bioRxiv}, pages 2022--11.

\bibitem[{Wu et~al.(2023{\natexlab{b}})Wu, Tucker, Nagler, and
  Messing}]{wu2023large}
P.~Y. Wu, J.~A. Tucker, J.~Nagler and S.~Messing. 2023{\natexlab{b}}.
\newblock \href {http://arxiv.org/abs/2303.12057} {Large language models can be
  used to estimate the ideologies of politicians in a zero-shot learning
  setting}.

\bibitem[{Wu et~al.(2021{\natexlab{b}})Wu, Zhao, Yu, Zhang, Shen, Liu, Li, Zhu,
  Luo, Xu, and Zhang}]{yuan}
S.~Wu, X.~Zhao, T.~Yu, R.~Zhang, C.~Shen, H.~Liu, F.~Li, H.~Zhu et~al.
  2021{\natexlab{b}}.
\newblock \href {https://doi.org/10.48550/ARXIV.2110.04725} {Yuan 1.0:
  Large-scale pre-trained language model in zero-shot and few-shot learning}.

\bibitem[{Wu et~al.(2023{\natexlab{c}})Wu, Irsoy, Lu, Dabravolski, Dredze,
  Gehrmann, Kambadur, Rosenberg, and Mann}]{wu2023bloomberggpt}
S.~Wu, O.~Irsoy, S.~Lu, V.~Dabravolski, M.~Dredze, S.~Gehrmann, P.~Kambadur,
  D.~Rosenberg et~al. 2023{\natexlab{c}}.
\newblock \href {http://arxiv.org/abs/2303.17564} {Bloomberggpt: A large
  language model for finance}.

\bibitem[{Wu et~al.(2016)Wu, Schuster, Chen, Le, Norouzi, Macherey, Krikun,
  Cao, Gao, Macherey, Klingner, Shah, Johnson, Liu, Kaiser, Gouws, Kato, Kudo,
  Kazawa, Stevens, Kurian, Patil, Wang, Young, Smith, Riesa, Rudnick, Vinyals,
  Corrado, Hughes, and Dean}]{wu2016wordpiece}
Y.~Wu, M.~Schuster, Z.~Chen, Q.~V. Le, M.~Norouzi, W.~Macherey, M.~Krikun,
  Y.~Cao et~al. 2016.
\newblock \href {https://doi.org/10.48550/ARXIV.1609.08144} {Google's neural
  machine translation system: Bridging the gap between human and machine
  translation}.

\bibitem[{Wu et~al.(2022{\natexlab{b}})Wu, Gardner, Stenetorp, and
  Dasigi}]{wu2022generating}
Y.~Wu, M.~Gardner, P.~Stenetorp and P.~Dasigi. 2022{\natexlab{b}}.
\newblock Generating data to mitigate spurious correlations in natural language
  inference datasets.
\newblock \emph{arXiv preprint arXiv:2203.12942}.

\bibitem[{Wu et~al.(2023{\natexlab{d}})Wu, Qiu, Ross, Akyürek, Chen, Wang,
  Kim, Andreas, and Kim}]{wuReasoningRecitingExploring2023}
Z.~Wu, L.~Qiu, A.~Ross, E.~Akyürek, B.~Chen, B.~Wang, N.~Kim, J.~Andreas
  et~al. 2023{\natexlab{d}}.
\newblock \href {http://arxiv.org/abs/2307.02477} {Reasoning or {Reciting}?
  {Exploring} the {Capabilities} and {Limitations} of {Language} {Models}
  {Through} {Counterfactual} {Tasks}}.
\newblock ArXiv:2307.02477 [cs].

\bibitem[{Xiao and Wang(2021)}]{xiaoHallucinationPredictiveUncertainty2021}
Y.~Xiao and W.~Y. Wang. 2021.
\newblock \href {http://arxiv.org/abs/2103.15025} {On {Hallucination} and
  {Predictive} {Uncertainty} in {Conditional} {Language} {Generation}}.
\newblock ArXiv:2103.15025 [cs].

\bibitem[{Xie et~al.(2023{\natexlab{a}})Xie, Luo, Wang, and
  Ananiadou}]{xie2023survey}
Q.~Xie, Z.~Luo, B.~Wang and S.~Ananiadou. 2023{\natexlab{a}}.
\newblock \href {http://arxiv.org/abs/2304.08763} {A survey on biomedical text
  summarization with pre-trained language model}.

\bibitem[{Xie et~al.(2023{\natexlab{b}})Xie, Pham, Dong, Du, Liu, Lu, Liang,
  Le, Ma, and Yu}]{xieDoReMiOptimizingData2023}
S.~M. Xie, H.~Pham, X.~Dong, N.~Du, H.~Liu, Y.~Lu, P.~Liang, Q.~V. Le et~al.
  2023{\natexlab{b}}.
\newblock \href {http://arxiv.org/abs/2305.10429} {{DoReMi}: {Optimizing}
  {Data} {Mixtures} {Speeds} {Up} {Language} {Model} {Pretraining}}.
\newblock ArXiv:2305.10429 [cs].

\bibitem[{Xie et~al.(2022)Xie, Raghunathan, Liang, and
  Ma}]{xieExplanationIncontextLearning2022}
S.~M. Xie, A.~Raghunathan, P.~Liang and T.~Ma. 2022.
\newblock \href {https://doi.org/10.48550/arXiv.2111.02080} {An {Explanation}
  of {In}-context {Learning} as {Implicit} {Bayesian} {Inference}}.
\newblock ArXiv:2111.02080 [cs].

\bibitem[{Xie et~al.(2023{\natexlab{c}})Xie, Santurkar, Ma, and
  Liang}]{xieDataSelectionLanguage2023}
S.~M. Xie, S.~Santurkar, T.~Ma and P.~Liang. 2023{\natexlab{c}}.
\newblock \href {http://arxiv.org/abs/2302.03169} {Data {Selection} for
  {Language} {Models} via {Importance} {Resampling}}.
\newblock ArXiv:2302.03169 [cs].

\bibitem[{Xu et~al.(2023{\natexlab{a}})Xu, Sun, Zheng, Geng, Zhao, Feng, Tao,
  and Jiang}]{xu2023wizardlm}
C.~Xu, Q.~Sun, K.~Zheng, X.~Geng, P.~Zhao, J.~Feng, C.~Tao and D.~Jiang.
  2023{\natexlab{a}}.
\newblock Wizardlm: Empowering large language models to follow complex
  instructions.
\newblock \emph{arXiv preprint arXiv:2304.12244}.

\bibitem[{Xu et~al.(2022)Xu, Alon, Neubig, and Hellendoorn}]{code_evaluation}
F.~F. Xu, U.~Alon, G.~Neubig and V.~J. Hellendoorn. 2022.
\newblock \href {https://doi.org/10.48550/ARXIV.2202.13169} {A systematic
  evaluation of large language models of code}.

\bibitem[{Xu et~al.(2023{\natexlab{b}})Xu, Yuan, Miret, and
  Tang}]{xu2023protst}
M.~Xu, X.~Yuan, S.~Miret and J.~Tang. 2023{\natexlab{b}}.
\newblock Protst: Multi-modality learning of protein sequences and biomedical
  texts.
\newblock \emph{arXiv preprint arXiv:2301.12040}.

\bibitem[{Xu et~al.(2021)Xu, Lee, Chen, Hechtman, Huang, Joshi, Krikun,
  Lepikhin, Ly, Maggioni et~al.}]{xu2021gspmd}
Y.~Xu, H.~Lee, D.~Chen, B.~Hechtman, Y.~Huang, R.~Joshi, M.~Krikun, D.~Lepikhin
  et~al. 2021.
\newblock Gspmd: general and scalable parallelization for ml computation
  graphs.
\newblock \emph{arXiv preprint arXiv:2105.04663}.

\bibitem[{Xue et~al.(2022{\natexlab{a}})Xue, Barua, Constant, Al-Rfou, Narang,
  Kale, Roberts, and Raffel}]{xueByT5TokenfreeFuture2022}
L.~Xue, A.~Barua, N.~Constant, R.~Al-Rfou, S.~Narang, M.~Kale, A.~Roberts and
  C.~Raffel. 2022{\natexlab{a}}.
\newblock \href {http://arxiv.org/abs/2105.13626} {{ByT5}: {Towards} a
  token-free future with pre-trained byte-to-byte models}.
\newblock ArXiv:2105.13626 [cs].

\bibitem[{Xue et~al.(2022{\natexlab{b}})Xue, Barua, Constant, Al-Rfou, Narang,
  Kale, Roberts, and Raffel}]{xue-etal-2022-byt5}
L.~Xue, A.~Barua, N.~Constant, R.~Al-Rfou, S.~Narang, M.~Kale, A.~Roberts and
  C.~Raffel. 2022{\natexlab{b}}.
\newblock \href {https://doi.org/10.1162/tacl_a_00461} {{B}y{T}5: Towards a
  token-free future with pre-trained byte-to-byte models}.
\newblock \emph{Transactions of the Association for Computational Linguistics},
  10:291--306.

\bibitem[{Xue et~al.(2021)Xue, Constant, Roberts, Kale, Al-Rfou, Siddhant,
  Barua, and Raffel}]{xue-etal-2021-mt5}
L.~Xue, N.~Constant, A.~Roberts, M.~Kale, R.~Al-Rfou, A.~Siddhant, A.~Barua and
  C.~Raffel. 2021.
\newblock \href {https://doi.org/10.18653/v1/2021.naacl-main.41} {m{T}5: A
  massively multilingual pre-trained text-to-text transformer}.
\newblock In \emph{Proceedings of the 2021 Conference of the North American
  Chapter of the Association for Computational Linguistics: Human Language
  Technologies}, pages 483--498, Online. Association for Computational
  Linguistics.

\bibitem[{Yan et~al.(2023)Yan, Sha, Zhao, Li, Martinez-Maldonado, Chen, Li,
  Jin, and Gašević}]{yan2023practical}
L.~Yan, L.~Sha, L.~Zhao, Y.~Li, R.~Martinez-Maldonado, G.~Chen, X.~Li, Y.~Jin
  et~al. 2023.
\newblock \href {http://arxiv.org/abs/2303.13379} {Practical and ethical
  challenges of large language models in education: A systematic literature
  review}.

\bibitem[{Yang et~al.(2021)Yang, Hu, Babuschkin, Sidor, Liu, Farhi, Ryder,
  Pachocki, Chen, and Gao}]{yang2021tuning}
G.~Yang, E.~Hu, I.~Babuschkin, S.~Sidor, X.~Liu, D.~Farhi, N.~Ryder,
  J.~Pachocki et~al. 2021.
\newblock Tuning large neural networks via zero-shot hyperparameter transfer.
\newblock \emph{Advances in Neural Information Processing Systems},
  34:17084--17097.

\bibitem[{Yang et~al.(2023{\natexlab{a}})Yang, Jin, Tang, Han, Feng, Jiang,
  Yin, and Hu}]{yang2023harnessing}
J.~Yang, H.~Jin, R.~Tang, X.~Han, Q.~Feng, H.~Jiang, B.~Yin and X.~Hu.
  2023{\natexlab{a}}.
\newblock \href {http://arxiv.org/abs/2304.13712} {Harnessing the power of llms
  in practice: A survey on chatgpt and beyond}.

\bibitem[{Yang and Klein(2021)}]{yang2021fudge}
K.~Yang and D.~Klein. 2021.
\newblock Fudge: Controlled text generation with future discriminators.
\newblock \emph{arXiv preprint arXiv:2104.05218}.

\bibitem[{Yang et~al.(2022{\natexlab{a}})Yang, Klein, Peng, and
  Tian}]{yang2022doc}
K.~Yang, D.~Klein, N.~Peng and Y.~Tian. 2022{\natexlab{a}}.
\newblock Doc: Improving long story coherence with detailed outline control.
\newblock \emph{arXiv preprint arXiv:2212.10077}.

\bibitem[{Yang et~al.(2022{\natexlab{b}})Yang, Peng, Tian, and
  Klein}]{yang2022re3}
K.~Yang, N.~Peng, Y.~Tian and D.~Klein. 2022{\natexlab{b}}.
\newblock Re3: Generating longer stories with recursive reprompting and
  revision.
\newblock \emph{arXiv preprint arXiv:2210.06774}.

\bibitem[{Yang et~al.(2023{\natexlab{b}})Yang, Chen, Zhang, Liu, Qi, Zhang,
  Fang, and Yu}]{yangWatermarkingTextGenerated2023}
X.~Yang, K.~Chen, W.~Zhang, C.~Liu, Y.~Qi, J.~Zhang, H.~Fang and N.~Yu.
  2023{\natexlab{b}}.
\newblock \href {http://arxiv.org/abs/2305.08883} {Watermarking {Text}
  {Generated} by {Black}-{Box} {Language} {Models}}.
\newblock ArXiv:2305.08883 [cs].

\bibitem[{Yao et~al.(2023{\natexlab{a}})Yao, Yu, Zhao, Shafran, Griffiths, Cao,
  and Narasimhan}]{yaoTreeThoughtsDeliberate2023}
S.~Yao, D.~Yu, J.~Zhao, I.~Shafran, T.~L. Griffiths, Y.~Cao and K.~Narasimhan.
  2023{\natexlab{a}}.
\newblock \href {http://arxiv.org/abs/2305.10601} {Tree of {Thoughts}:
  {Deliberate} {Problem} {Solving} with {Large} {Language} {Models}}.
\newblock ArXiv:2305.10601 [cs].

\bibitem[{Yao et~al.(2022{\natexlab{a}})Yao, Zhao, Yu, Du, Shafran, Narasimhan,
  and Cao}]{yao2022react}
S.~Yao, J.~Zhao, D.~Yu, N.~Du, I.~Shafran, K.~Narasimhan and Y.~Cao.
  2022{\natexlab{a}}.
\newblock React: Synergizing reasoning and acting in language models.
\newblock \emph{arXiv preprint arXiv:2210.03629}.

\bibitem[{Yao et~al.(2022{\natexlab{b}})Yao, Zheng, Yang, and
  Yang}]{yaoNLPScratchLargeScale2022}
X.~Yao, Y.~Zheng, X.~Yang and Z.~Yang. 2022{\natexlab{b}}.
\newblock \href {https://proceedings.mlr.press/v162/yao22c.html} {{NLP} {From}
  {Scratch} {Without} {Large}-{Scale} {Pretraining}: {A} {Simple} and
  {Efficient} {Framework}}.
\newblock In \emph{Proceedings of the 39th {International} {Conference} on
  {Machine} {Learning}}, pages 25438--25451. PMLR.
\newblock ISSN: 2640-3498.

\bibitem[{Yao et~al.(2023{\natexlab{b}})Yao, Wang, Tian, Cheng, Li, Deng, Chen,
  and Zhang}]{yaoEditingLargeLanguage2023}
Y.~Yao, P.~Wang, B.~Tian, S.~Cheng, Z.~Li, S.~Deng, H.~Chen and N.~Zhang.
  2023{\natexlab{b}}.
\newblock \href {http://arxiv.org/abs/2305.13172} {Editing {Large} {Language}
  {Models}: {Problems}, {Methods}, and {Opportunities}}.
\newblock ArXiv:2305.13172 [cs].

\bibitem[{Yao et~al.(2022{\natexlab{c}})Yao, Aminabadi, Zhang, Wu, Li, and
  He}]{yao2022zeroquant}
Z.~Yao, R.~Y. Aminabadi, M.~Zhang, X.~Wu, C.~Li and Y.~He. 2022{\natexlab{c}}.
\newblock Zeroquant: Efficient and affordable post-training quantization for
  large-scale transformers.
\newblock \emph{arXiv preprint arXiv:2206.01861}.

\bibitem[{Yasunaga et~al.(2022)Yasunaga, Bosselut, Ren, Zhang, Manning, Liang,
  and Leskovec}]{yasunaga2022deep}
M.~Yasunaga, A.~Bosselut, H.~Ren, X.~Zhang, C.~D. Manning, P.~Liang and
  J.~Leskovec. 2022.
\newblock Deep bidirectional language-knowledge graph pretraining.
\newblock \emph{arXiv preprint arXiv:2210.09338}.

\bibitem[{Yi et~al.(2019)Yi, Goel, Khatri, Cervone, Chung, Hedayatnia,
  Venkatesh, Gabriel, and Hakkani-Tur}]{yi-etal-2019-towards}
S.~Yi, R.~Goel, C.~Khatri, A.~Cervone, T.~Chung, B.~Hedayatnia, A.~Venkatesh,
  R.~Gabriel et~al. 2019.
\newblock \href {https://doi.org/10.18653/v1/W19-8608} {Towards coherent and
  engaging spoken dialog response generation using automatic conversation
  evaluators}.
\newblock In \emph{Proceedings of the 12th International Conference on Natural
  Language Generation}, pages 65--75, Tokyo, Japan. Association for
  Computational Linguistics.

\bibitem[{Yogatama et~al.(2021)Yogatama, de~Masson~d{'}Autume, and
  Kong}]{yogatama-etal-2021-adaptive}
D.~Yogatama, C.~de~Masson~d{'}Autume and L.~Kong. 2021.
\newblock \href {https://doi.org/10.1162/tacl_a_00371} {Adaptive semiparametric
  language models}.
\newblock \emph{Transactions of the Association for Computational Linguistics},
  9:362--373.

\bibitem[{Yoneda et~al.(2023)Yoneda, Fang, Li, Zhang, Jiang, Lin, Picker,
  Yunis, Mei, and Walter}]{yoneda2023statler}
T.~Yoneda, J.~Fang, P.~Li, H.~Zhang, T.~Jiang, S.~Lin, B.~Picker, D.~Yunis
  et~al. 2023.
\newblock \href {http://arxiv.org/abs/2306.17840} {Statler: State-maintaining
  language models for embodied reasoning}.

\bibitem[{Yoo et~al.(2021)Yoo, Park, Kang, Lee, and
  Park}]{yoo-etal-2021-gpt3mix-leveraging}
K.~M. Yoo, D.~Park, J.~Kang, S.-W. Lee and W.~Park. 2021.
\newblock \href {https://doi.org/10.18653/v1/2021.findings-emnlp.192}
  {{GPT}3{M}ix: Leveraging large-scale language models for text augmentation}.
\newblock In \emph{Findings of the Association for Computational Linguistics:
  EMNLP 2021}, pages 2225--2239, Punta Cana, Dominican Republic. Association
  for Computational Linguistics.

\bibitem[{Yoo et~al.(2023)Yoo, Ahn, Jang, and
  Kwak}]{yooRobustNaturalLanguage2023}
K.~Yoo, W.~Ahn, J.~Jang and N.~Kwak. 2023.
\newblock \href {http://arxiv.org/abs/2305.01904} {Robust {Natural} {Language}
  {Watermarking} through {Invariant} {Features}}.
\newblock ArXiv:2305.01904 [cs].

\bibitem[{You et~al.(2021)You, Liu, Mamitsuka, and Zhu}]{you2021bertmesh}
R.~You, Y.~Liu, H.~Mamitsuka and S.~Zhu. 2021.
\newblock Bertmesh: deep contextual representation learning for large-scale
  high-performance mesh indexing with full text.
\newblock \emph{Bioinformatics}, 37(5):684--692.

\bibitem[{Yu et~al.(2022{\natexlab{a}})Yu, Quartey, and Schilder}]{yu2022legal}
F.~Yu, L.~Quartey and F.~Schilder. 2022{\natexlab{a}}.
\newblock Legal prompting: Teaching a language model to think like a lawyer.
\newblock \emph{arXiv preprint arXiv:2212.01326}.

\bibitem[{Yu et~al.(2023)Yu, Simig, Flaherty, Aghajanyan, Zettlemoyer, and
  Lewis}]{yu2023megabyte}
L.~Yu, D.~Simig, C.~Flaherty, A.~Aghajanyan, L.~Zettlemoyer and M.~Lewis. 2023.
\newblock Megabyte: Predicting million-byte sequences with multiscale
  transformers.
\newblock \emph{arXiv preprint arXiv:2305.07185}.

\bibitem[{Yu et~al.(2022{\natexlab{b}})Yu, Artetxe, Ott, Shleifer, Gong,
  Stoyanov, and Li}]{smlp}
P.~Yu, M.~Artetxe, M.~Ott, S.~Shleifer, H.~Gong, V.~Stoyanov and X.~Li.
  2022{\natexlab{b}}.
\newblock \href {https://doi.org/10.48550/ARXIV.2203.06850} {Efficient language
  modeling with sparse all-mlp}.

\bibitem[{Yu et~al.(2022{\natexlab{c}})Yu, Wang, Golovneva, Alkhamissy, Ghosh,
  Diab, and Celikyilmaz}]{yu2022alert}
P.~Yu, T.~Wang, O.~Golovneva, B.~Alkhamissy, G.~Ghosh, M.~Diab and
  A.~Celikyilmaz. 2022{\natexlab{c}}.
\newblock Alert: Adapting language models to reasoning tasks.
\newblock \emph{arXiv preprint arXiv:2212.08286}.

\bibitem[{Yunxiang et~al.(2023)Yunxiang, Zihan, Kai, Ruilong, and
  You}]{yunxiang2023chatdoctor}
L.~Yunxiang, L.~Zihan, Z.~Kai, D.~Ruilong and Z.~You. 2023.
\newblock \href {http://arxiv.org/abs/2303.14070} {Chatdoctor: A medical chat
  model fine-tuned on llama model using medical domain knowledge}.

\bibitem[{Zelikman et~al.(2022)Zelikman, Wu, Mu, and
  Goodman}]{zelikman2022star}
E.~Zelikman, Y.~Wu, J.~Mu and N.~Goodman. 2022.
\newblock \href {https://openreview.net/forum?id=_3ELRdg2sgI} {{ST}ar:
  Bootstrapping reasoning with reasoning}.
\newblock In \emph{Advances in Neural Information Processing Systems}.

\bibitem[{Zellers et~al.(2019)Zellers, Holtzman, Rashkin, Bisk, Farhadi,
  Roesner, and Choi}]{zellers2019defending}
R.~Zellers, A.~Holtzman, H.~Rashkin, Y.~Bisk, A.~Farhadi, F.~Roesner and
  Y.~Choi. 2019.
\newblock Defending against neural fake news.
\newblock \emph{Advances in neural information processing systems}, 32.

\bibitem[{Zeng et~al.(2022)Zeng, Liu, Du, Wang, Lai, Ding, Yang, Xu, Zheng,
  Xia, Tam, Ma, Xue, Zhai, Chen, Zhang, Dong, and Tang}]{zeng2022glm130b}
A.~Zeng, X.~Liu, Z.~Du, Z.~Wang, H.~Lai, M.~Ding, Z.~Yang, Y.~Xu et~al. 2022.
\newblock \href {https://doi.org/10.48550/ARXIV.2210.02414} {Glm-130b: An open
  bilingual pre-trained model}.

\bibitem[{Zeng et~al.(2021)Zeng, Ren, Su, Wang, Liao, Wang, Jiang, Yang, Wang,
  Zhang, Li, Gong, Yao, Huang, Wang, Yu, Guo, Yu, Zhang, Wang, Tao, Yan, Yi,
  Peng, Jiang, Zhang, Deng, Zhang, Lin, Zhang, Zhang, Guo, Gu, Fan, Wang, Jin,
  Liu, and Tian}]{pangu}
W.~Zeng, X.~Ren, T.~Su, H.~Wang, Y.~Liao, Z.~Wang, X.~Jiang, Z.~Yang et~al.
  2021.
\newblock \href {https://doi.org/10.48550/ARXIV.2104.12369} {Pangu-$\alpha$:
  Large-scale autoregressive pretrained chinese language models with
  auto-parallel computation}.

\bibitem[{Zhang et~al.(2023{\natexlab{a}})Zhang, Chen, Zhang, Liu, Zan, Mao,
  Lou, and Chen}]{zhang2023repocoder}
F.~Zhang, B.~Chen, Y.~Zhang, J.~Liu, D.~Zan, Y.~Mao, J.-G. Lou and W.~Chen.
  2023{\natexlab{a}}.
\newblock \href {http://arxiv.org/abs/2303.12570} {Repocoder: Repository-level
  code completion through iterative retrieval and generation}.

\bibitem[{Zhang et~al.(2022{\natexlab{a}})Zhang, Li, Meng, Chang, and
  Broeck}]{zhangParadoxLearningReason2022a}
H.~Zhang, L.~H. Li, T.~Meng, K.-W. Chang and G.~V.~d. Broeck.
  2022{\natexlab{a}}.
\newblock \href {http://arxiv.org/abs/2205.11502} {On the {Paradox} of
  {Learning} to {Reason} from {Data}}.
\newblock ArXiv:2205.11502 [cs].

\bibitem[{Zhang et~al.(2021{\natexlab{a}})Zhang, Duckworth, Ippolito, and
  Neelakantan}]{zhang-etal-2021-trading}
H.~Zhang, D.~Duckworth, D.~Ippolito and A.~Neelakantan. 2021{\natexlab{a}}.
\newblock \href {https://aclanthology.org/2021.humeval-1.3} {Trading off
  diversity and quality in natural language generation}.
\newblock In \emph{Proceedings of the Workshop on Human Evaluation of NLP
  Systems (HumEval)}, pages 25--33, Online. Association for Computational
  Linguistics.

\bibitem[{Zhang and He(2020)}]{zhang2020accelerating}
M.~Zhang and Y.~He. 2020.
\newblock \href {http://arxiv.org/abs/2010.13369} {Accelerating training of
  transformer-based language models with progressive layer dropping}.

\bibitem[{Zhang et~al.(2023{\natexlab{b}})Zhang, Press, Merrill, Liu, and
  Smith}]{zhangHowLanguageModel2023}
M.~Zhang, O.~Press, W.~Merrill, A.~Liu and N.~A. Smith. 2023{\natexlab{b}}.
\newblock \href {http://arxiv.org/abs/2305.13534} {How {Language} {Model}
  {Hallucinations} {Can} {Snowball}}.
\newblock ArXiv:2305.13534 [cs].

\bibitem[{Zhang(2023)}]{Suchenzang_2023}
S.~Zhang. 2023.
\newblock [...] that's an unhelpful order of magnitude difference in how large
  of a model you should be training in order to be considered ``compute
  optimal''.
\newblock \url{https://twitter.com/suchenzang/status/1616752494608007171?s=20}.
\newblock Accessed: 2023-06-06.

\bibitem[{Zhang et~al.(2022{\natexlab{b}})Zhang, Roller, Goyal, Artetxe, Chen,
  Chen, Dewan, Diab, Li, Lin, Mihaylov, Ott, Shleifer, Shuster, Simig, Koura,
  Sridhar, Wang, and Zettlemoyer}]{zhang2022opt}
S.~Zhang, S.~Roller, N.~Goyal, M.~Artetxe, M.~Chen, S.~Chen, C.~Dewan, M.~Diab
  et~al. 2022{\natexlab{b}}.
\newblock \href {https://doi.org/10.48550/ARXIV.2205.01068} {Opt: Open
  pre-trained transformer language models}.

\bibitem[{Zhang et~al.(2019)Zhang, Kishore, Wu, Weinberger, and
  Artzi}]{zhang2019bertscore}
T.~Zhang, V.~Kishore, F.~Wu, K.~Q. Weinberger and Y.~Artzi. 2019.
\newblock Bertscore: Evaluating text generation with bert.
\newblock \emph{arXiv preprint arXiv:1904.09675}.

\bibitem[{Zhang et~al.(2023{\natexlab{c}})Zhang, Ladhak, Durmus, Liang,
  McKeown, and Hashimoto}]{news_summary}
T.~Zhang, F.~Ladhak, E.~Durmus, P.~Liang, K.~McKeown and T.~B. Hashimoto.
  2023{\natexlab{c}}.
\newblock \href {https://doi.org/10.48550/ARXIV.2301.13848} {Benchmarking large
  language models for news summarization}.

\bibitem[{Zhang et~al.(2021{\natexlab{b}})Zhang, Gu, Han, Chen, Xiao, Sun, Yao,
  Qi, Guan, Ke, Cai, Zeng, Tan, Liu, Huang, Han, Liu, Zhu, and Sun}]{cpm2}
Z.~Zhang, Y.~Gu, X.~Han, S.~Chen, C.~Xiao, Z.~Sun, Y.~Yao, F.~Qi et~al.
  2021{\natexlab{b}}.
\newblock \href {https://doi.org/10.48550/ARXIV.2106.10715} {Cpm-2: Large-scale
  cost-effective pre-trained language models}.

\bibitem[{Zhang et~al.(2022{\natexlab{c}})Zhang, Lin, Liu, Li, Sun, and
  Zhou}]{zhang2022moefication}
Z.~Zhang, Y.~Lin, Z.~Liu, P.~Li, M.~Sun and J.~Zhou. 2022{\natexlab{c}}.
\newblock \href {http://arxiv.org/abs/2110.01786} {Moefication: Transformer
  feed-forward layers are mixtures of experts}.

\bibitem[{Zhang et~al.(2022{\natexlab{d}})Zhang, Zhang, Li, and
  Smola}]{zhang2022autocot}
Z.~Zhang, A.~Zhang, M.~Li and A.~Smola. 2022{\natexlab{d}}.
\newblock \href {https://doi.org/10.48550/ARXIV.2210.03493} {Automatic chain of
  thought prompting in large language models}.

\bibitem[{Zhao et~al.(2023{\natexlab{a}})Zhao, Wen, Tuan, Zhao, and
  Fu}]{zhao2023prompt}
S.~Zhao, J.~Wen, L.~A. Tuan, J.~Zhao and J.~Fu. 2023{\natexlab{a}}.
\newblock Prompt as triggers for backdoor attack: Examining the vulnerability
  in language models.
\newblock \emph{arXiv preprint arXiv:2305.01219}.

\bibitem[{Zhao et~al.(2023{\natexlab{b}})Zhao, Zhou, Li, Tang, Wang, Hou, Min,
  Zhang, Zhang, Dong, Du, Yang, Chen, Chen, Jiang, Ren, Li, Tang, Liu, Liu,
  Nie, and Wen}]{zhaoSurveyLargeLanguage2023}
W.~X. Zhao, K.~Zhou, J.~Li, T.~Tang, X.~Wang, Y.~Hou, Y.~Min, B.~Zhang et~al.
  2023{\natexlab{b}}.
\newblock \href {http://arxiv.org/abs/2303.18223} {A {Survey} of {Large}
  {Language} {Models}}.
\newblock ArXiv:2303.18223 [cs].

\bibitem[{Zhao et~al.(2023{\natexlab{c}})Zhao, Gu, Varma, Luo, Huang, Xu,
  Wright, Shojanazeri, Ott, Shleifer et~al.}]{zhao2023pytorch}
Y.~Zhao, A.~Gu, R.~Varma, L.~Luo, C.-C. Huang, M.~Xu, L.~Wright, H.~Shojanazeri
  et~al. 2023{\natexlab{c}}.
\newblock Pytorch fsdp: experiences on scaling fully sharded data parallel.
\newblock \emph{arXiv preprint arXiv:2304.11277}.

\bibitem[{Zhao et~al.(2021)Zhao, Wallace, Feng, Klein, and
  Singh}]{zhao-etal-2021-calibrate}
Z.~Zhao, E.~Wallace, S.~Feng, D.~Klein and S.~Singh. 2021.
\newblock \href {https://proceedings.mlr.press/v139/zhao21c.html} {Calibrate
  before use: Improving few-shot performance of language models}.
\newblock In \emph{Proceedings of the 38th International Conference on Machine
  Learning}, volume 139 of \emph{Proceedings of Machine Learning Research},
  pages 12697--12706. PMLR.

\bibitem[{Zheng et~al.(2021)Zheng, Dong, Huang, Singhal, Che, Liu, Song, and
  Wei}]{zheng2021allocating}
B.~Zheng, L.~Dong, S.~Huang, S.~Singhal, W.~Che, T.~Liu, X.~Song and F.~Wei.
  2021.
\newblock Allocating large vocabulary capacity for cross-lingual language model
  pre-training.
\newblock \emph{arXiv preprint arXiv:2109.07306}.

\bibitem[{Zheng et~al.(2022)Zheng, Li, Zhang, Zhuang, Chen, Huang, Wang, Xu,
  Zhuo, Xing, Gonzalez, and Stoica}]{zheng2022alpa}
L.~Zheng, Z.~Li, H.~Zhang, Y.~Zhuang, Z.~Chen, Y.~Huang, Y.~Wang, Y.~Xu et~al.
  2022.
\newblock \href
  {https://www.usenix.org/conference/osdi22/presentation/zheng-lianmin} {Alpa:
  Automating inter- and {Intra-Operator} parallelism for distributed deep
  learning}.
\newblock In \emph{16th USENIX Symposium on Operating Systems Design and
  Implementation (OSDI 22)}, pages 559--578, Carlsbad, CA. USENIX Association.

\bibitem[{Zheng et~al.(2023)Zheng, Dou, Gao, Shen, Wang, Liu, Jin, Liu, Xiong,
  Chen, Xi, Zhou, Xu, Lai, Zhu, Weng, Cheng, Chang, Yin, Hua, Huang, Sun, Yan,
  Gui, Zhang, Qiu, and Huang}]{zhengSecretsRLHFLarge2023}
R.~Zheng, S.~Dou, S.~Gao, W.~Shen, B.~Wang, Y.~Liu, S.~Jin, Q.~Liu et~al. 2023.
\newblock \href {http://arxiv.org/abs/2307.04964} {Secrets of {RLHF} in {Large}
  {Language} {Models} {Part} {I}: {PPO}}.
\newblock ArXiv:2307.04964 [cs].

\bibitem[{Zhong et~al.(2023)Zhong, Cui, Guo, Liang, Lu, Wang, Saied, Chen, and
  Duan}]{zhong2023agieval}
W.~Zhong, R.~Cui, Y.~Guo, Y.~Liang, S.~Lu, Y.~Wang, A.~Saied, W.~Chen et~al.
  2023.
\newblock Agieval: A human-centric benchmark for evaluating foundation models.
\newblock \emph{arXiv preprint arXiv:2304.06364}.

\bibitem[{Zhou et~al.(2021)Zhou, Ma, Zhu, Liu, Zhang, Yuan, Sun, and
  Li}]{zhou2021sparsity}
A.~Zhou, Y.~Ma, J.~Zhu, J.~Liu, Z.~Zhang, K.~Yuan, W.~Sun and H.~Li. 2021.
\newblock \href {https://openreview.net/forum?id=K9bw7vqp\_s} {Learning {N:}
  {M} fine-grained structured sparse neural networks from scratch}.
\newblock In \emph{9th International Conference on Learning Representations,
  {ICLR} 2021, Virtual Event, Austria, May 3-7, 2021}. OpenReview.net.

\bibitem[{Zhou et~al.(2023{\natexlab{a}})Zhou, Liu, Xu, Iyer, Sun, Mao, Ma,
  Efrat, Yu, Yu, Zhang, Ghosh, Lewis, Zettlemoyer, and
  Levy}]{zhouLIMALessMore2023}
C.~Zhou, P.~Liu, P.~Xu, S.~Iyer, J.~Sun, Y.~Mao, X.~Ma, A.~Efrat et~al.
  2023{\natexlab{a}}.
\newblock \href {http://arxiv.org/abs/2305.11206} {{LIMA}: {Less} {Is} {More}
  for {Alignment}}.
\newblock ArXiv:2305.11206 [cs].

\bibitem[{Zhou et~al.(2022)Zhou, Schärli, Hou, Wei, Scales, Wang, Schuurmans,
  Cui, Bousquet, Le, and Chi}]{zhou2022least}
D.~Zhou, N.~Schärli, L.~Hou, J.~Wei, N.~Scales, X.~Wang, D.~Schuurmans, C.~Cui
  et~al. 2022.
\newblock \href {https://doi.org/10.48550/ARXIV.2205.10625} {Least-to-most
  prompting enables complex reasoning in large language models}.

\bibitem[{Zhou et~al.(2023{\natexlab{b}})Zhou, Muresanu, Han, Paster, Pitis,
  Chan, and Ba}]{zhou2023large}
Y.~Zhou, A.~I. Muresanu, Z.~Han, K.~Paster, S.~Pitis, H.~Chan and J.~Ba.
  2023{\natexlab{b}}.
\newblock \href {https://openreview.net/forum?id=92gvk82DE-} {Large language
  models are human-level prompt engineers}.
\newblock In \emph{International Conference on Learning Representations}.

\bibitem[{Zhu et~al.(2015)Zhu, Kiros, Zemel, Salakhutdinov, Urtasun, Torralba,
  and Fidler}]{bookcorpus_1}
Y.~Zhu, R.~Kiros, R.~Zemel, R.~Salakhutdinov, R.~Urtasun, A.~Torralba and
  S.~Fidler. 2015.
\newblock \href {https://doi.org/10.48550/ARXIV.1506.06724} {Aligning books and
  movies: Towards story-like visual explanations by watching movies and reading
  books}.

\bibitem[{Zhuang et~al.(2023)Zhuang, Liu, Pan, He, Weng, and
  Shen}]{zhuang2023survey}
B.~Zhuang, J.~Liu, Z.~Pan, H.~He, Y.~Weng and C.~Shen. 2023.
\newblock A survey on efficient training of transformers.
\newblock \emph{arXiv preprint arXiv:2302.01107}.

\bibitem[{Ziegler et~al.(2019)Ziegler, Stiennon, Wu, Brown, Radford, Amodei,
  Christiano, and Irving}]{ziegler2019fine}
D.~M. Ziegler, N.~Stiennon, J.~Wu, T.~B. Brown, A.~Radford, D.~Amodei,
  P.~Christiano and G.~Irving. 2019.
\newblock Fine-tuning language models from human preferences.
\newblock \emph{arXiv preprint arXiv:1909.08593}.

\bibitem[{Zoph et~al.(2022)Zoph, Bello, Kumar, Du, Huang, Dean, Shazeer, and
  Fedus}]{stmoe}
B.~Zoph, I.~Bello, S.~Kumar, N.~Du, Y.~Huang, J.~Dean, N.~Shazeer and W.~Fedus.
  2022.
\newblock \href {https://doi.org/10.48550/ARXIV.2202.08906} {St-moe: Designing
  stable and transferable sparse expert models}.

\bibitem[{Zvyagin et~al.(2022)Zvyagin, Brace, Hippe, Deng, Zhang, Bohorquez,
  Clyde, Kale, Perez-Rivera, Ma et~al.}]{zvyagin2022genslms}
M.~Zvyagin, A.~Brace, K.~Hippe, Y.~Deng, B.~Zhang, C.~O. Bohorquez, A.~Clyde,
  B.~Kale et~al. 2022.
\newblock Genslms: Genome-scale language models reveal sars-cov-2 evolutionary
  dynamics.
\newblock \emph{bioRxiv}, pages 2022--10.

\end{thebibliography}
